\documentclass[10pt,twocolumn,letterpaper]{article}
\usepackage[pagenumbers]{cvpr}
\usepackage[dvipsnames]{xcolor}
\definecolor{cvprblue}{rgb}{0.21,0.49,0.74}
\usepackage[pagebackref,breaklinks,colorlinks,citecolor=cvprblue]{hyperref}
\usepackage{multirow}

\newcommand{\mypar}[1]{\noindent\textbf{#1}.}
\newcommand{\vfnerf}{VF-NeRF }
\newcommand{\monosdf}{\textsc{\small{MonoSDF}}\xspace}
\newcommand{\msdf}{\textsc{\small{ManhattanSDF}}\xspace}

\renewcommand{\Re}{\mathbb{R}}
\newcommand{\loc}{\mathbf{x}}

\newcommand{\direction}{\mathbf{d}}

\newcommand{\pixel}{\mathbf{p}}
\newcommand{\ray}{\mathbf{r}}
\newcommand{\surfaceLoc}{\mathbf{x}_S}
\DeclareMathOperator*{\argmin}{arg\,min}
\newcommand{\vf}{\mathbf{v}}
\newcommand{\loss}{\mathcal{L}}
\newcommand{\radiance}{\mathbf{c}}
\newcommand{\cossim}{c_{sim}}
\newcommand{\slidingWindow}{\mathbf{w}}
\newcommand{\pixelBatch}{\mathcal{P}}
\newcommand{\pointsExterior}{\mathcal{P}_{ext}}
\newcommand{\pointsCenter}{\mathcal{P}_{cen}}
\newcommand{\vfNetwork}{\mathbf{f}_{\phi}}
\newcommand{\radianceNetwork}{\mathbf{c}_{\psi}}
\newcommand{\vfFunction}{\mathbf{f}}
\newcommand{\featVec}{\mathbf{z}}
\newcommand{\NaPos}{\mathbb{N}_{\geq 0}}
\newcommand\boldblue[1]{\textcolor{RoyalBlue}{\textbf{#1}}}
\newcommand\boldgreen[1]{\textcolor{OliveGreen}{\textbf{#1}}}

\title{VF-NeRF: Learning Neural Vector Fields for Indoor Scene Reconstruction}

\author{Albert Gassol Puigjaner$^{*1}$
\qquad
Edoardo Mello Rella$^{*1}$
\qquad
Erik Sandström$^1$
\\
Ajad Chhatkuli$^{1,3}$
\qquad
Luc Van Gool$^{1,2,3}$\\
$^1$Computer Vision Lab, ETH Zurich \qquad $^2$VISICS, KU Leuven \qquad $^3$INSAIT, Sofia
}

\let\oldtwocolumn\twocolumn
\renewcommand\twocolumn[1][]{%
    \oldtwocolumn[{#1}{
    \begin{center}
    \vspace{-2em}
        \begin{tabular}{cccc}
        \centering
        {\includegraphics[width=.475\columnwidth]{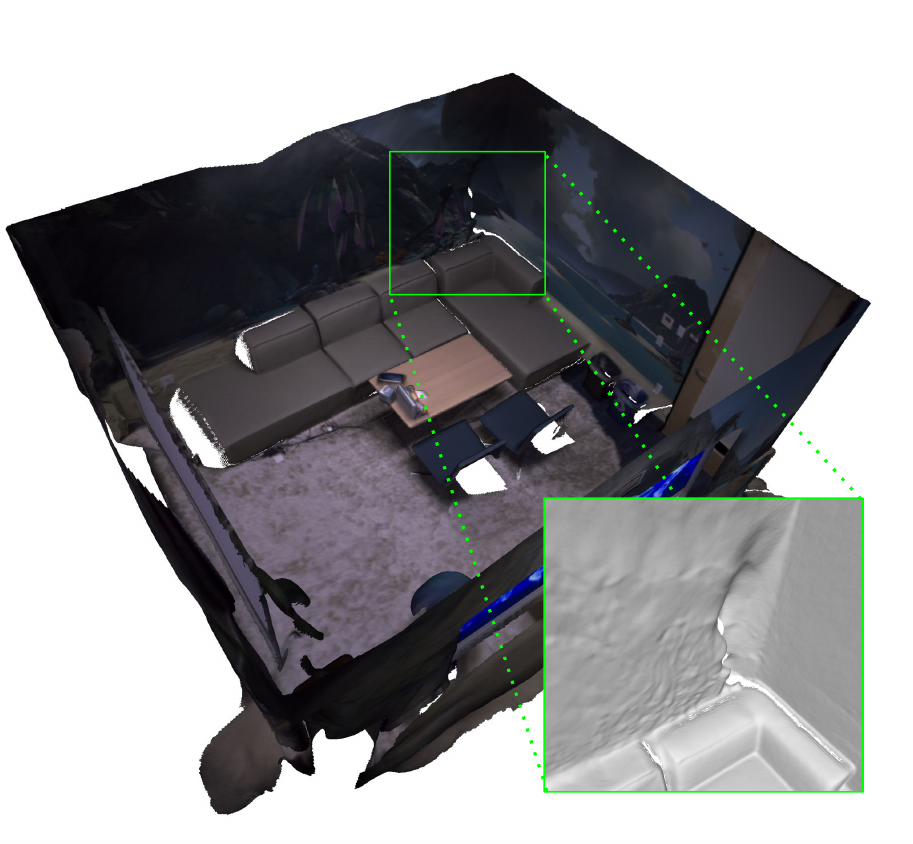}} & {\includegraphics[width=.475\columnwidth]{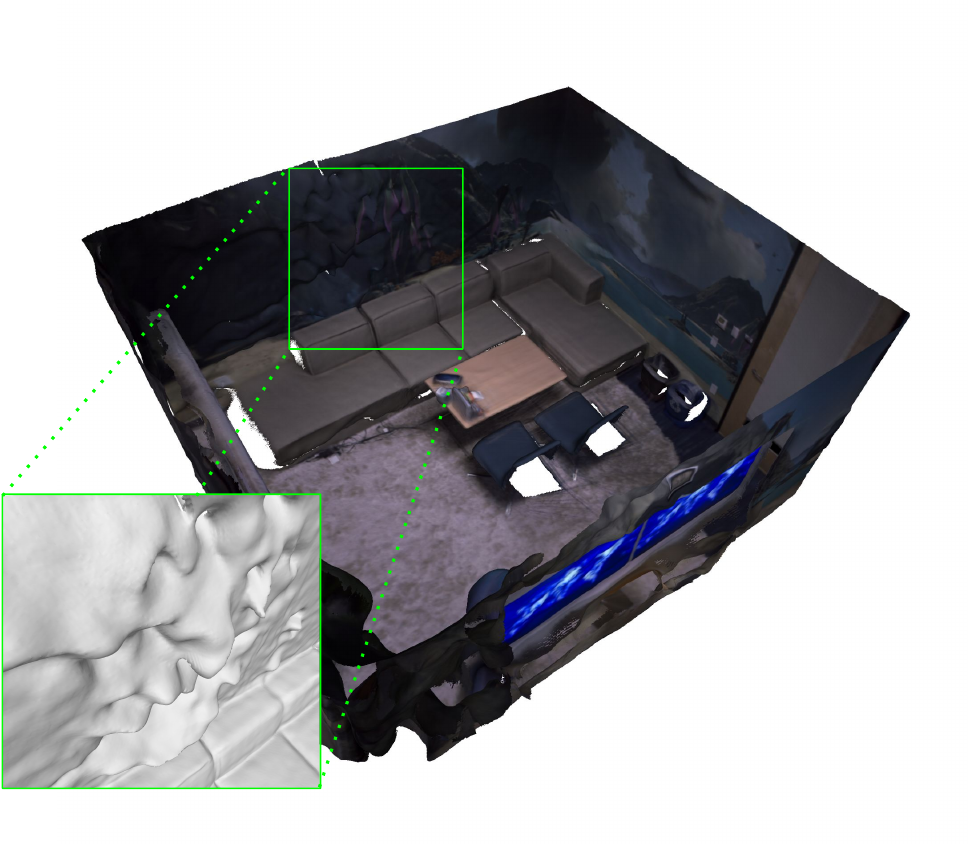} } &
         
        {\includegraphics[width=.475\columnwidth]{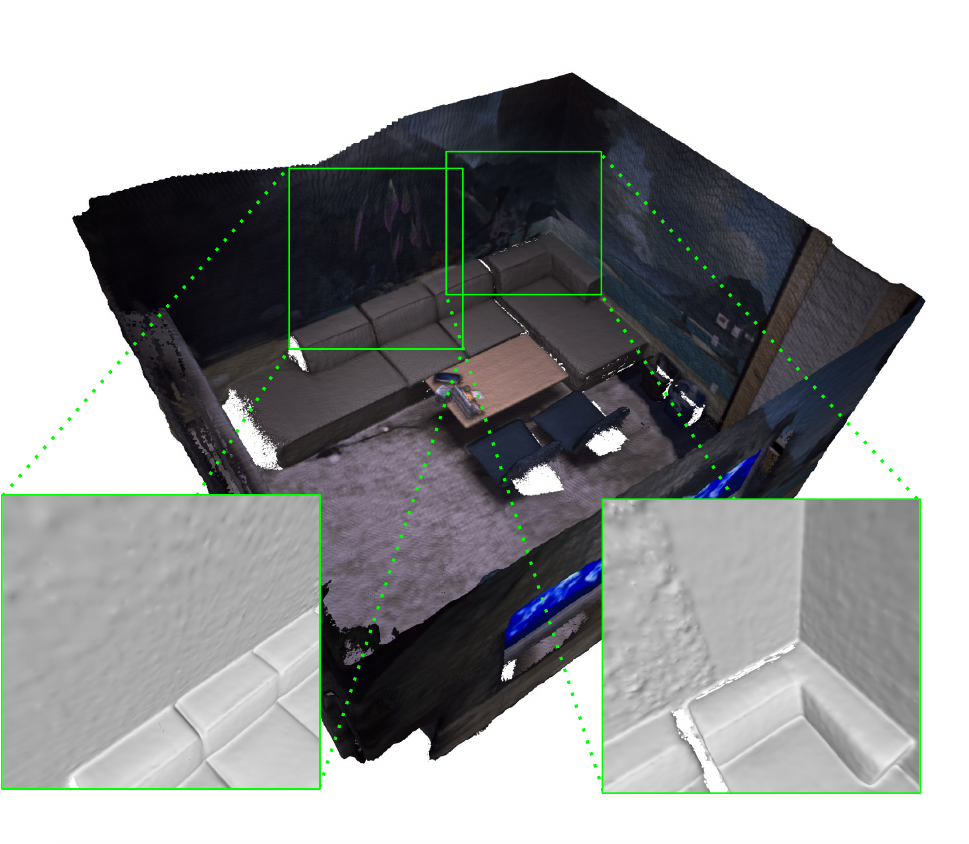}} & {\includegraphics[width=.475\columnwidth]{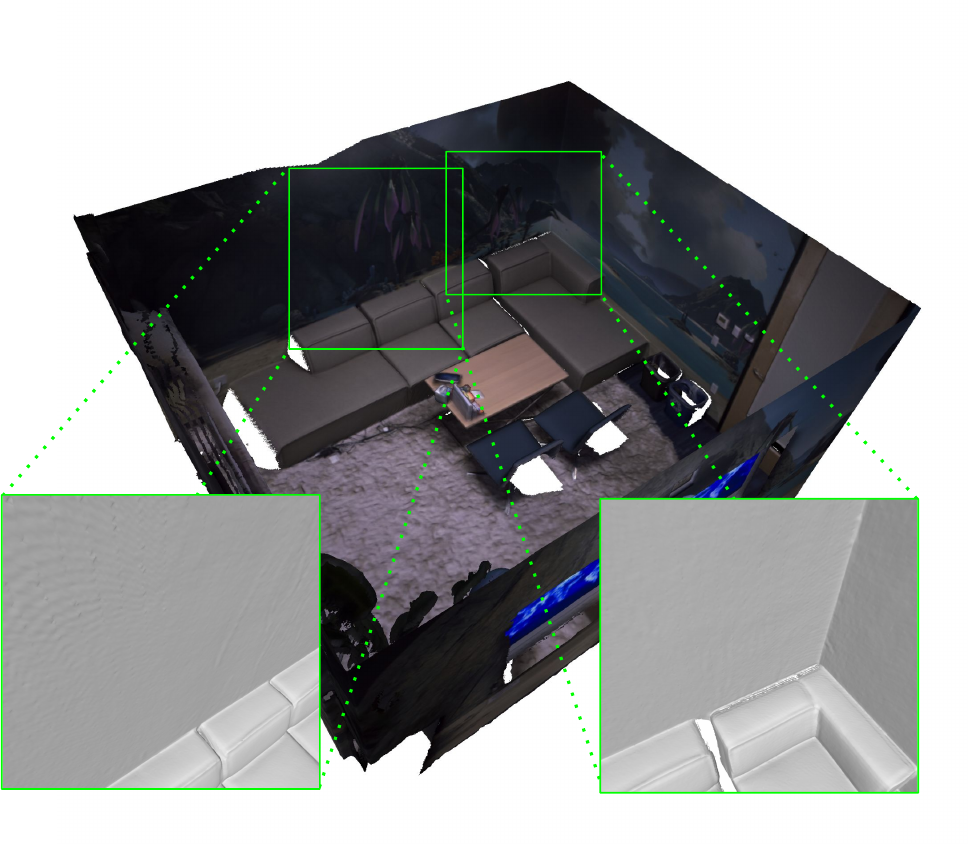} }\\
        \monosdf~\cite{yu2022monosdf} & \msdf~\cite{guo2022manhattan} &
        \vfnerf (Ours) & Ground Truth \\
        F-score: 0.790 & F-score: 0.757 & F-score: \textbf{0.894} &  \\
    \end{tabular}
    \vspace*{-0.5em} 
    \captionof{figure}{\textbf{\vfnerf.} Using the recently proposed Vector Field (VF)~\cite{Rella2022NeuralVF} representation, our method reconstructs indoor scenes in the NeRF setting. Due to the planar inductive bias of VF, we can generally recover indoor scenes with high fidelity, providing State-of-the-Art (SOTA) performance.}
    \label{fig:object-centric-meshes}
        \end{center}
    }]
}

\begin{document}
\maketitle
\def\thefootnote{*}\footnotetext{The authors contributed equally to the work}\def\thefootnote{\arabic{footnote}}
\begin{abstract}
Implicit surfaces via neural radiance fields (NeRF) have shown surprising accuracy in surface reconstruction. Despite their success in reconstructing richly textured surfaces, existing methods struggle with planar regions with weak textures, which account for the majority of indoor scenes. In this paper, we address indoor dense surface reconstruction by revisiting key aspects of NeRF in order to use the recently proposed Vector Field (VF) as the implicit representation.
VF is defined by the unit vector directed to the nearest surface point. It therefore flips direction at the surface and equals to the explicit surface normals. Except for this flip, VF remains constant along planar surfaces and provides a strong inductive bias in representing planar surfaces.
Concretely, we develop a novel density-VF relationship and a training scheme that allows us to learn VF via volume rendering.
By doing this, VF-NeRF can model large planar surfaces and sharp corners accurately.
We show that, when depth cues are available, our method further improves and achieves state-of-the-art results in reconstructing indoor scenes and rendering novel views. We extensively evaluate VF-NeRF on indoor datasets and run ablations of its components. We release the code at: \url{https://github.com/albertgassol1/vf-nerf}.
\end{abstract}
    
\section{Introduction}
\label{sec:intro}

Multi-view image-based 3D scene reconstruction is a cornerstone challenge in computer vision~\cite{hartley_zisserman_2004_mv_geom,snavely2008modeling,schonberger2016_sfm}.
Traditional multi-view stereo (MVS) algorithms~\cite{furukawa2010robust_stereopsis, schonberger206pixelwise_mvs, schonberger2016_sfm, tola2011large_mvs, zheng2014patchmatch} leverage matching and triangulation to derive 3D point coordinates from given input images. Nonetheless, they often struggle in regions characterized by uniform low-texture or repetitive patterns. Equipped with volume rendering, Neural Radiance Fields (NeRF)~\cite{mildenhall2020nerf,yariv2021volsdf,wang2021neus} and its variants~\cite{Brualla2021NerfWild,li2023dynibar,liu2020nsvf} have established themselves as powerful alternatives to previous methods for surface reconstruction. However, NeRF methods still struggle with low-texture indoor surfaces, even when using Manhattan normal priors~\cite{coughlan1999manhattan,guo2022manhattan}.

NeRF for indoor scene reconstruction has currently two significant challenges. The first is that the classical NeRF surface density~\cite{mildenhall2020nerf}, which provides high-quality view rendering, stumbles significantly when it comes to scene geometry reconstruction. Even when an SDF~\cite{yariv2021volsdf,wang2021neus} representation is used for the geometry, any surface regularization for planar surfaces has to rely on the gradients of the SDF~\cite{guo2022manhattan}. Note that these gradients are often noisy and unreliable for regularization. An additional downside of SDF is that its representation power is limited to water-tight surfaces. Therefore, it may not be able to faithfully reconstruct thin or open surfaces. The second challenge stems from poor texture in indoor surfaces, which provides weak multi-view constraints for the indirect triangulation in NeRF or direct triangulation in MVS approaches.

In this paper, we address the first challenge, that of the implicit scene representation in NeRF. In the process, we also push towards mitigating the challenge of weak texture through an improved inductive bias towards planar surfaces.
To that end, we make use of the recently proposed Vector Field (VF) representation~\cite{Rella2022NeuralVF,yang2023neural} in order to encode the scene geometry. This involves associating each position in the 3D space with a unit vector directed towards the nearest surface. It has been shown that VF may exhibit superior performance to SDF even on closed surfaces, particularly on sharp corners, thin objects and planar surfaces. This is due to the properties of VF and its inductive bias towards planar surfaces as they exhibit a constant normal along flat surfaces. However, the study confines itself to a supervised learning paradigm, and the self-supervised learning with NeRF poses significant challenges.
Notably, without the ground-truth VF, a pair of points is required to compute the surface density given the VF predictions.

In the VF optimization, we use a dual MLP network, one to predict the VF and the other to predict the RGB color values. We learn the VF and the color through a training scheme via volume rendering on multi-view posed images similarly to VolSDF~\cite{yariv2021volsdf,wang2021neus}. Specifically, we express the surface density via the cosine similarity of the VF predictions in the ray samples, which are obtained in a hierarchical manner. This novel VF-density relationship allows us to use neural volume rendering in order to train the VF as in \cite{yariv2021volsdf,wang2021neus}.
As a first study on VF for NeRF, we consider its use for learning indoor scene geometry. We rigorously evaluate our method against leading benchmarks for indoor scenes, including ManhattanSDF~\cite{guo2022manhattan}, MonoSDF~\cite{yu2022monosdf}, and Neuralangelo~\cite{li2023neuralangelo}, on indoor datasets such as Replica~\cite{Straub2019TheRD} and ScanNet~\cite{dai2017scannet}, showing superior performance on both reconstruction and novel view rendering.

In summary, our contributions are threefold:

\begin{itemize}
    \item We propose to learn the VF representation of 3D scenes with multi-view images via volume rendering.
    \item We design a function for the differentiable derivation of surface density from the VF, along with a training approach utilizing weight annealing and hierarchical sampling to optimize it.
    \item We demonstrate the effectiveness of our method on different indoor scene datasets, showing state-of-the-art results.
\end{itemize}

 \begin{figure*}[t]
        \centering
         \includegraphics[width=\textwidth]{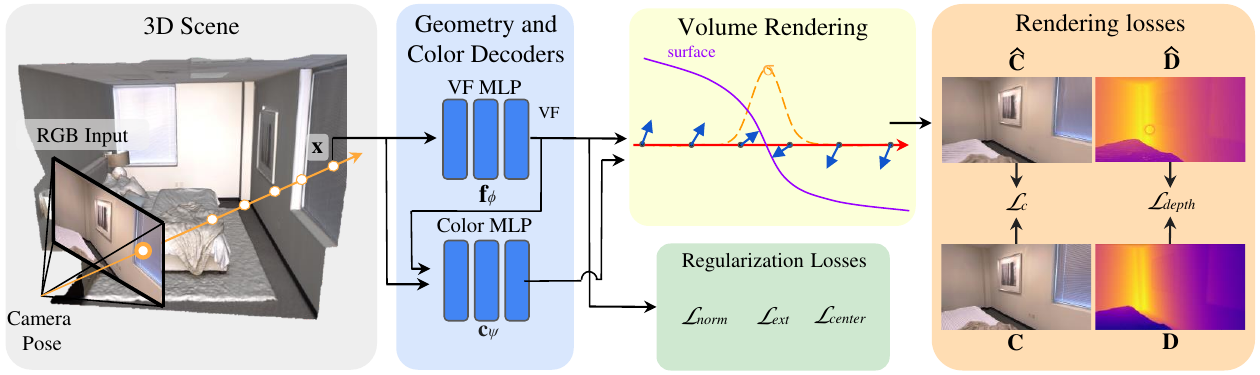}
    \caption[\vfnerf overview]{\textbf{\vfnerf overview.} We use VF to represent the geometry of a scene. Specifically, given an input image taken from the camera view position, we shoot a batch of rays onto the 3D scene. We predict the VF and color of the points along the ray using geometry and color decoders. By computing the cosine similarity between neighboring points on the ray, we can identify the surface as the locations where the value equals $-1$, i.e. when the two predicted vectors have opposing directions. From the cosine similarity, we differentiably compute the surface density. We render the RGB and depth in order to compute the re-rendering losses. }
    \label{fig:pipeline}
\end{figure*}

\section{Related work}
\label{sec:related_work}

\mypar{Multi-view Surface Reconstruction}
Traditional MVS approaches have often relied on feature matching for depth estimation~\cite{Barnes2009PAR, Bleyer2011PatchMatchS, Broadhurst2001SpaceCarving, Galliani2015MassivelyMvs, Kutulakos1999TheorySpaceCarving, schonberger206pixelwise_mvs, schonberger2016_sfm, Seitz2006ComparisonMVS}. These classical methods extract image features, match them across views for depth estimation, and then fuse the obtained depth maps to form dense point clouds. Voxel-based representations~\cite{Agrawal2001probFrameworkMvs, furukawa2010robust_stereopsis, Seitz1997Photorealistic} rely on color consistency between the projected images to generate an occupancy grid of voxels. Subsequently, meshing techniques, like Poisson surface reconstruction~\cite{Kazhdan2006Poisson, Kazhdan2013Poisson} are applied to delineate the surface. However, these methods typically fail to reconstruct low-textured regions and non-Lambertian surfaces. Additionally, the reconstructed point clouds or meshes are often noisy and may fail to reconstruct some surfaces. 

Recently, learning-based methods have gained attention, offering replacements for classic MVS methods. Methods like~\cite{Chen209PointBasedNet, im2019dpsnet, Yao2019rmvsnet, Yao2018MVSNetDI} leverage 3D CNNs to extract features and predict depth maps, while others~\cite{Cheng2020DeepStereo, gu2020cascade} construct cost volumes hierarchically, yielding high-resolution results. However, these methods often fail to accurately reconstruct the scene geometry due to the limited resolution of the cost volume.

\mypar{Neural Radiance Fields (NeRF)} 
In recent studies~\cite{liu2020nsvf, Brualla2021NerfWild, mildenhall2020nerf, pumarola2020dnerf} the potential of MLPs to represent scenes both in terms of density and appearance has been explored. While these techniques can produce photorealistic results for novel view synthesis, determining an isosurface for the volume density to reconstruct scene geometry remains a challenge. Commonly, NeRF uses thresholding techniques to derive surfaces from the predicted density. However, these extracted surfaces can often exhibit noise and inaccuracies.

\mypar{Neural Scene Representations} Approaches based on neural scene representations employ deep learning to learn properties of 3D points and to generate geometry. Traditional methods like point clouds~\cite{Fan2017PointSet3DReconstruction, Lin2018LearniongPointCloud} and voxel grids~\cite{Choy20163dr2n2, Xie2019Pix2vox} have been primary choices for representing scene geometry. More recently, implicit functions, such as occupancy grids~\cite{Niemeyer2020diff_vol_rendering, Oechsle2021UNISURFUN}, SDF~\cite{Jiang2019SDFDiffDR, li2023neuralangelo,Liu2019DISTRD, Park_2019_DeepSDF, wang2021neus, yariv2021volsdf, yariv2020multiview}, and VF \cite{Rella2022NeuralVF, yang2023neural} have gained popularity due to their precision in capturing scene geometry. For instance, in~\cite{Liu2019DISTRD, Niemeyer2020diff_vol_rendering} a novel differentiable renderer to learn the scene geometry from images is proposed, while~\cite{yariv2020multiview} focuses on modeling view-dependent appearance, which proves successful on non-Lambertian surfaces. \cite{lin2020sdf}, instead, utilizes 2D silhouettes from single images to reconstruct their underlying 3D shape. However, these methods rely on masks to accurately reconstruct the geometry from multi-view images. Consequent works~\cite{yariv2021volsdf,wang2021neus} introduce a second MLP in the NeRF context to represent the geometry as the SDF, further leveraging volume rendering to learn the geometry from images. Building upon these methods, \cite{li2023neuralangelo} takes inspiration from Instant Neural Graphics Primitives (Instant NGP)~\cite{mueller2022instant} to introduce hash encodings in neural SDF models, enhancing surface reconstruction resolution. However, a challenge persists as these methods tend to fail in large indoor planar scenes with low-texture regions, leading to inaccurate surface reconstructions.

\mypar{Priors for Neural Scene Representations} Several works have explored the integration of priors during optimization to improve the reconstruction of indoor scenes. For instance, \cite{guo2022manhattan} suggests incorporating dense depth maps from COLMAP~\cite{schonberger2016_sfm} to facilitate the learning of 3D geometry and employs Manhattan world~\cite{coughlan1999manhattan} priors to address the challenges posed by low-textured planar surfaces. A limitation of this approach is its reliance on semantic segmentation masks to pinpoint planar regions, adhering to the Manhattan world assumption. This dependency can lead to added complexity and potential inaccuracies in regions where segmentations are less accurate. More recently, NeuRIS~\cite{Wang2022NeuRIS} proposes to use normal priors to guide the reconstruction of indoor scenes. Expanding on this work, MonoSDF~\cite{yu2022monosdf} introduces both normal and depth monocular cues into the optimization.

\section{Method}

Given a set of posed images of an indoor scene, our goal is to reconstruct the dense scene geometry. We represent the surface geometry in NeRF with VF~\cite{Rella2022NeuralVF, yang2023neural}, and describe its properties in~\cref{sec:vf}. We then introduce the surface density as a parametrization of the VF in~\cref{sec:density} and describe our hierarchical ray sampling method in~\cref{sec:sampling}. Finally, in~\cref{sec:vf-optimization}, we formulate the optimization problem and introduce the loss terms. We provide an overview in~\cref{fig:pipeline}.

 \subsection{Vector Field Representation}
\label{sec:vf}
In \vfnerf, the scene geometry is defined using unit vectors that point towards the nearest surface. Let $\Omega \subset \Re^3$ be the surface of an object in $\Re^3$ and $\Gamma \subset \Re^3$ be the set of unit norm 3-vectors. We make use of the VF definition~\cite{Rella2022NeuralVF}: VF is a function  $\vfFunction : \Re^3 \to \Gamma$ that maps a point in space to a unit vector directed to the closest surface point of $\Omega$:

\begin{equation}
    \vfFunction(\loc) = \begin{cases}
        \dfrac{\surfaceLoc - \loc}{|| \surfaceLoc - \loc ||_2} & \text{if } \loc \notin \Omega\\
        \dfrac{\widehat{\loc}_S - \widehat{\loc}}{|| \widehat{\loc}_S - \widehat{\loc} ||_2} & \text{if } \loc \in \Omega,
    \end{cases}
    \label{eq:vf_representation}
\end{equation}
where $\surfaceLoc = \argmin_{\mathbf{s} \in \Omega} || \loc - \mathbf{s} ||_2$ is the closest surface point with respect to $\loc$, and $\widehat{\loc} = \lim\limits_{||\epsilon||_2 \to 0} \loc + \epsilon$ is a point close to the surface, with $\epsilon \in  \Re^3$ being an infinitesimal 3D vector.

Given the definition of the VF representation, we identify a surface $\Omega$ between a point $\loc \in \Re^3$ and an infinitesimally close neighbor using the cosine similarity between the VF at the two points. When the two points are on opposite sides of the surface $\Omega$, their cosine similarity approaches $-1$. Conversely, it is close to $1$ everywhere else, except at diverging discontinuities of the field.

\begin{equation}
\begin{aligned}
    \Omega = \{ \loc_1, \loc_2 = \loc_1 + \epsilon  |  \cos{(\vfFunction(\loc_1), \vfFunction(\loc_2)) < \tau}\}, \\ \cos{(\vf_1, \vf_2)} = \dfrac{\vf_1 \cdot \vf_2}{||\vf_1||_2||\vf_2||_2},
\end{aligned}    
\end{equation}
where  $\epsilon \in \Re^3$ is an infinitesimal displacement and $|\tau| \leq 1$ is a cosine similarity threshold. Ideally, $\tau = -1$ for infinitesimally close neighbors.

From these definitions, we notice a similarity to the surface density $\sigma : \Re^3 \to \Re_{\geq 0}$, a function that indicates the rate at which a ray is occluded at location $\loc$. Ideally, for non-translucid surfaces, $\sigma(\loc)$ behaves as a delta function, being zero everywhere except at the surface. To model this function typically used in volume rendering~\cite{Kajiya1984RayTV,mildenhall2020nerf}, a simple transformation of the cosine similarity can be used. In fact, the cosine similarity between the VF of infinitesimally close neighbors is a delta function itself, yielding approximately $1$ everywhere and $-1$ at the surface. However, as we show, a smooth function is necessary in order to ease the learning of VF through volume rendering.

\subsection{Density as Transformed VF}
\label{sec:density}

\begin{figure}[t]
        \centering
         \includegraphics[width=0.8\columnwidth]{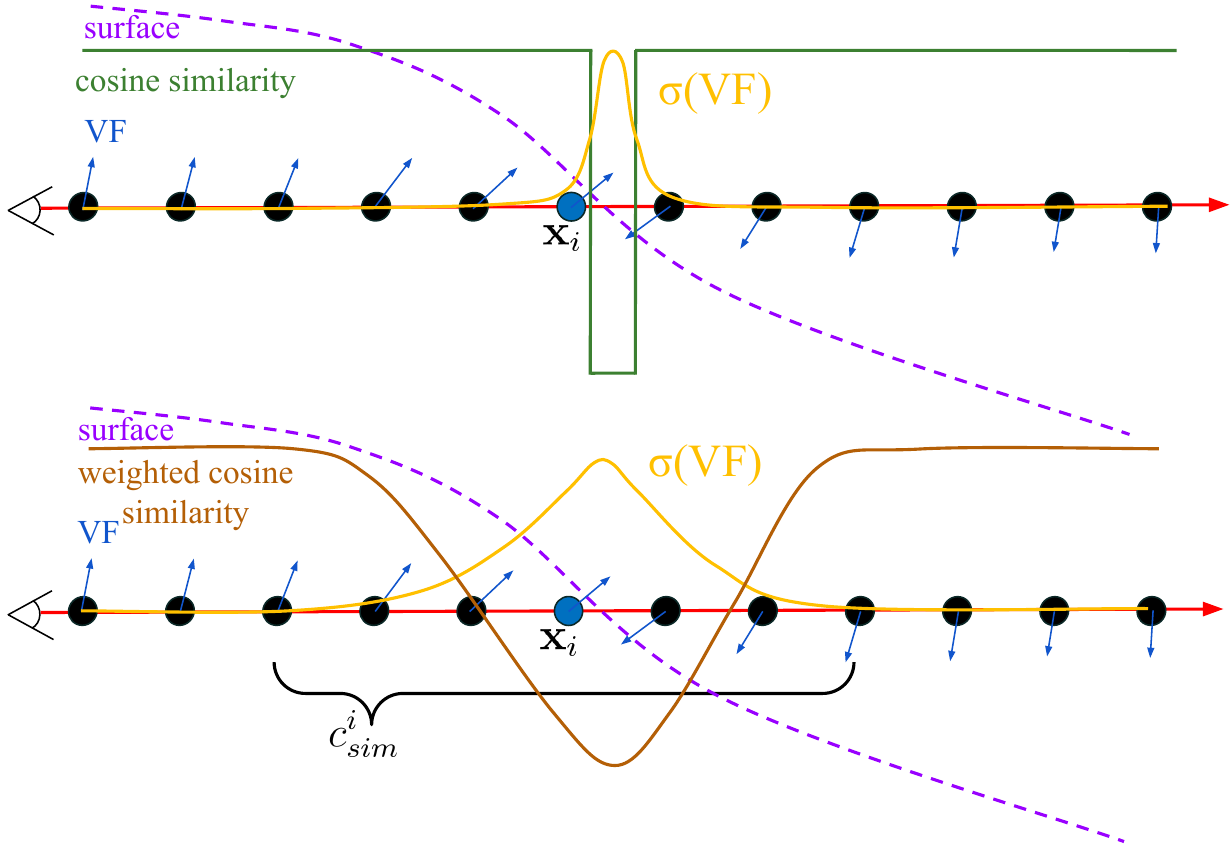}
    \caption[Density and weighted cosine similarity]{\textbf{Density using non-averaged and averaged cosine similarity}. The figures show the VF, cosine similarity and density of a ray crossing a surface. Top: density as a transformation of the cosine similarity. This yields a sharp function similar to the delta function centered at the surface. Bottom: Density as a transformation of the weighted average cosine similarity. This produces a smoother function with the maximum centered at the surface.}
    \label{fig:ray_density}
\end{figure}

We draw inspiration from existing methods~\cite{wang2021neus, yariv2021volsdf}, which use neural volume rendering to learn the geometry of a scene as an implicit function. Contrary to these previous methods that use SDF, we propose to model the surface density as a function of the learnable VF.
Given a viewing ray and the VF sampled at multiple points along the ray, we use a differentiable process to estimate the surface density.
As previously highlighted, the cosine similarity of the VF at neighboring points along the ray can be used to indicate whether there is a surface between them. The resulting surface density function, showcased in~\cref{fig:ray_density} top row, closely resembles a delta function. This behavior is desirable in order to obtain sharp reconstructions; however, due to its discontinuity, the desired convergence is hard to achieve. In order to make the gradient-based optimization tractable, we first need a smoothing transformation. To this end, we adopt a sliding window approach and compute a weighted average cosine similarity. We thus smooth the function at points near the surface. The effect of the sliding window can be seen in the bottom row of~\cref{fig:ray_density}.

 \begin{figure}[t]
        \centering
         \includegraphics[width=0.8\columnwidth]{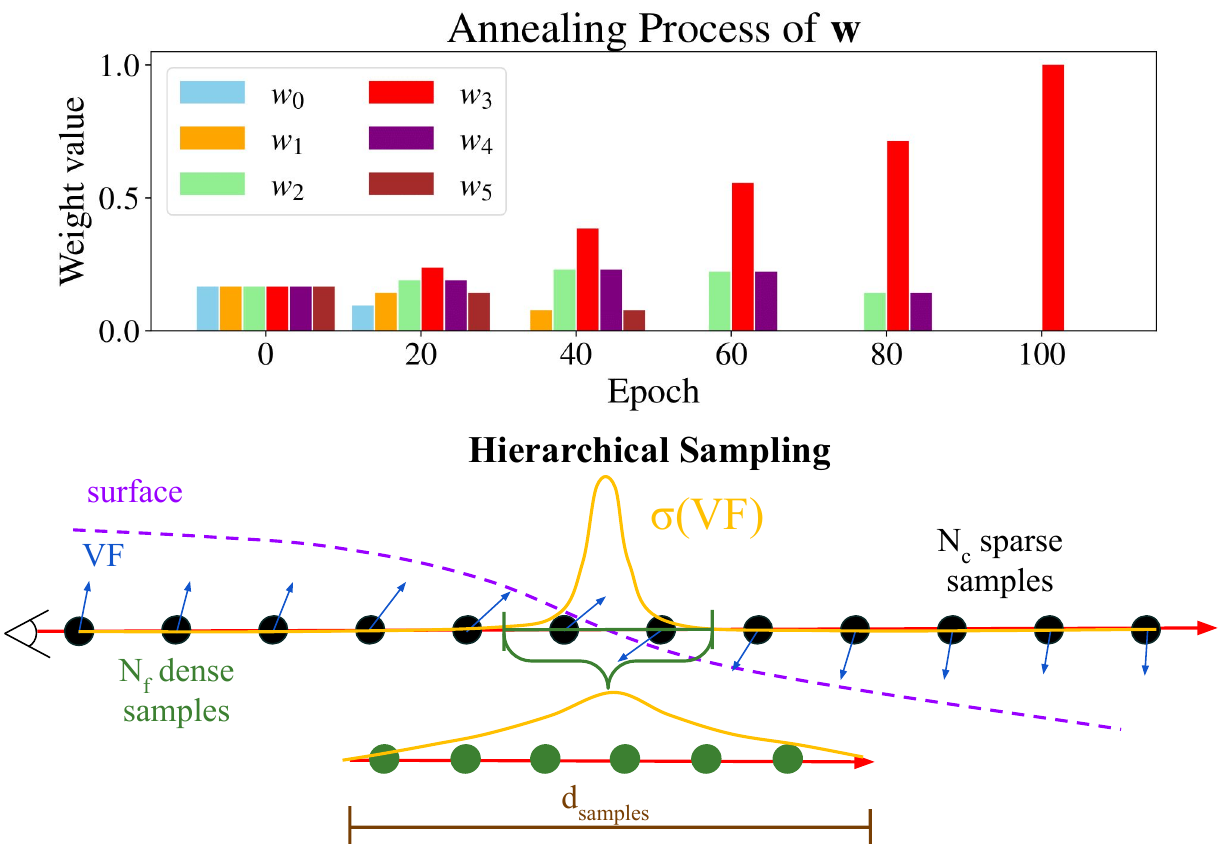}
    \caption[Density and weighted cosine similarity]{\textbf{Sliding window weights annealing example and hierarchical sampling}. \textbf{Top: }Example of $M=6$ weights at different stages of the annealing. At the beginning of the training (epoch 0), the weights for each neighbor are equal. At the end of the training (epoch 100) the cosine similarity is computed only with respect to the closest next neighbor. \textbf{Bottom: } Initially, we sample uniform points along the ray and compute the surface density through the predicted VF. We then densely sample points within a range $d_{samples}$ centered at the maximum of the surface density.}
    \label{fig:annealing_sampling}
\end{figure}

Given a set of samples in a ray, we initially predict the VF at each point of the ray. We then define the weights of a sliding window of size $M$, where the size is an even number, as $\slidingWindow = \left[w_0, w_1, ..., w_{M-1} \right]$. The sliding window and the predicted VF are used to compute the weighted average cosine similarity associated with each point. The smoothed cosine similarity of a point is computed as the weighted average of the cosine similarities using multiple forward and backward neighbors of the ray. Therefore,  given a ray of $N+1$ points $\ray = \left[ \loc_0, \loc_1, \cdots, \loc_N \right]$, we can compute $N$ averaged cosine similarities as:
\begin{equation}    
\begin{aligned}
 \cossim^i(\ray) =  \sum_{j=0}^{M/2-1}[w_j \cos{(\vfFunction(\loc_i), \vfFunction(\loc_{i-j-1}))} + \\
 w_{j+M/2} \cos{(\vfFunction(\loc_i), \vfFunction(\loc_{i+j+1}))} ],
 \label{eq:ray_cos_sim}
\end{aligned}
\end{equation}

For simplicity, in~\cref{eq:ray_cos_sim} we do not consider the boundary cases of computing the weighted average cosine similarity of the first and last samples along rays. However, note that the cosine similarity of the first and last points of the ray is not smooth because the sliding window would go out of range. Additionally, given N+1 points, we can only compute N cosine similarities since the last point of the ray does not have a successor.

The effect of the weighted sliding window can be changed by modifying its weights. We start with a uniform distribution where all the weights are equal and sum up to 1. We introduce an annealing process to progressively add more weight to the closest neighbors with the final objective to end with a one-hot vector where all the weight is located at the closest next neighbor. Thanks to this approach, the network can be easily optimized, while preserving the desired sharpness during inference. The annealing process is depicted in~\cref{fig:annealing_sampling}. This process is linear and depends on the training epoch. Specifically, the weights of the sliding window are computed at the beginning of every epoch using the following equation:

\begin{equation}
     \widehat{w}_i = \dfrac{M}{2} \text{ReLU}\left(1 - \dfrac{n |i-M/2|}{N_{epochs}} \right), \quad w_i = \dfrac{\widehat{w}_i}{|| \widehat{\slidingWindow} ||}.
\end{equation}

Using sliding window cosine similarity, we redefine the surface density as a transformation that maps a point in the ray $\ray \in \Re^{(N+1) \times 3}$ to a scalar value, $\sigma~:~\Re^{(N+1) \times 3}~\times~\NaPos~\to~ \Re_{\geq 0}$. Leveraging the cosine similarity, we define the surface density as follows:
\begin{equation}
    \sigma(\ray, i) = \text{ReLU}(\alpha\Psi_{\mu, \beta}(-\cossim^i(\ray)) - \alpha\Psi_{\mu, \beta}(\xi)).
    \label{eq:density}
\end{equation}
where $\alpha, \mu, \beta > 0$ are learnable parameters and $\xi$ is a cosine similarity threshold value left as a hyperparameter. ReLU is the rectified linear unit and $\Psi_{\mu, \beta}$ represents the Cumulative Distribution Function (CDF) of the Laplace distribution. $\mu$ denotes the Laplacian mean, while $\beta$ is Laplacian "diversity" and $\alpha$ is a scaling factor. Formally, the Laplacian CDF is defined as follows:
\begin{equation}
    \Psi_{\mu, \beta}(x) = \begin{cases} 1 -  \exp{\left ( -\frac{|x - \mu|}{\beta} \right )} &\text{if } x > \mu \\
    \exp{\left ( -\frac{|x - \mu|}{\beta} \right )} &\text{if } x \leq \mu.
    \end{cases}
    \label{eq:laplacian_cdf}
\end{equation}

With this definition of the density function, we can accumulate the densities and colors using numerical quadrature~\cite{mildenhall2020nerf} to render the color and depth of the pixel associated with the ray:

\begin{equation}
C(\pixel) = \sum_{i=1}^N T_i (1 - \exp{(-\sigma_i\delta_i)}) \radiance_i
\label{eq:color_numerical_volume_rendering}\\
\end{equation}
\begin{equation}
D(\pixel) = \sum_{i=1}^N T_i (1 - \exp{(-\sigma_i\delta_i)}) t_i,
\label{eq:depth_numerical_volume_rendering}
\end{equation}

where $T_i = \exp{\left( -\sum_{j=1}^{i-1}\sigma_j\delta_j \right)}$ is the accumulated transmittance and $\delta_i = t_{i+1} - t_i$ is the distance between samples along a ray. Note that \cref{eq:color_numerical_volume_rendering,eq:depth_numerical_volume_rendering} can be seen as traditional alpha compositing with alpha values $\alpha_i = 1 - \exp{(\sigma_i\delta_i)}$.

\subsection{Hierarchical Ray Sampling}
\label{sec:sampling}
Sampling rays densely in a uniform manner proves highly inefficient due to the prevalence of free space and occluded regions along the ray, which do not contribute significantly to volume rendering. To address this challenge, similarly to prior works~\cite{mildenhall2020nerf, wang2021neus, yariv2021volsdf, li2023neuralangelo, yu2022monosdf, Wang2022NeuRIS, guo2022manhattan}, we propose a hierarchical sampling strategy to allocate samples selectively in regions likely to contain surfaces. Initially, we sparsely sample $N_c=100$ points along a ray and predict their corresponding VFs. Given these predictions, we compute the surface density $\sigma$ along the ray and densely resample around its maximum. As shown in~\cref{fig:annealing_sampling} bottom row, our dense sampling approach involves uniformly sampling $N_f$ points in a window of size $d_{samples} = 30 \text{cm}$ centered at the point yielding maximum surface density $\sigma$. The number of points $N_f$ sampled during this step increases every $n^{inc} = 50$ epochs using a fixed step size $N_f^{inc} = 5$ until reaching a maximum $N_f^{max} = 100$. Consequently, after some epochs we make use of a total of $N_c + N_f^{max} = 200$ points to render the predicted color $C(\pixel)$ and depth $D(\pixel)$. Note that while some works optimize coarse and fine networks simultaneously to predict the surface density~\cite{mildenhall2020nerf, pumarola2020dnerf}, our approach employs a single network to predict the VF. Hierarchical sampling allows the network to progressively refine the 3D representation of the scene in a coarse-to-fine manner.

\subsection{Training}
\label{sec:vf-optimization}

Our approach leverages a dual-MLP structure. First, $\vfNetwork: \Re^3 \rightarrow \Re^{3+256}$ predicts the VF of the scene alongside a global geometry feature vector $\featVec \in \Re^{256}$. Here, $\phi$ represents the network learnable parameters. Second, $\radianceNetwork: \Re^{3+3+3+256} \rightarrow \Re^3$ approximates the radiance field color values based on a given spatial point, viewing direction, VF, and global feature vector. Here, $\psi$ represents the radiance field network learnable parameters. Consequently, for a specific point on a ray $\loc$ and its viewing direction $\direction$ , we can predict the VF as $(\vf, \featVec) = \vfNetwork(\loc)$ and the radiance field as $\radiance = \radianceNetwork(\loc, \vf, \direction, \featVec)$. Our model also incorporates three learnable parameters for the density function as described in \cref{eq:density}, namely $\alpha$, $\mu$ and $\beta$. 

During training, a batch of pixels $\pixelBatch$ and their corresponding rays are sampled to minimize the difference between the rendered images $\widehat{C}(\pixel)$ and the reference images $C(\pixel)$:
\begin{equation}
    \loss_c = \dfrac{1}{|\pixelBatch|} \sum_{\pixel \in \pixelBatch} || \widehat{C}(\pixel) - C(\pixel) ||_1.
    \label{eq:color_loss}
\end{equation}
Learning the geometry of indoor scenes solely from images presents a challenge in reconstructing accurate geometries, even in textured regions. To address this, we enhance the learning of scene representation by introducing a depth consistency loss similarly to~\cite{yu2022monosdf, Wang2022NeuRIS}. This loss compares the rendered depth, $\widehat{D}(\pixel)$, with a reference depth, symbolized as $D(\pixel)$. Depending on the availability of data, the depth $D(\pixel)$ can be derived from multi-view stereo methods~\cite{schonberger206pixelwise_mvs, schonberger2016_sfm, zheng2014patchmatch} or by monocular depth estimation~\cite{godard2017unsupervised, godard2019digging}.
\begin{equation}
    \loss_{depth} = \dfrac{1}{|\pixelBatch|} \sum_{\pixel \in \pixelBatch} || \widehat{D}(\pixel) - D(\pixel) ||_1.
    \label{eq:depth_loss}
\end{equation}

In addition to the rendering losses, we add three regularization terms to impose the known properties of VF. First, we impose that the VF has a unit vector property by applying the unit norm loss $\loss_{norm}$:
\begin{equation}
    \loss_{norm} = \dfrac{1}{N+1}\sum_{i=0}^{N} (||\vfFunction(\loc_i)||_2-1)^2.
    \label{eq:norm_loss}
\end{equation}
Additionally, in object-centric scenes, the VF at outer, distant points resembles a vector directed toward the center. Therefore, we incorporate a loss that guides the VF for points outside the scene, denoted as $\pointsExterior$, to point towards the object's center, represented by $\mathbf{c}_{scene}$.
\begin{equation}
    \loss_{ext} = \dfrac{1}{|\pointsExterior|} \sum_{\loc \in \pointsExterior} \left|\left| \vfFunction(\loc) - \dfrac{\mathbf{c}_{scene} - \loc}{|| \mathbf{c}_{scene} - \loc||_2}\right|\right|_2.
    \label{eq:loss_exterior}
\end{equation}
Finally, considering that in indoor scenes, images are captured from within the scene's geometry, we introduce a loss function that guides points near the scene's center, represented as $\pointsCenter$, to point outwards: 
\begin{equation}
    \loss_{cen} = \dfrac{1}{|\pointsCenter|} \sum_{\loc \in \pointsCenter} \left|\left| \vfFunction(\loc) - \dfrac{\loc - \mathbf{c}_{scene}}{|| \loc - \mathbf{c}_{scene} ||_2}\right|\right|_2.
    \label{eq:loss_center}
\end{equation}

The overall loss is defined as a weighted sum of the individual losses:
\begin{equation}
\begin{aligned}
    \loss = w_c\loss_c + w_{norm}\loss_{norm} + w_{ext}\loss_{ext} + \\
    w_{depth}\loss_{depth} + w_{cen}\loss_{center}, 
    \label{eq:loss}
\end{aligned}
\end{equation}
where  $w_c$, $w_{norm}$, $w_{ext}$, $w_{depth}$ and $w_{cen}$ are hyperparameters.

\section{Experiments}

\setlength{\tabcolsep}{4pt}
\begin{table*}[t]
    \centering
    \label{tab:quantitative} 
    \begin{tabular}{l|ccccc|ccccc}
        \toprule
         & \multicolumn{5}{c|}{\textbf{Replica}} & \multicolumn{5}{c}{\textbf{ScanNet}}\\
        & PSNR $\uparrow$ & CD $\downarrow$ &Precision $\uparrow$  & Rrecall $\uparrow$ & F1-score $\uparrow$  & PSNR $\uparrow$ & CD $\downarrow$ & Precision $\uparrow$  & Recall $\uparrow$ & F1-score $\uparrow$\\
        \midrule
        COLMAP~\cite{schonberger2016_sfm} &  - & - & 0.760 & 0.403 & 0.527 & - & - & 0.604 & 0.485 & 0.538 \\
        NeRF~\cite{mildenhall2020nerf} & - & - & 0.153 & 0.295 & 0.201 & - & - & 0.085 & 0.166  & 0.112 \\
        UNISURF~\cite{Oechsle2021UNISURFUN}  & - & - & 0.195 & 0.338 & 0.247 & - & - & 0.298 & 0.335 & 0.315 \\ 
        NeuS~\cite{wang2021neus} &  - & - & 0.524 & 0.465 & 0.493 & - & - & 0.406 & 0.437 & 0.421 \\
        VolSDF~\cite{yariv2021volsdf} &  - & - & 0.317 & 0.442 & 0.369 & - & - & 0.489 & 0.546 & 0.516 \\
        N-Angelo~\cite{li2023neuralangelo} &  31.44 & 611 & 0.243 &  0.323 & 0.262 & 17.83 & 103 & 0.269 & 0.188 & 0.220\\
        M-SDF~\cite{guo2022manhattan}  & 27.48 & 5.6 & 0.723  & \boldgreen{0.856} & 0.779 & 20.78 & 1.45 & 0.778 & 0.694 & 0.730\\
        NeuRIS~\cite{Wang2022NeuRIS}  & - & - & - & - & - & \boldgreen{24.40} & 1.71 & 0.773 & 0.682 & 0.723\\
        MonoSDF~\cite{yu2022monosdf} &  \boldblue{32.25} & \boldgreen{0.37} &\boldgreen{0.906} & \boldblue{0.889} & \boldgreen{0.897} & 23.84 & \boldgreen{1.42} &\boldgreen{0.863} & \boldgreen{0.730} & \boldgreen{0.788} \\
        \midrule
        \vfnerf  & \boldgreen{31.49} & \boldblue{0.13} & \boldblue{0.976} & 0.842 & \boldblue{0.904} & \boldblue{26.21} & \boldblue{0.258} & \boldblue{0.928} & \boldblue{0.821} & \boldblue{0.865} \\
        \bottomrule
    \end{tabular}
    \caption[quantitative results]{\textbf{Quantitative results.} Our method outperforms all baselines in terms of the averaged F-score and  median Chamfer Distance. On novel view synthesis, \vfnerf outperforms all baselines on ScanNet, and renders high-quality images on Replica, being second only to MonoSDF by a small margin. \boldblue{Best result}. \boldgreen{Second best result}.}
\end{table*}

\setlength{\tabcolsep}{.1pt}
\begin{figure*}[t]
        \centering
        \begin{tabular}{ccccc}
            \rotatebox[origin=b]{90}{ \texttt{Room 1}} & \raisebox{-0.5\height}{\includegraphics[width=.24\textwidth]{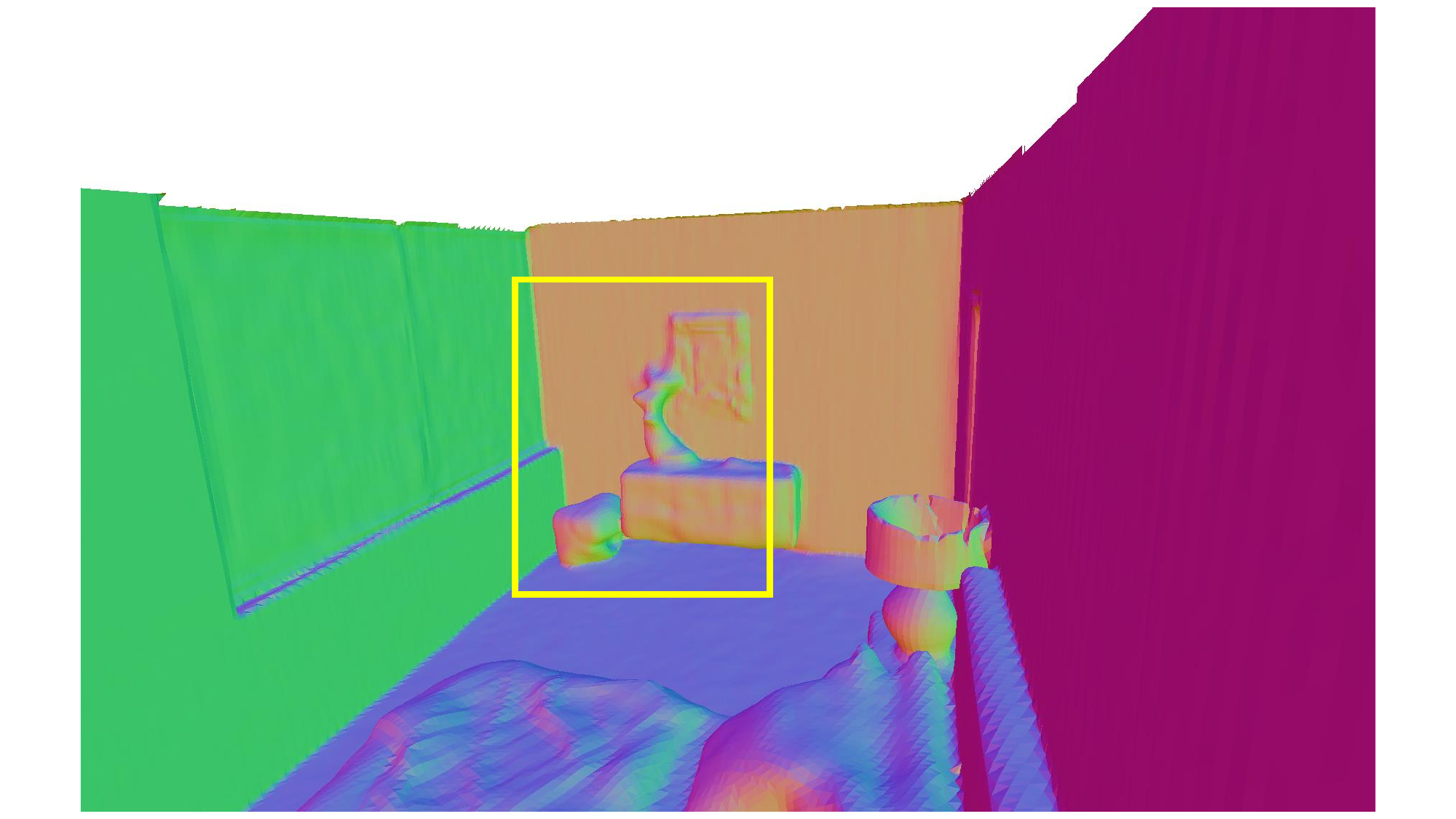} } & \raisebox{-0.5\height}{\includegraphics[width=.24\textwidth]{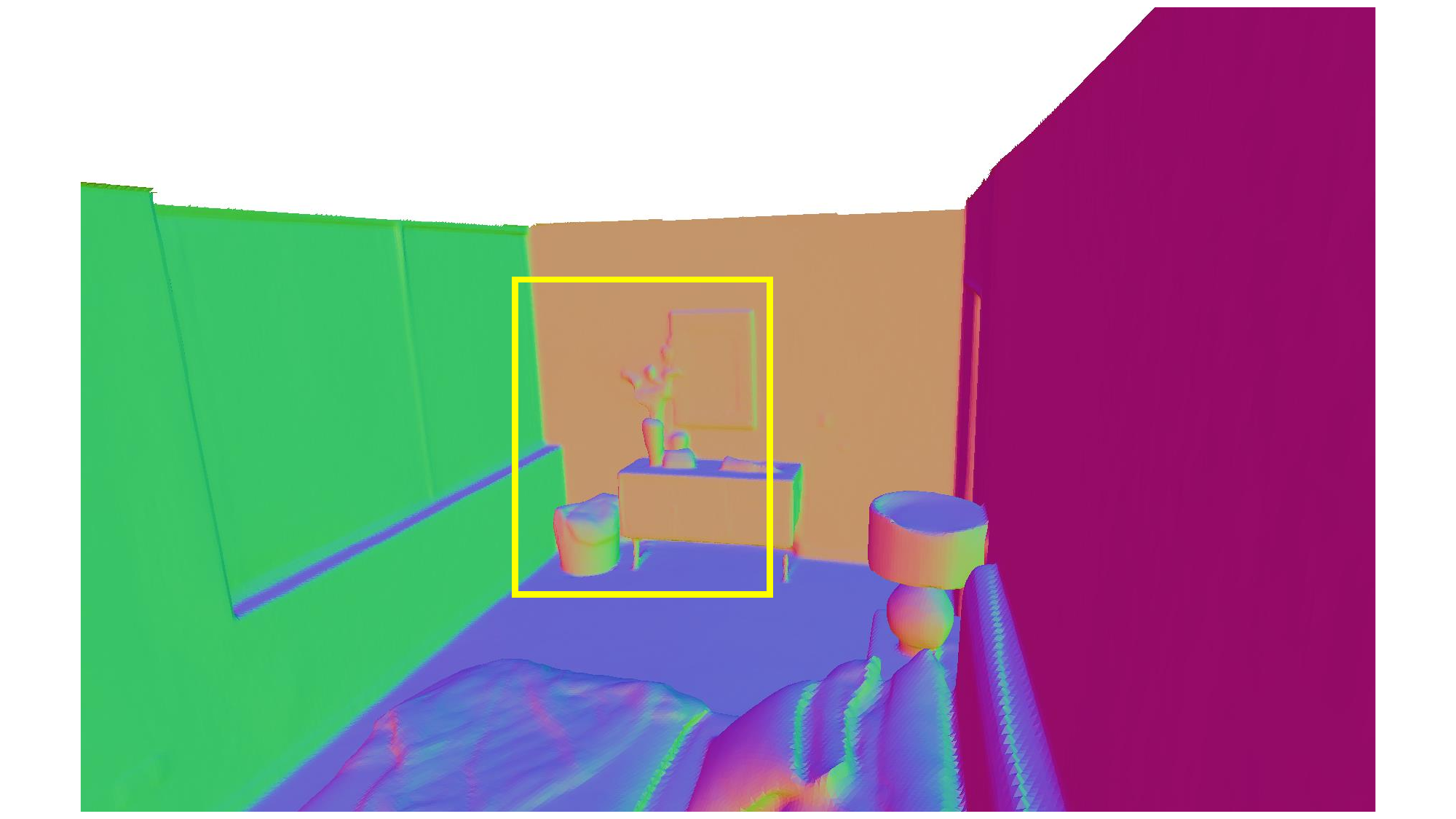} } & \raisebox{-0.5\height}{\includegraphics[width=.24\textwidth]{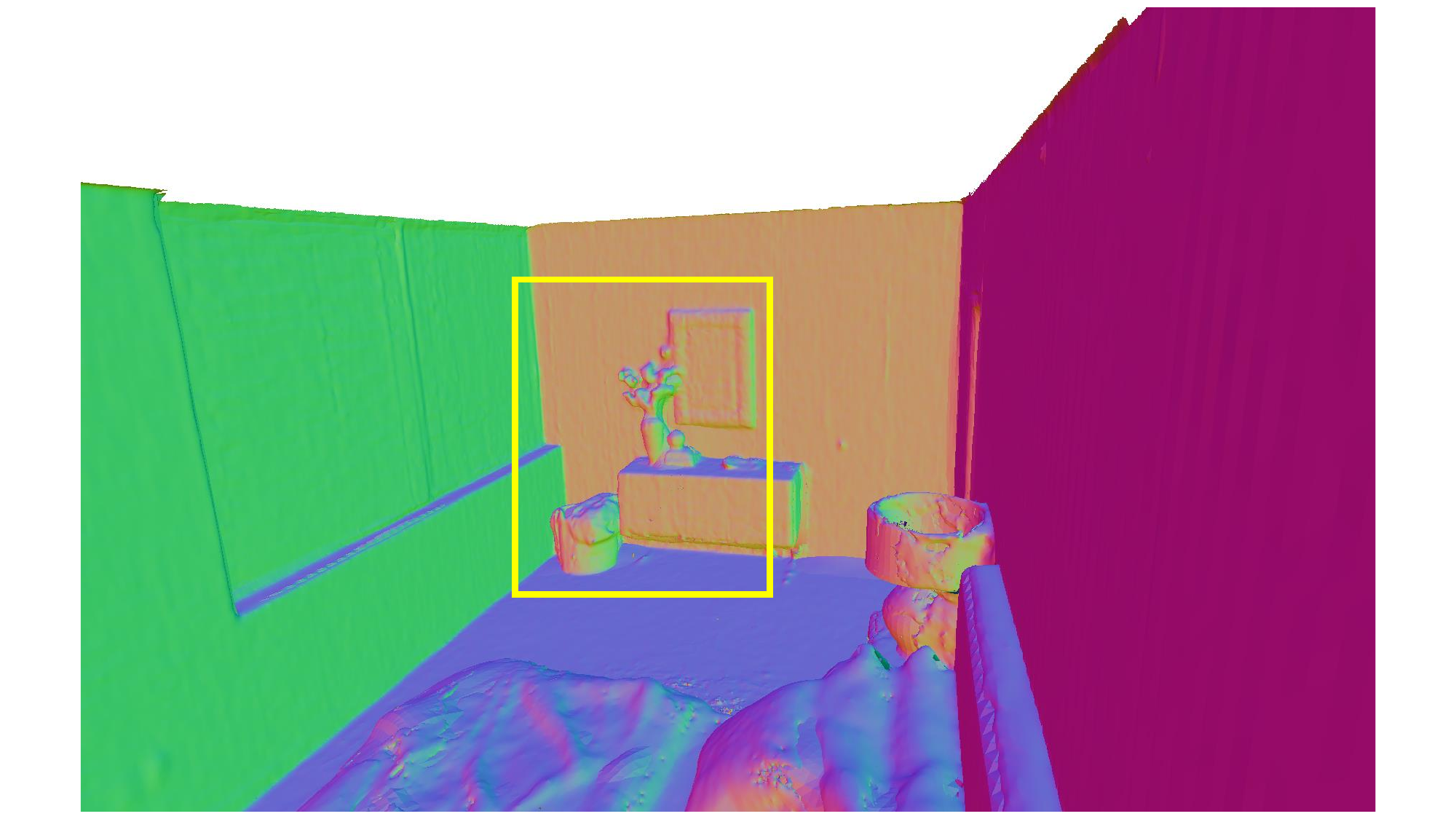} } & \raisebox{-0.5\height}{\includegraphics[width=.24\textwidth]{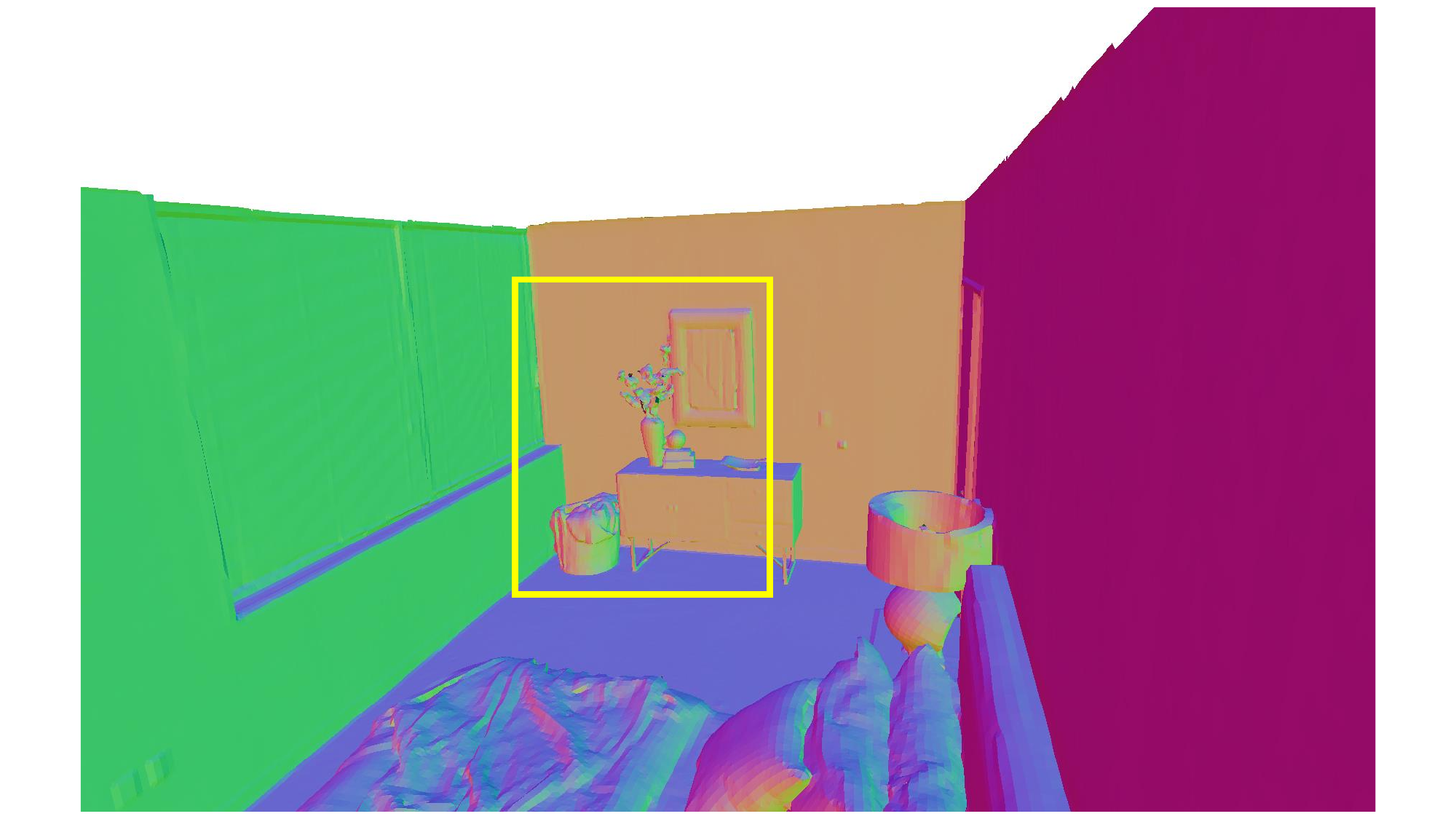} } \\
            \rotatebox[origin=b]{90}{ \texttt{Office 0}} & \raisebox{-0.5\height}{\includegraphics[width=.24\textwidth]{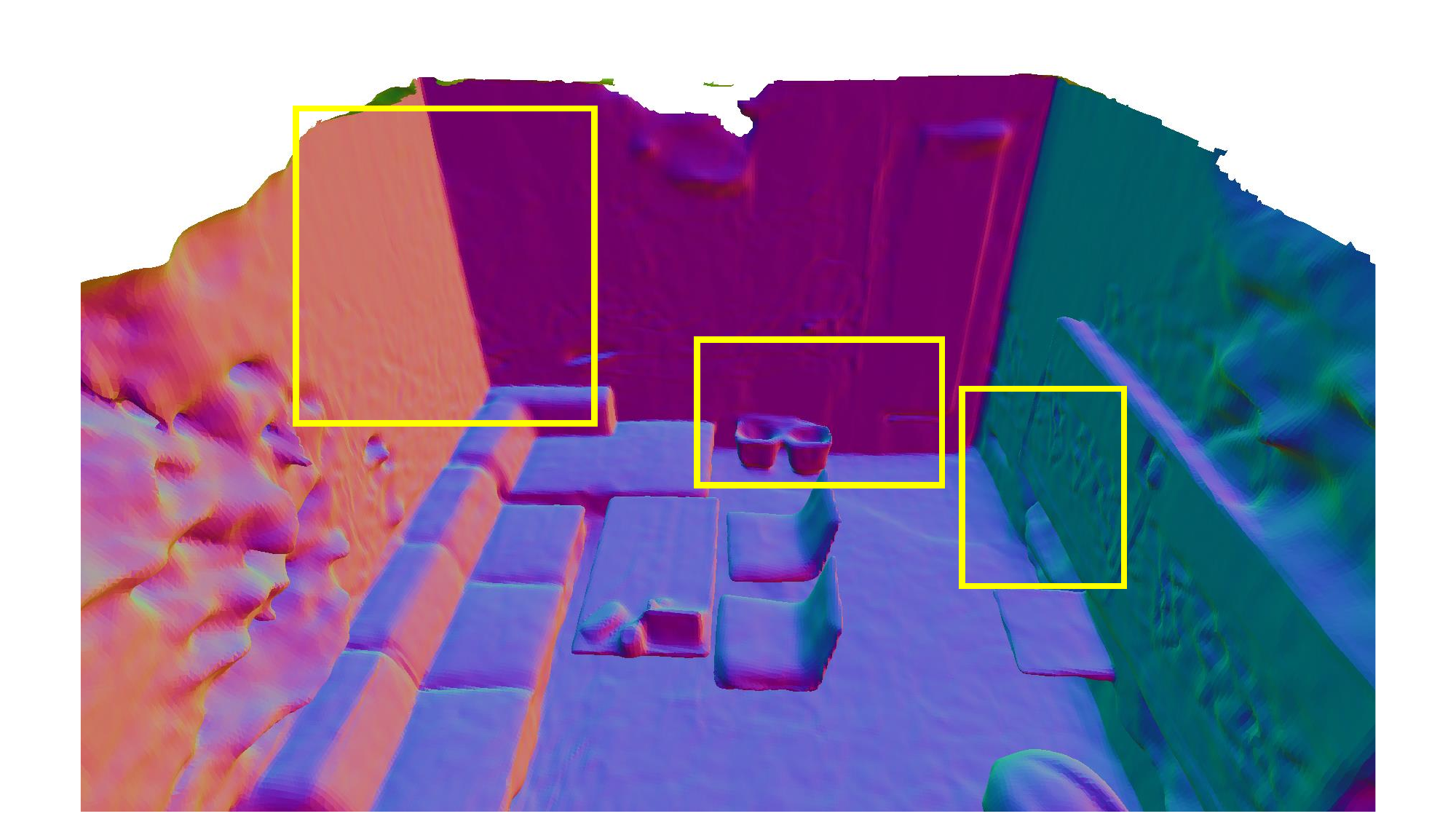} } & \raisebox{-0.5\height}{\includegraphics[width=.24\textwidth]{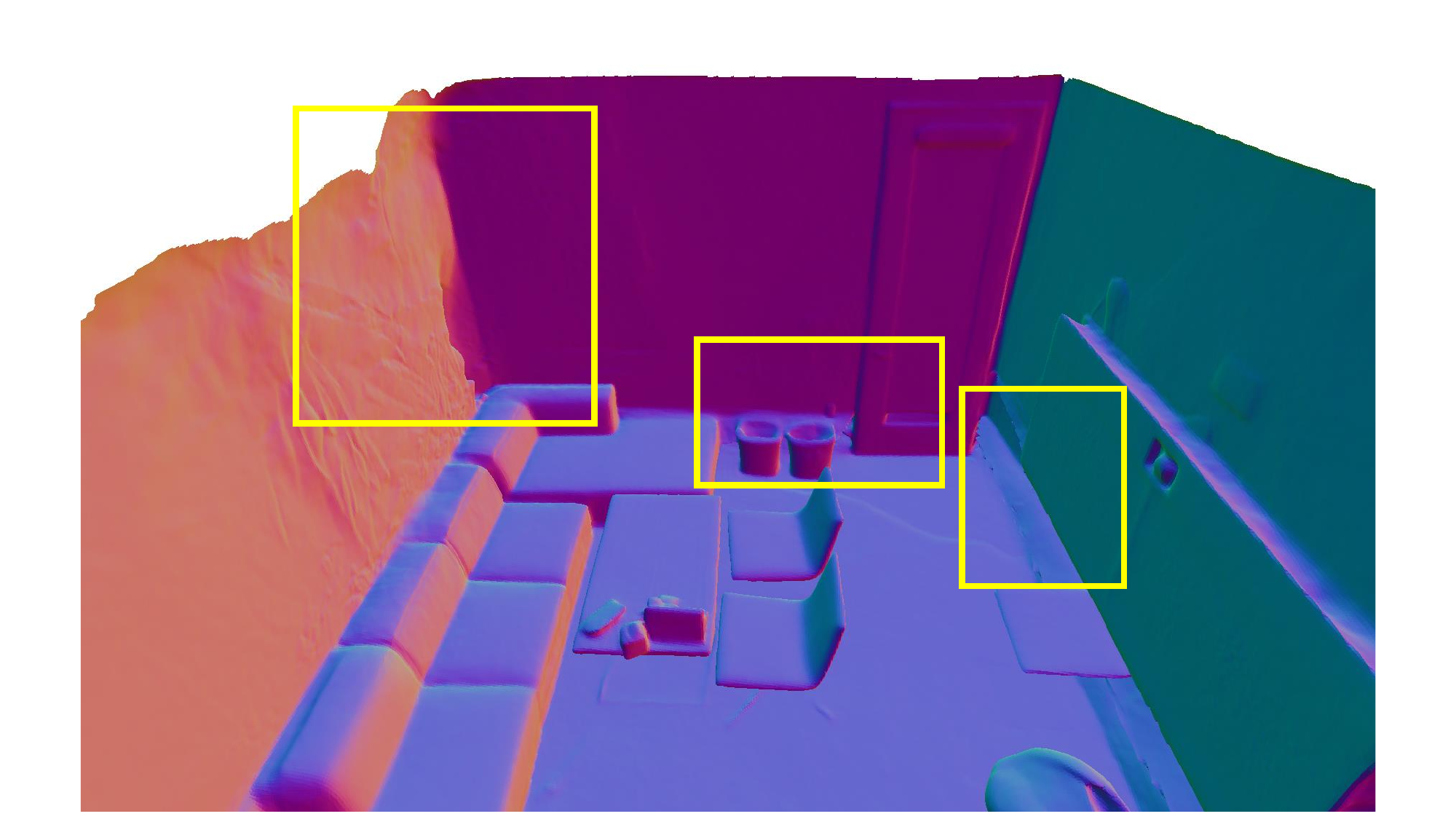} } & \raisebox{-0.5\height}{\includegraphics[width=.24\textwidth]{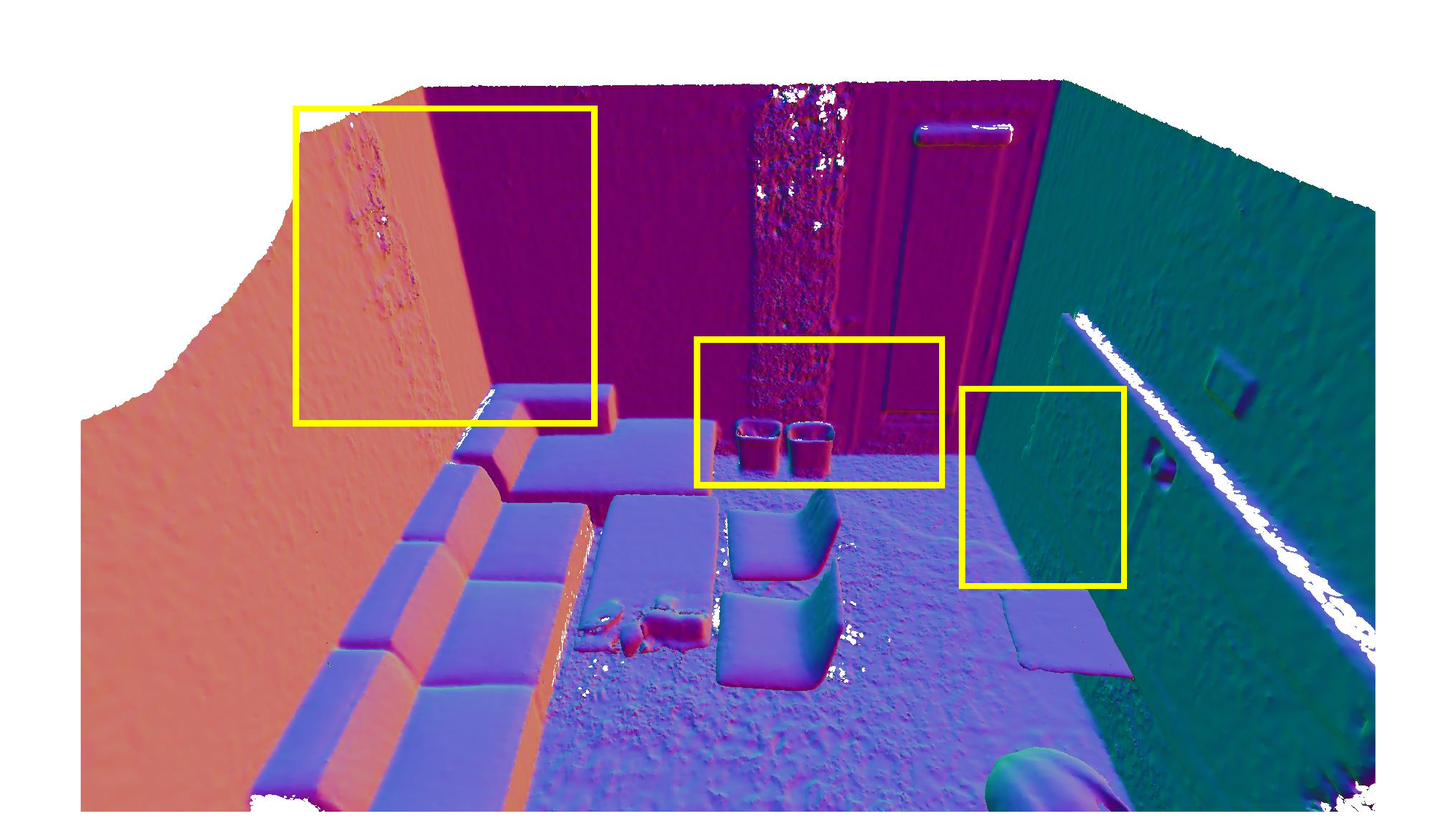} } & \raisebox{-0.5\height}{\includegraphics[width=.24\textwidth]{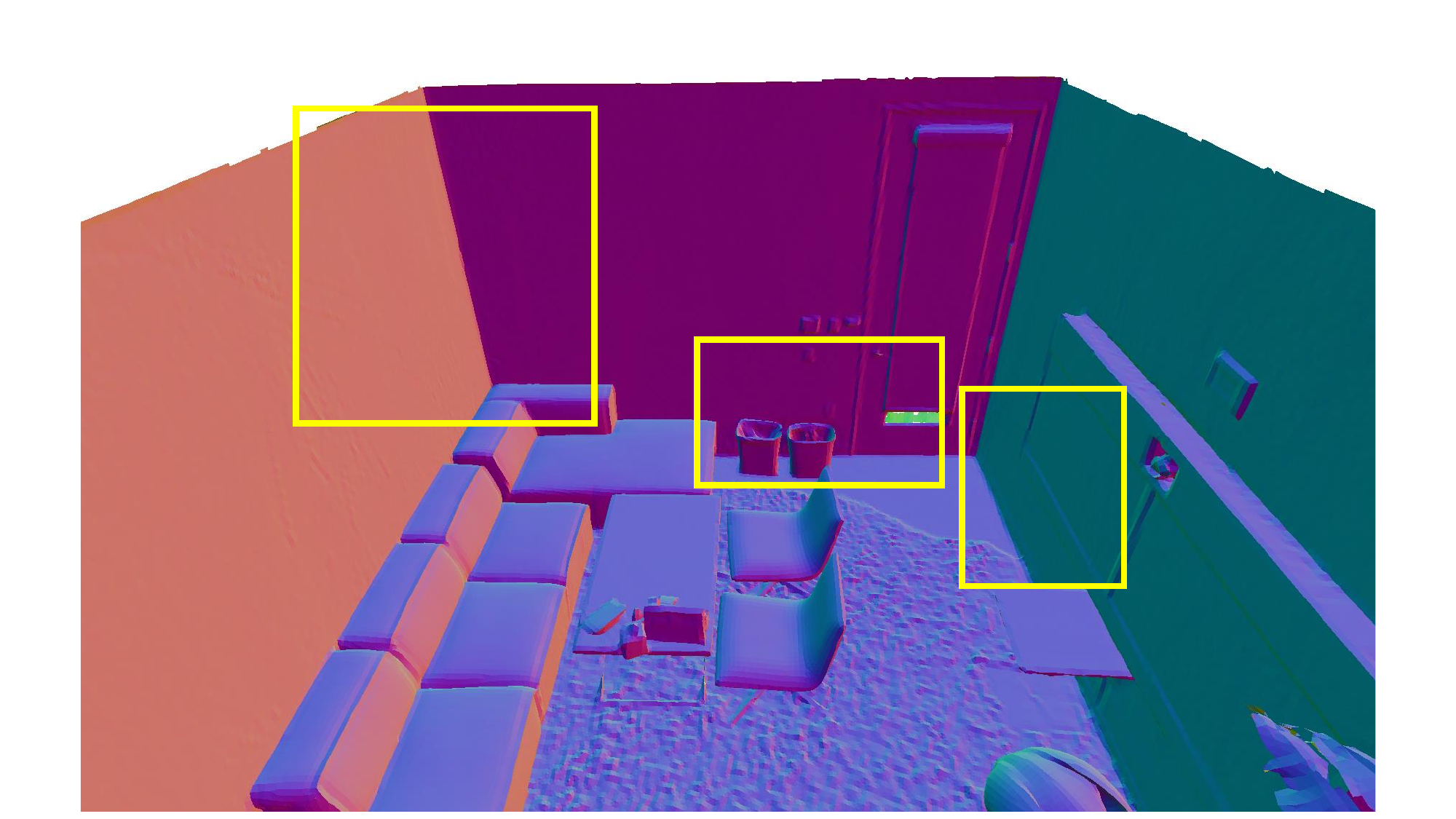} } \\
            & M-SDF~\cite{guo2022manhattan} & MonoSDF~\cite{yu2022monosdf} & \textbf{\vfnerf}  & GT \\[3mm]
        \end{tabular}
        \setlength{\tabcolsep}{1.6pt}
        \begin{tabular}{ccccc}
            \rotatebox[origin=b]{90}{ \texttt{0580}} & \raisebox{-0.5\height}{\includegraphics[trim=3cm 0cm 3cm 0cm, clip, width=.24\textwidth]{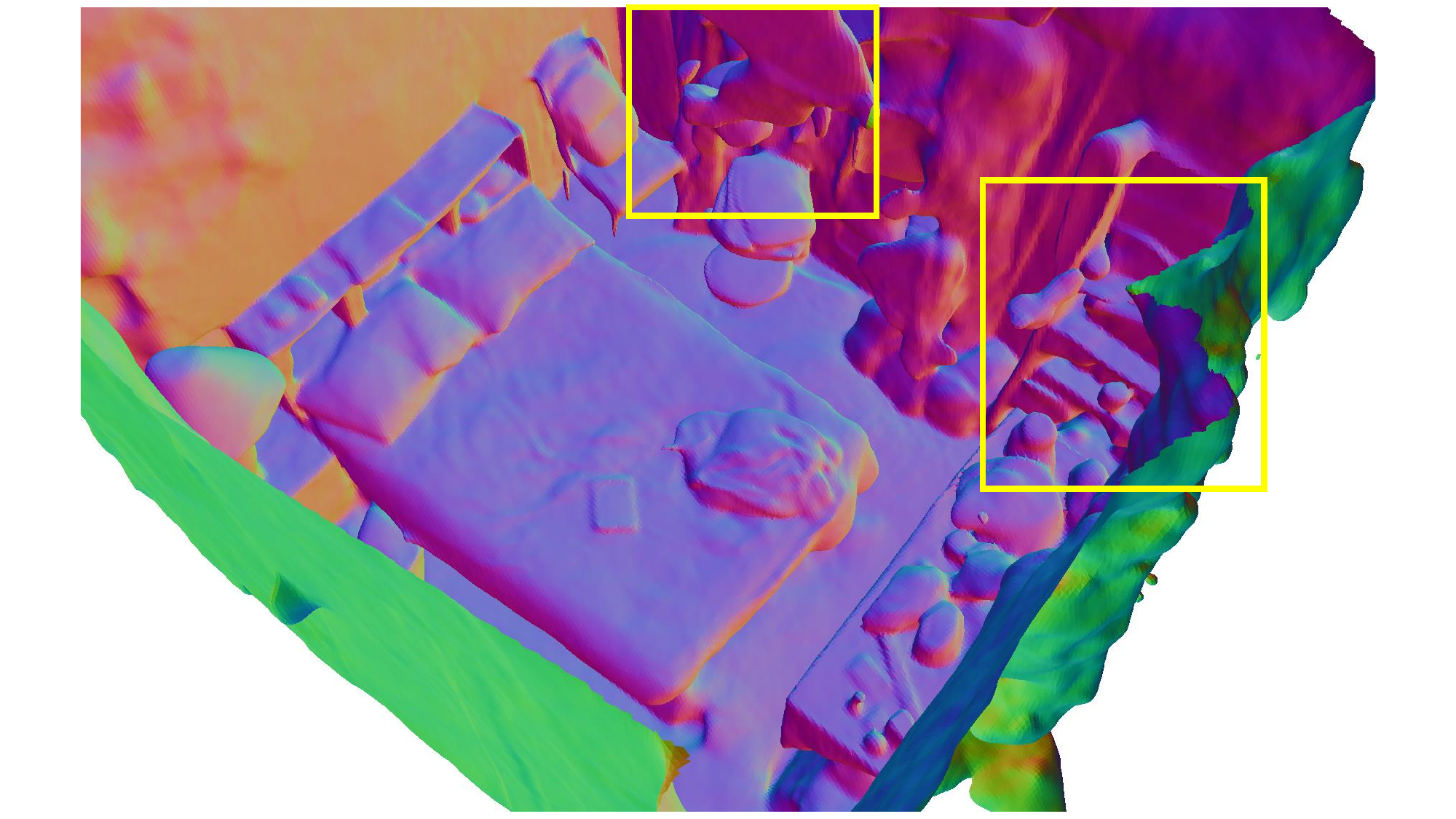}} & \raisebox{-0.5\height}{\includegraphics[trim=3cm 0cm 3cm 0cm, clip, width=.24\textwidth]{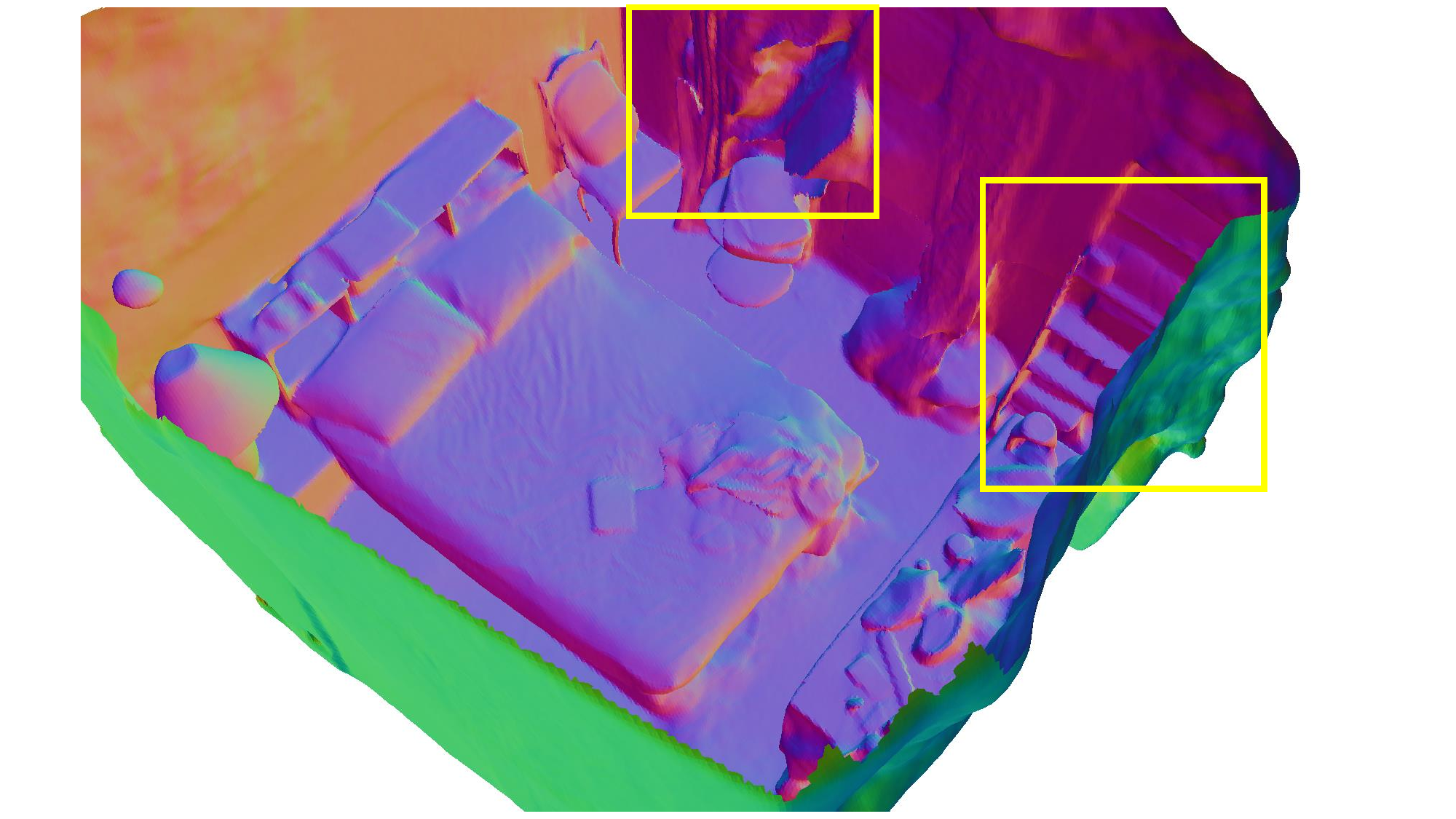}}
            &\raisebox{-0.5\height}{\includegraphics[trim=3cm 0cm 3cm 0cm, clip, width=.24\textwidth]{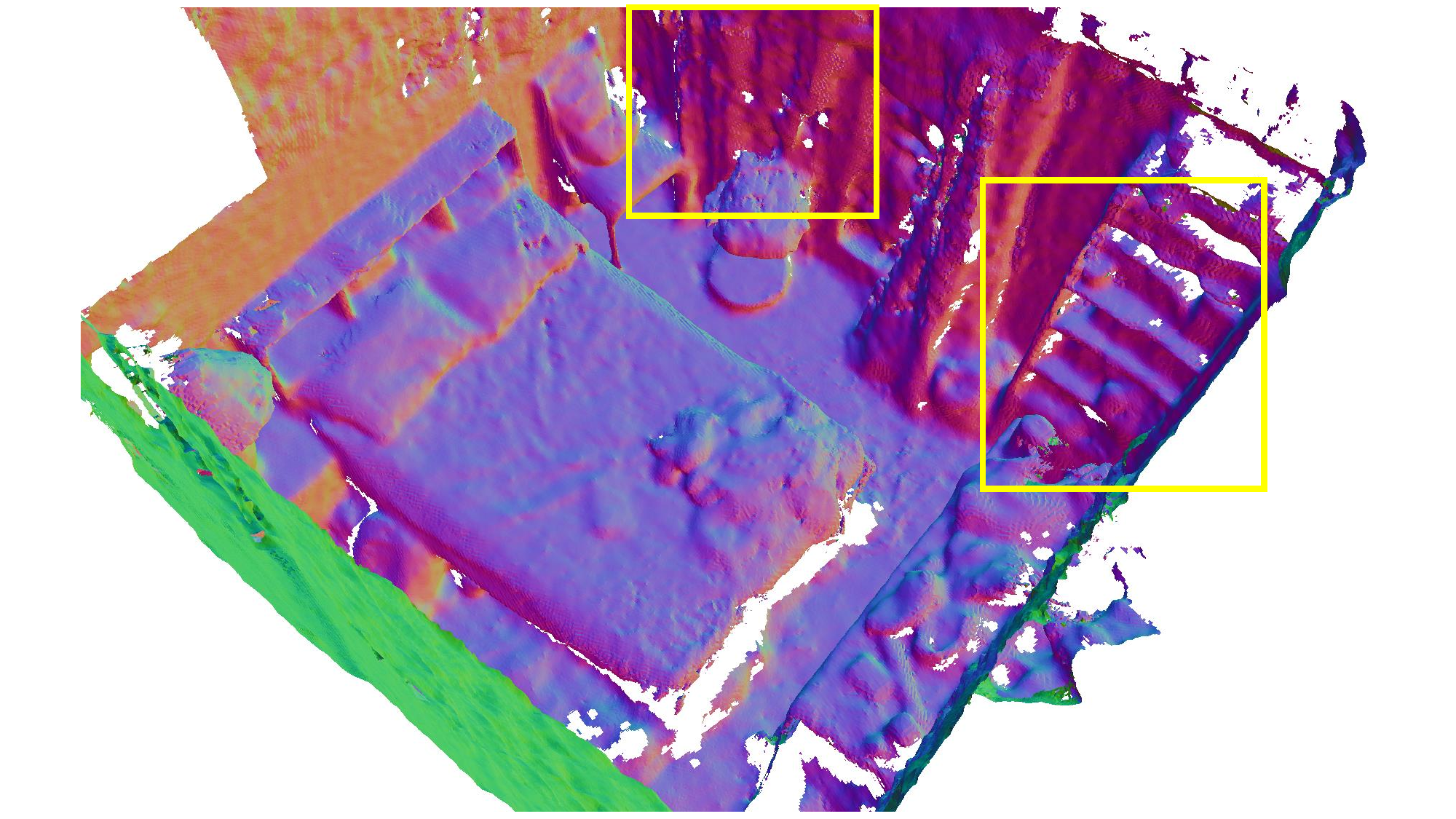} } & \raisebox{-0.5\height}{\includegraphics[trim=3cm 0cm 3cm 0cm, clip, width=.24\textwidth]{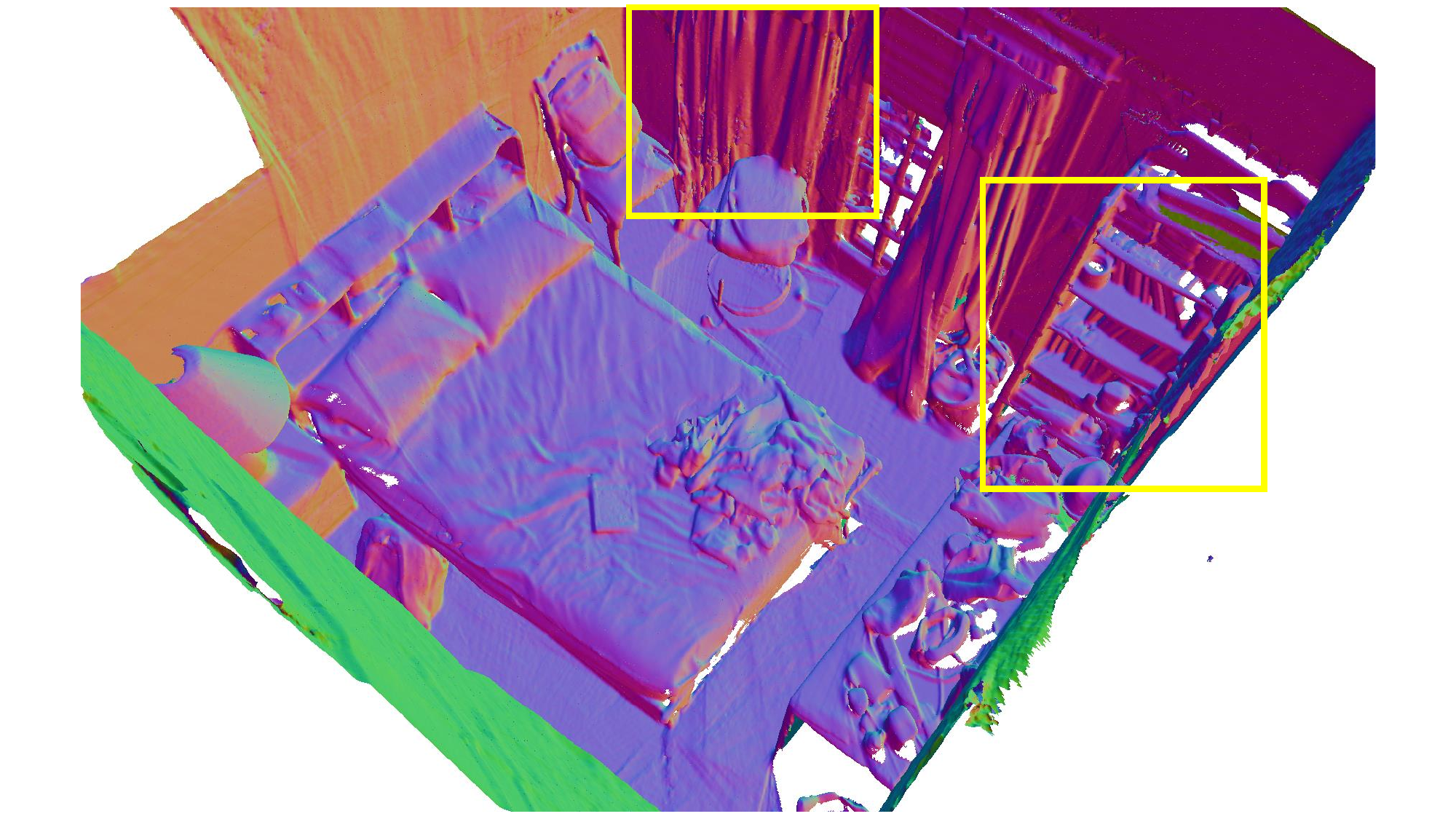} } \\
            & NeuRIS~\cite{Wang2022NeuRIS} & MonoSDF~\cite{yu2022monosdf} & \textbf{\vfnerf}  & GT \\
        \end{tabular}
        \caption[3D reconstruction qualitative results]{\textbf{3D reconstruction qualitative results.} \vfnerf outperforms the SOTA in planar regions of the scenes such as walls and floors as well as in several details. We highlight regions where \vfnerf outperforms the other methods with yellow boxes. }
        \label{fig:qualitative}
    \end{figure*}

\mypar{Implementation details}
Our method is developed using PyTorch~\cite{Paszke2010pytorch} and trained using the Adam optimizer~\cite{Bengio2015Adam}. The VF and color functions are designed as MLPs consisting of 8 and 4 hidden layers, respectively. Positional encodings~\cite{mildenhall2020nerf} are used for the spatial positions $\loc$ and viewing directions $\direction$ to address the challenge of learning high-frequency details of the scene. Furthermore, we find that initializing the VF network by pretraining it to point toward the center of the scene eases the training process. The learning rate is initialized at $5\times 10^{-4}$ and is decreased using an exponential decay approach~\cite{Li2020ExponentialLR}. The training process spans 3000 epochs. Notably, weight annealing for the sliding window technique is executed between the 700th and 1400th epochs. Additionally, we use Truncated Signed Distance Function (TSDF) fusion to extract the surface mesh from the predicted depth maps and images. We set the following weights of the loss function: $w_c=1.0$, $w_{norm}=0.05$, $w_{ext} = w_{cen} = 0.5$, $w_{depth}=0.25$. Regarding the density function parameters, we set the cosine similarity threshold to $\xi=-0.5$ and initialize the learnable parameters to $\mu=0.7$, $\beta=0.5$ and $\alpha=100$.

\mypar{Datasets} We test the performance on Replica~\cite{Straub2019TheRD} and ScanNet~\cite{dai2017scannet}. The Replica dataset consists of 18 synthetic indoor scenes, where each scene contains a dense ground truth mesh, and 2000 RGB and depth images. Similarly to MonoSDF~\cite{yu2022monosdf}, we focus on only seven scenes from this dataset for comparison purposes. The ScanNet dataset contains 16113 indoor scenes with 2.5 million views, with each view containing RGB-D images. Additionally, a fused mesh is provided for each scene. We select the four scenes from this dataset used by ManhattanSDF~\cite{guo2022manhattan} and MonoSDF to evaluate our method. For replica, we sample 1 of every 20 posed images for training, while in Scannet we sample 1 of every 40.

\mypar{Metrics}  We evaluate following standard protocol \cite{yu2022monosdf}. For 3D surface reconstruction, we focus on evaluating our method with median Chamfer distance (CD) and F1-score~\cite{Sun2021NeuralReconRC} with a threshold of $5\text{cm}$. We also provide the peak signal-to-noise ratio (PSNR) to evaluate view synthesis. The detailed definitions of these metrics are included in the supplementary material.

\mypar{Baselines} We compare our method against the State of the Art (SOTA), which use volume rendering for indoor scene reconstruction: ManhattanSDF~\cite{guo2022manhattan}, MonoSDF~\cite{yu2022monosdf}, NeuRIS~\cite{Wang2022NeuRIS} and Neuralangelo~\cite{li2023neuralangelo}. We use Marching Cubes~\cite{Lorensen987MC} to extract the meshes rendered by the baselines.

\subsection{Comparisons with baselines}

\mypar{3D reconstruction}
We evaluate our method on the Replica and ScanNet datasets. The qualitative results are illustrated in~\cref{fig:qualitative}. Quantitative results on both datasets are shown in~\cref{tab:quantitative}. Additional detailed qualitative and quantitative results are included in the supplementary material. Our method outperforms volume rendering based benchmarks in terms of F1-score and median CD on both datasets. Most interestingly, the gap in performance is significantly higher in ScanNet, a more challenging dataset containing noisy depth maps. The ability to perform extremely well on real depths/images might be explained by the strong inductive bias offered by VF which allows it to learn planar regions even in the presence of noisy data.

\setlength{\tabcolsep}{.5pt}
\begin{figure*}[t]
        \centering
        \begin{tabular}{cccc}
             {\includegraphics[width=.2\textwidth]{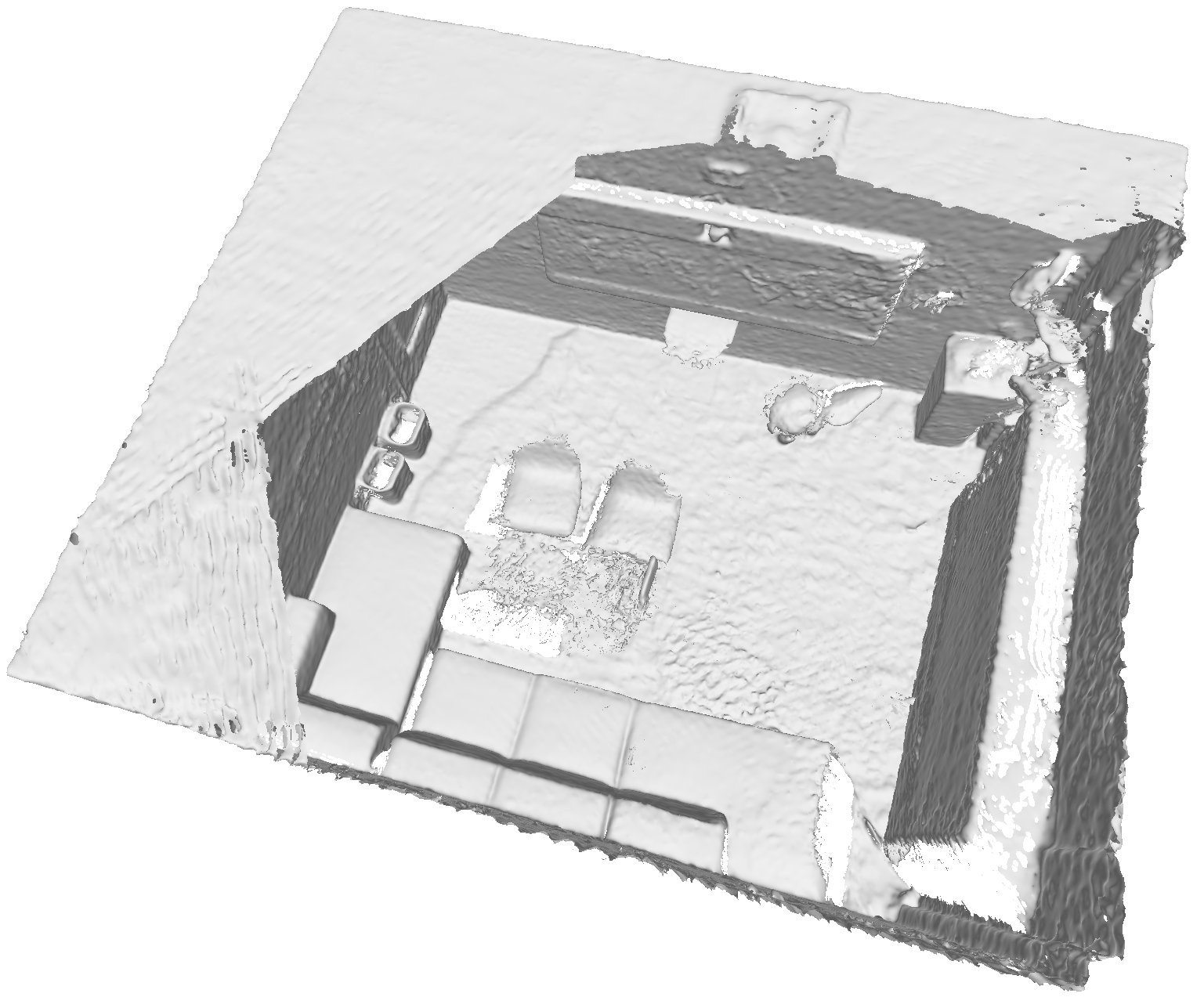}} & {\includegraphics[width=.2\textwidth]{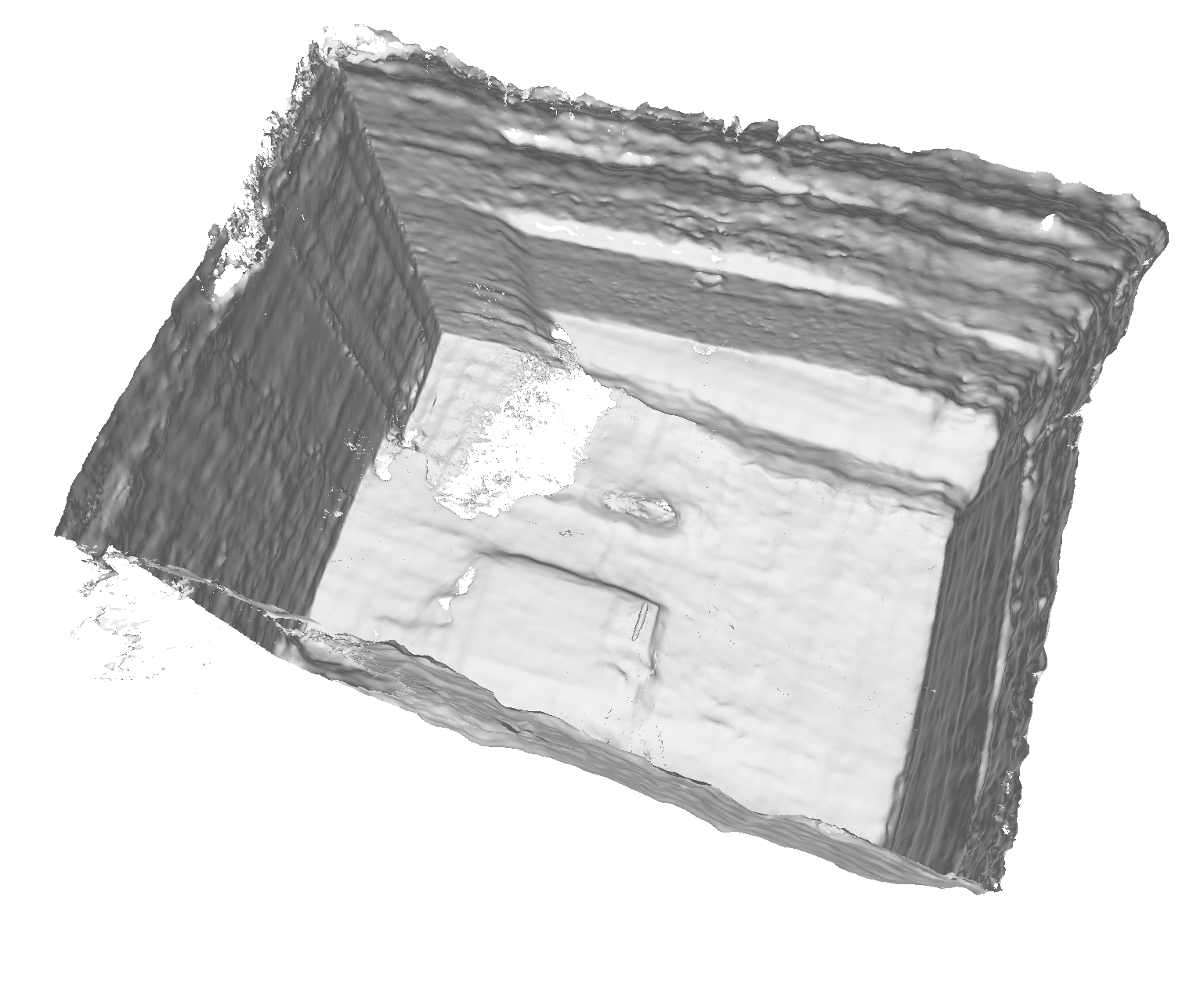} } & {\includegraphics[width=.2\textwidth]{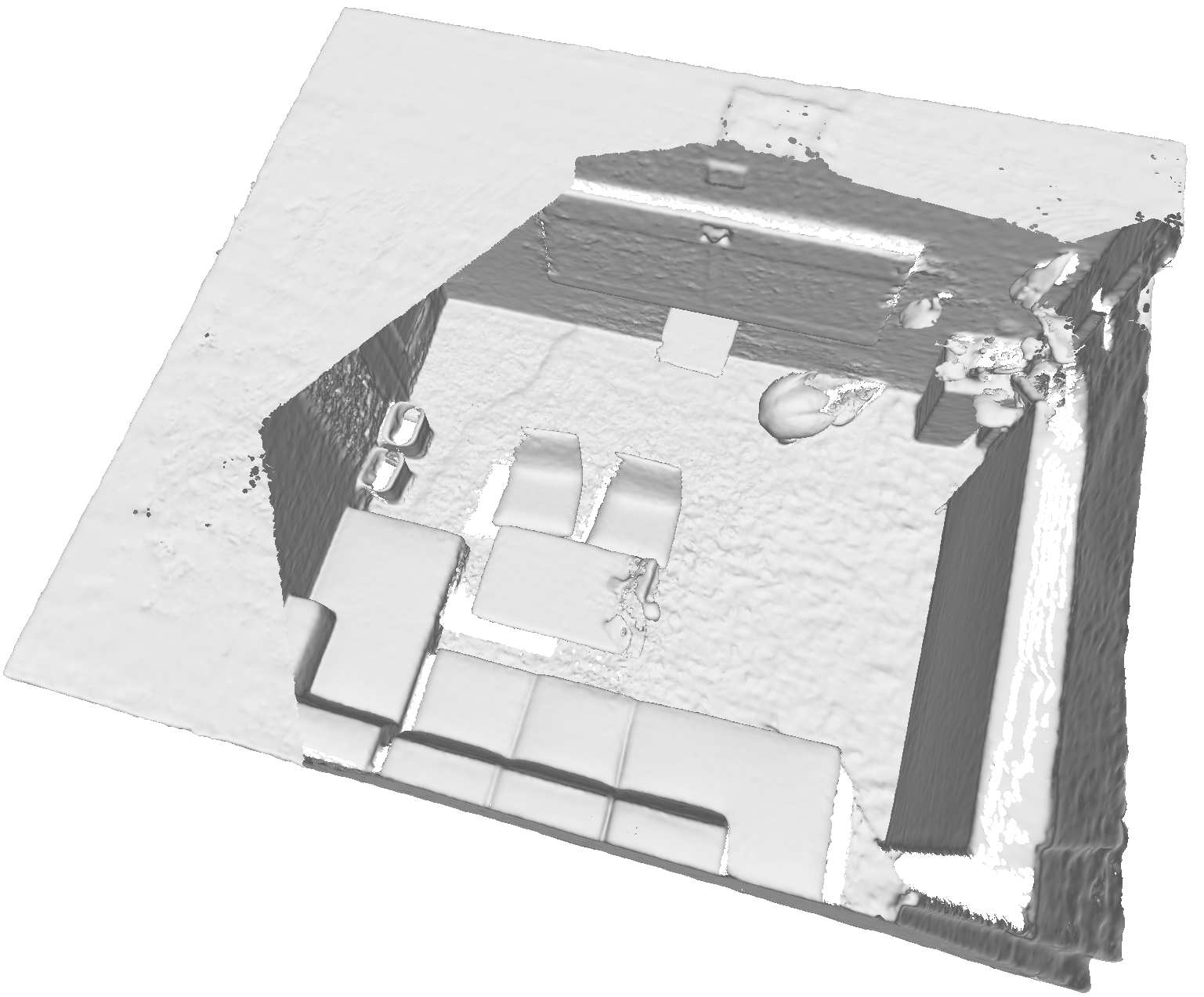} } &{\includegraphics[width=.2\textwidth]{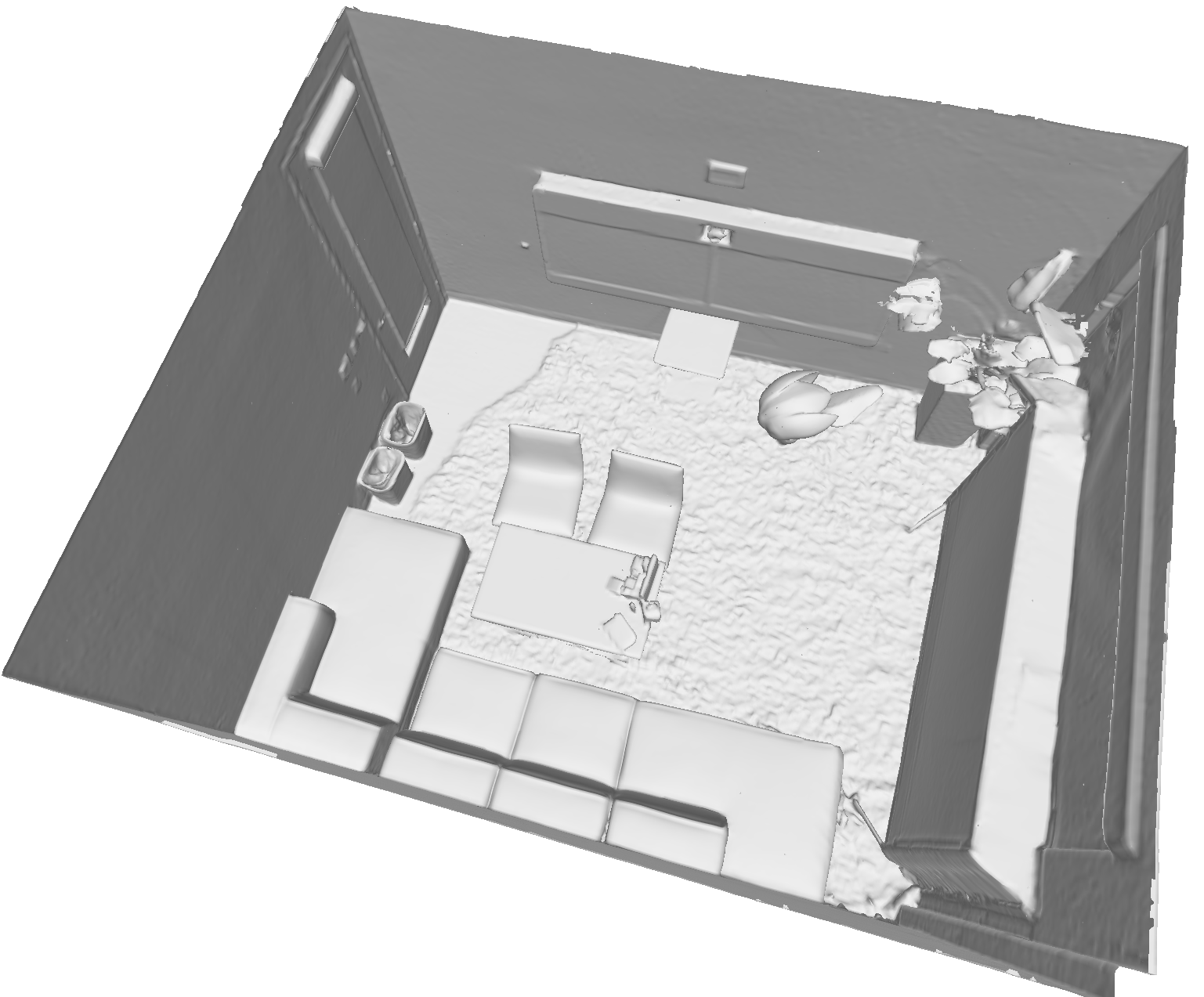} } \\
        w$\boldsymbol{\setminus}$o hierarchical & w$\boldsymbol{\setminus}$o $\loss_{depth}$ & \textbf{\vfnerf} & Ground\\  
        sampling & & \textbf{(Ours)}& Truth\\
        \end{tabular}
        \caption[Ablations]{\textbf{Ablations.} Removing the hierarchical sampling generates holes and artifacts (see table in the meshes). Our method without $\loss_{depth}$ is less accurate, as most regions of the scene are low-textured. Nonetheless, it still captures the overall scene coarse geometry.}
        \label{fig:loss-ablation}
\end{figure*}

\setlength{\tabcolsep}{.5pt}
\begin{figure}[t]
        \centering
        \begin{tabular}{cccc}
            {\includegraphics[width=.25\columnwidth]{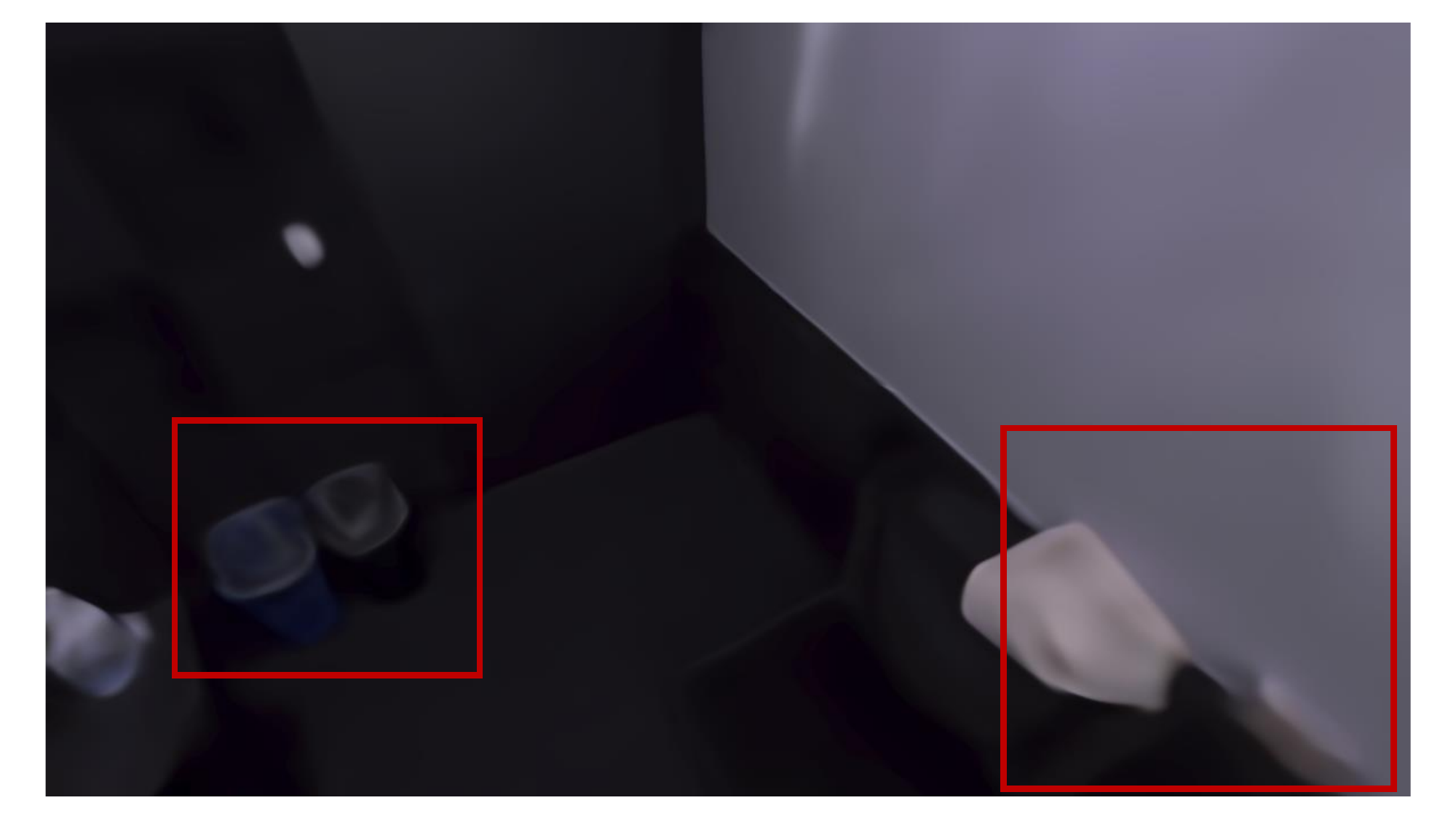} } & {\includegraphics[width=.25\columnwidth]{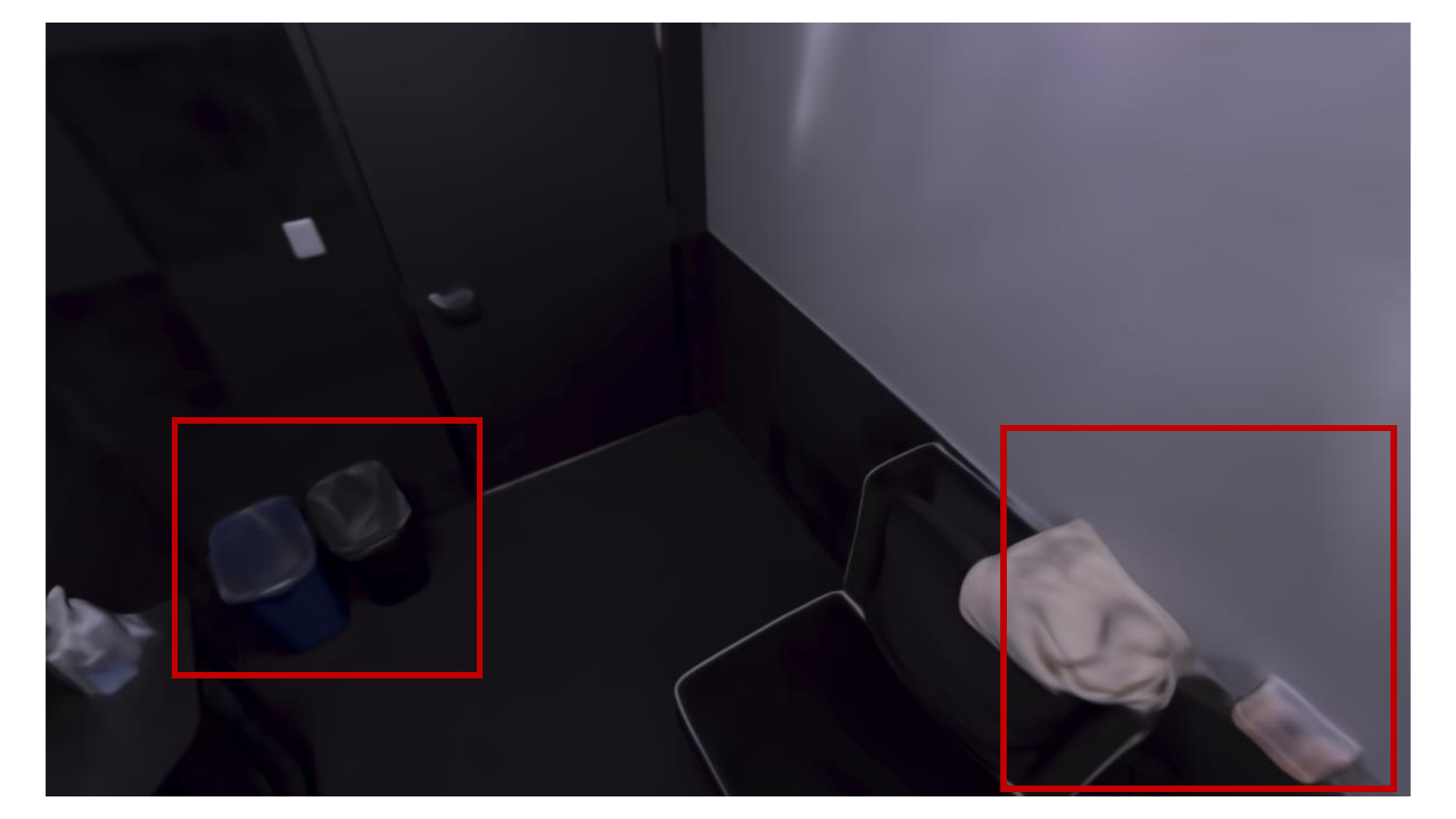} } & {\includegraphics[width=.25\columnwidth]{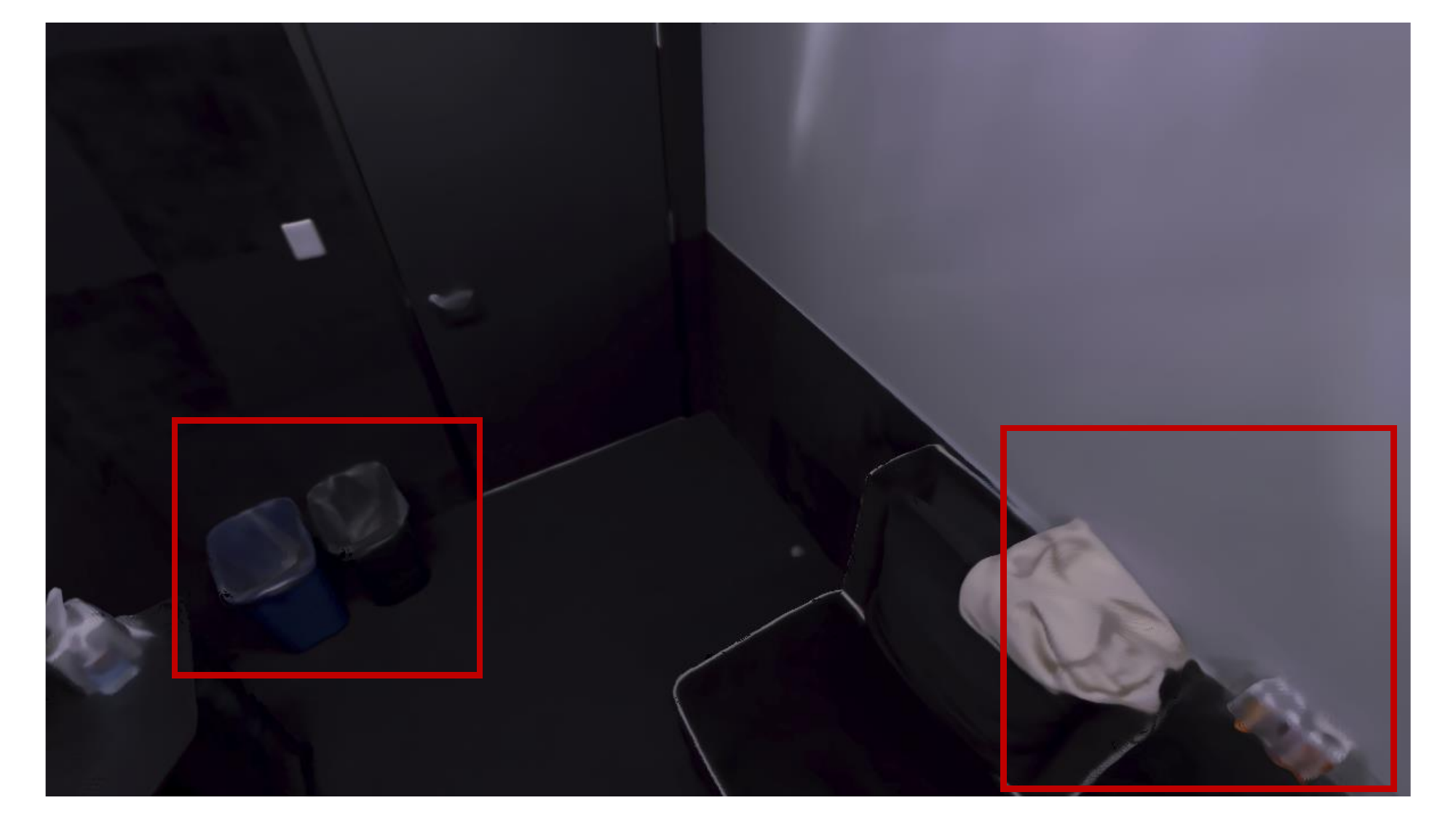} } & {\includegraphics[width=.25\columnwidth]{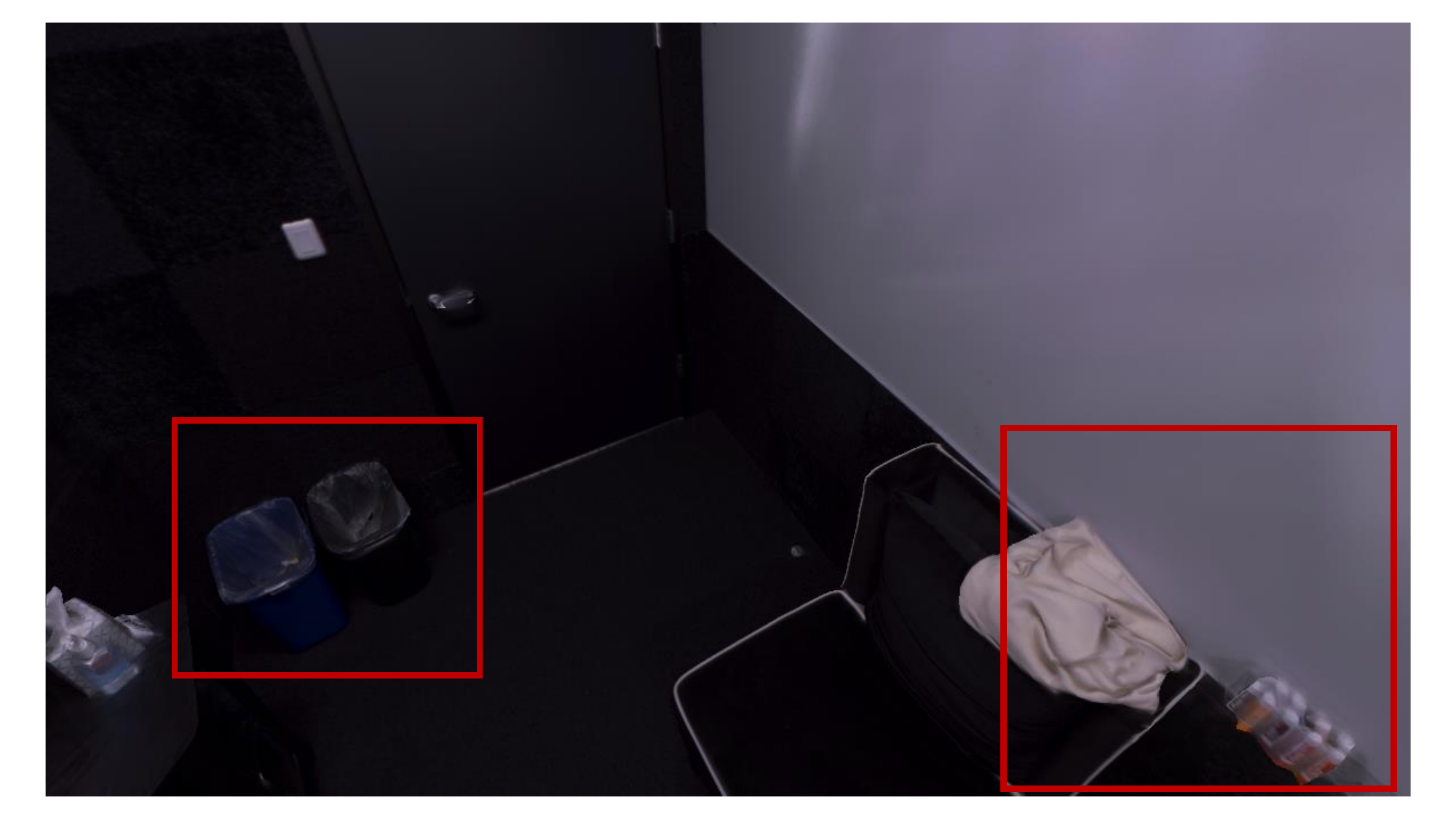} } \\
            {\includegraphics[width=.25\columnwidth]{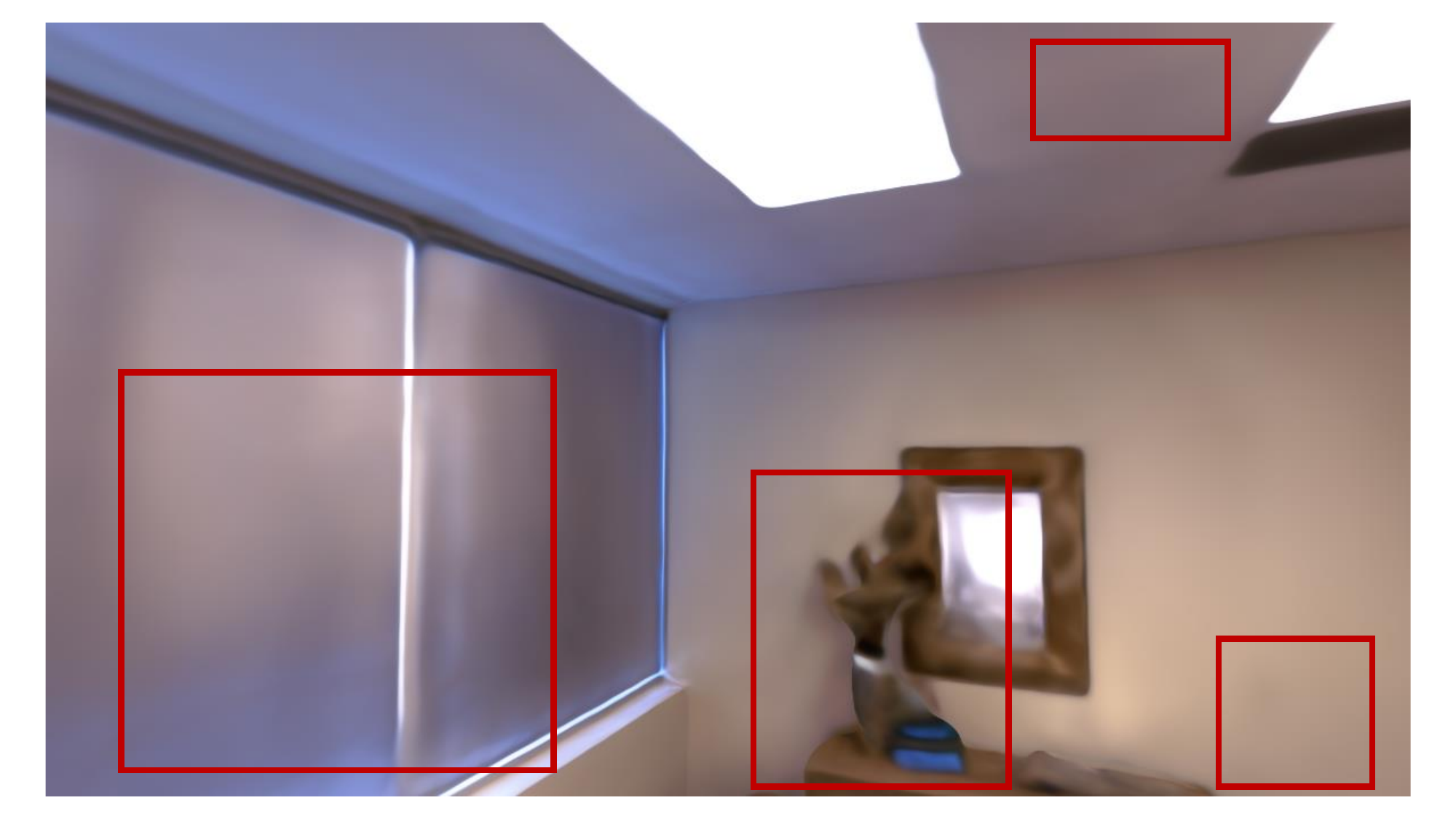} } & {\includegraphics[width=.25\columnwidth]{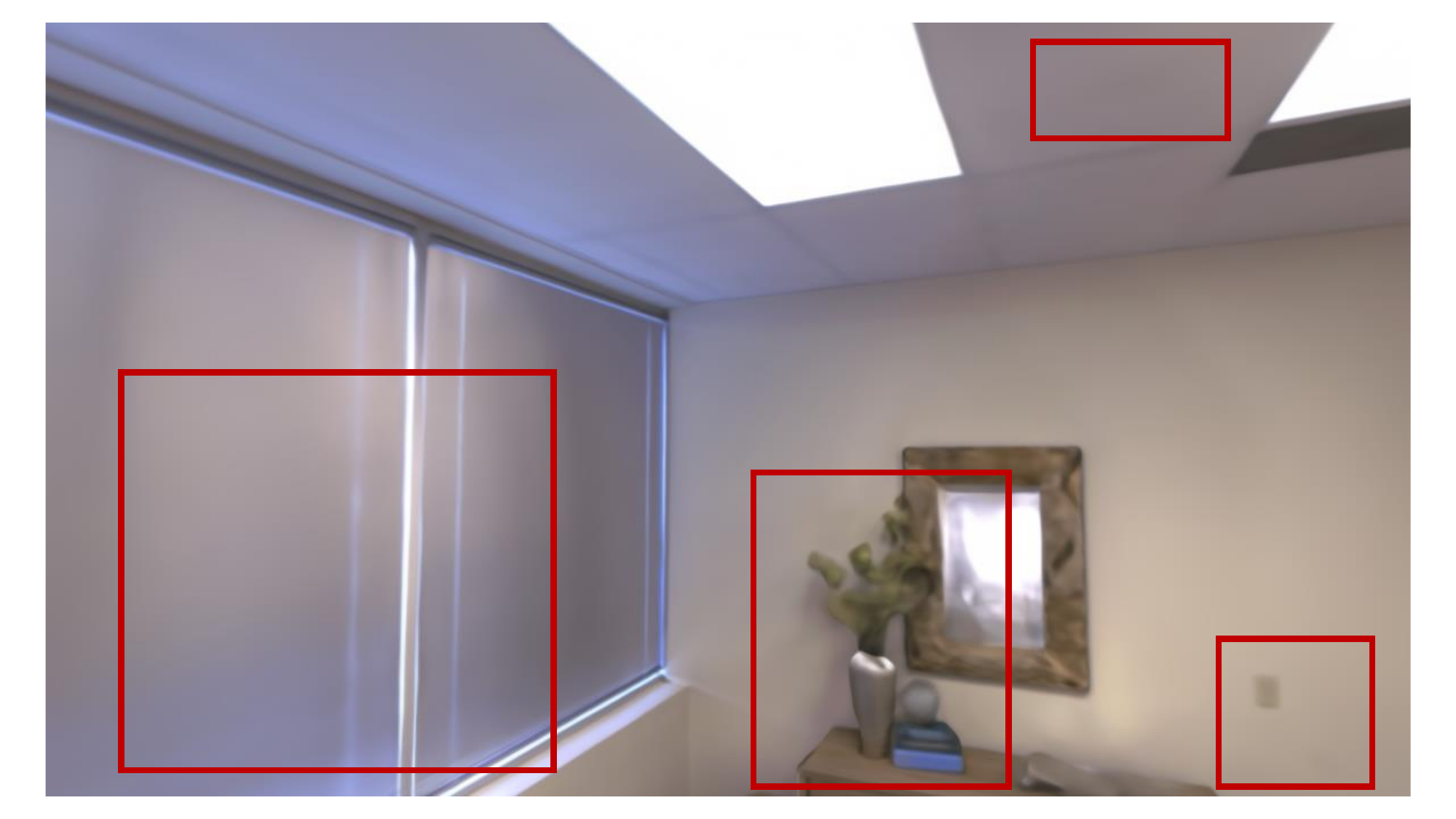} } & {\includegraphics[width=.25\columnwidth]{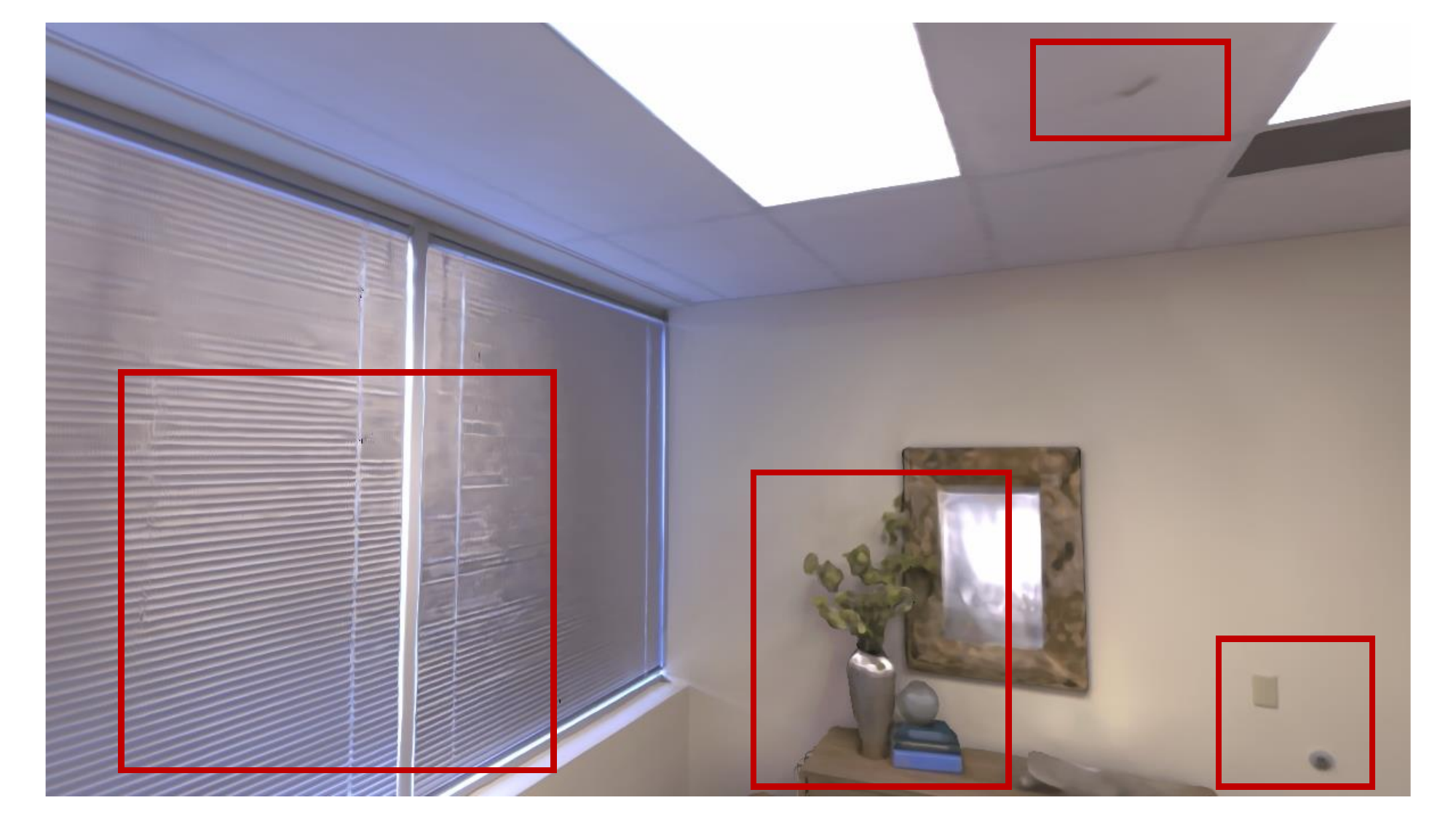} } & {\includegraphics[width=.25\columnwidth]{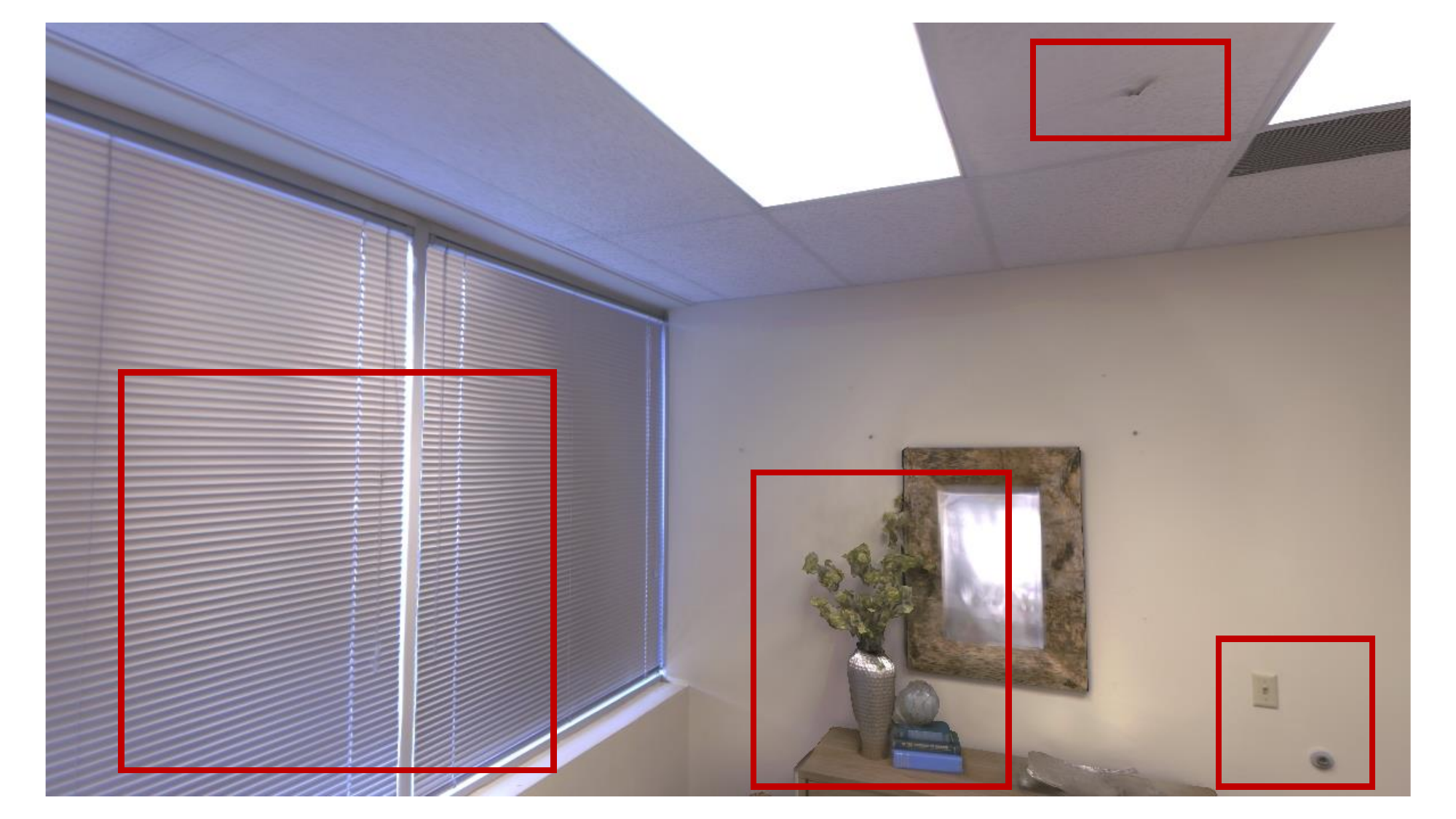} } \\
            {\includegraphics[width=.25\columnwidth]{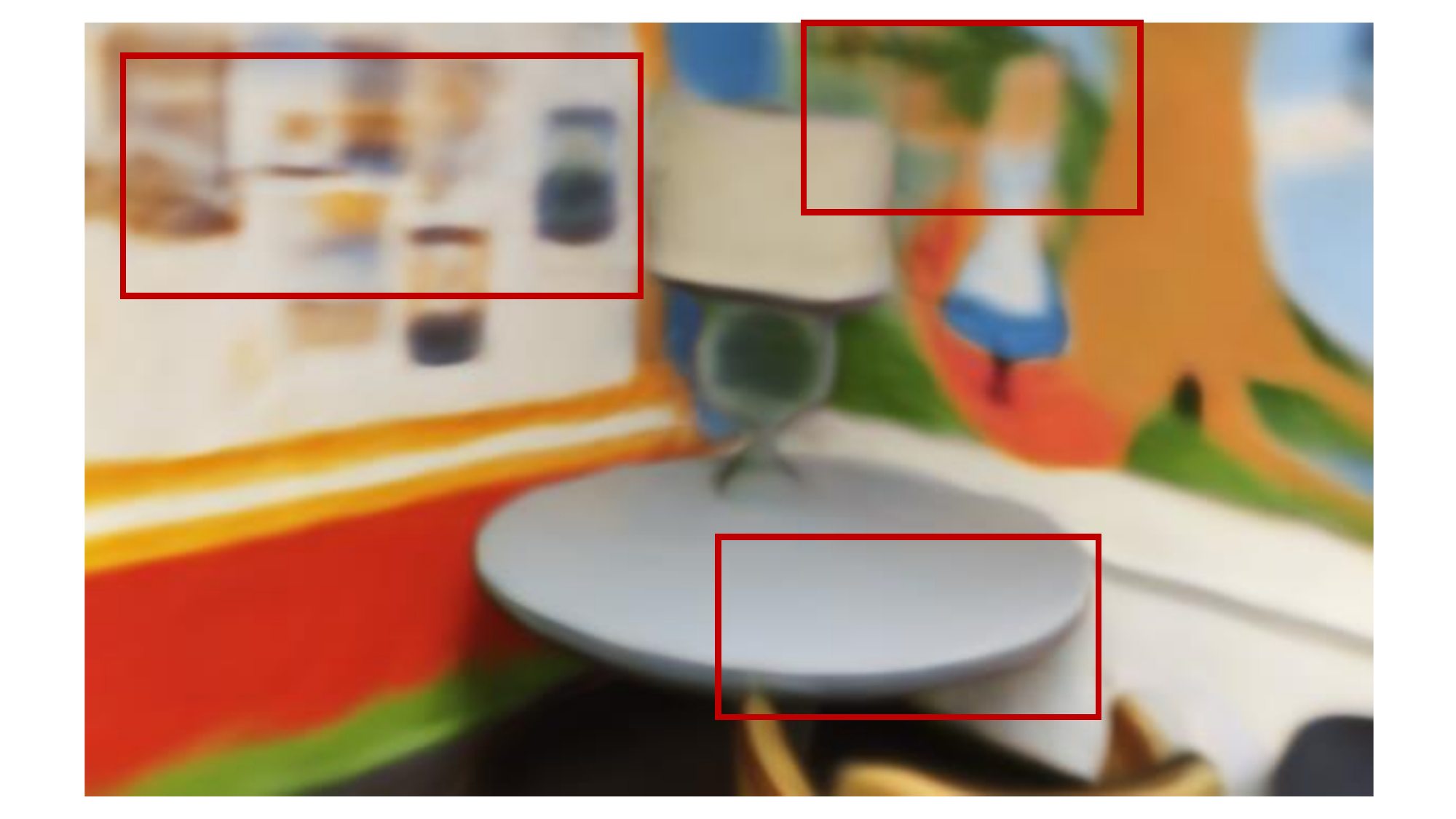} } & {\includegraphics[width=.25\columnwidth]{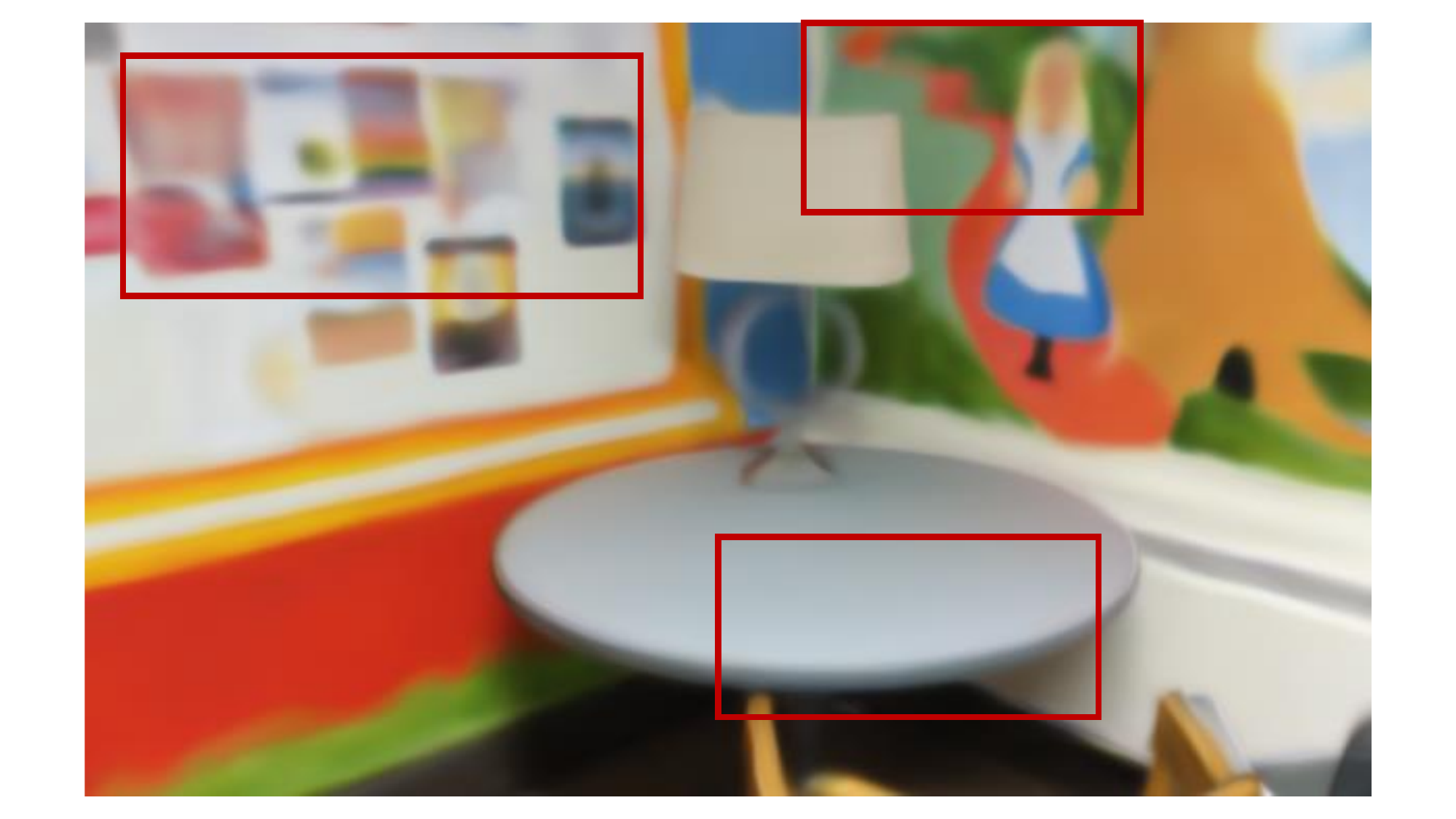} } & {\includegraphics[width=.25\columnwidth]{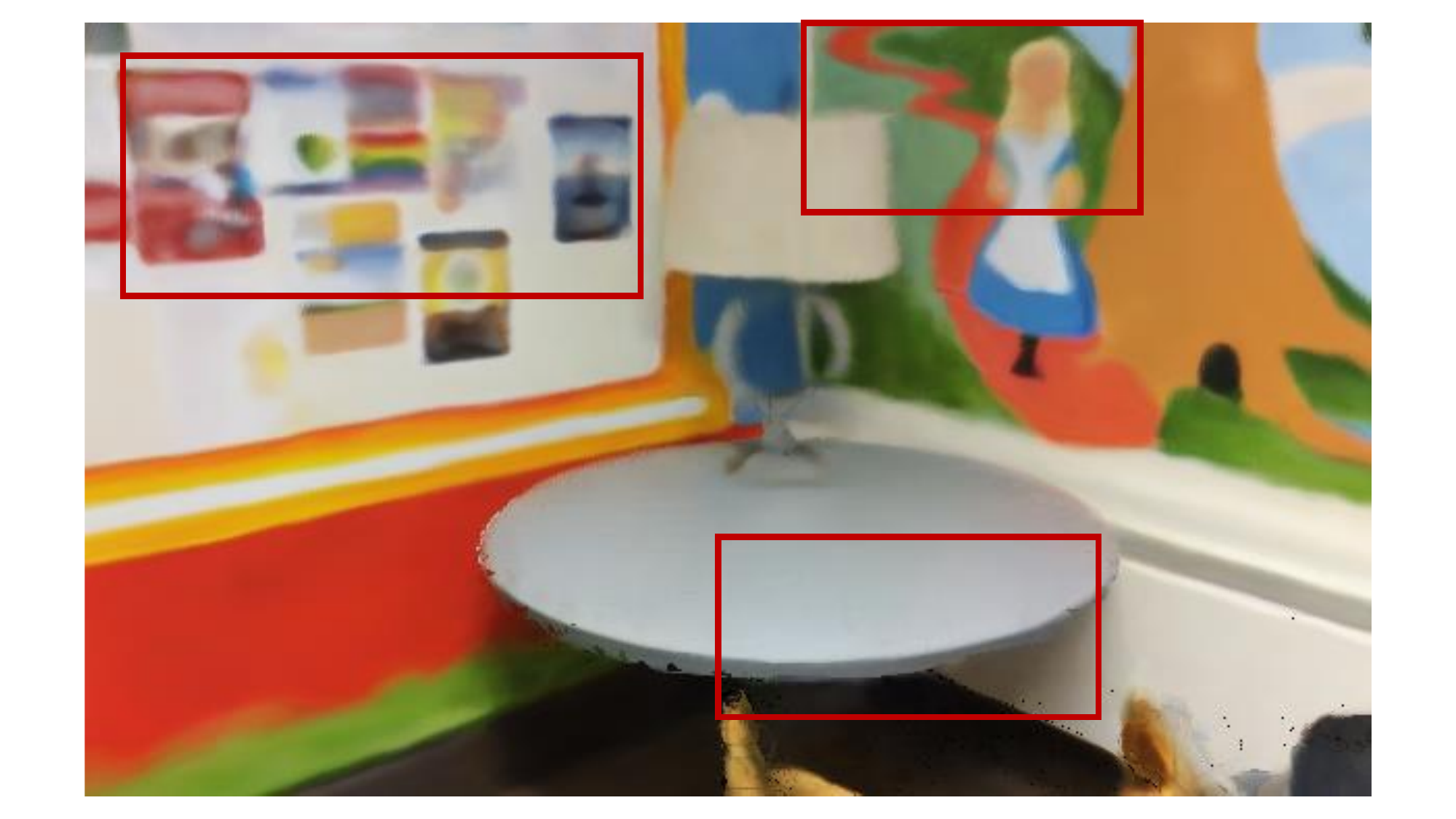} } & {\includegraphics[width=.25\columnwidth]{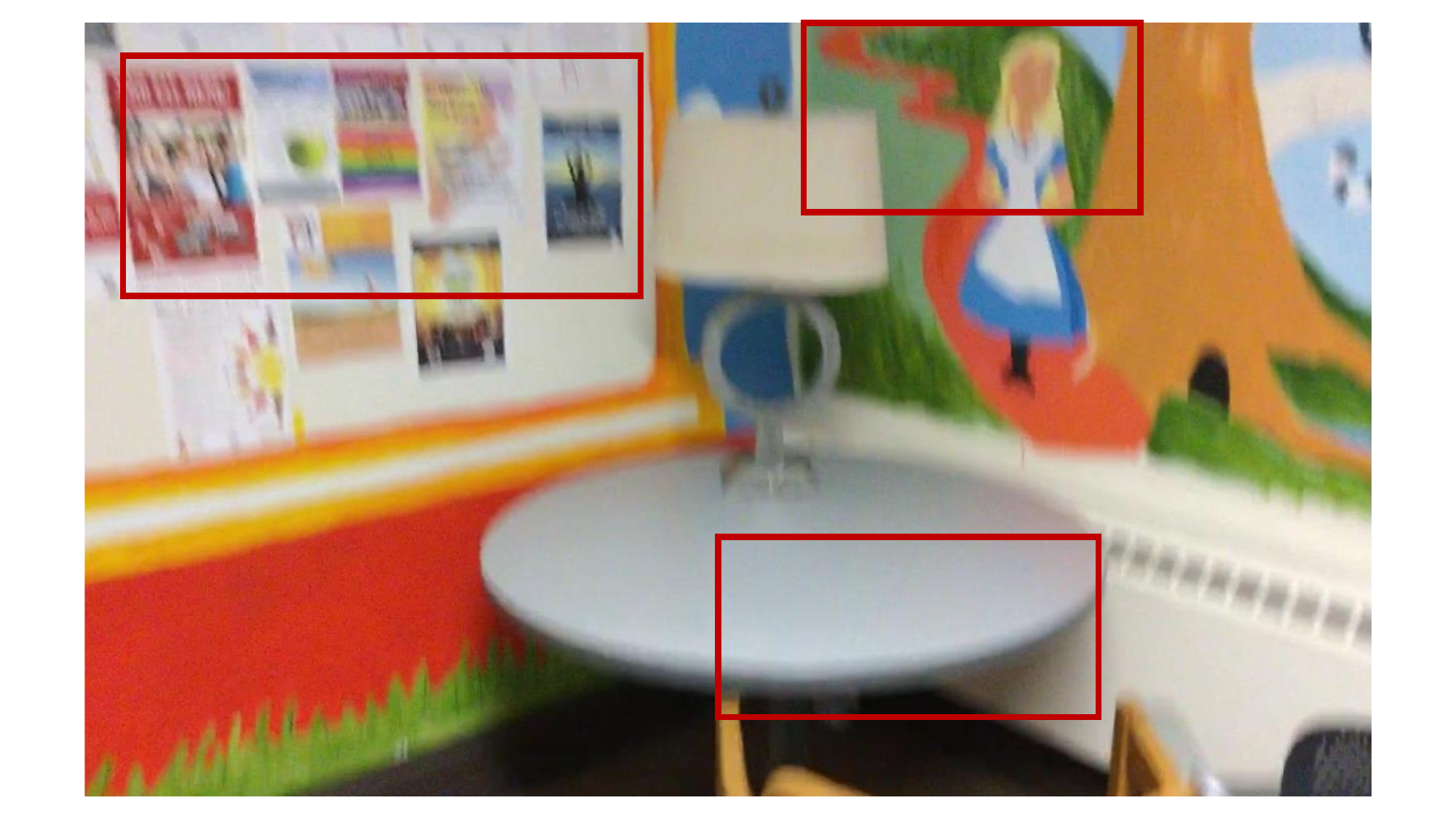} } \\
            \textsc{\small{Manhattan}}\xspace & \monosdf~\cite{yu2022monosdf} & \textbf{\vfnerf} & Ground \\
            \textsc{\small{SDF}}\xspace~\cite{guo2022manhattan} &  & \textbf{(Ours)} & Truth \\
        \end{tabular}
        \caption[Novel view synthesis qualitative results]{\textbf{Novel view synthesis qualitative results.} Our method renders accurate images with high-frequency details. Compared to ManhattanSDF, our method is more accurate and introduces less smoothness. Additionally, \vfnerf is more effective than MonoSDF in preserving high frequency details (e.g. the blinds). Interestingly, we observe that in the bottom example, \vfnerf renders an image that, at spots (e.g. the drawing on the wall), is sharper than the GT image which suffered from motion blur.}
        \label{fig:qualitative-syntesis}
    \end{figure}

    \setlength{\tabcolsep}{.1pt}
\begin{figure}[ht!]
        \centering
        \begin{tabular}{cccc}
            {\includegraphics[width=.24\columnwidth]{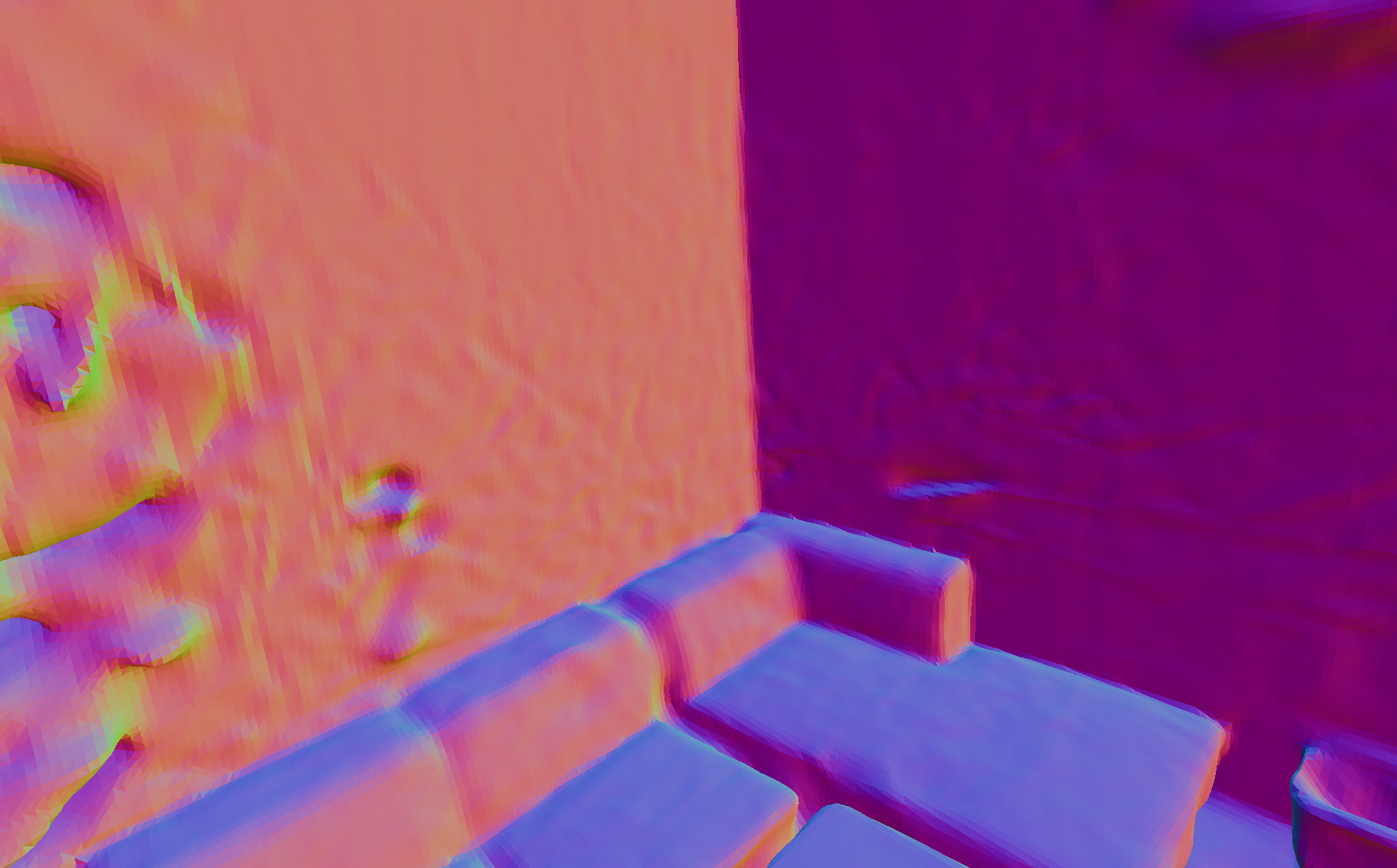} } & {\includegraphics[width=.24\columnwidth]{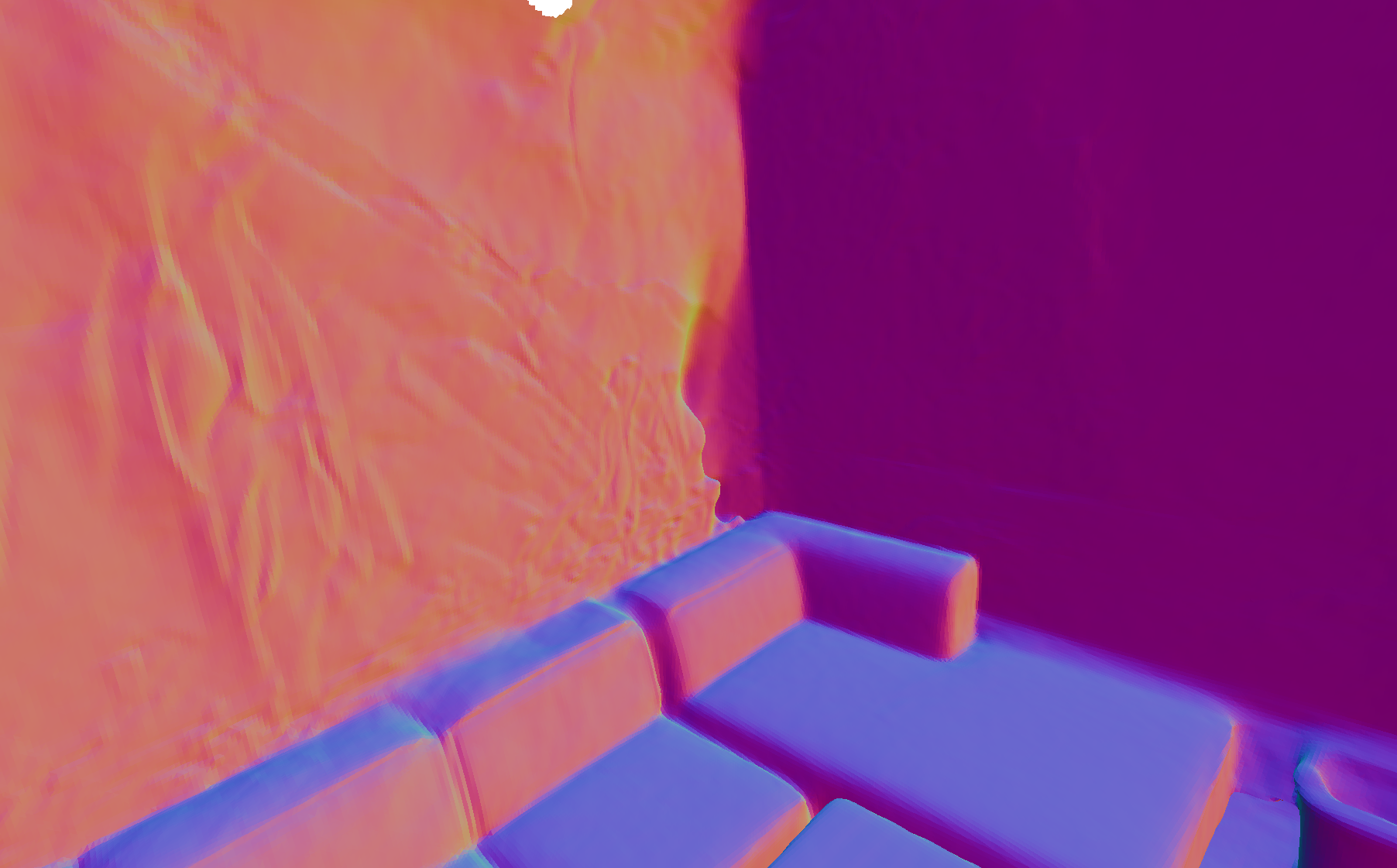} } & {\includegraphics[width=.24\columnwidth]{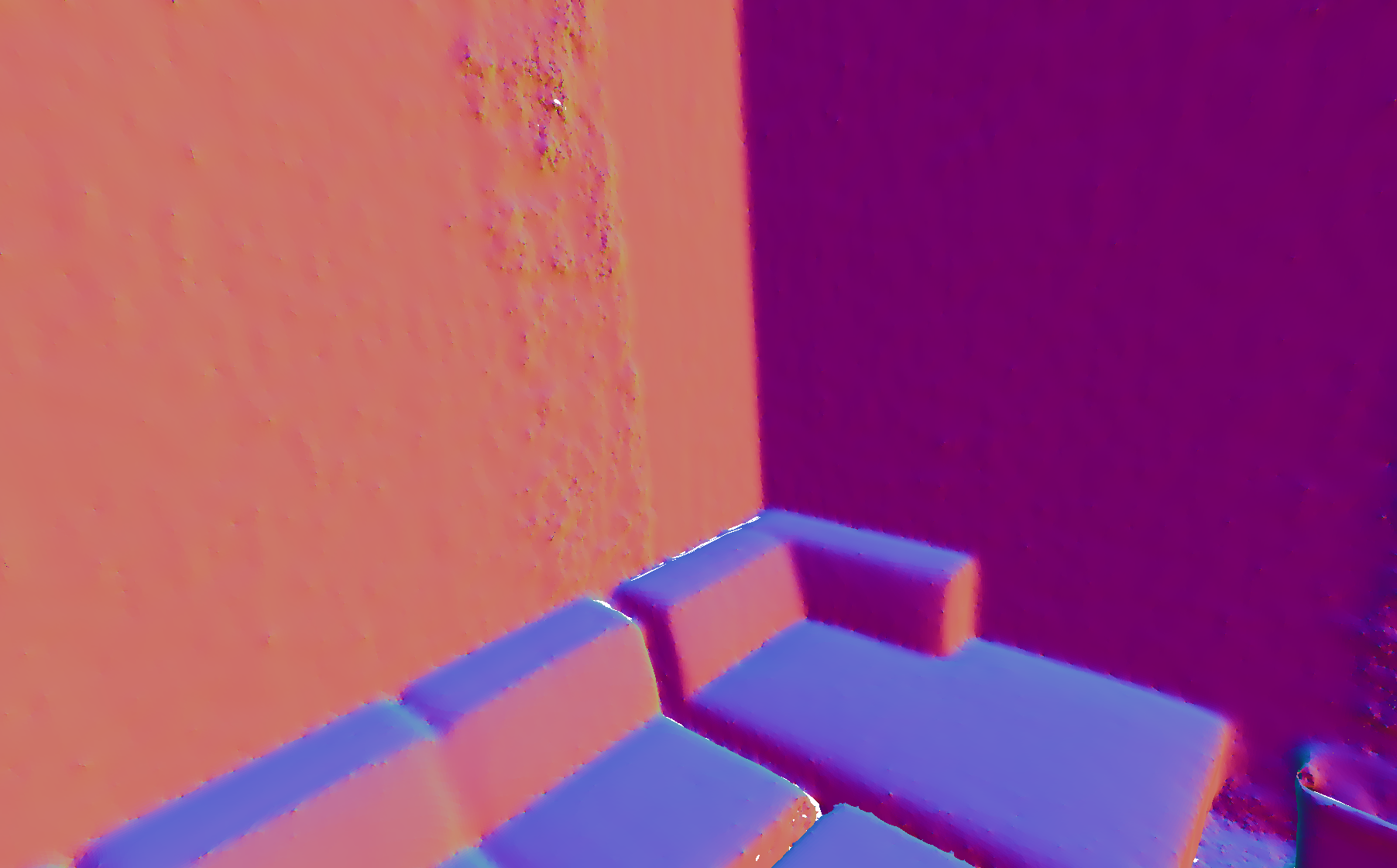} } & {\includegraphics[width=.24\columnwidth]{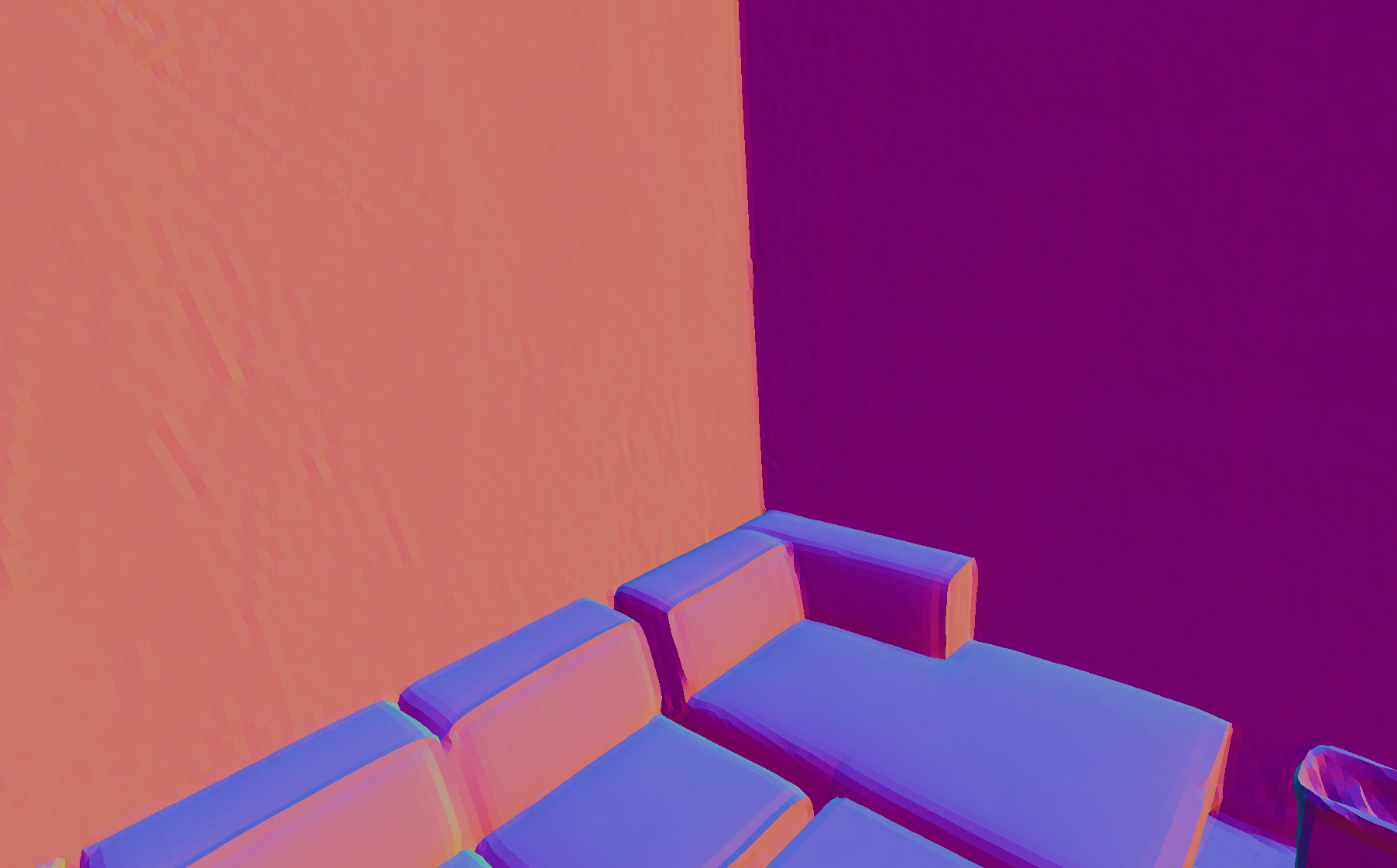}}  \\
            {\includegraphics[width=.24\columnwidth]{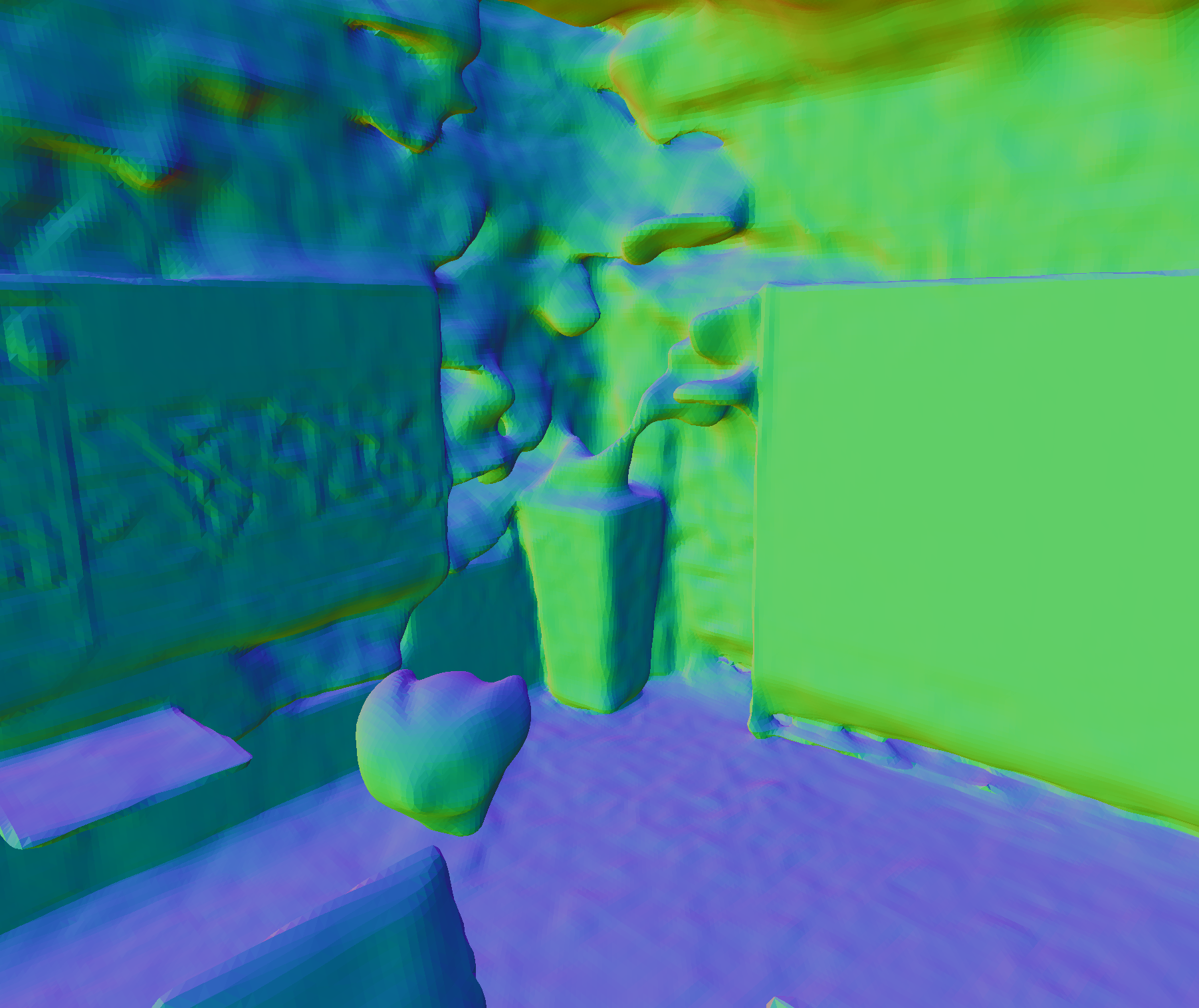} } & {\includegraphics[width=.24\columnwidth]{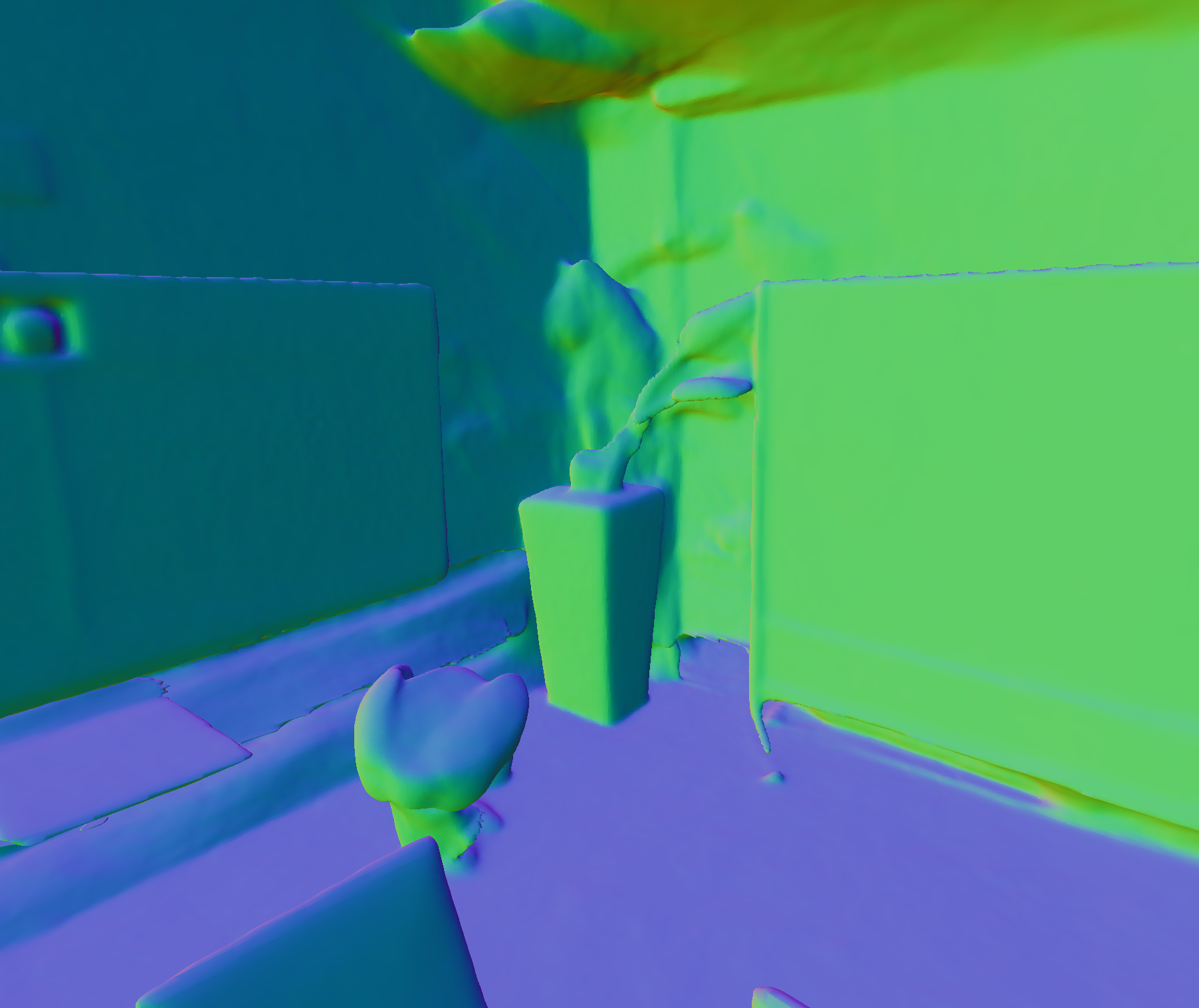} } & {\includegraphics[width=.24\columnwidth]{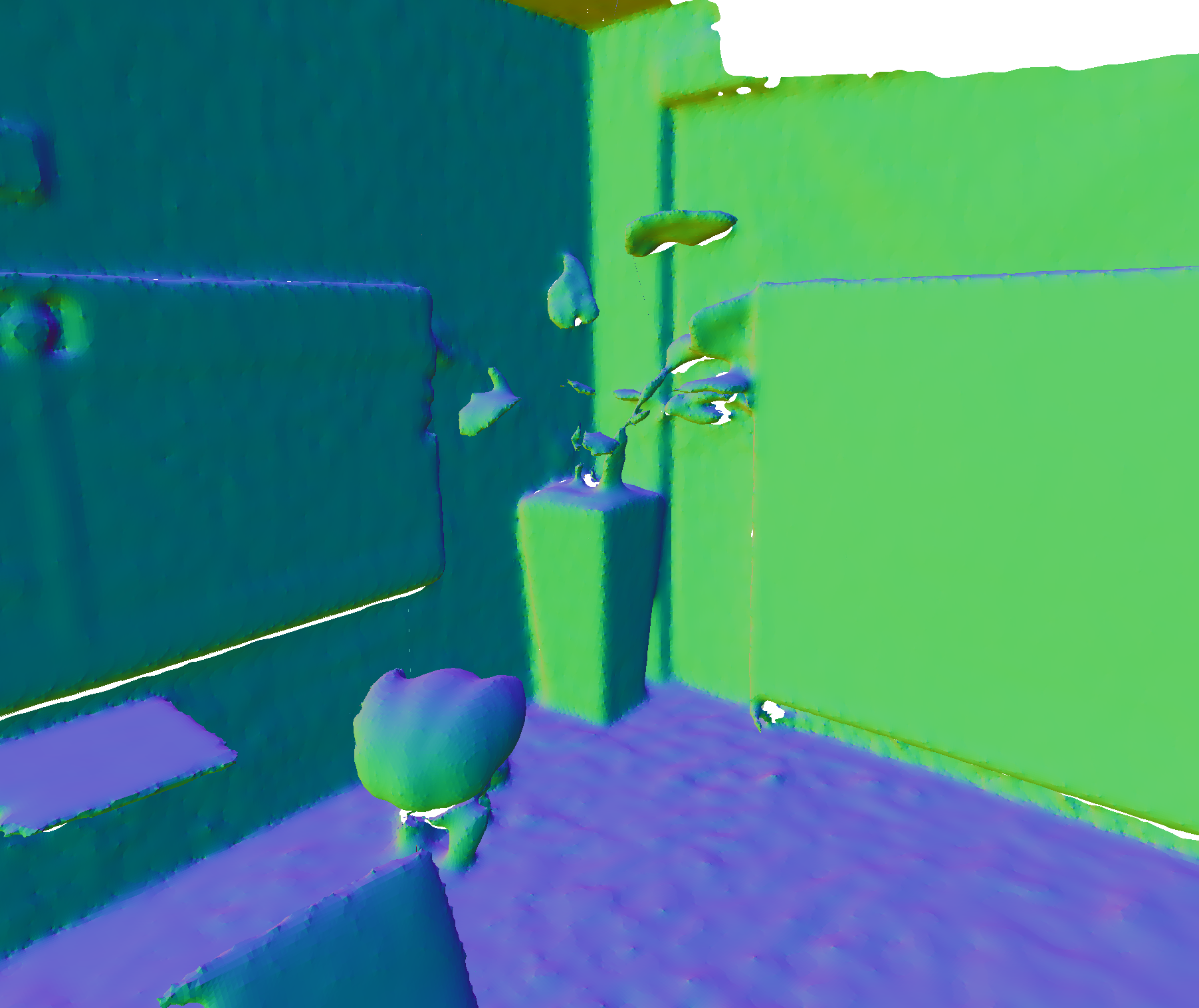} } & {\includegraphics[width=.24\columnwidth]{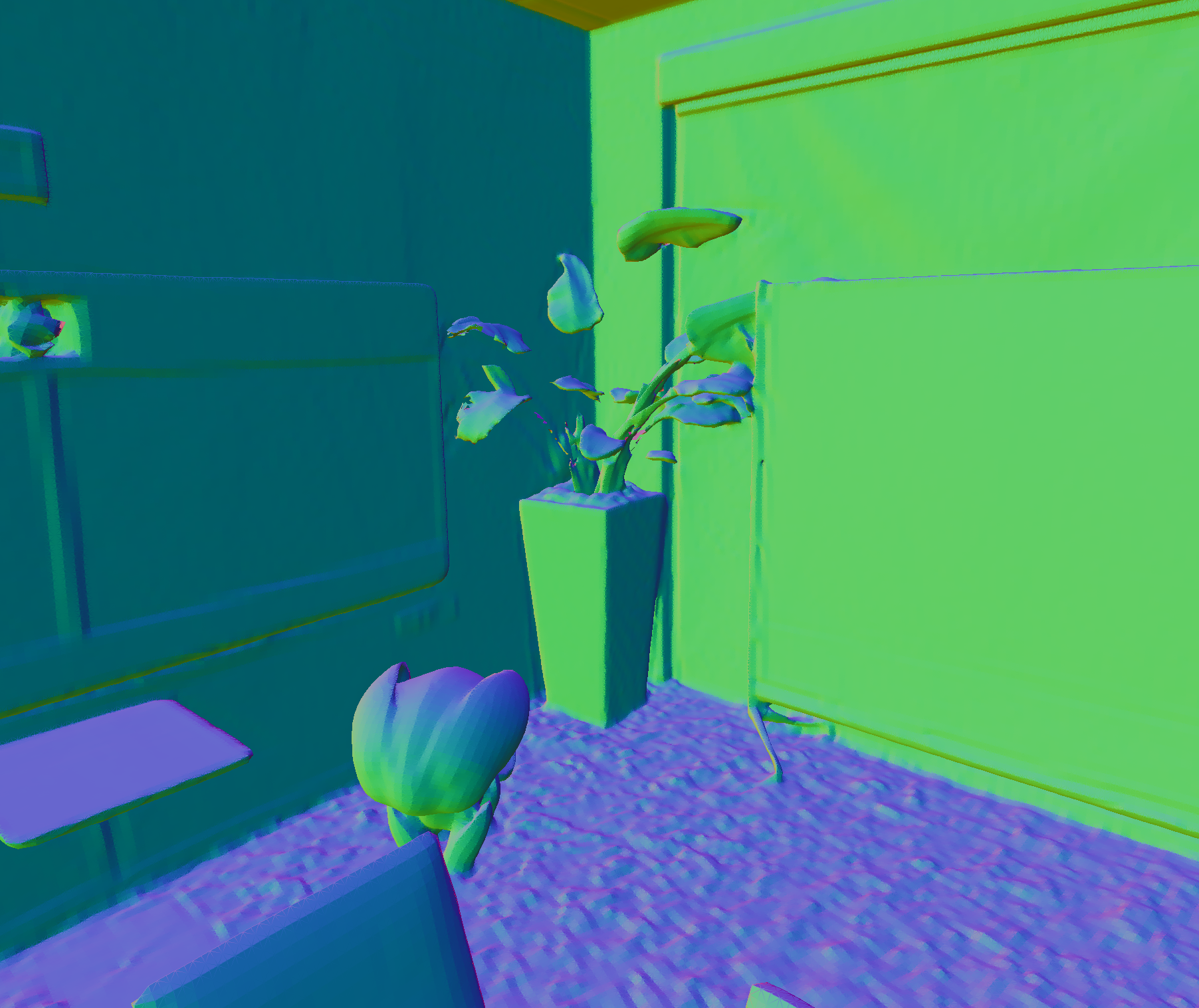}}  \\
            \textsc{\small{Manhattan}}\xspace & \monosdf~\cite{yu2022monosdf} & \textbf{\vfnerf} & Ground \\
            \textsc{\small{SDF}}\xspace~\cite{guo2022manhattan} & & \textbf{(Ours)} & Truth\\
        \end{tabular}
        \caption[3D reconstruction of planar regions]{\textbf{3D reconstruction of planar regions.} \vfnerf represents planar surfaces with higher accuracy and fewer artifacts. Besides planar regions, note that high-frequency details are still preserved in our method (see the plant in second-row meshes).}
        \label{fig:planar-comparisons}
        \vspace{-1.5em}
    \end{figure}

The performance of Neuralangelo drops due to the lack of texture in indoor scenes, since it only makes use of color images as supervision signal throughout the optimization. ManhattanSDF can generally recover high-quality scenes, although it struggles in some planar areas due to its dependency on semantic segmentation masks. MonoSDF generally achieves impressive results, although several artifacts appear in some planar regions. In contrast, our method can recover planar surfaces with great fidelity as well as many fine details (the flowers and picture on the wall or the trash bins in~\cref{fig:qualitative}). The capacity of our method to represent planar surfaces compared to the baselines is depicted in~\cref{fig:planar-comparisons}.

\mypar{Novel view synthesis}
We present qualitative and quantitative novel view synthesis comparisons in~\cref{fig:qualitative-syntesis} and \cref{tab:quantitative}. \vfnerf generally renders high-quality views that preserve high-frequency details and outperforms ManhattanSDF and Neuralangelo in terms of PSNR. Additionally, \vfnerf outperforms all baselines in ScanNet, a more realistic setup with real data, and comes second to MonoSDF in Replica by a small margin. More qualitative results can be found in the supplementary material.

\subsection{Ablations}

\mypar{Loss terms} We analyze the impact of different loss terms on the surface representation and provide quantitative results. Specifically, we study their effects of: center and exterior supervision $\loss_{center},\,\loss_{ext}$, unit norm $\loss_{norm}$ and depth $\loss_{depth}$. We demonstrate that removing these losses decreases the performance, as presented in~\cref{tab:ablations}. Additionally, removing the depth loss significantly decreases the performance, although the coarse geometry is still preserved, as demonstrated in~\cref{fig:loss-ablation} since most of the surfaces of the scene do not have enough texture.

\setlength{\tabcolsep}{1.5pt}
\begin{table}[t]
    \centering
    \label{tab:ablations}
    \begin{tabular}{l|ccccc|}
        \toprule
         & Precision$\uparrow$ & Recall$\uparrow$ & \textbf{F1-score}$\uparrow$ & \textbf{CD (mm)}$\downarrow$ \\
        \midrule
         w$\setminus$o annealing  & 0.964 & 0.809 & 0.880  & 0.10 \\
        w$\setminus$o initialization  & 0.901 & 0.793 & 0.844 & 0.13\\
        Uniform sampling & 0.942 & 0.818 & 0.876 & 0.14\\
        w$\setminus$o $\loss_{center}, \loss_{ext}$  & 0.952 & 0.804 & 0.872 & 0.11\\
        w$\setminus$o $\loss_{norm}$  & 0.928 & 0.801 & 0.860 & 0.11 \\
        w$\setminus$o $\loss_{depth}$  & 0.421 & 0.263 & 0.324 & 63.5 \\
        VF-NeRF  & \textbf{0.986} & \textbf{0.817} & \textbf{0.894} & \textbf{0.09} \\
        \bottomrule
    \end{tabular}
    \caption[Ablations quantitative results]{\textbf{Ablations quantitative results.} We showcase the importance of our loss terms, the sliding window annealing, the VF network initialization and the hierarchical sampling. }
    \vspace{-1.5em}
\end{table}

\mypar{Sliding window annealing and initialization} ~\cref{tab:ablations} presents the results of ablating the sliding window cosine similarity and the custom VF network initialization. In the case of the sliding window annealing, we use cosine similarity with the next point of the ray instead of the weighted average. The results show that both elements enhance the surface reconstruction.

\mypar{Sampling} We investigate the importance of our sampling strategy introduced in~\cref{sec:sampling}. We showcase the results achieved when using only uniform sampling in~\cref{tab:ablations}. We find that using a hierarchical sampling strategy enhances the performance of our method. Additionally, we find that removing the hierarchical sampling and just using uniform sampling generates many artifacts as depicted in~\cref{fig:loss-ablation}. 

\subsection{Limitations} 
One limitation of our method is its inherent smoothing bias, which can make it difficult to represent high-frequency details. Although our method is generally capable of representing details, it struggles in some cases. Additionally, VF-NeRF assumes homogeneous density with three hyperparameters. Future works could explore using different hyperparameters depending on the geometric characteristics. 

\section{Conclusion}

In this work, we presented VF-NeRF, a novel NERF approach for multiview surface reconstruction, utilizing Vector Fields (VFs) to encapsulate the scene's geometry. By transforming the VF, we can represent the volume density of each point in the scene. The key idea is to learn the VF of the scene through volume rendering. Additionally, we proposed a hierarchical sampling approach that enables us to sample more densely near surfaces, improving efficiency and precision. The experiments demonstrate the performance of our method to reconstruct indoor scenes, outperforming state-of-the-art methods on indoor datasets. Furthermore, our method is capable of rendering novel views that preserve high-frequency details and outperforms several baselines. Finally, we showcased the effectiveness of our method to reconstruct planar surfaces, while preserving details present in the scenes.

{
    \small
    \bibliographystyle{ieeenat_fullname}
    \bibliography{main}
}
\clearpage 
\maketitle
\appendix

\section{Detailed networks' architecture}
\begin{figure*}[th!]
    \centering
    \includegraphics[width=\textwidth]{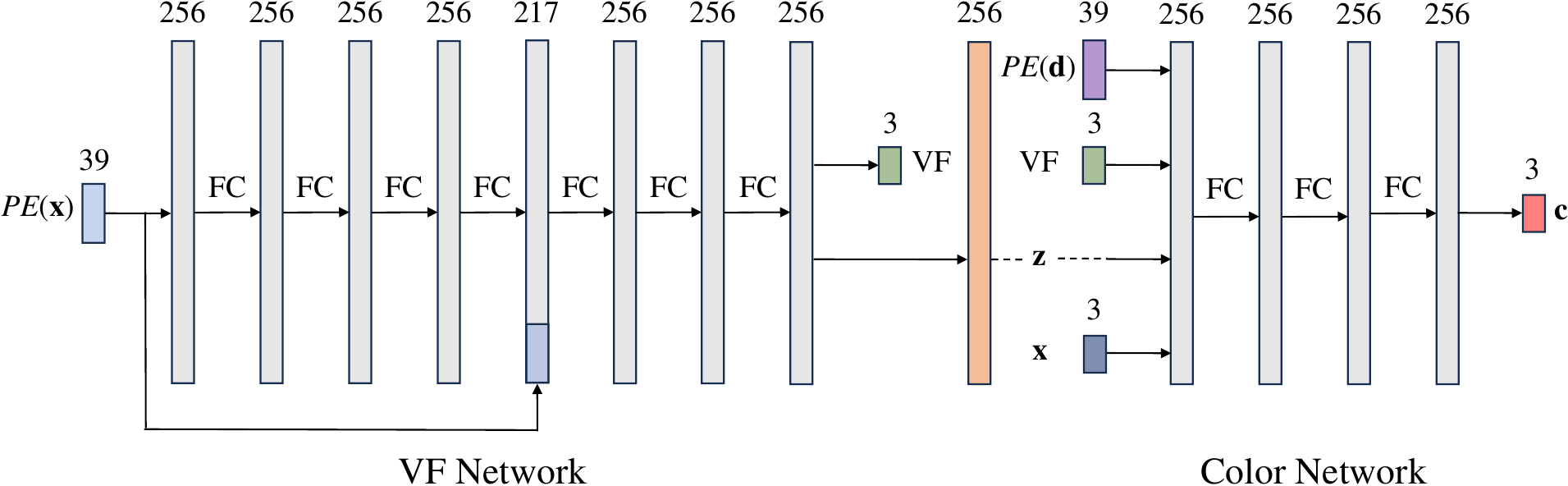}
    \caption{\textbf{Networks' architecture.} The VF network takes a point in space $\loc$ as input and applies positional encoding ($PE(\cdot)$) before feeding it to the MLP. The color network takes the spatial point, the predicted VF, the feature vector $\featVec$, and the viewing direction $\direction$ with positional encoding as inputs to predict the color.}
    \label{fig:nets}
\end{figure*}

We present an illustration of the VF and color networks in~\Cref{fig:nets}. As the first MLP, the VF network predicts the VF at a location in space $\loc$ and a feature vector $\featVec$ that is later used to predict the color at that point. The second smaller MLP is the color network. It takes as input the VF, the position in space $\loc$, the direction $\direction$\footnote{Positional encoding $PE(\cdot)$ is applied to it} and the predicted feature vector $\featVec$. As explained in the main paper, the two networks are applied in a rendering pipeline similar to \cite{mildenhall2020nerf} and optimized jointly.

\section{Metrics}
To evaluate the method, we use the standard metrics for the task \cite{yu2022monosdf, guo2022manhattan, yariv2021volsdf, li2023neuralangelo}. To evaluate the reconstruction accuracy, we use the precision, recall and F1-score together with the Chamfer Distance (CD). Given that the CD is sensitive to outliers, we report both its mean and median value. To evaluate the rendering capability of the network, we use the standard Peak Signal-to-Noise Ratio (PSNR).
The definitions of metrics are reported in \Cref{tab:3d-metrics-definitions}.

\setlength{\tabcolsep}{4pt}
\begin{table}[ht]
    \caption[Metric definitions]{\textbf{Metric definitions.} $P$ and $P^{\ast}$ are the point clouds sampled from the rendered and ground truth meshes. $M$ and $N$ are the height and weight of the images. $C$ and $\widehat{C}$ are the ground truth and rendered images.}
    \label{tab:3d-metrics-definitions}
    \begin{center}
        \begin{tabular}{c|c}
            \toprule
            \textbf{Metric} &\textbf{Definition}
            \\ \hline \\
            Precision &$\text{mean}_{p \in P}(\text{min}_{p^{\ast}\in P^{\ast}}||p-p^{\ast}|| < 0.05)$\\[1.5mm]
            Recall &$\text{mean}_{p^{\ast} \in P^{\ast}}(\text{min}_{p \in P}||p-p^{\ast}|| < 0.05)$\\[1.5mm]
            F1-score &$\dfrac{2 \cdot \text{Precision} \cdot \text{Recall}}{\text{Precision} + \text{Recall}}$\\[1.5mm]
            \multirow{2}{*}{CD} &$\sum_{p \in P} \text{min}_{p^{\ast} \in P^{\ast}}||p^{\ast} - p||_{2}^{2} + $\\
            & $\sum_{p^{\ast} \in P^{\ast}} \text{min}_{p \in P}||p^{\ast} - p||_{2}^{2}$ \\[1.5mm]
            MSE &$\frac{1}{MN}\sum_{i=0}^{M-1}\sum_{j=0}^{N-1}||C(i, j) - \widehat{C}(i, j)||_{2}^{2}$\\[1.5mm]
            PSNR &$-10 \cdot \log_{10}(\text{MSE})$ \\
            \bottomrule
        \end{tabular}
    \end{center}

    \end{table}

\section{Baselines}

We use the official Manhattan-SDF~\footnote{\url{https://github.com/zju3dv/manhattan\_sdf}}, MonoSDF~\footnote{\url{https://github.com/autonomousvision/monosdf}} and NeuRIS~\footnote{\url{https://github.com/jiepengwang/NeuRIS}} implementations as baselines. We adapted the Replica dataset to use it in Manhattan-SDF, while unfortunately NeuRIS does not support it. Additionally, we use SDFStudio's~\cite{Yu2022SDFStudio} implementation~\footnote{\url{https://github.com/autonomousvision/sdfstudio}} of Neuralangelo.

\section{3D reconstruction quantitative results}

We quantitatively evaluate the capacity of our method to reconstruct Replica and ScanNet scenes and compare it against state-of-the-art neural volume rendering methods. We present these quantitative results for each scene in~\Cref{tab:quantitative_app}. \vfnerf is among the 2 best methods on every Replica scene and the best method on every ScanNet scene in terms of F-score. Additionally, we observe that our method outperforms by a large margin all the others on the median CD and is among the best in mean CD. We note that median CD is more representative of the overall performance of the methods due to the high sensitivity to outliers of the mean CD. The large difference between mean CD and median CD (over 2 orders of magnitude in many cases) is caused by the overall poor performance of every method in representing the ceiling of the rooms. This is mainly due to the lack of observations that actually include the ceiling.

\begin{table*}[ht]
    \centering
\label{tab:quantitative_app}
    \begin{tabular}{l|ccccccc|c|cccc|c}
        \toprule
         & \multicolumn{8}{c|}{Replica} & \multicolumn{5}{c}{ScanNet}\\
         \cline{2-14}
         & r0 & r1 & r2 & o0 & o1 & o3 & o4  & Mean & 0050 & 0084 & 0580 & 0616  & Mean \\
        \cline{2-14}
         & \multicolumn{13}{c}{\textbf{Precision}$\uparrow$} \\
        \midrule
        M-SDF & 0.674 & 0.867 & 0.746 & 0.703 & 0.382 & 0.905 & 0.784 & 0.723 & 0.819 & 0.892 & 0.685 & 0.714 & 0.778 \\
        N-Angelo & 0.265 & 0.458 & 0.176 & 0.261 & 0.269 & 0.122 & 0.153 & 0.243 & 0.359 & - & 0.310 & 0.138 & 0.269 \\
        MonoSDF & \boldgreen{0.924} & \boldgreen{0.959} & \boldgreen{0.944} & \boldgreen{0.778} & \boldgreen{0.883} & \boldgreen{0.915} & \boldblue{0.941} & \boldgreen{0.906} & \boldgreen{0.857} & \boldblue{0.928} & \boldgreen{0.814} & \boldgreen{0.854} & \boldblue{0.863} \\
        NeuRIS & - & - & - & - & - & - & - & - & 0.822 & 0.776 & 0.740 & 0.755 & 0.773 \\
        \midrule
        \vfnerf & \boldblue{0.974} & \boldblue{0.988} & \boldblue{0.981} & \boldblue{0.986} & \boldblue{0.992} & \boldblue{0.973} & \boldgreen{0.940} & \boldblue{0.976} & \boldblue{0.942} & \boldgreen{0.927} & \boldblue{0.949} & \boldblue{0.893} & \boldblue{0.928}\\
        \midrule
        & \multicolumn{13}{c}{\textbf{Recall}$\uparrow$} \\
        \midrule
        M-SDF & \boldgreen{0.924} & \boldblue{0.926} & \boldgreen{0.854} & \boldgreen{0.819} & 0.691 & \boldblue{0.882} & \boldgreen{0.899} & \boldgreen{0.856} & 0.662 & 0.854 & \boldgreen{0.735} & 0.523 & 0.694 \\
        N-Angelo & 0.338 & 0.370 & 0.279 & 0.417 & 0.207 & 0.482 & 0.166 & 0.323 & 0.233 & - & 0.262 & 0.070 & 0.188 \\
        MonoSDF & \boldblue{0.964} & \boldgreen{0.912} & \boldblue{0.934} & 0.802 & \boldblue{0.829} & \boldgreen{0.878} & \boldblue{0.904} & \boldblue{0.889} & 0.660 & \boldgreen{0.896} & 0.725 & \boldgreen{0.637} & \boldgreen{0.730} \\
        NeuRIS & - & - & - & - & - & - & - & - &  \boldgreen{0.699} & 0.741 & 0.719 & 0.568 & 0.682 \\
        \midrule
        \vfnerf & 0.873 & 0.875 & 0.848 & \boldgreen{0.817} & \boldgreen{0.784} & 0.825 & 0.872 & 0.842 & \boldblue{0.779} & \boldblue{0.920} & \boldblue{0.888} & \boldblue{0.696} & \boldblue{0.821} \\
        \midrule
        & \multicolumn{13}{c}{\textbf{F-score}$\uparrow$} \\
        \midrule
        M-SDF & 0.778 & 0.896 & 0.796 & 0.757 & 0.492 & 0.893 & 0.838 & 0.779 & 0.732 & 0.873 & 0.709 & 0.604 & 0.730 \\
        N-Angelo & 0.297 & 0.410 & 0.216 & 0.321 & 0.234 & 0.195 & 0.159 & 0.262 & 0.283 & - & 0.284 & 0.093 & 0.220 \\
        MonoSDF & \boldblue{0.944} & \boldblue{0.935} & \boldblue{0.939} & \boldgreen{0.790} & \boldgreen{0.855} & \boldblue{0.896} & \boldblue{0.922} & \boldgreen{0.897} & 0.745 & \boldgreen{0.911} & \boldgreen{0.767} & \boldgreen{0.730} & \boldgreen{0.788} \\
        NeuRIS & - & - & - & - & - & - & - & - & \boldgreen{0.755} & 0.758 & 0.729 & 0.648 & 0.723 \\
        \midrule
        \vfnerf & \boldgreen{0.921} & \boldgreen{0.928} & \boldgreen{0.910} & \boldblue{0.894} & \boldblue{0.876} & \boldgreen{0.893} & \boldgreen{0.905} & \boldblue{0.904} & \boldblue{0.853} & \boldblue{0.923} & \boldblue{0.918} & \boldblue{0.765} & \boldblue{0.865}\\
        \midrule
        & \multicolumn{13}{c}{\textbf{Mean Chamfer Distance (mm)}$\downarrow$} \\
        \midrule
        M-SDF & 494 & 65.2 & 392 & 77.5 & 1266 & 7.68 & 149 & 350 & \boldgreen{11.0} & 9.18 & 23.1 & 47.9 & 22.80 \\
        N-Angelo & 1113 & 67.9 & 1107 & 317 & 280 & 5464 & 1002 & 1336 & 95.6 & - & 196 & 523 & 272 \\
        MonoSDF & \boldblue{2.72} & \boldblue{3.44} & \boldblue{2.95} & \boldblue{9.23} & \boldblue{12.0} & \boldblue{4.50} & \boldblue{2.49} & \boldblue{5.33} & 12.1 & \boldgreen{5.80} & \boldgreen{12.6} & \boldgreen{40.8} & \boldblue{17.83} \\
        NeuRIS & - & - & - & - & - & - & - & - &  11.3 & 10.6 & 24.6 & \boldblue{34.3} & 20.2 \\
        \midrule
        \vfnerf & \boldgreen{8.59} & \boldgreen{6.84} & \boldgreen{15.2} & \boldgreen{44.2} & \boldgreen{57.0} & \boldgreen{7.34} & \boldgreen{6.50} & \boldgreen{20.81} & \boldblue{8.37} & \boldblue{5.14} & \boldblue{7.34} & 55.2 & \boldgreen{19.0} \\
        \midrule
         & \multicolumn{13}{c}{\textbf{Median Chamfer Distance (mm)}$\downarrow$} \\
        \midrule
        M-SDF & 0.87 & 0.34 & 0.91 & \boldgreen{0.42} & 35.6 & 0.41 & 0.63 & 5.60 & 1.25 & 1.09 & 1.76 & 2.71 & 1.45 \\
        N-Angelo & 39.0 & 12.5 & 215 & 39.0 & 44.5 & 3754 & 174 & 611 & 28.8 & - & 29.2 & 251 & 103 \\
        MonoSDF & \boldgreen{0.21} & \boldgreen{0.23} & \boldgreen{0.46} & 0.82 & \boldgreen{0.34} & \boldgreen{0.33} & \boldgreen{0.22} & \boldgreen{0.37} & 2.13 & \boldgreen{0.90} & \boldgreen{1.45} & \boldgreen{1.19} & \boldgreen{1.42} \\
        NeuRIS & - & - & - & - & - & - & - & - &  \boldgreen{0.57} & 2.35 & 1.48 & 2.42 & 1.71 \\
        \midrule
        \vfnerf & \boldblue{0.19} & \boldblue{0.09} & \boldblue{0.15} & \boldblue{0.09} & \boldblue{0.05} & \boldblue{0.21} & \boldblue{0.14} & \boldblue{0.13} & \boldblue{0.32} & \boldblue{0.07} & \boldblue{0.18} & \boldblue{0.46} & \boldblue{0.258} \\
        \bottomrule
    \end{tabular}
            \caption[3D reconstruction quantitative results]{\textbf{3D reconstruction quantitative results of individual scenes on Replica and ScanNet.} \boldblue{Best result}. \boldgreen{Second best result}. Note Neuralangelo fails to reconstruct a valid geometry for scene 0580 of ScanNet.}
\end{table*}

\begin{table*}[ht!]
    \centering
    \begin{tabular}{l|ccccccc|c|cccc|c}
        \toprule
         & \multicolumn{13}{c}{PSNR$\uparrow$} \\
         \cline{2-14}
         & \multicolumn{8}{c|}{Replica} & \multicolumn{5}{c}{ScanNet}\\
         \cline{2-14}
         & r0 & r1 & r2 & o0 & o1 & o3 & o4  & Mean & 0050 & 0084 & 0580 & 0616  & Mean \\
        \midrule
        M-SDF & 25.06 & 26.38 & 29.36 & 28.87 & 26.39 & 28.35 & 27.92 & 27.48 & 22.44 & 18.92 & 22.87 & 18.90 & 20.78 \\
        N-Angelo & \boldblue{28.22} & \boldgreen{30.45} & \boldgreen{29.59} & 36.02 & 36.15 & \boldgreen{29.54} & \boldgreen{30.14} & 31.44 & 17.48 & 18.66 & 18.40& 16.78 & 17.83\\
        MonoSDF & \boldgreen{27.91} & 30.29 & \boldblue{31.16} & \boldgreen{36.26} & \boldgreen{36.80} & \boldblue{30.70} & \boldblue{32.63} & \boldblue{32.25} & 17.61 & \boldblue{33.11} & \boldblue{27.16} & 17.46 & 23.84 \\
        NeuRIS & - & - & - & - & - & - & - & - & \boldgreen{23.29} & 27.63 & 23.56 & \boldgreen{23.14} & \boldgreen{24.41} \\
        \midrule
        VF-NeRF & 27.05 & \boldblue{31.23} & 28.90 & \boldblue{36.72} & \boldblue{38.09} & 27.57 & 29.89 & \boldgreen{31.49} & \boldblue{25.60} & \boldgreen{28.43} & \boldgreen{25.64} & \boldblue{25.20} & \boldblue{26.21} \\
        \bottomrule
    \end{tabular}
     \caption[Novel view synthesis quantitative results]{\textbf{Novel view synthesis quantitative results.} \boldblue{Best result}. \boldgreen{Second best result}.}
    \label{tab:psnr}
\end{table*}

\section{3D reconstruction qualitative results}

\begin{figure*}[ht]
        \centering
        \begin{tabular}{cccccc}
             \rotatebox[origin=b]{90}{ \texttt{Room 1}} & \raisebox{-0.5\height}{\includegraphics[width=.19\textwidth]{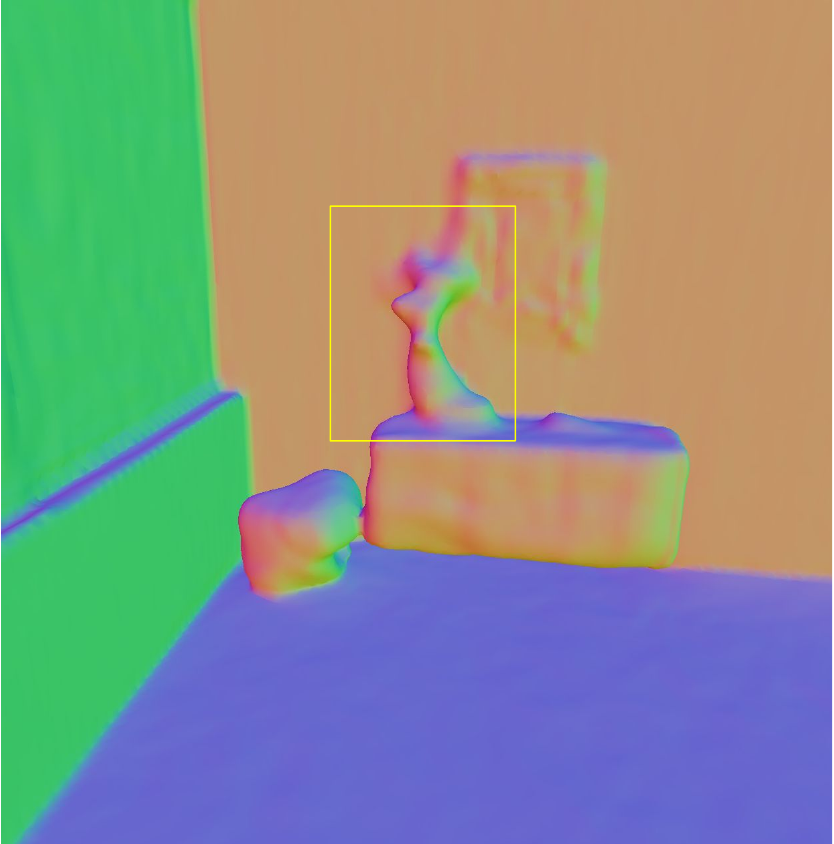} } & \raisebox{-0.5\height}{\includegraphics[width=.19\textwidth]{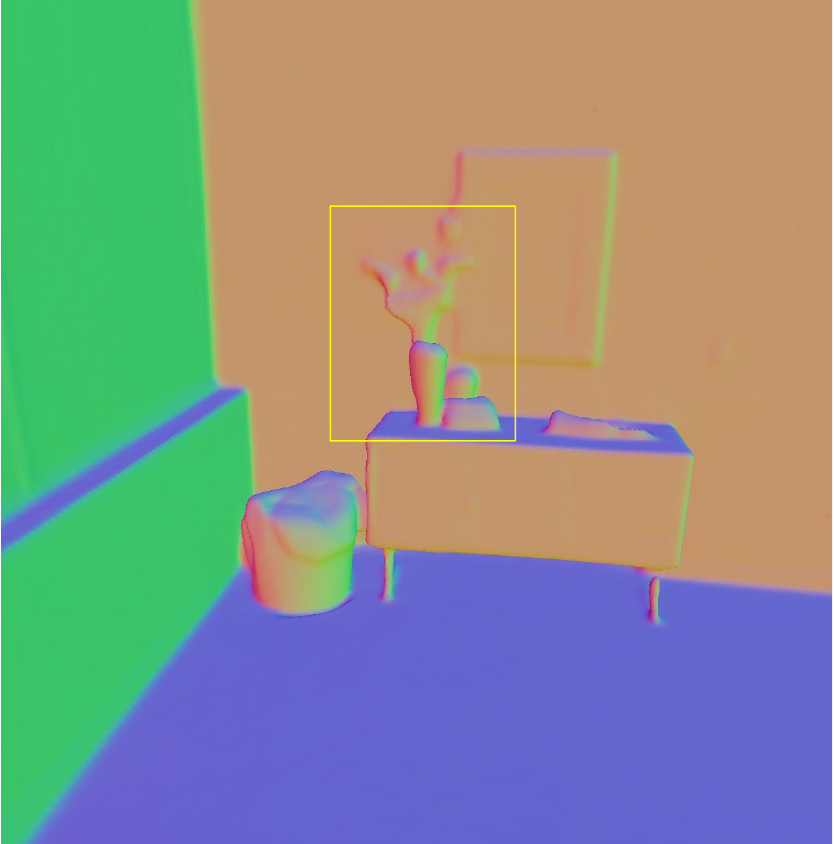} }
             &\raisebox{-0.5\height}{\includegraphics[width=.19\textwidth]{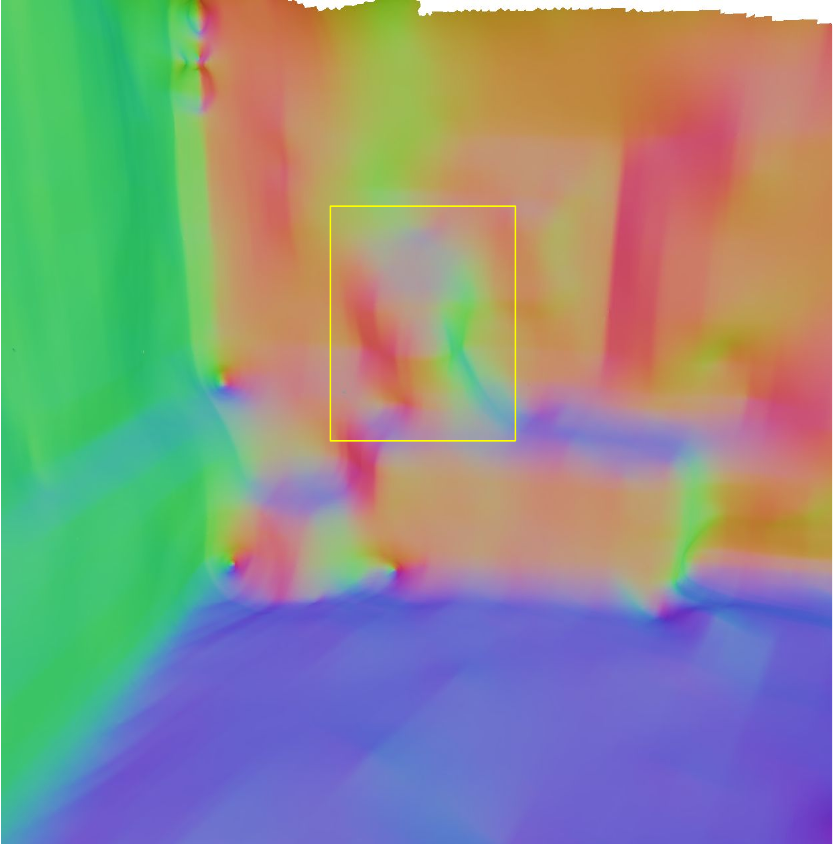} } & \raisebox{-0.5\height}{\includegraphics[width=.19\textwidth]{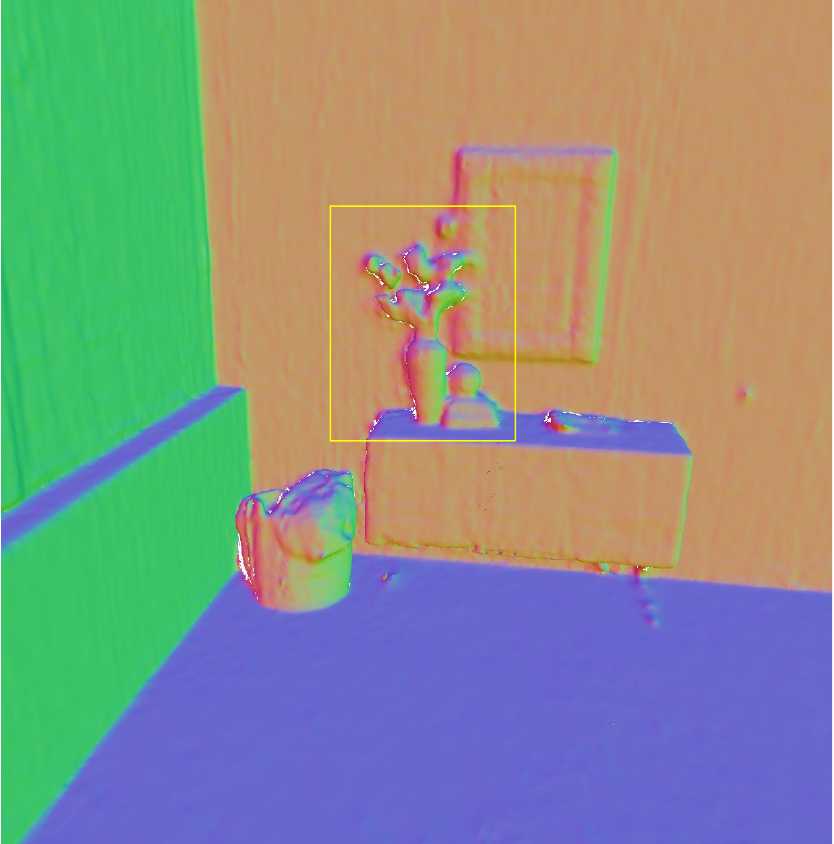} } & \raisebox{-0.5\height}{\includegraphics[width=.19\textwidth]{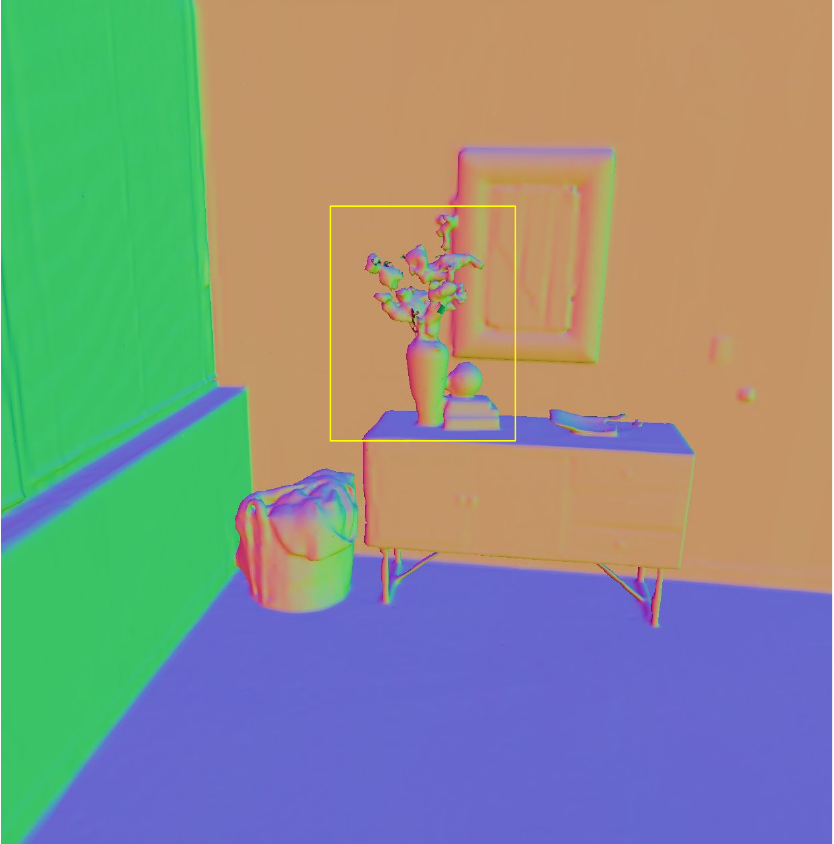} } \\
             \rotatebox[origin=b]{90}{ \texttt{Office 1}} & \raisebox{-0.5\height}{\includegraphics[width=.19\textwidth]{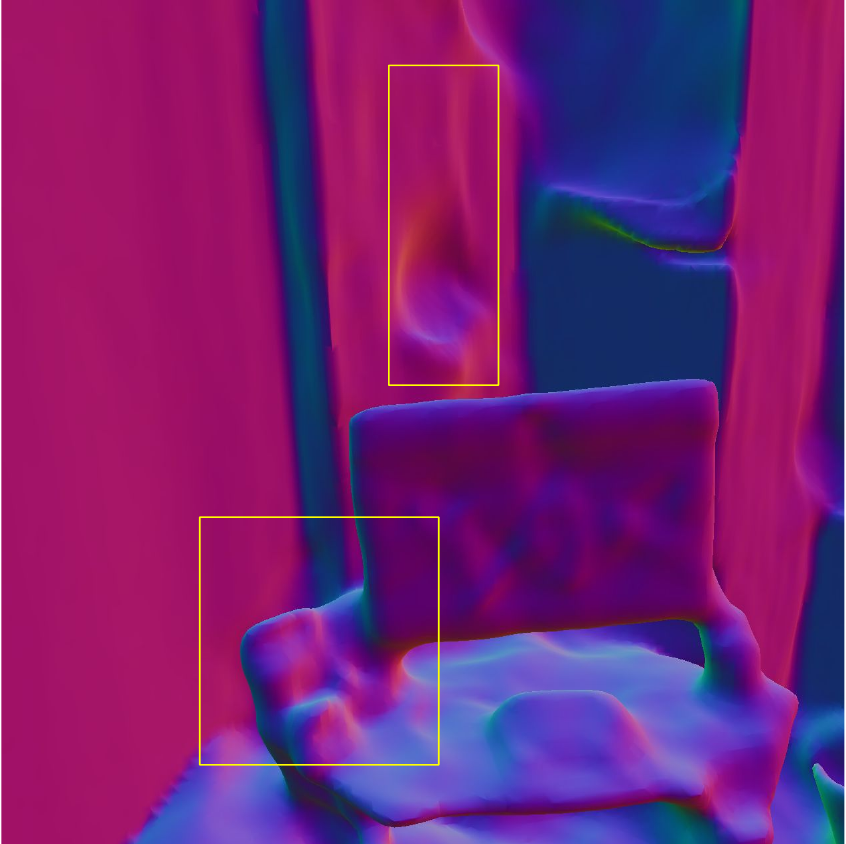} } & \raisebox{-0.5\height}{\includegraphics[width=.19\textwidth]{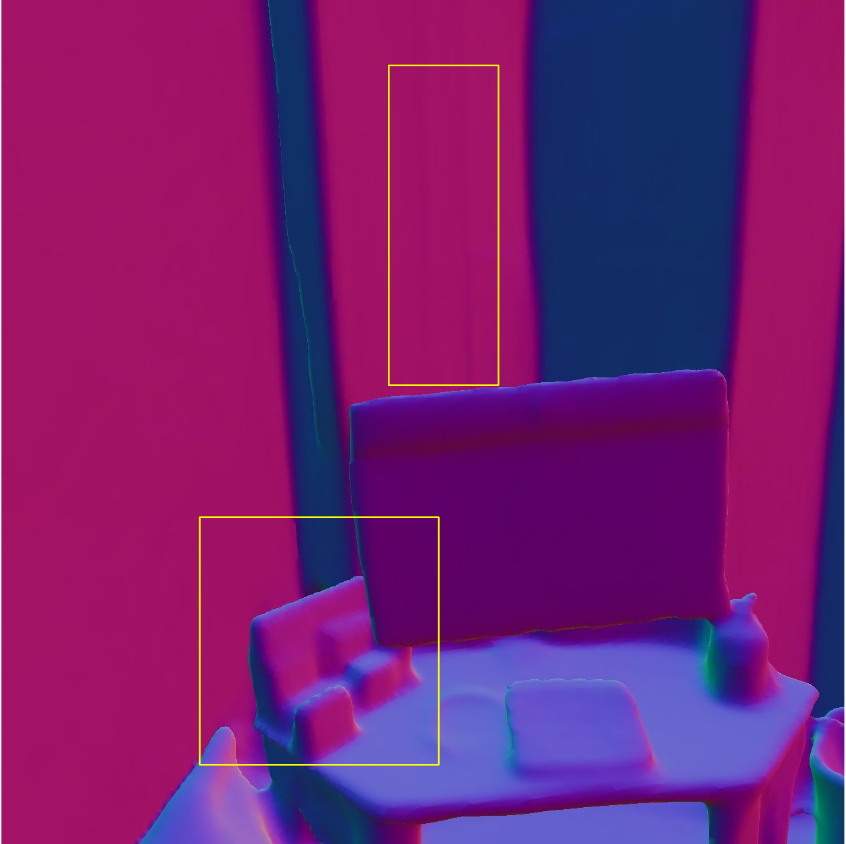} }
             &\raisebox{-0.5\height}{\includegraphics[width=.19\textwidth]{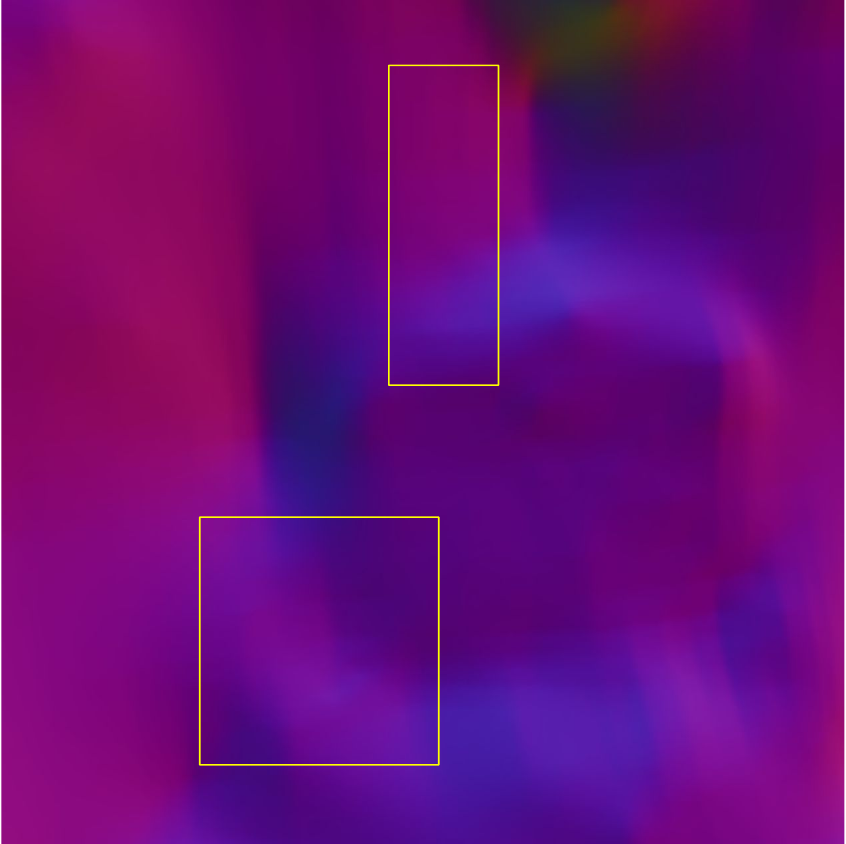} } & \raisebox{-0.5\height}{\includegraphics[width=.19\textwidth]{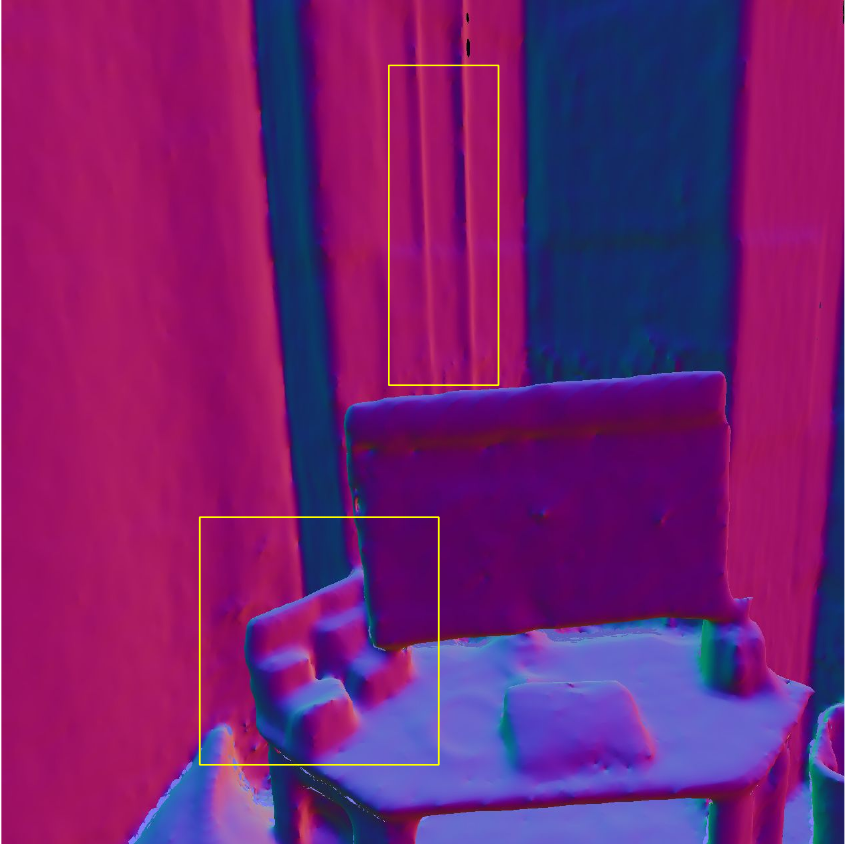} } & \raisebox{-0.5\height}{\includegraphics[width=.19\textwidth]{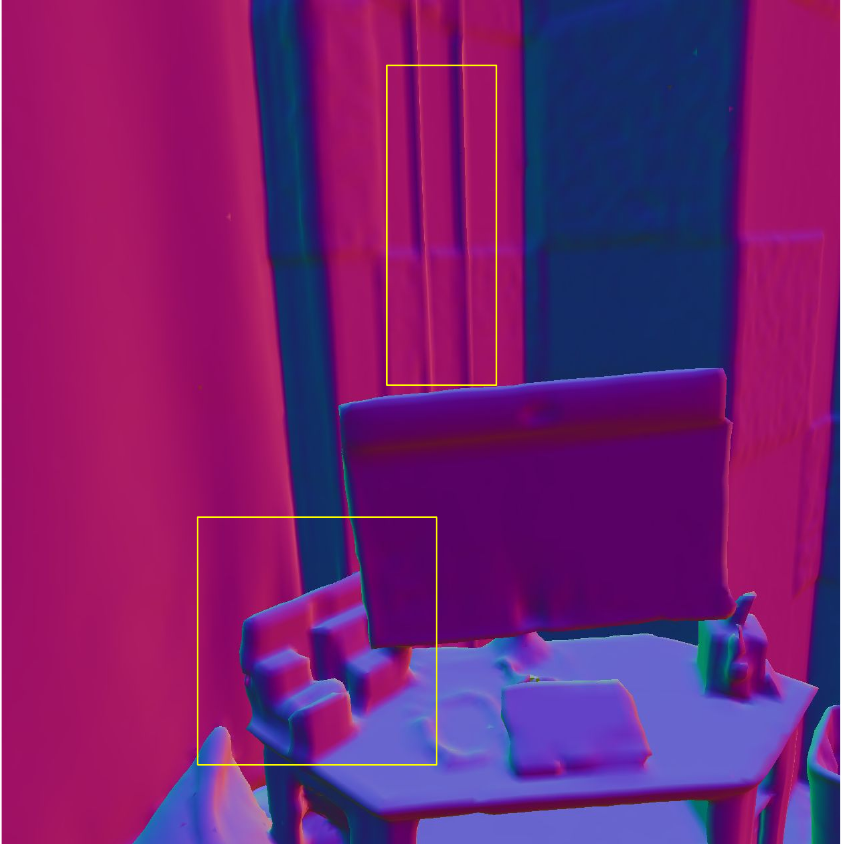} } \\
             \rotatebox[origin=b]{90}{ \texttt{Office 3}} & \raisebox{-0.5\height}{\includegraphics[width=.19\textwidth]{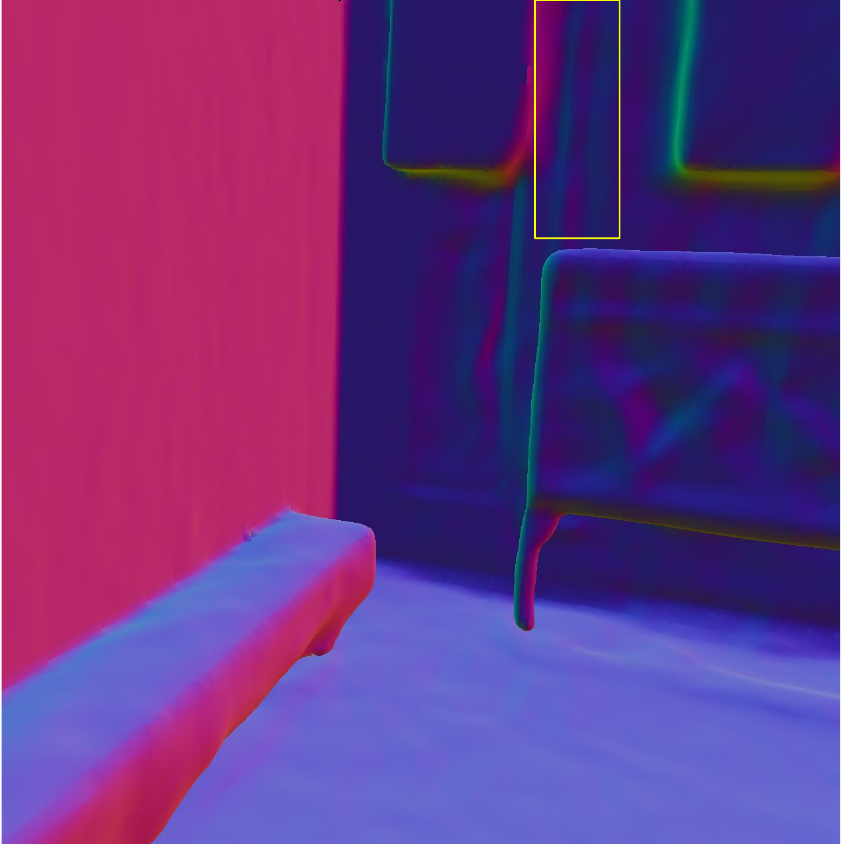} } & \raisebox{-0.5\height}{\includegraphics[width=.19\textwidth]{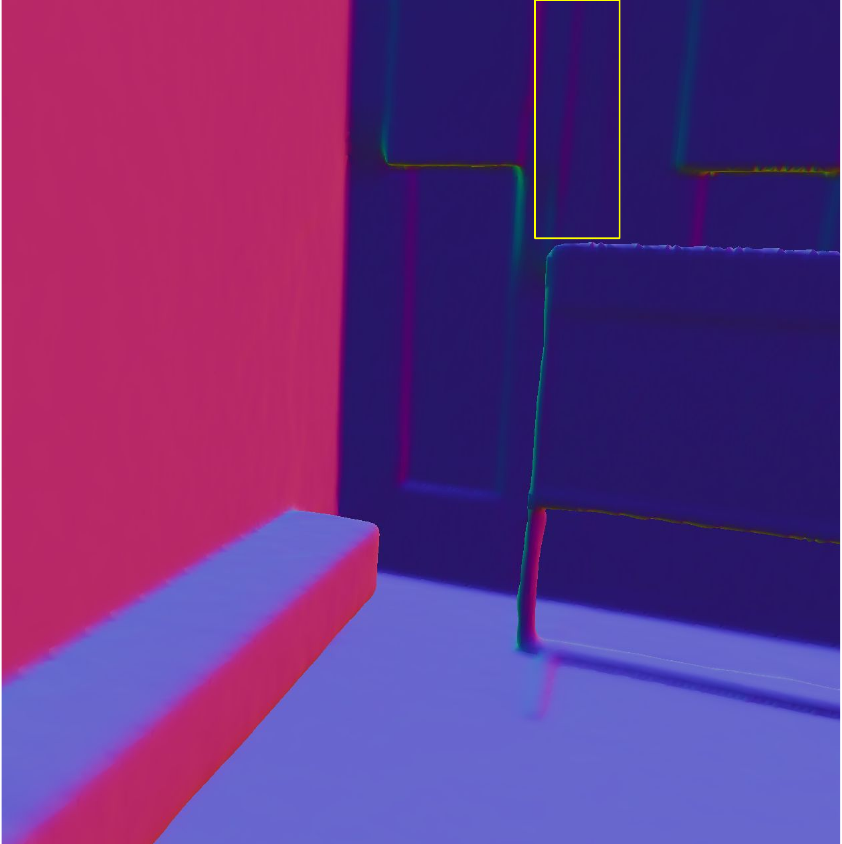} }
             &\raisebox{-0.5\height}{\includegraphics[width=.19\textwidth]{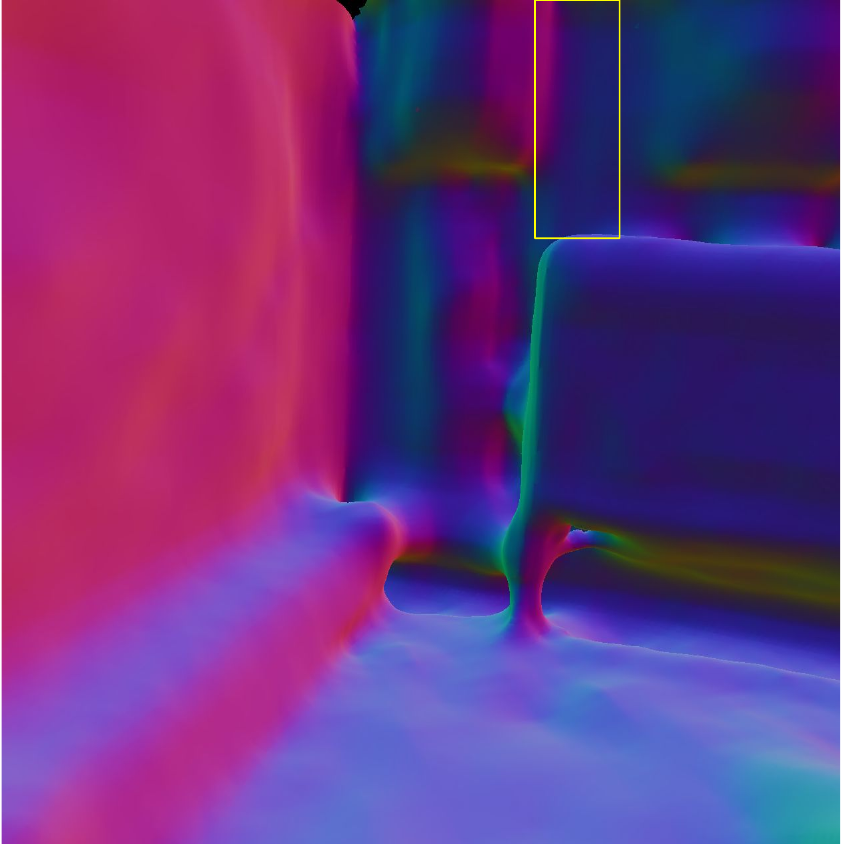} } & \raisebox{-0.5\height}{\includegraphics[width=.19\textwidth]{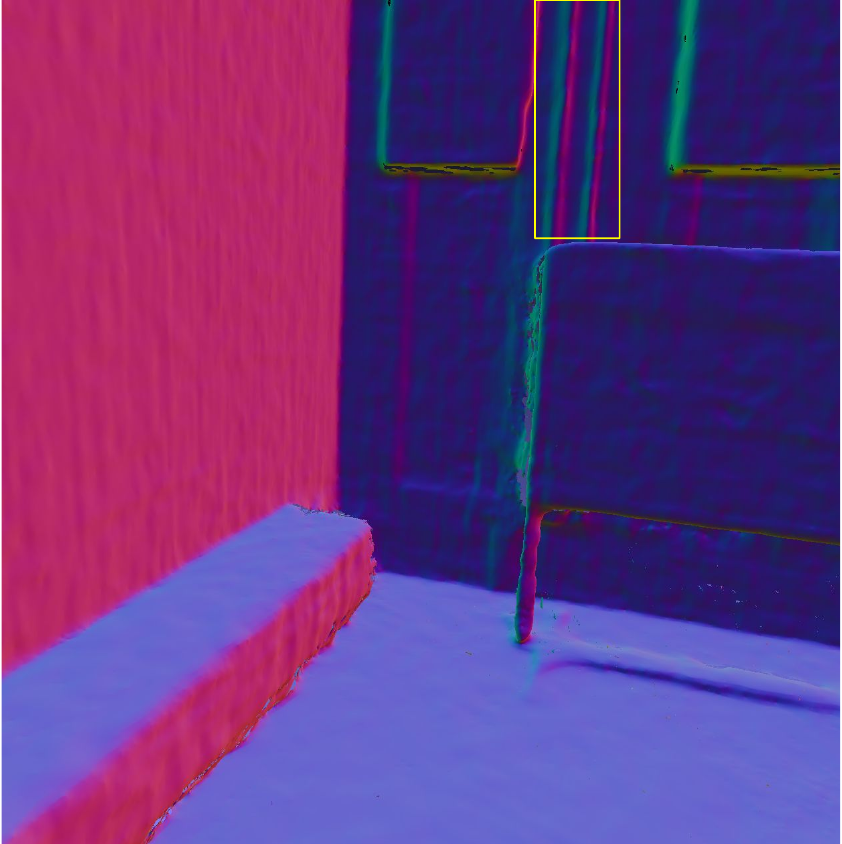} } & \raisebox{-0.5\height}{\includegraphics[width=.19\textwidth]{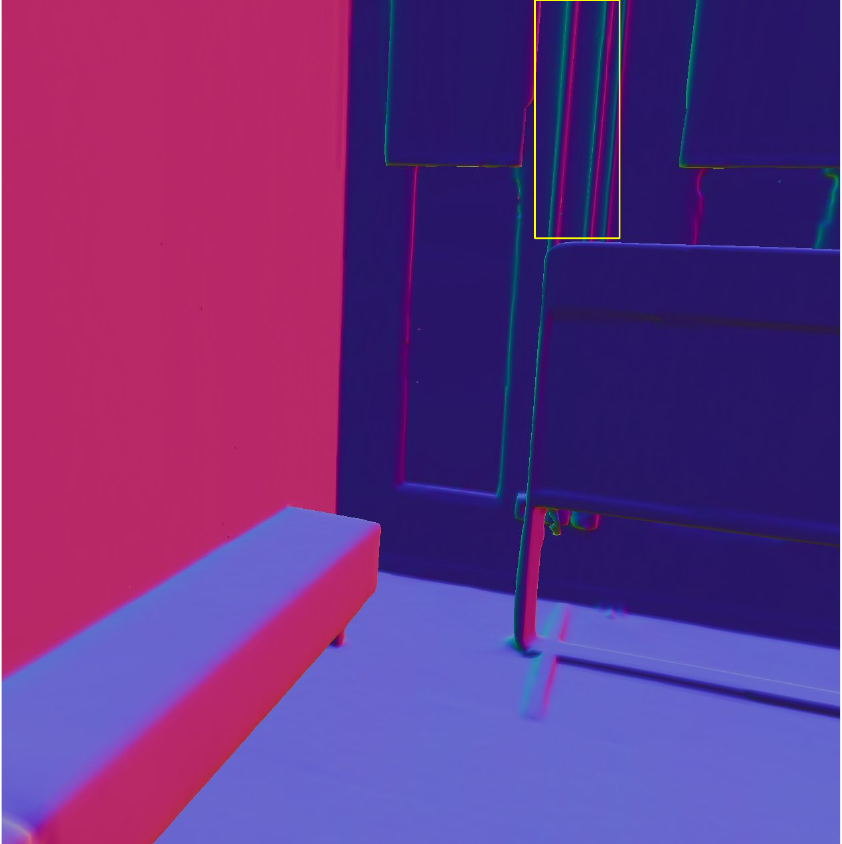} } \\
             \rotatebox[origin=b]{90}{ \texttt{Office 4}} & \raisebox{-0.5\height}{\includegraphics[width=.19\textwidth]{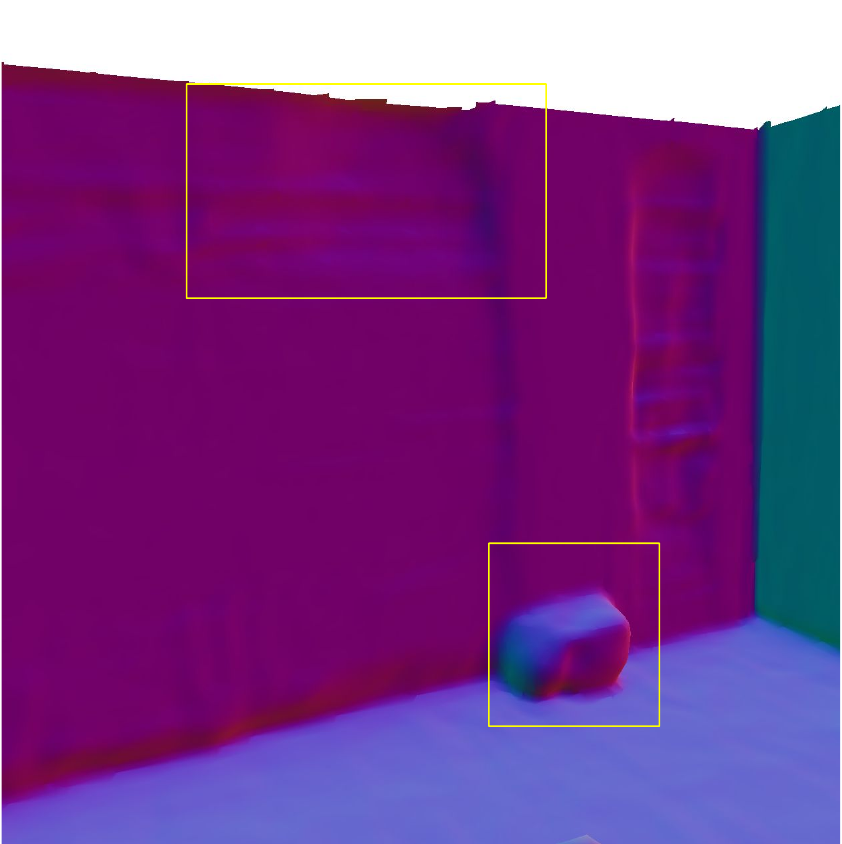} } & \raisebox{-0.5\height}{\includegraphics[width=.19\textwidth]{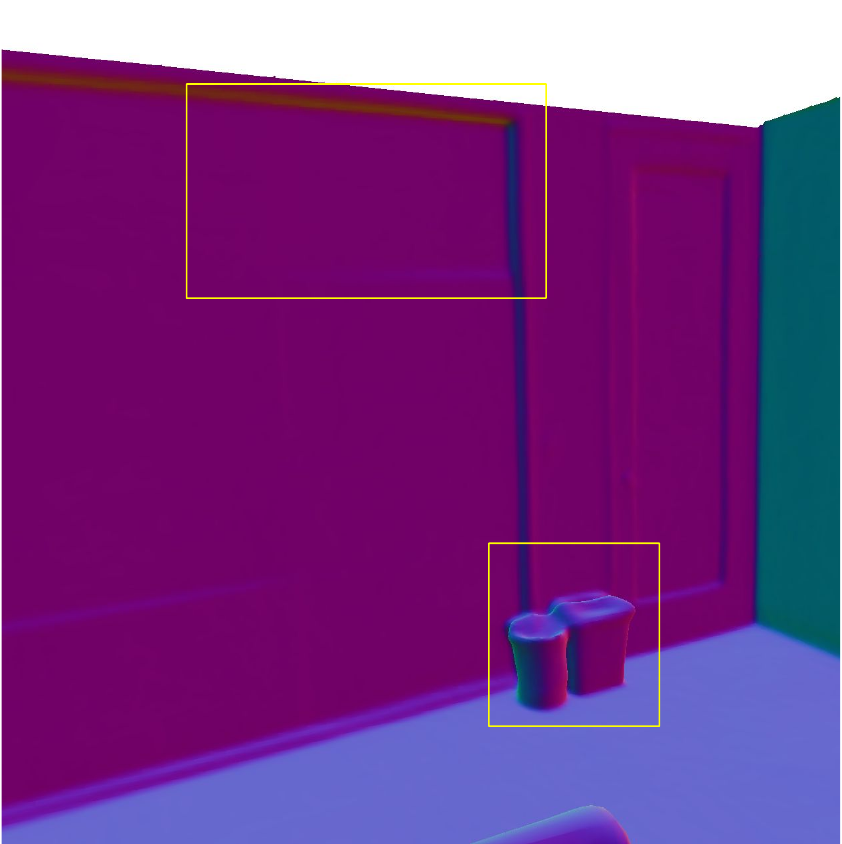} }
             &\raisebox{-0.5\height}{\includegraphics[width=.19\textwidth]{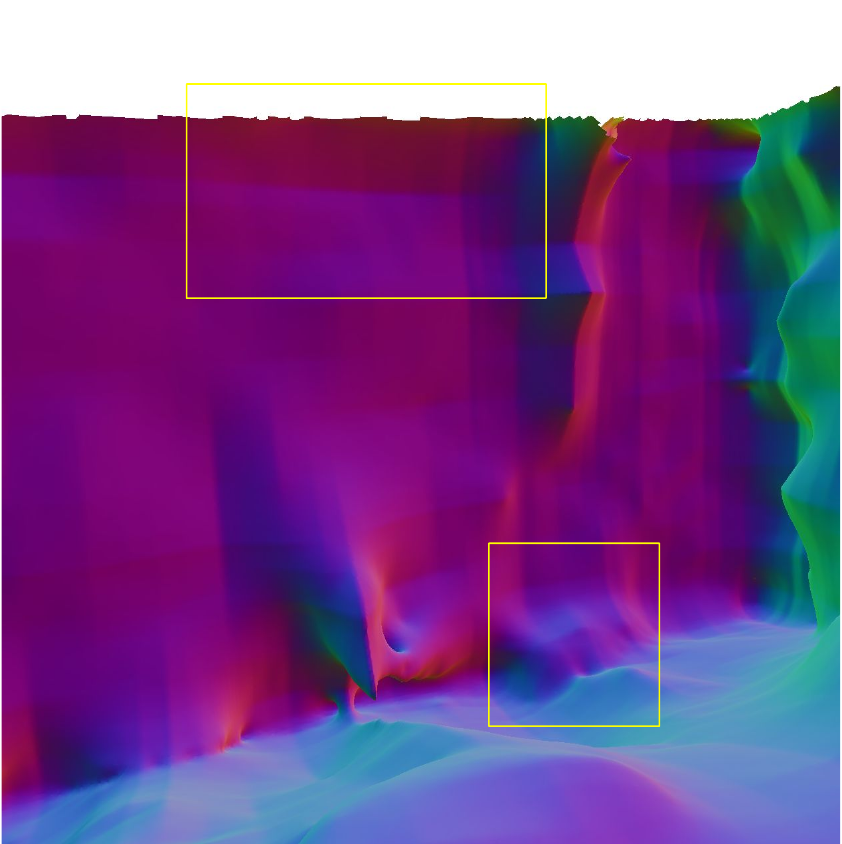} } & \raisebox{-0.5\height}{\includegraphics[width=.19\textwidth]{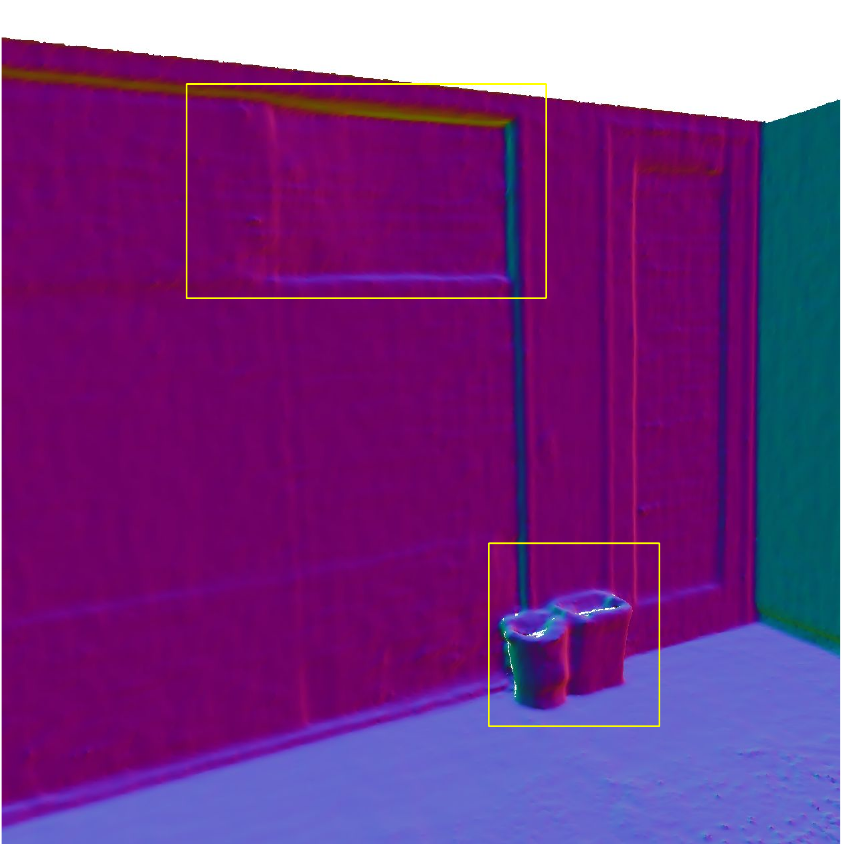} } & \raisebox{-0.5\height}{\includegraphics[width=.19\textwidth]{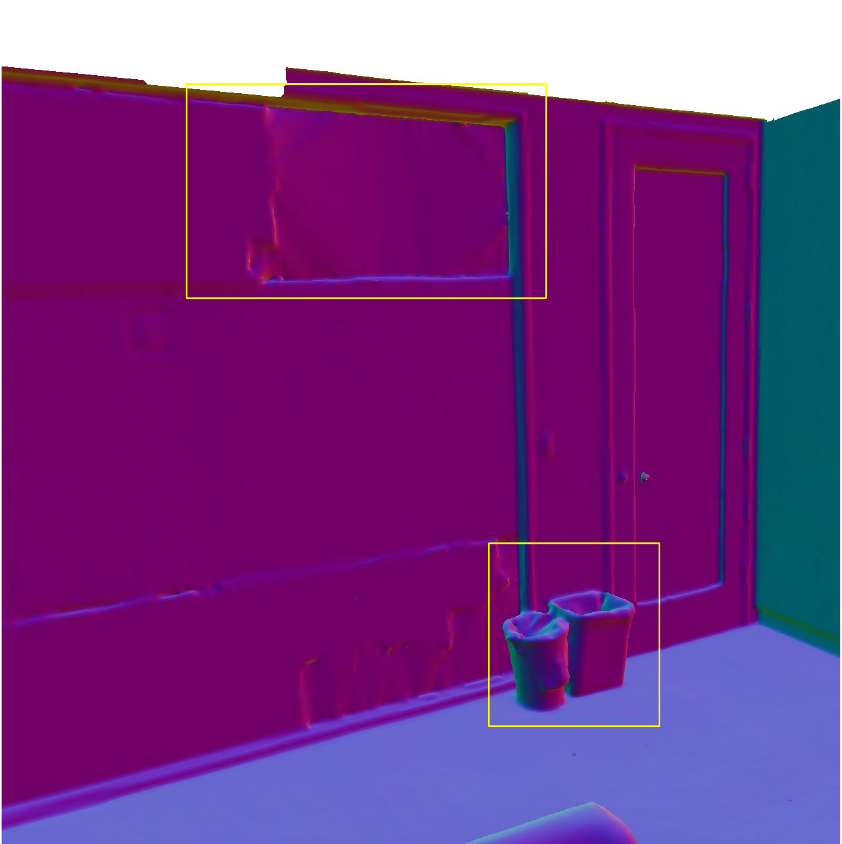} } \\
             & \textbf{Manhattan-} & \textbf{MonoSDF} & \textbf{Neuralangelo} & \textbf{VF-NeRF} & \textbf{Ground} \\
              & \textbf{SDF} &  &  & \textbf{(Ours)} & \textbf{truth} \\
            \rotatebox[origin=b]{90}{ \texttt{0050}} & \raisebox{-0.5\height}{\includegraphics[width=.19\textwidth]{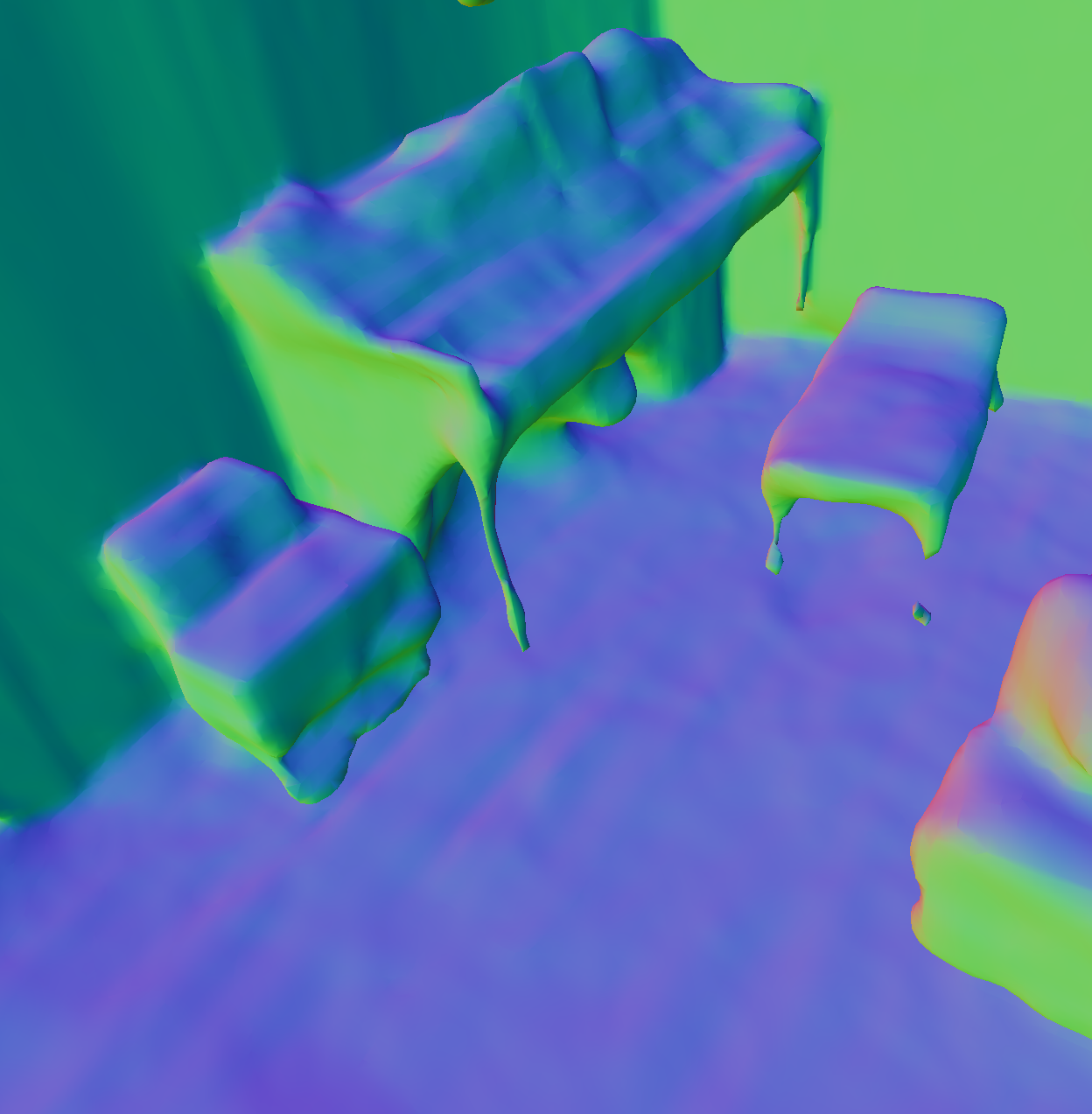} } & \raisebox{-0.5\height}{\includegraphics[width=.19\textwidth]{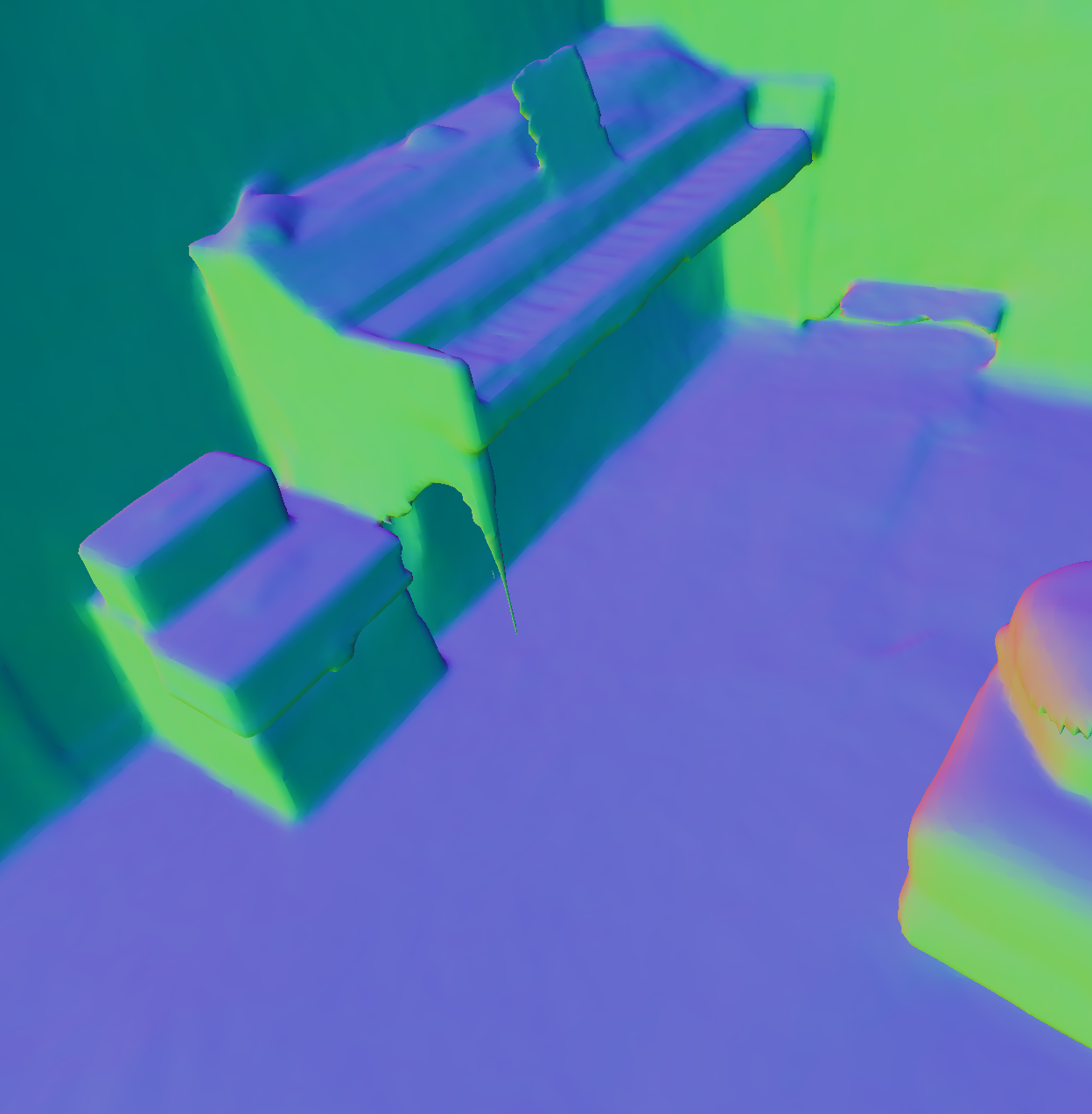} }
             &\raisebox{-0.5\height}{\includegraphics[width=.19\textwidth]{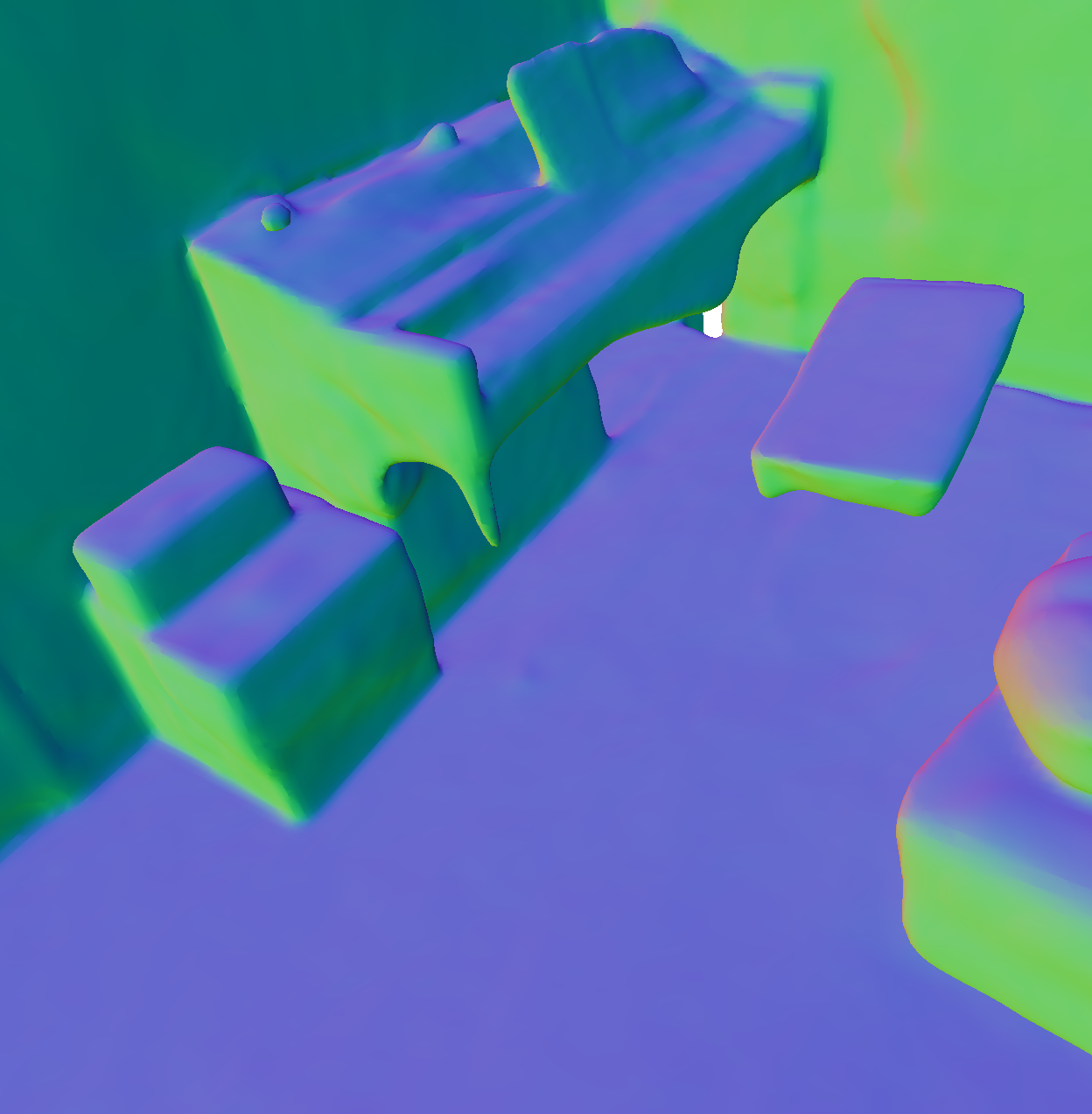} } & \raisebox{-0.5\height}{\includegraphics[width=.19\textwidth]{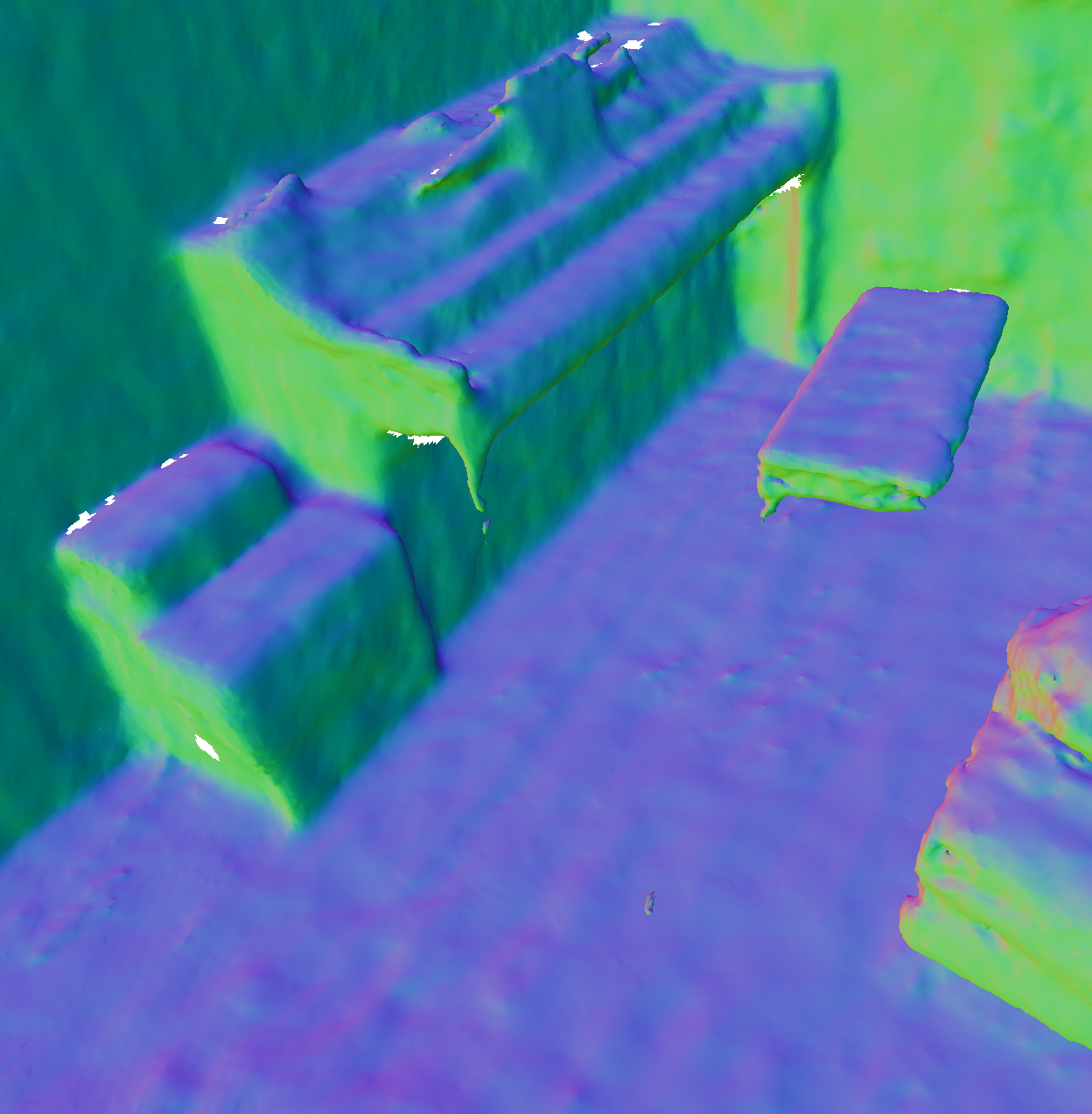} }  & \raisebox{-0.5\height}{\includegraphics[width=.19\textwidth]{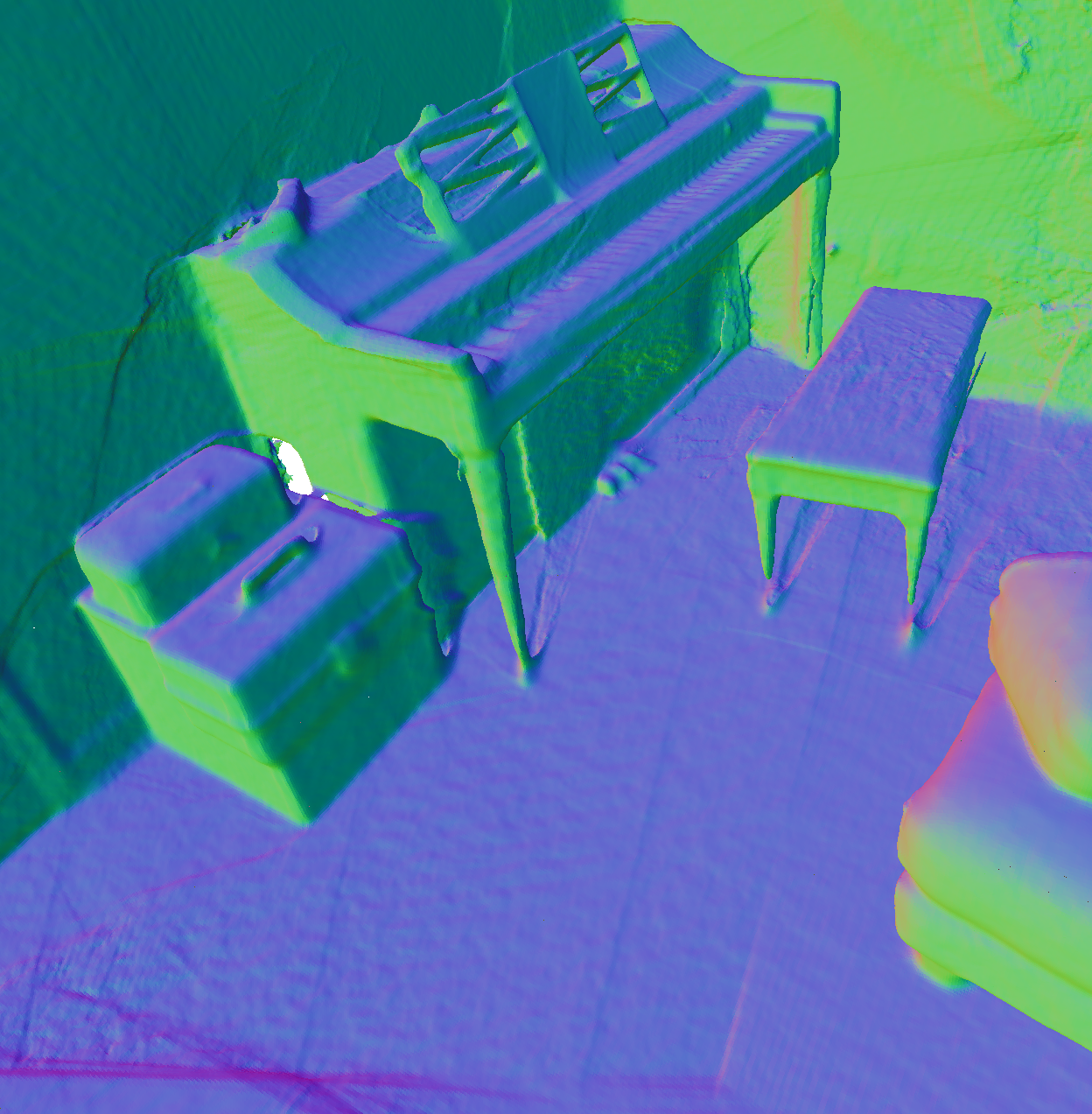} } \\
             \rotatebox[origin=b]{90}{ \texttt{0580}} & \raisebox{-0.5\height}{\includegraphics[width=.19\textwidth]{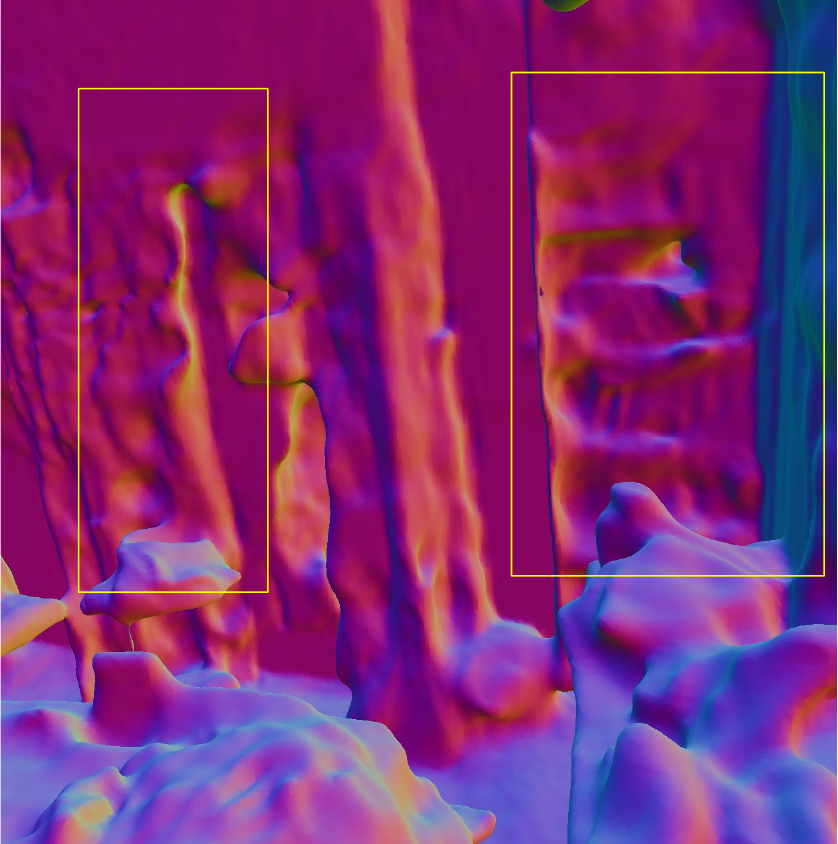} } & \raisebox{-0.5\height}{\includegraphics[width=.19\textwidth]{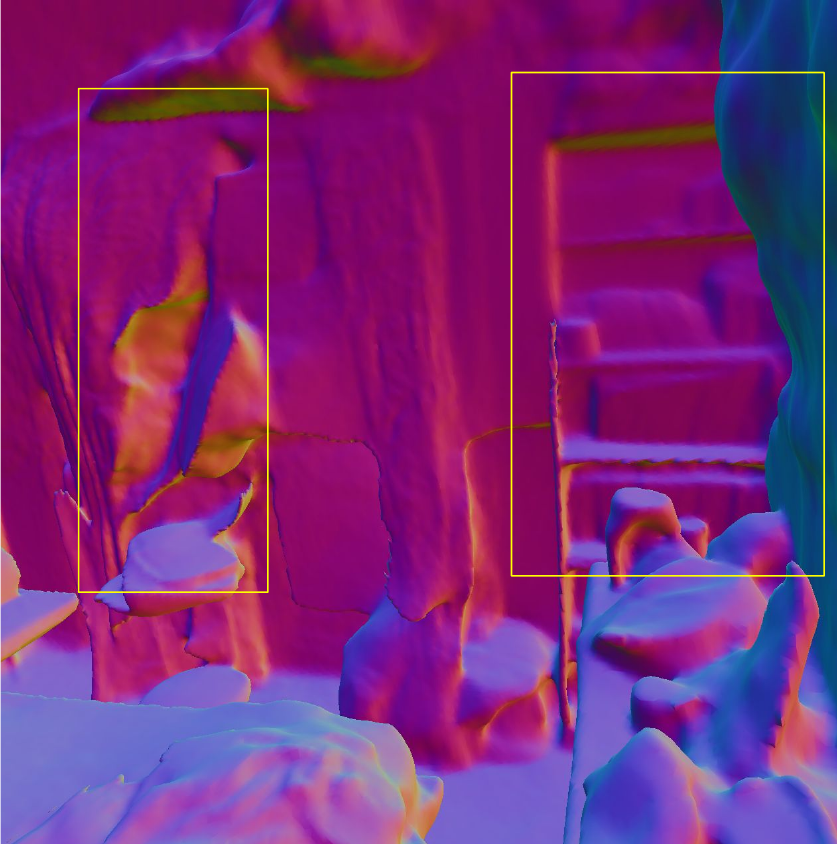} }
             &\raisebox{-0.5\height}{\includegraphics[width=.19\textwidth]{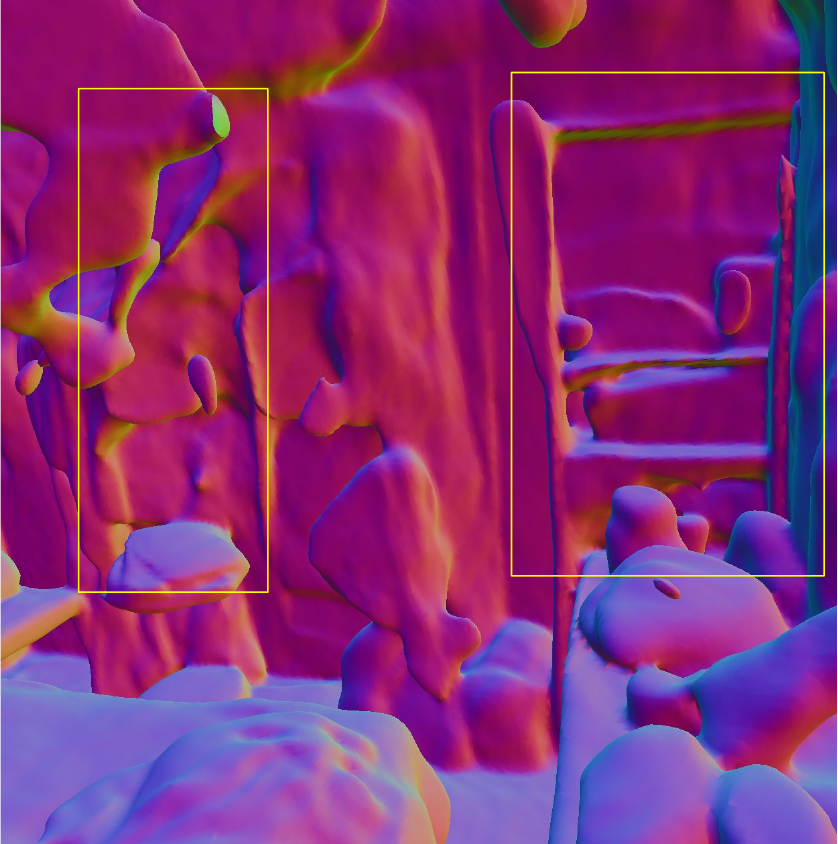} } & \raisebox{-0.5\height}{\includegraphics[width=.19\textwidth]{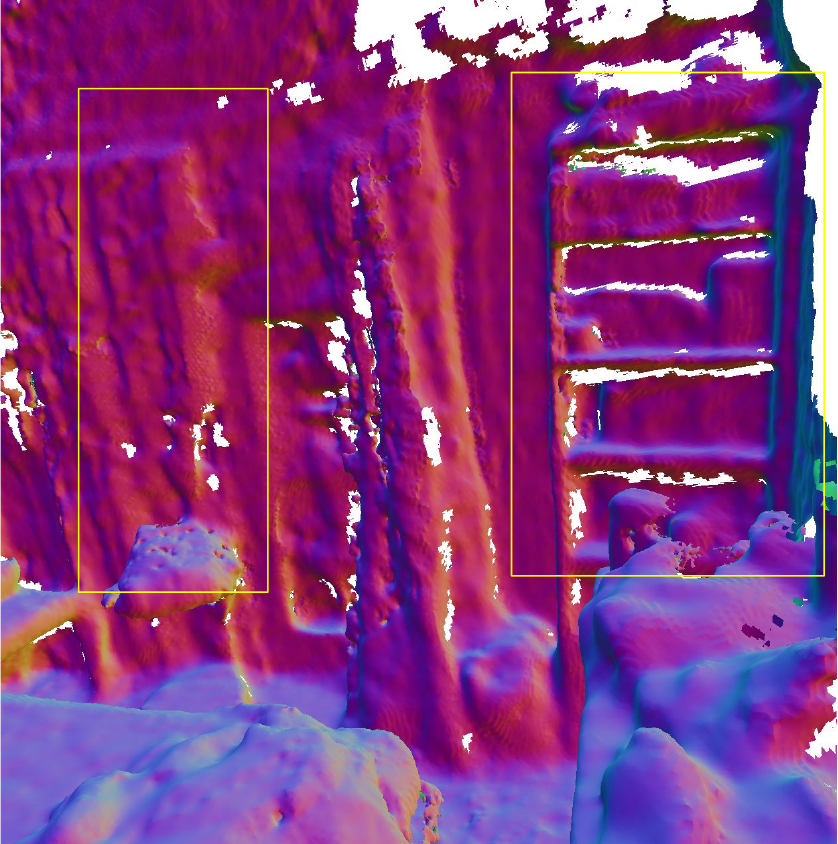} }  & \raisebox{-0.5\height}{\includegraphics[width=.19\textwidth]{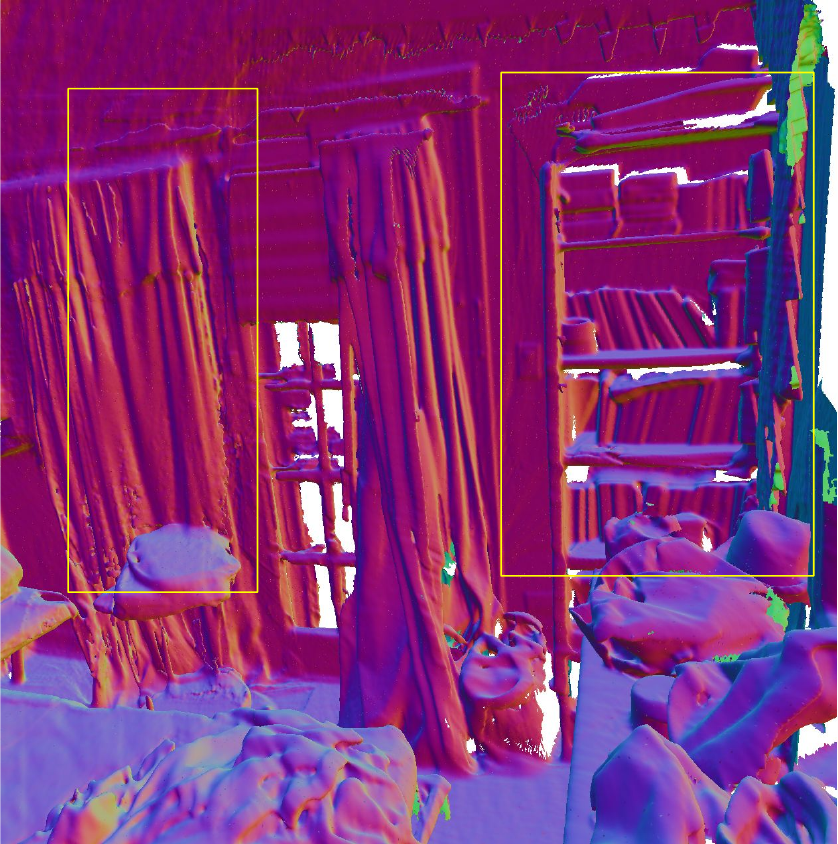} } \\
             & \textbf{Manhattan-} & \textbf{MonoSDF} & \textbf{NeuRIS} & \textbf{VF-NeRF} & \textbf{Ground} \\
              & \textbf{SDF} &  & &  \textbf{(Ours)} & \textbf{truth} \\
        \end{tabular}
        \caption[3D reconstruction qualitative results]{\textbf{3D reconstruction qualitative results.}}
        \label{fig:scene-qualitative-3d-replica}
    \end{figure*}

We provide qualitative results for each scene of Replica and ScanNet in~\Cref{fig:scene-qualitative-3d-replica}. We include visualizations of state-of-the-art 3D reconstruction methods as comparisons. \vfnerf always achieves very accurate results and avoids generating artifacts on the walls, something that often happens in SDF-based methods.

\section{Novel view synthesis qualitative results}

\begin{figure*}[ht]
        \centering
        \begin{tabular}{cccccc}
             \rotatebox[origin=b]{90}{ \texttt{Room 0}} & \raisebox{-0.5\height}{\includegraphics[width=.19\textwidth]{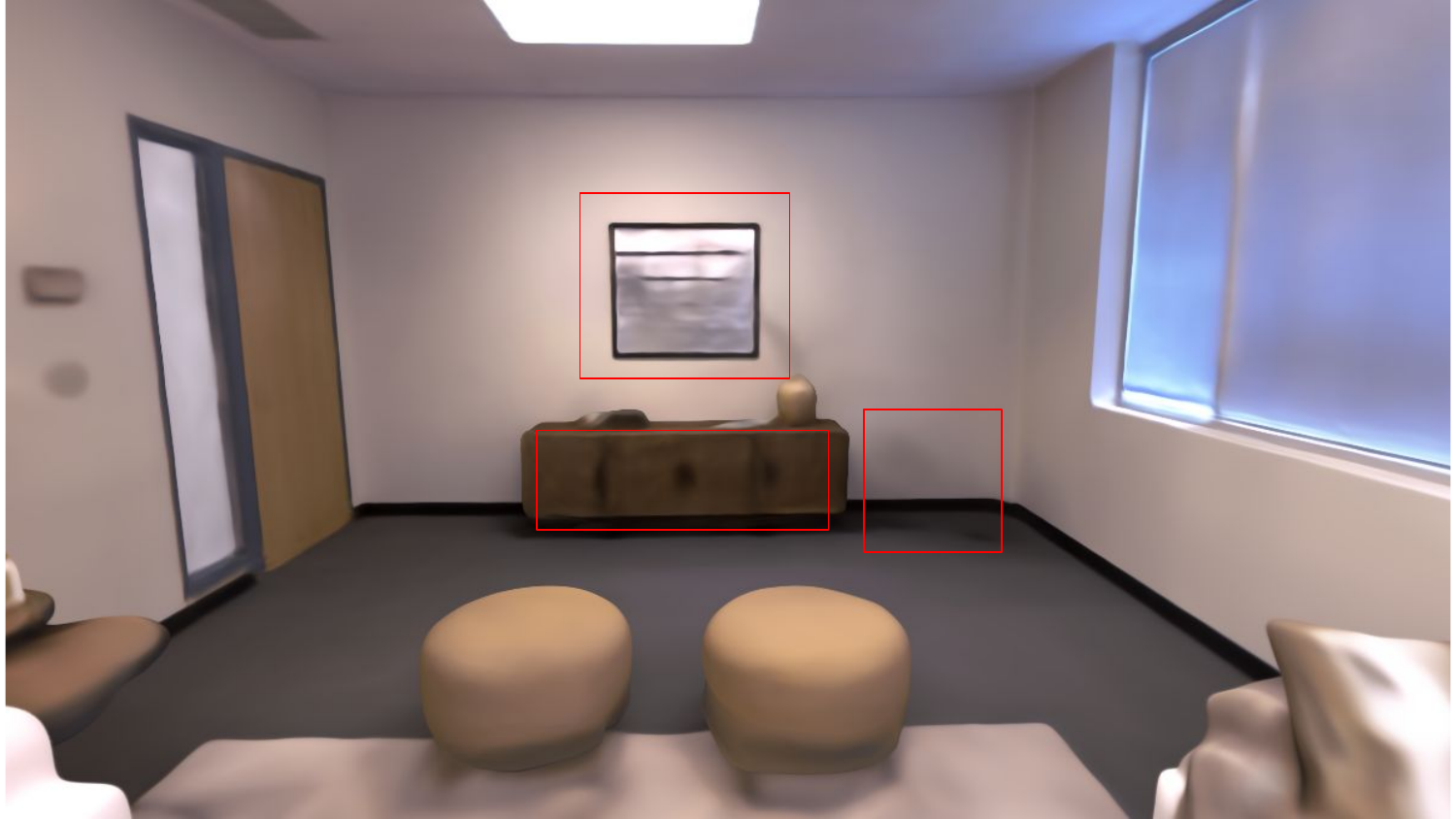} } & \raisebox{-0.5\height}{\includegraphics[width=.19\textwidth]{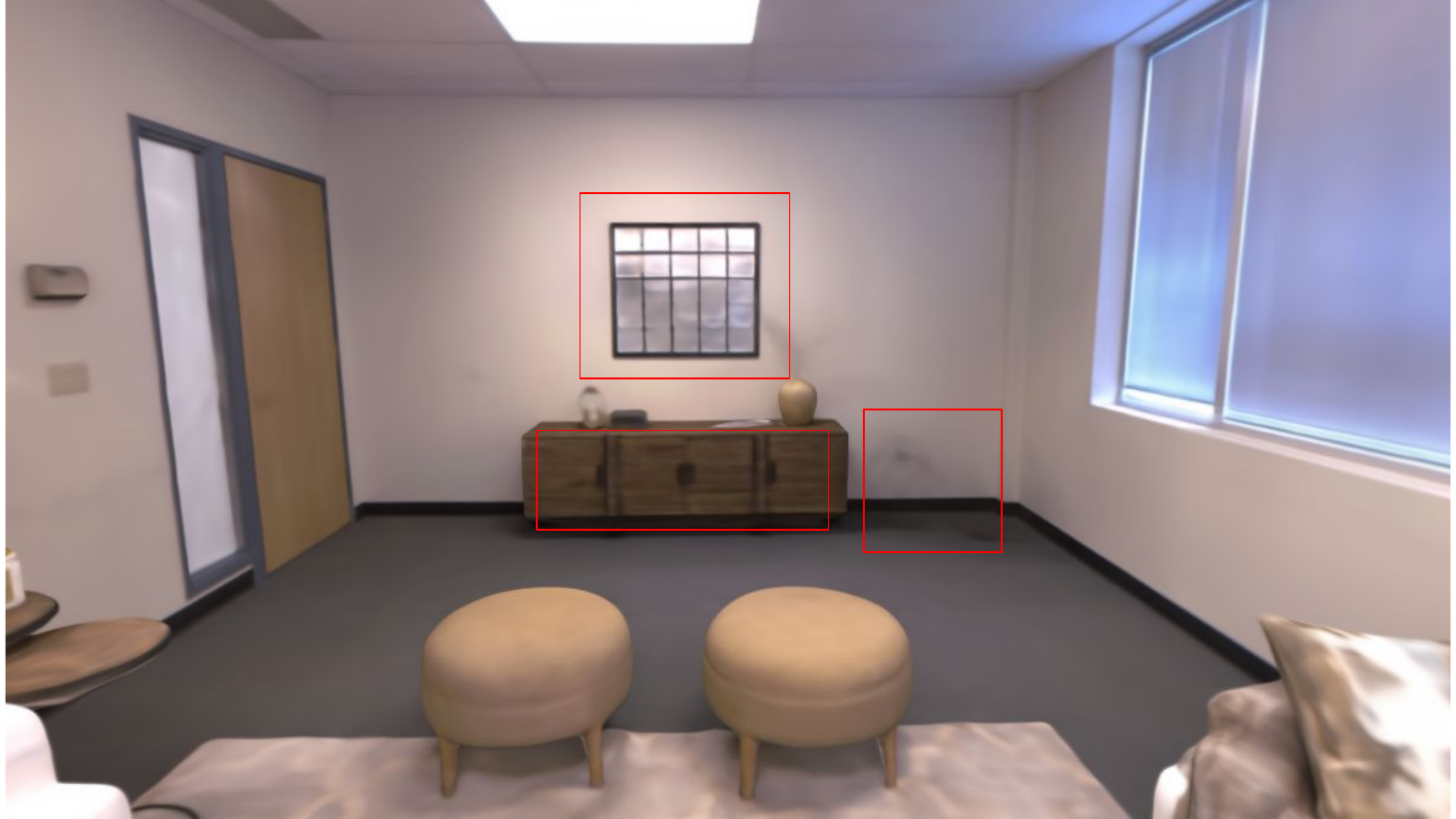} }
             &\raisebox{-0.5\height}{\includegraphics[width=.19\textwidth]{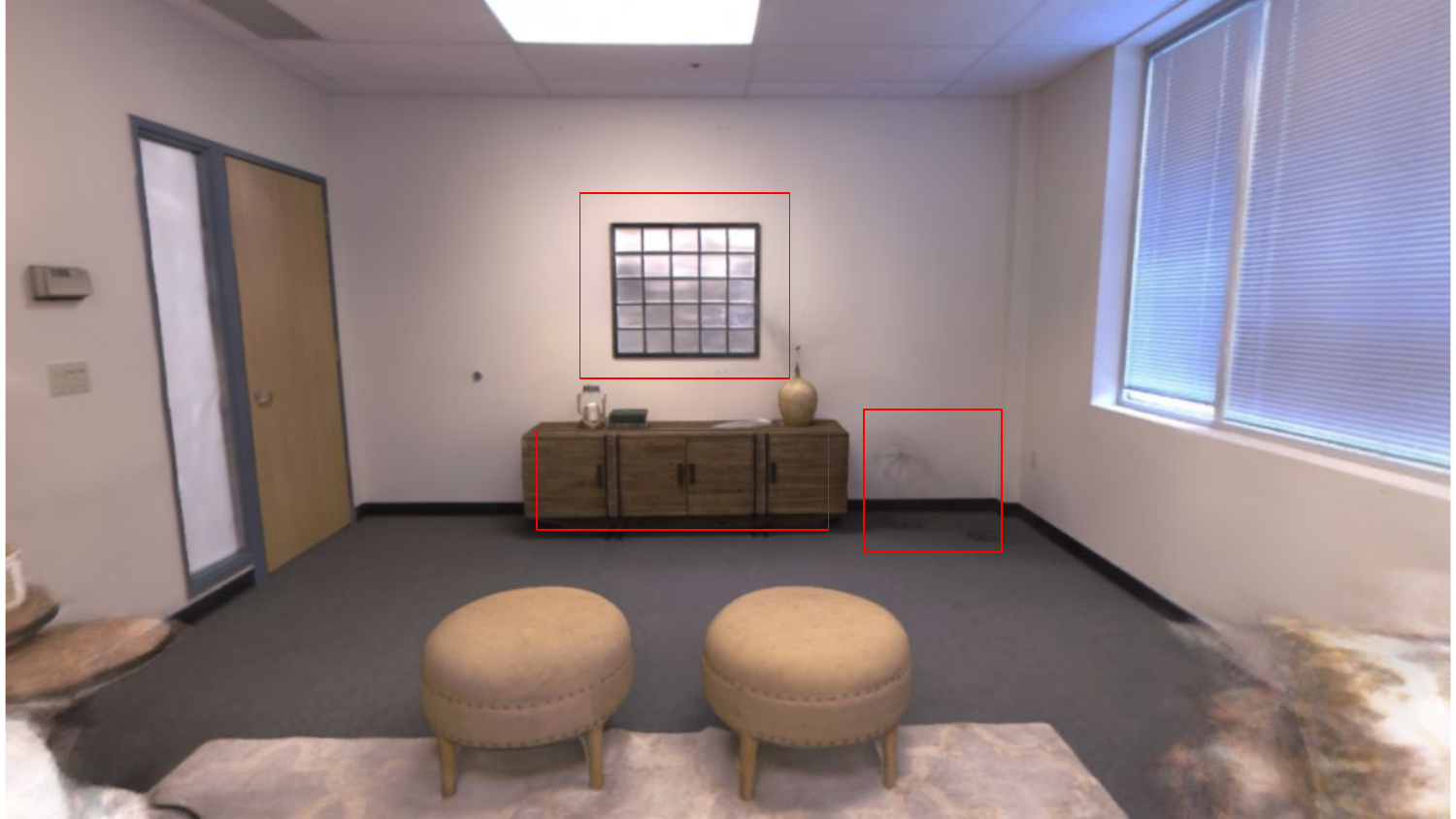} } & \raisebox{-0.5\height}{\includegraphics[width=.19\textwidth]{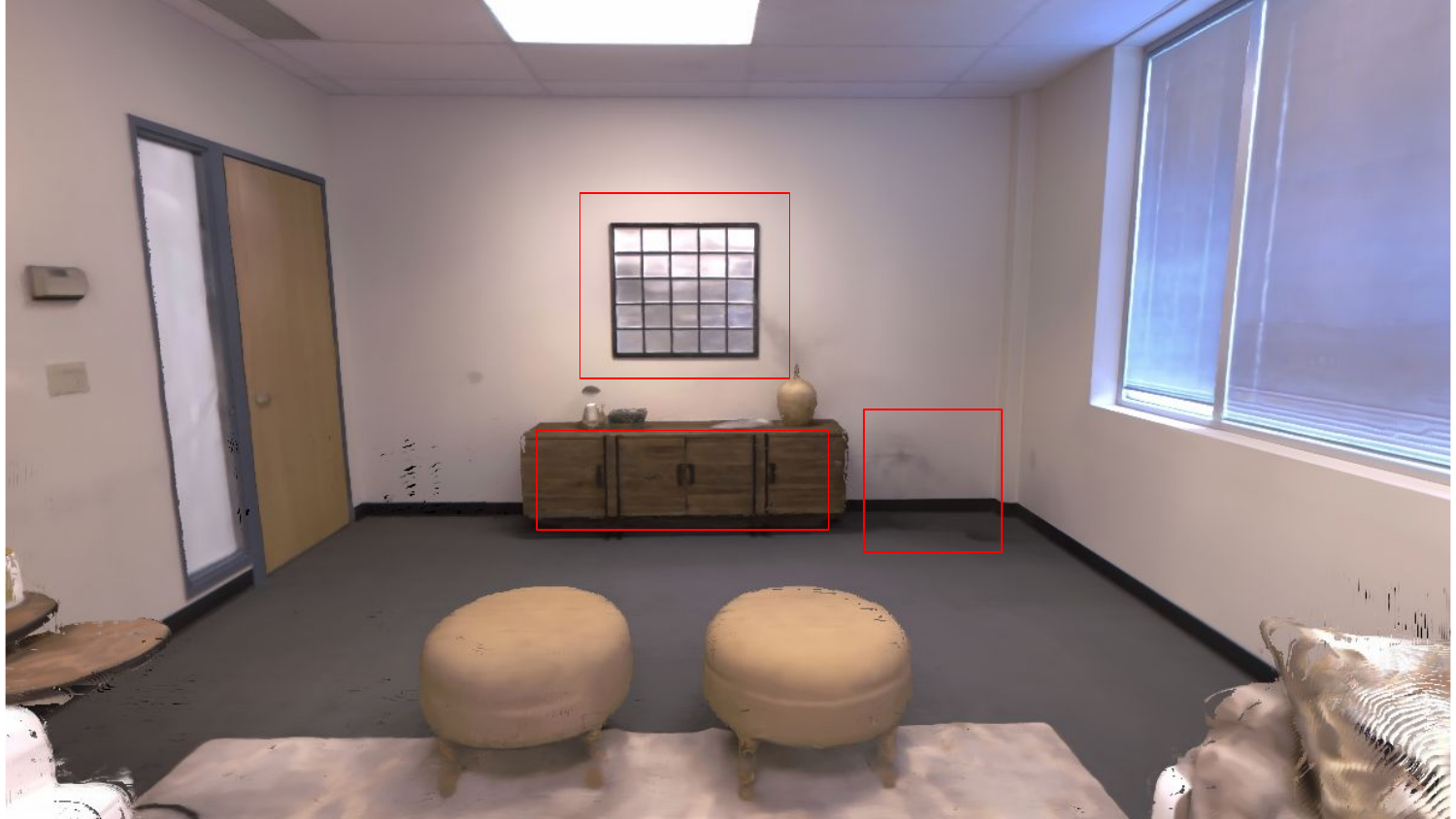} } & \raisebox{-0.5\height}{\includegraphics[width=.19\textwidth]{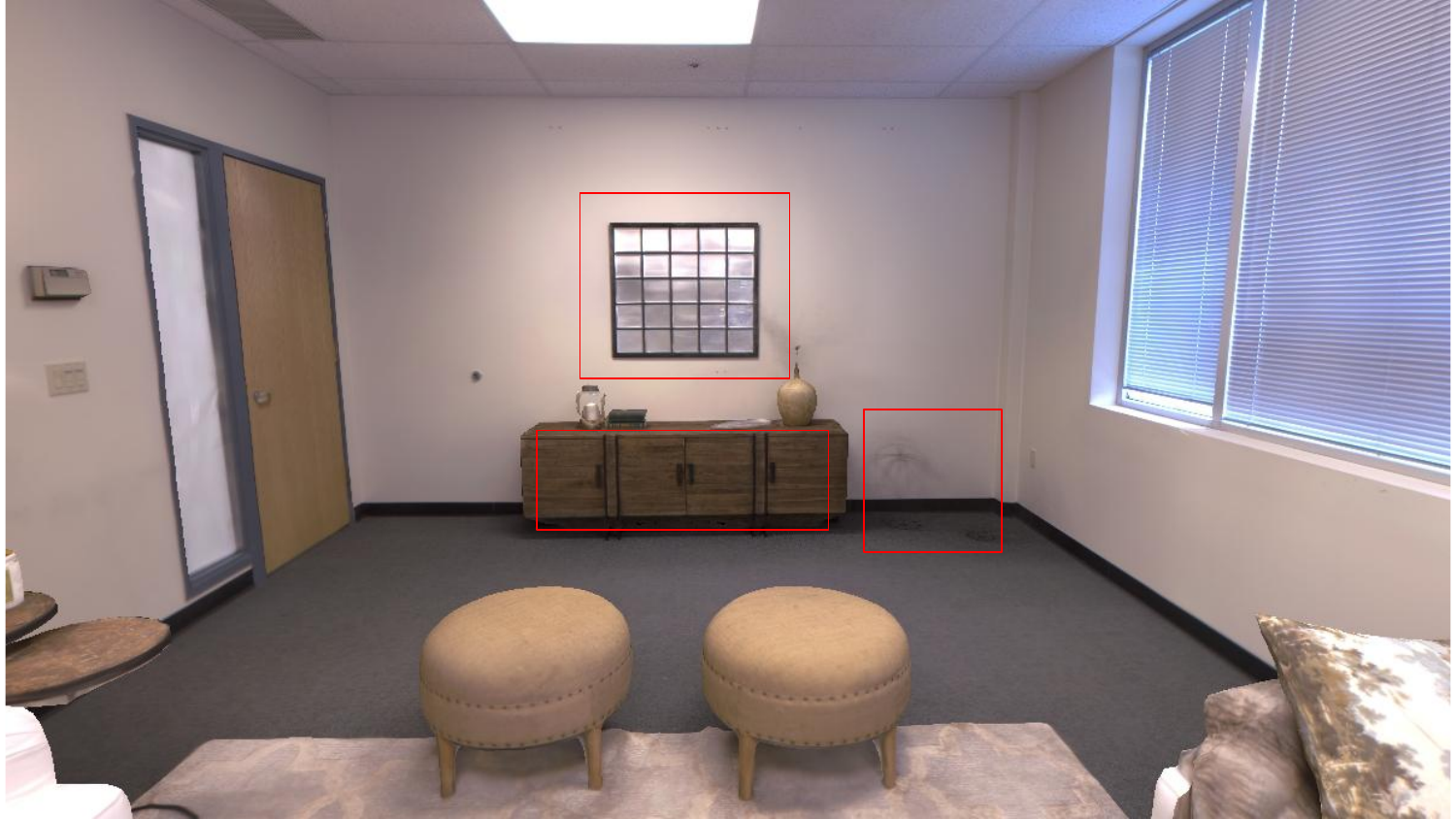} } \\
             \rotatebox[origin=b]{90}{ \texttt{Room 1}} & \raisebox{-0.5\height}{\includegraphics[width=.19\textwidth]{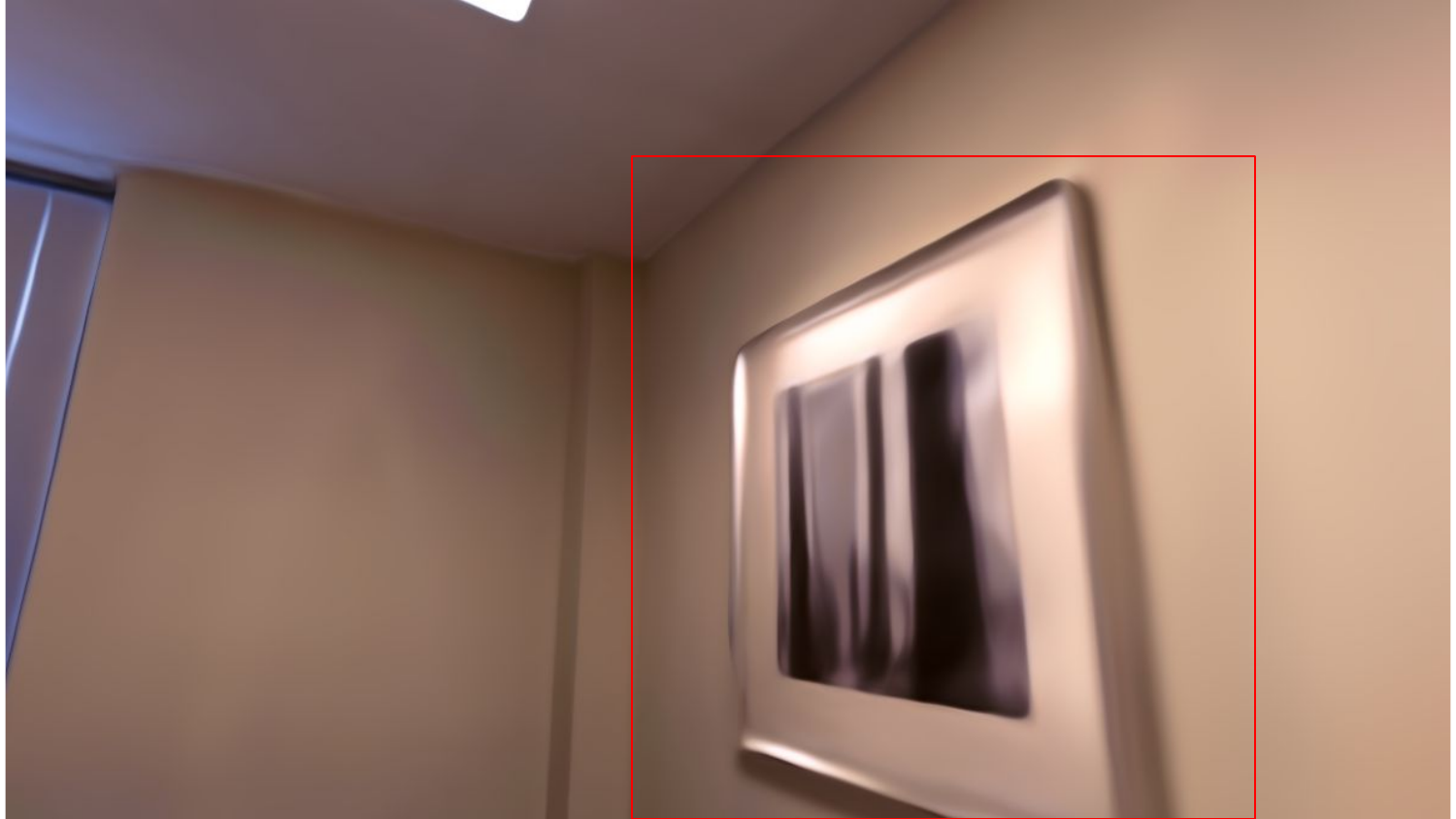} } & \raisebox{-0.5\height}{\includegraphics[width=.19\textwidth]{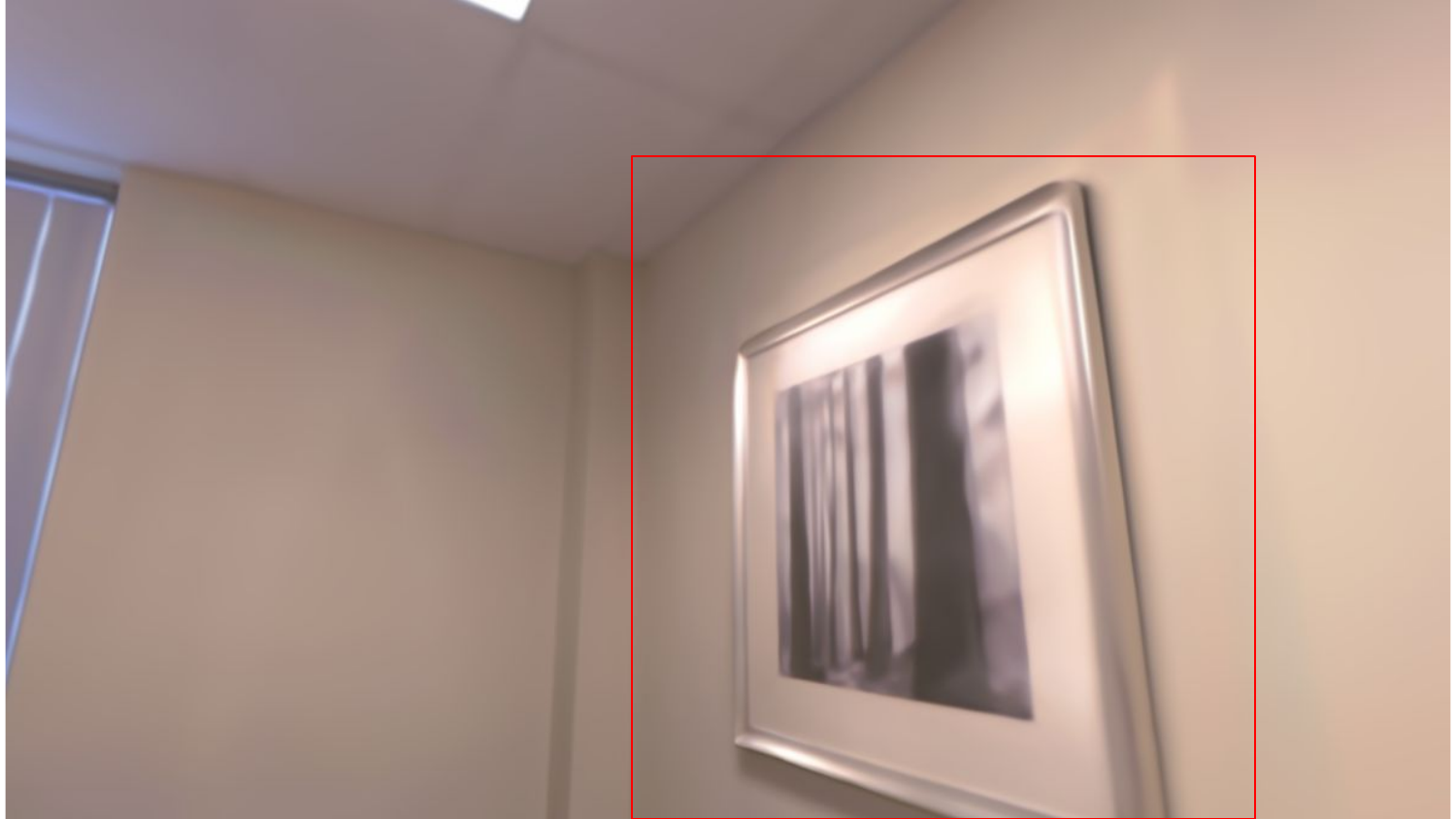} }
             &\raisebox{-0.5\height}{\includegraphics[width=.19\textwidth]{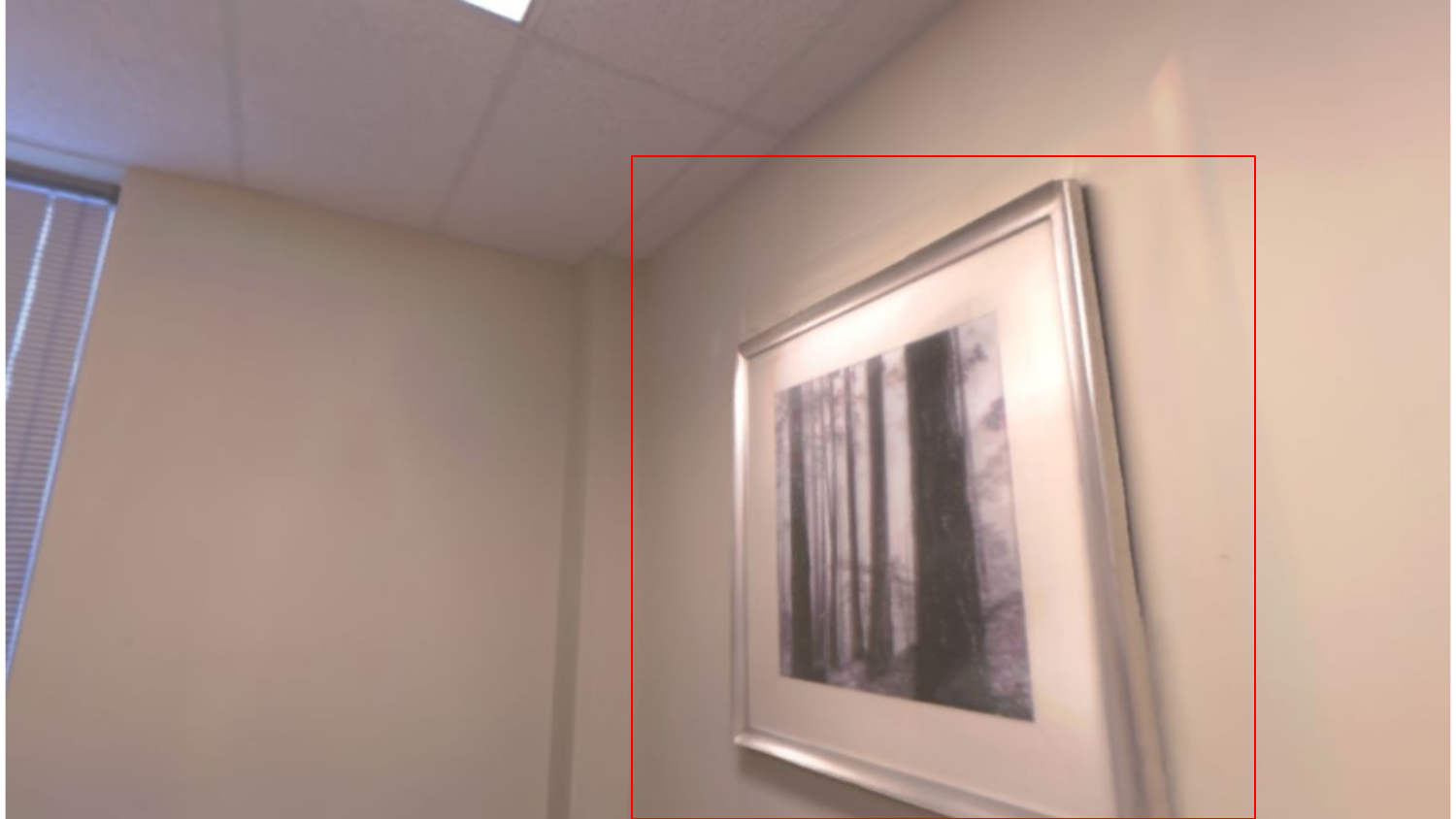} } & \raisebox{-0.5\height}{\includegraphics[width=.19\textwidth]{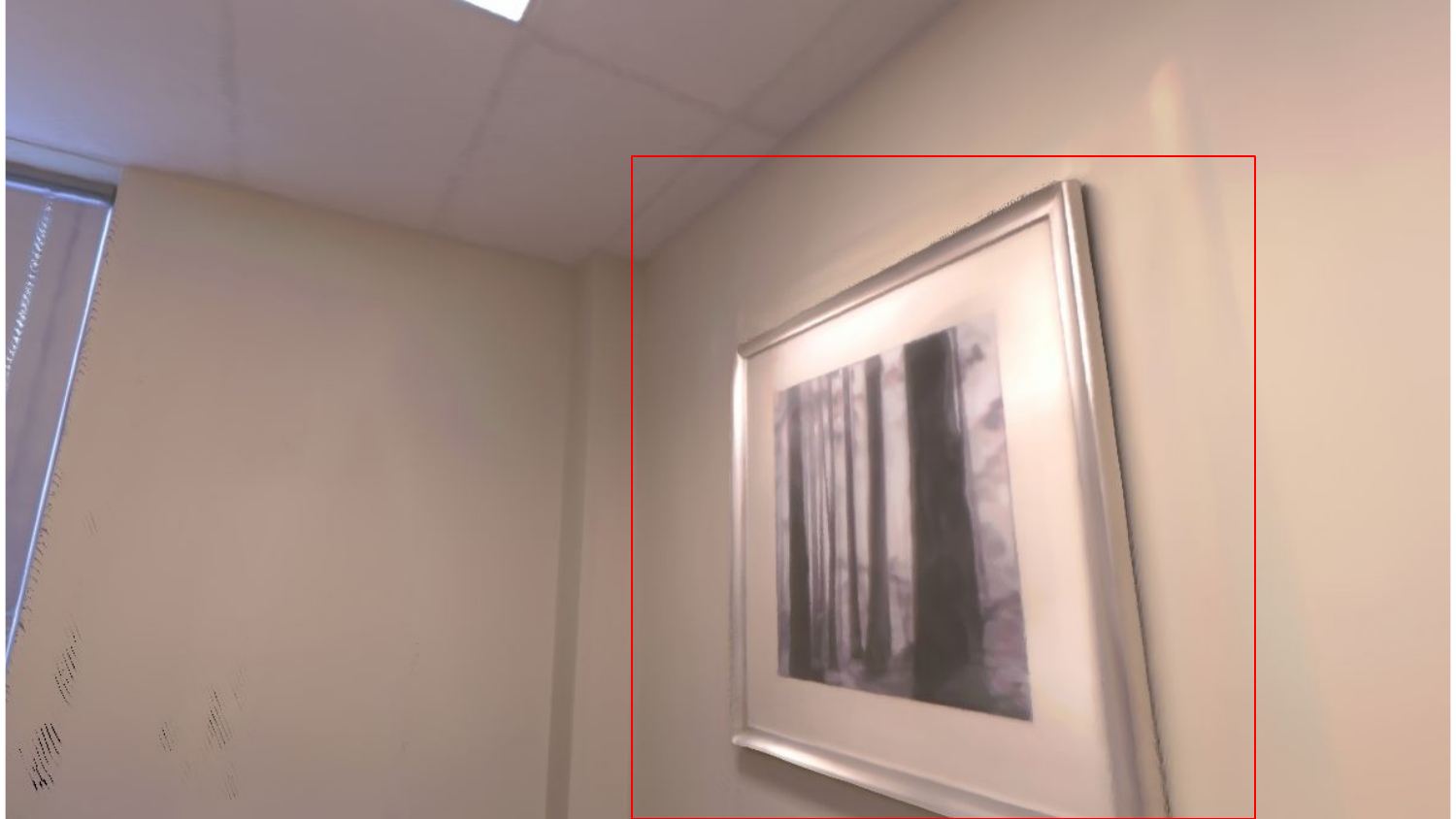} } & \raisebox{-0.5\height}{\includegraphics[width=.19\textwidth]{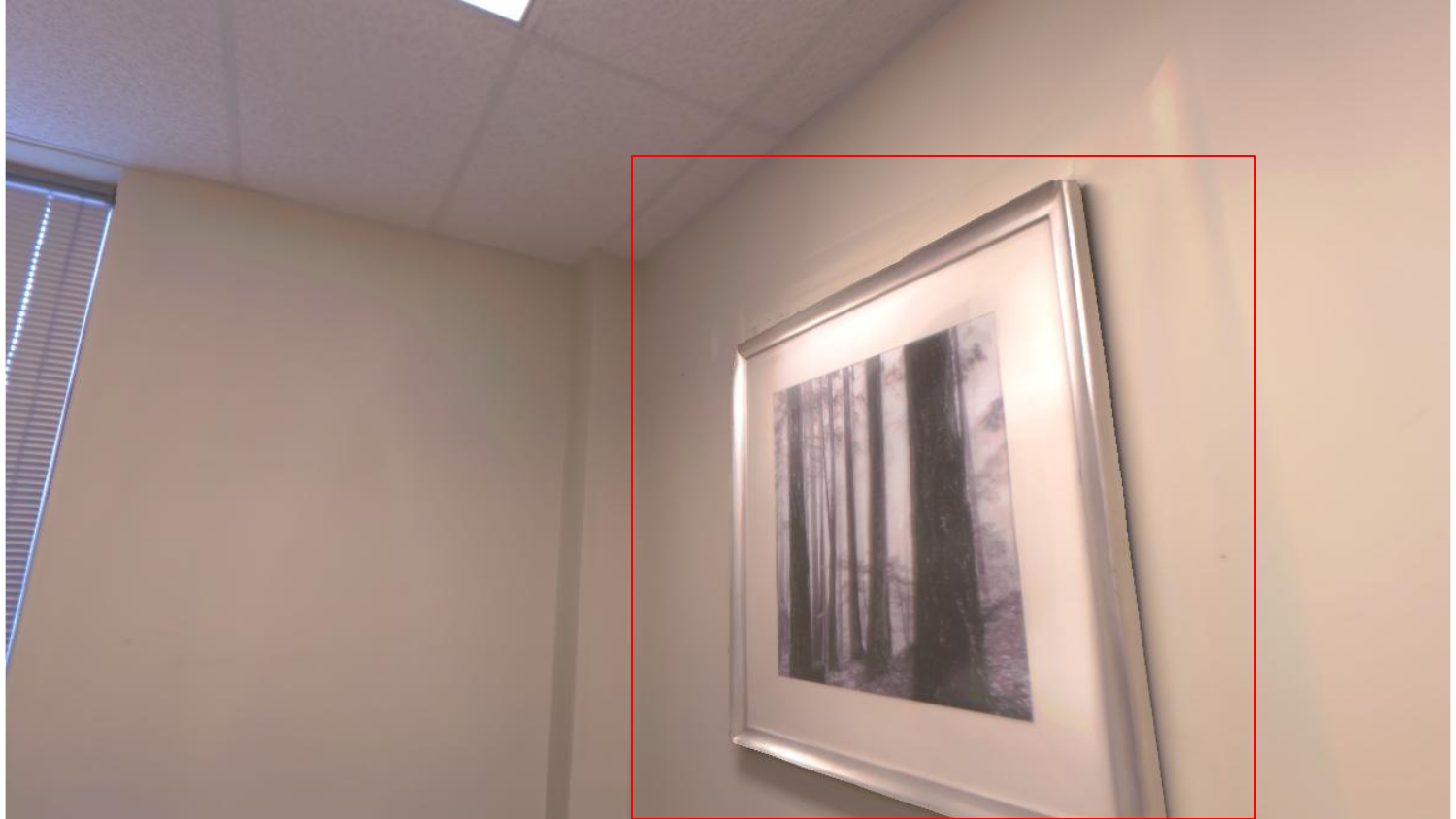} } \\
             \rotatebox[origin=b]{90}{ \texttt{Room 2}} & \raisebox{-0.5\height}{\includegraphics[width=.19\textwidth]{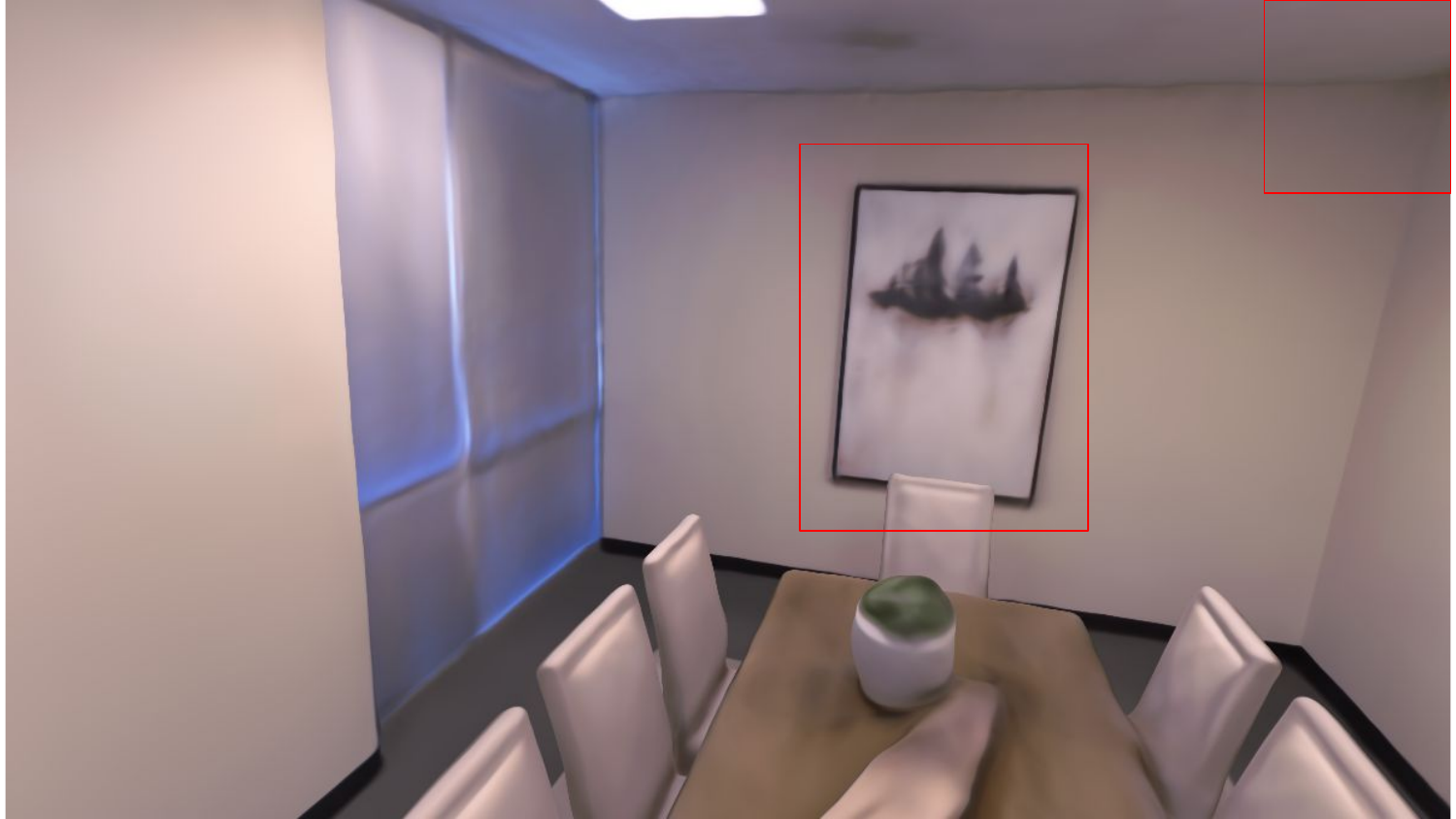} } & \raisebox{-0.5\height}{\includegraphics[width=.19\textwidth]{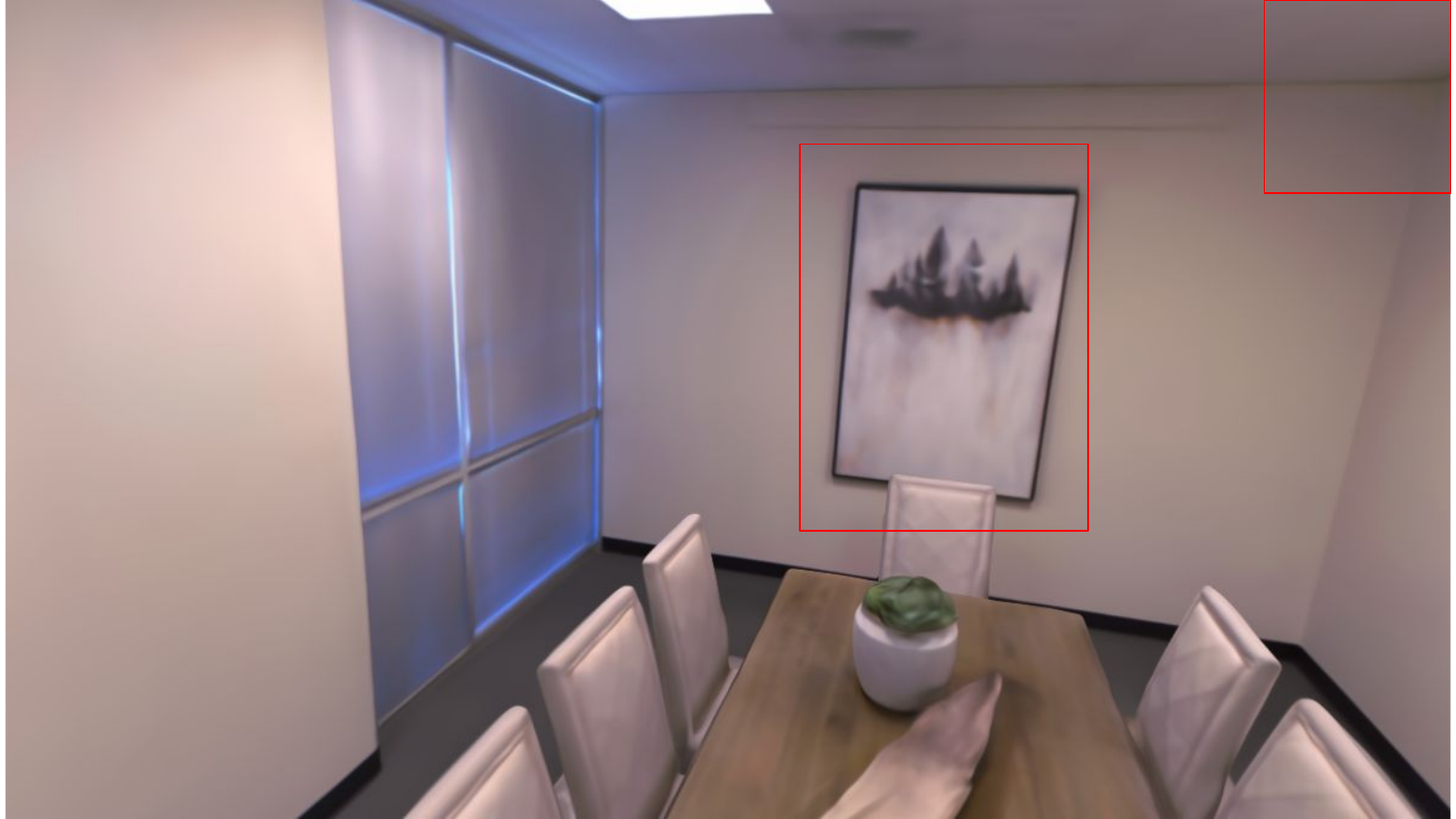} }
             &\raisebox{-0.5\height}{\includegraphics[width=.19\textwidth]{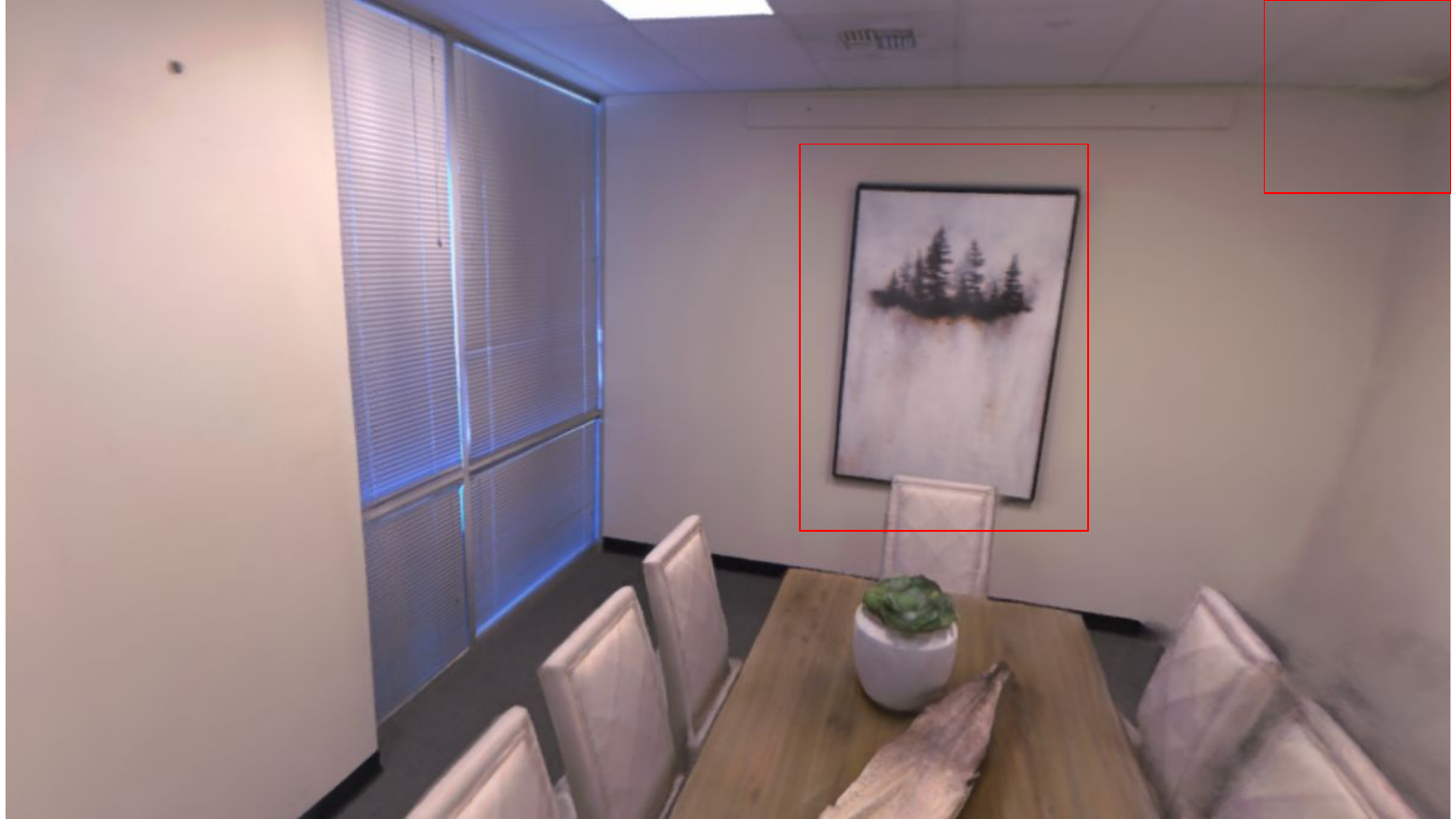} } & \raisebox{-0.5\height}{\includegraphics[width=.19\textwidth]{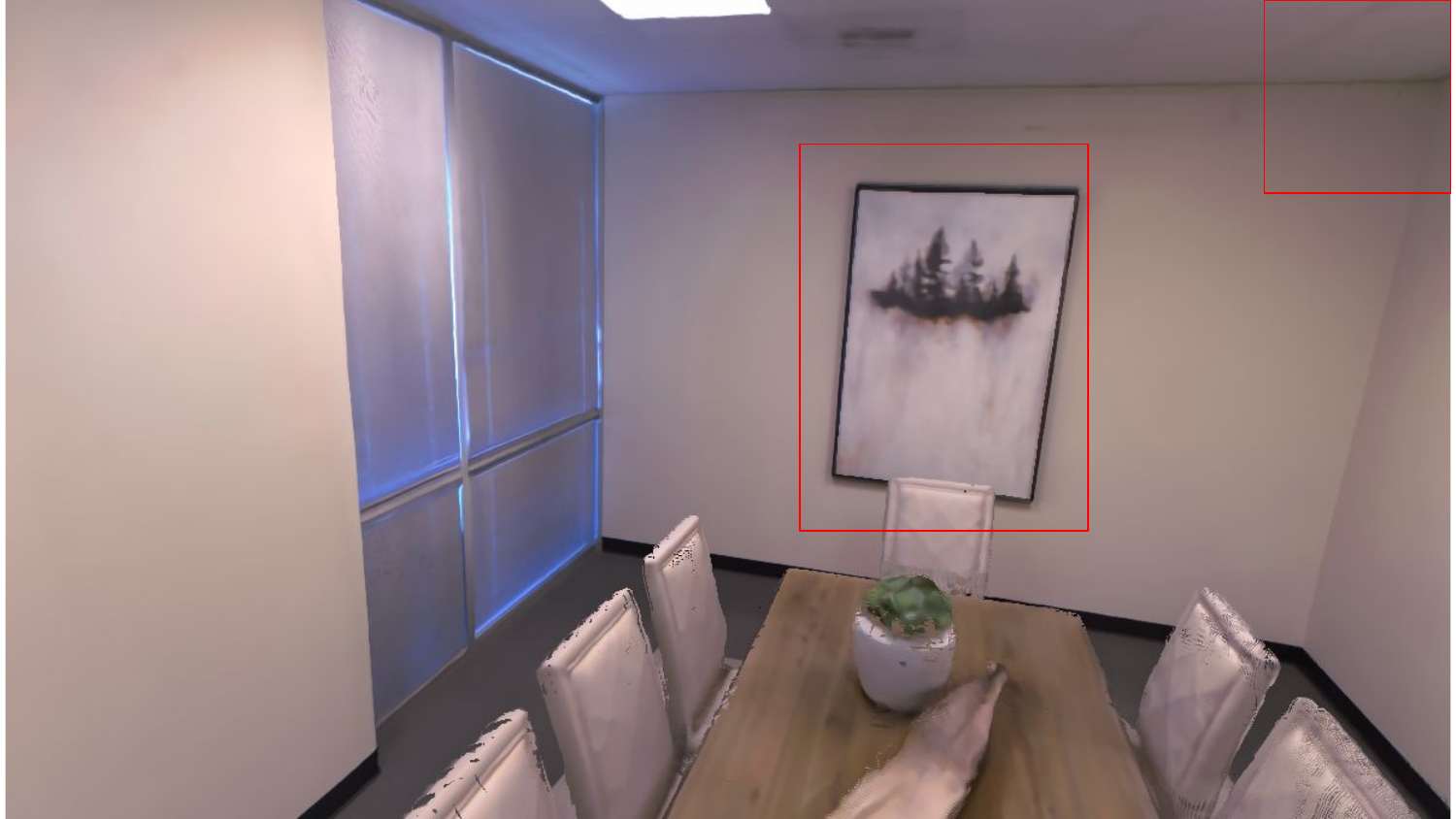 }} & \raisebox{-0.5\height}{\includegraphics[width=.19\textwidth]{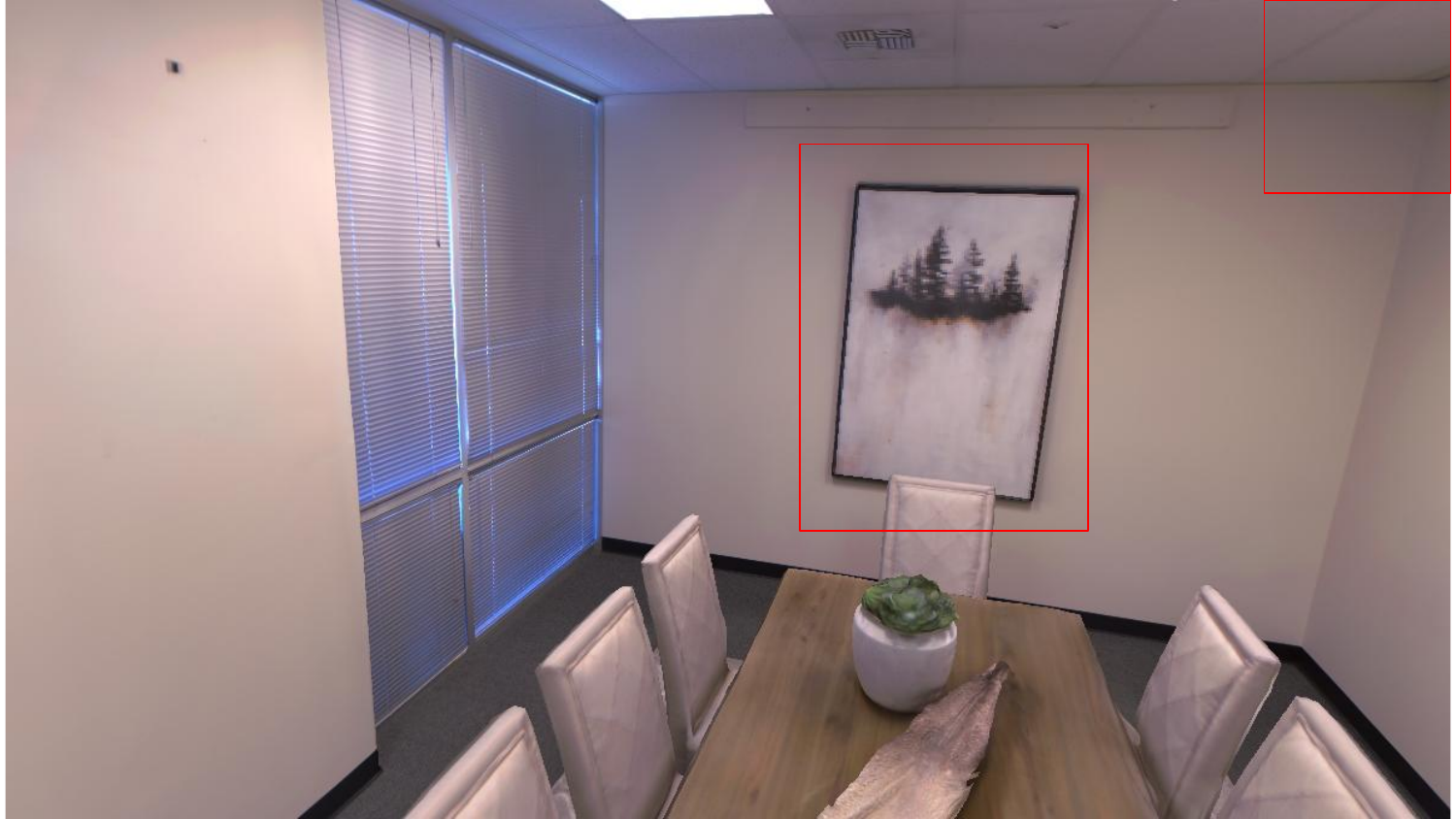} } \\
             \rotatebox[origin=b]{90}{ \texttt{Office 0}} & \raisebox{-0.5\height}{\includegraphics[width=.19\textwidth]{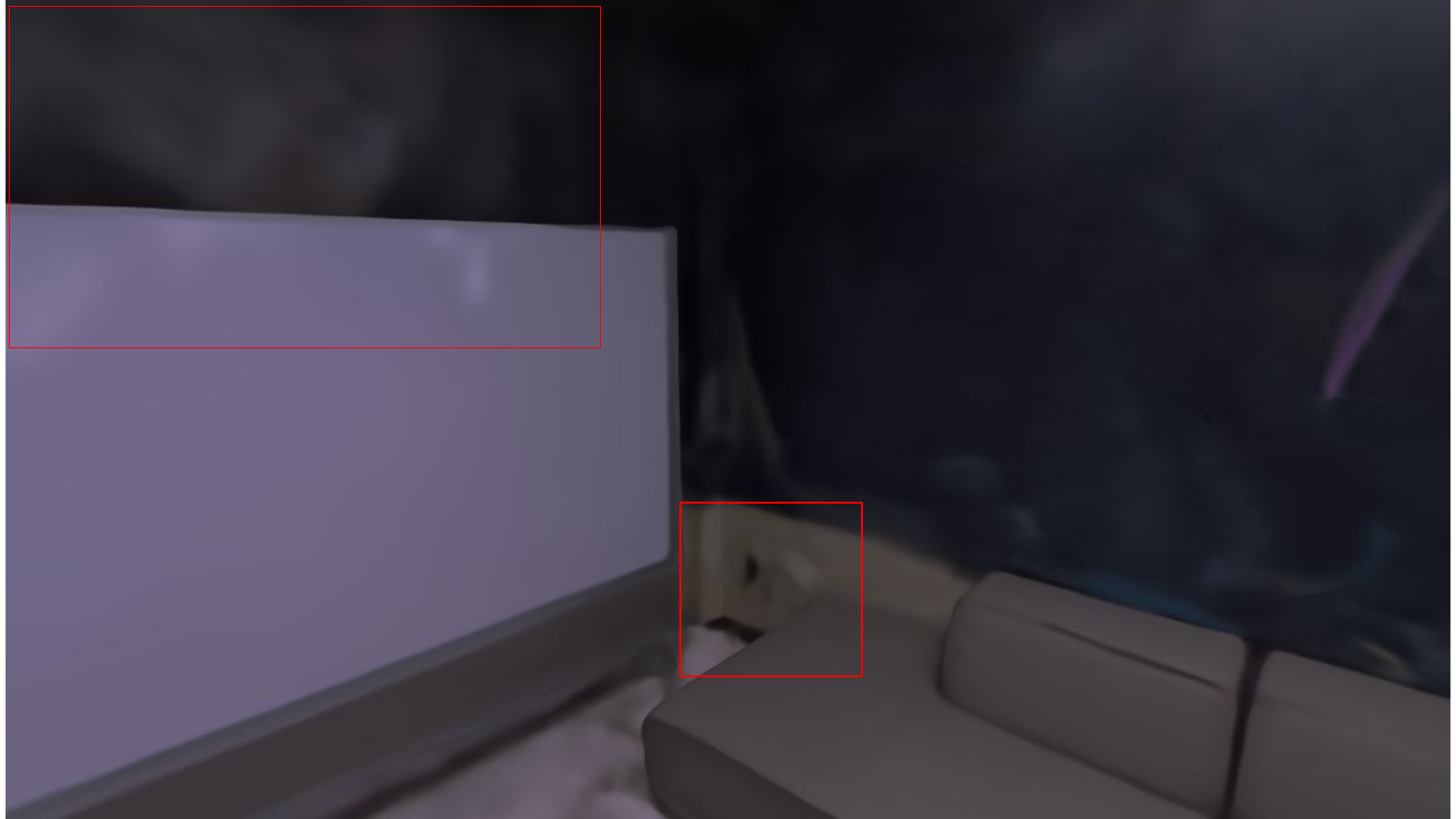} } & \raisebox{-0.5\height}{\includegraphics[width=.19\textwidth]{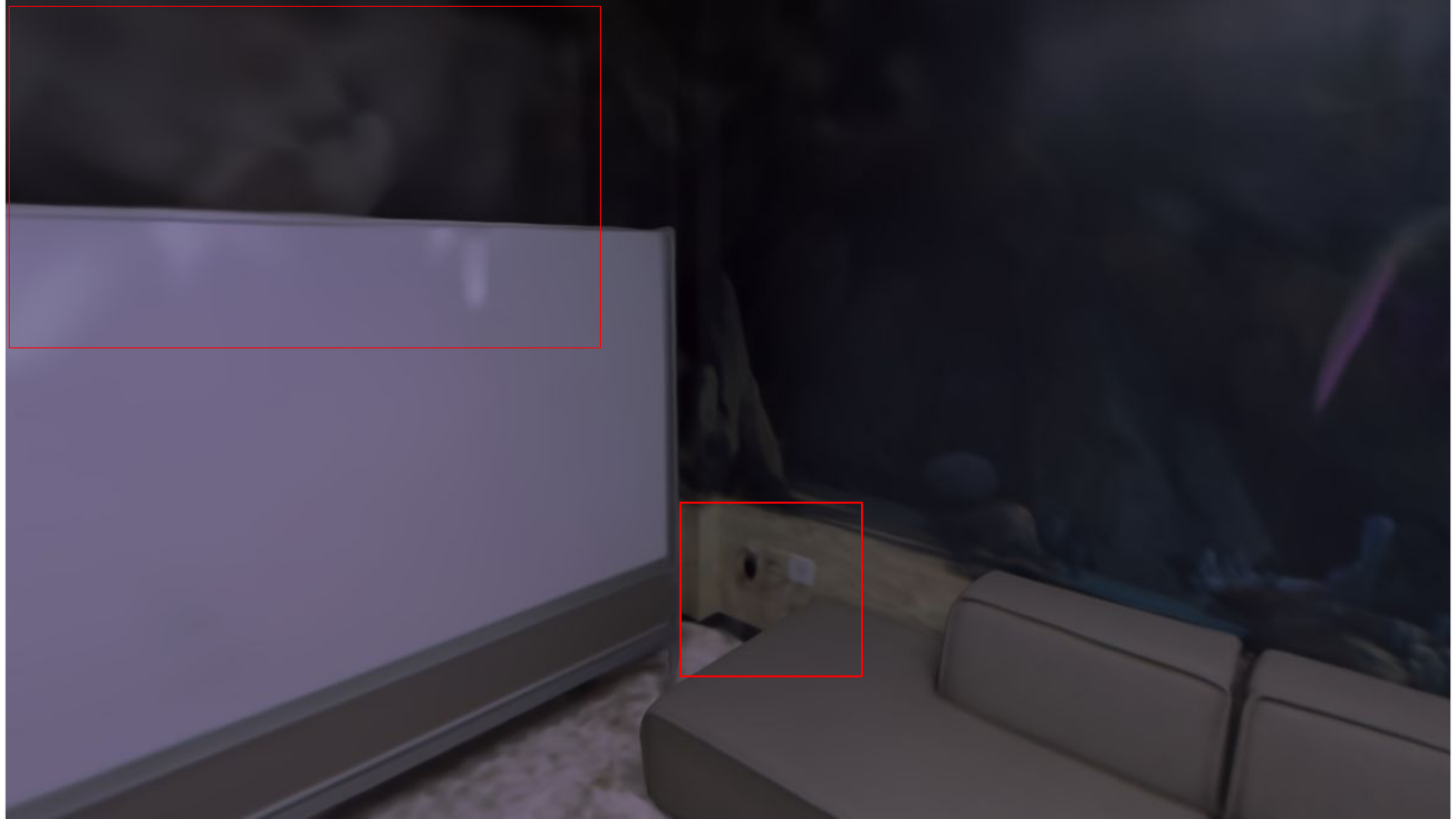} }
             &\raisebox{-0.5\height}{\includegraphics[width=.19\textwidth]{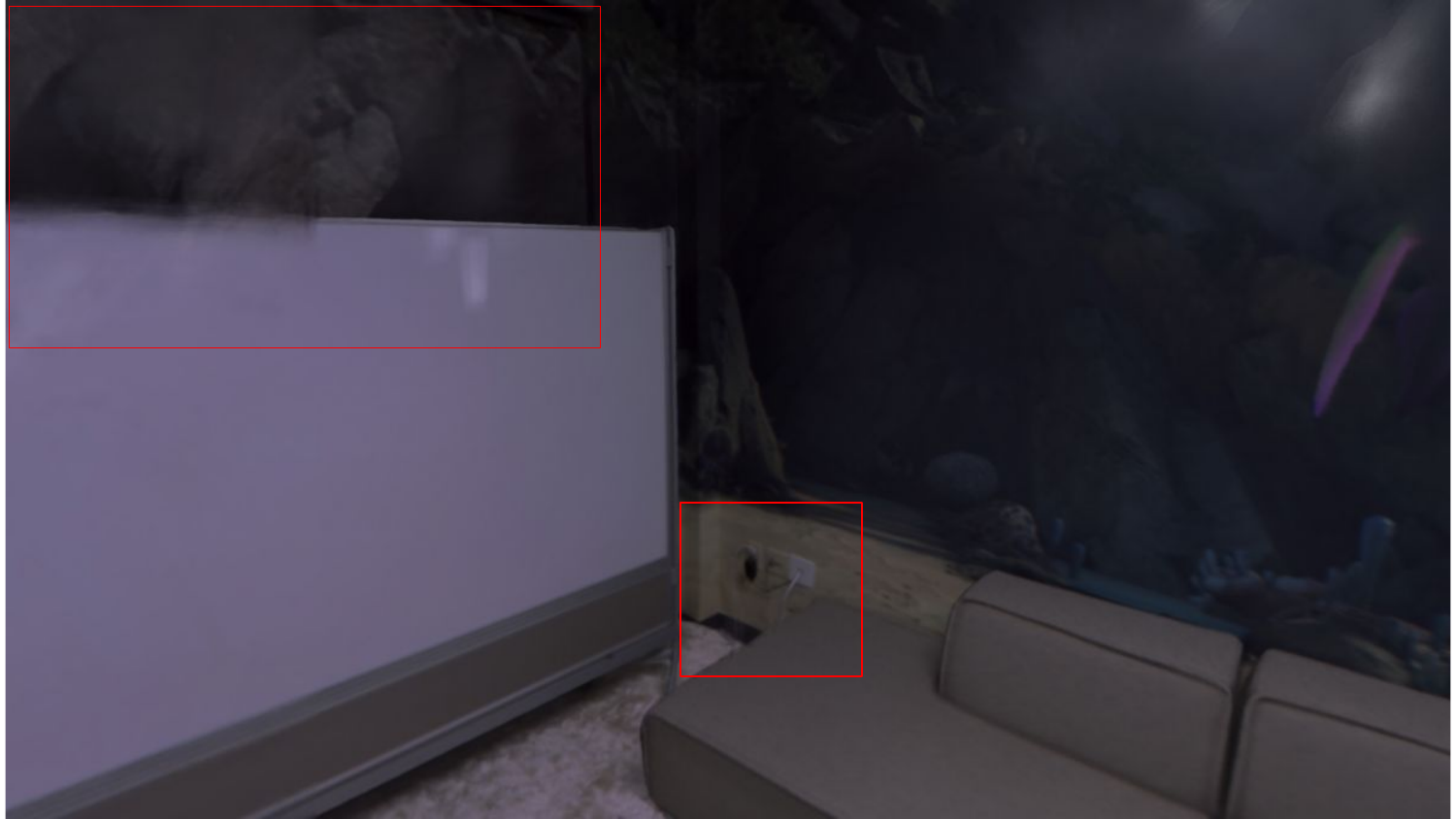} } & \raisebox{-0.5\height}{\includegraphics[width=.19\textwidth]{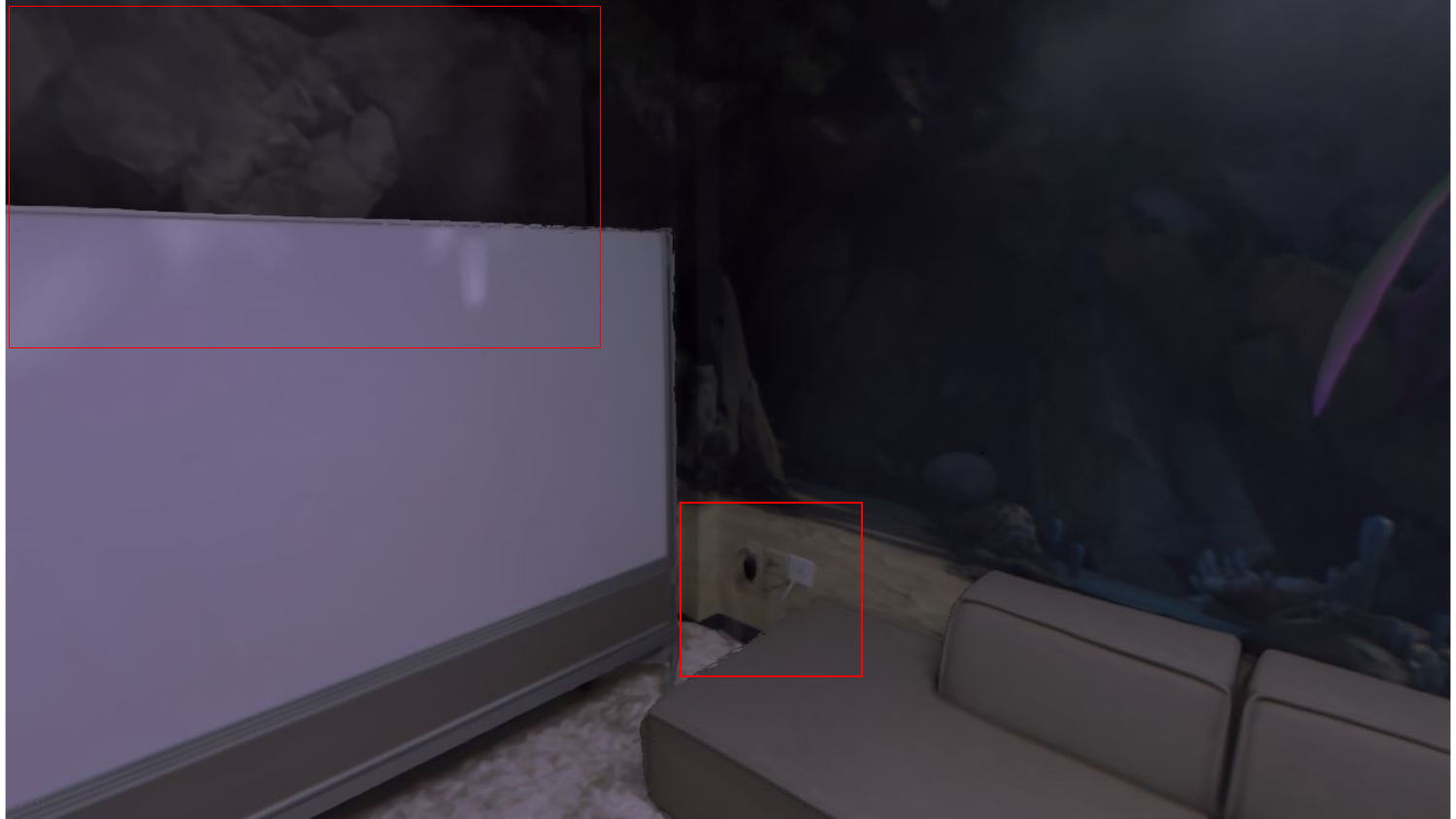} } & \raisebox{-0.5\height}{\includegraphics[width=.19\textwidth]{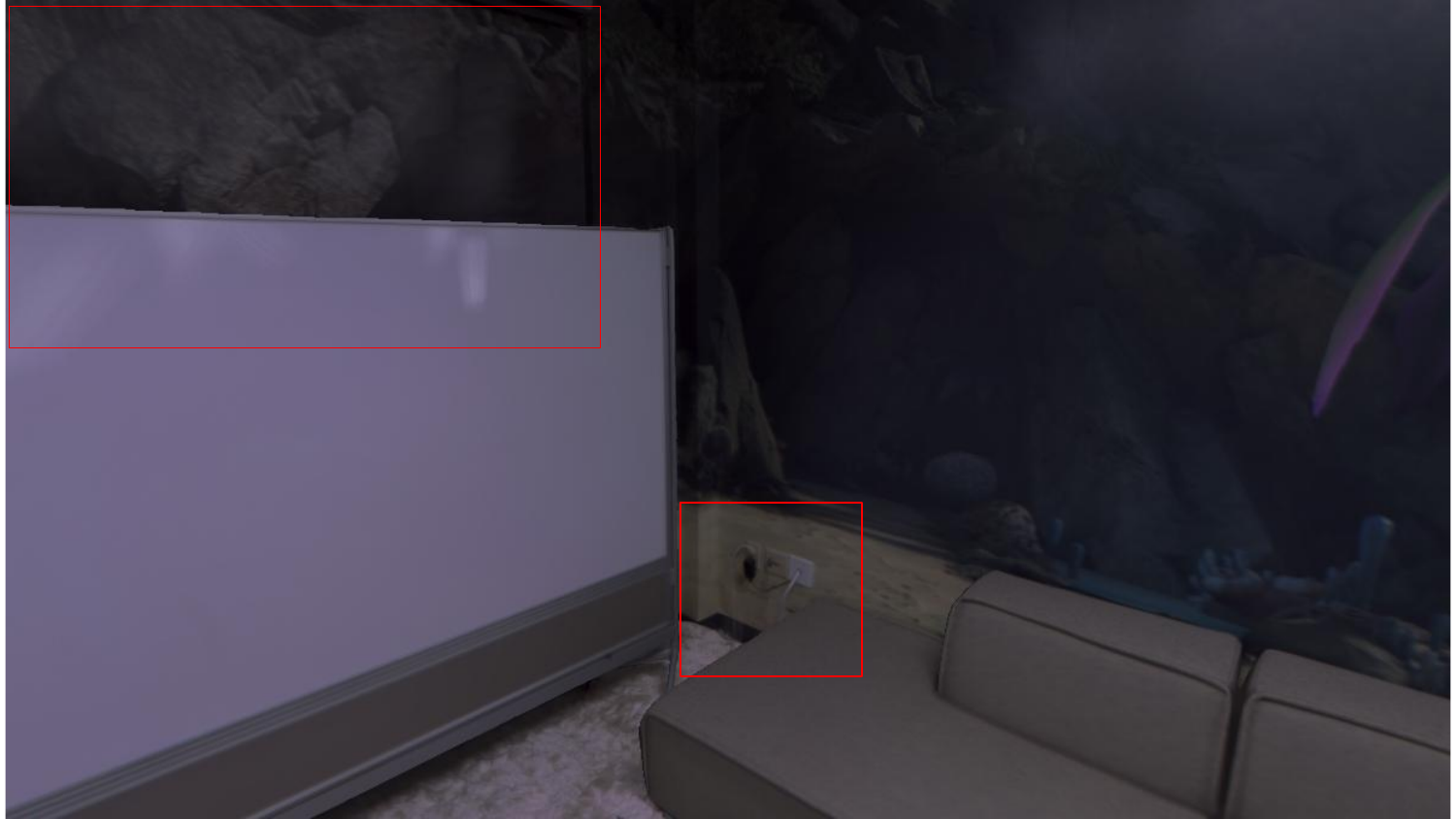} } \\
             \rotatebox[origin=b]{90}{ \texttt{Office 1}} & \raisebox{-0.5\height}{\includegraphics[width=.19\textwidth]{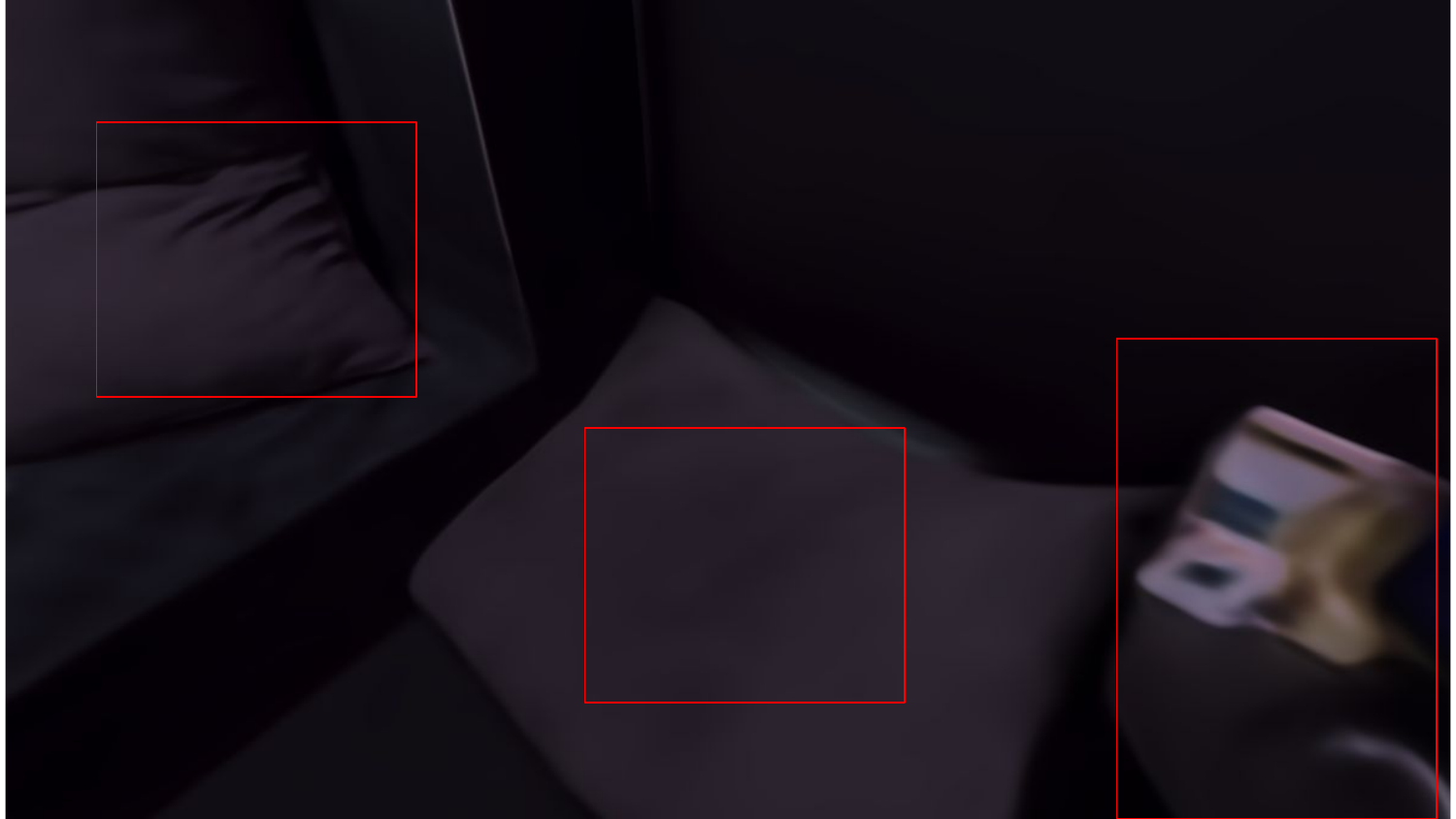} } & \raisebox{-0.5\height}{\includegraphics[width=.19\textwidth]{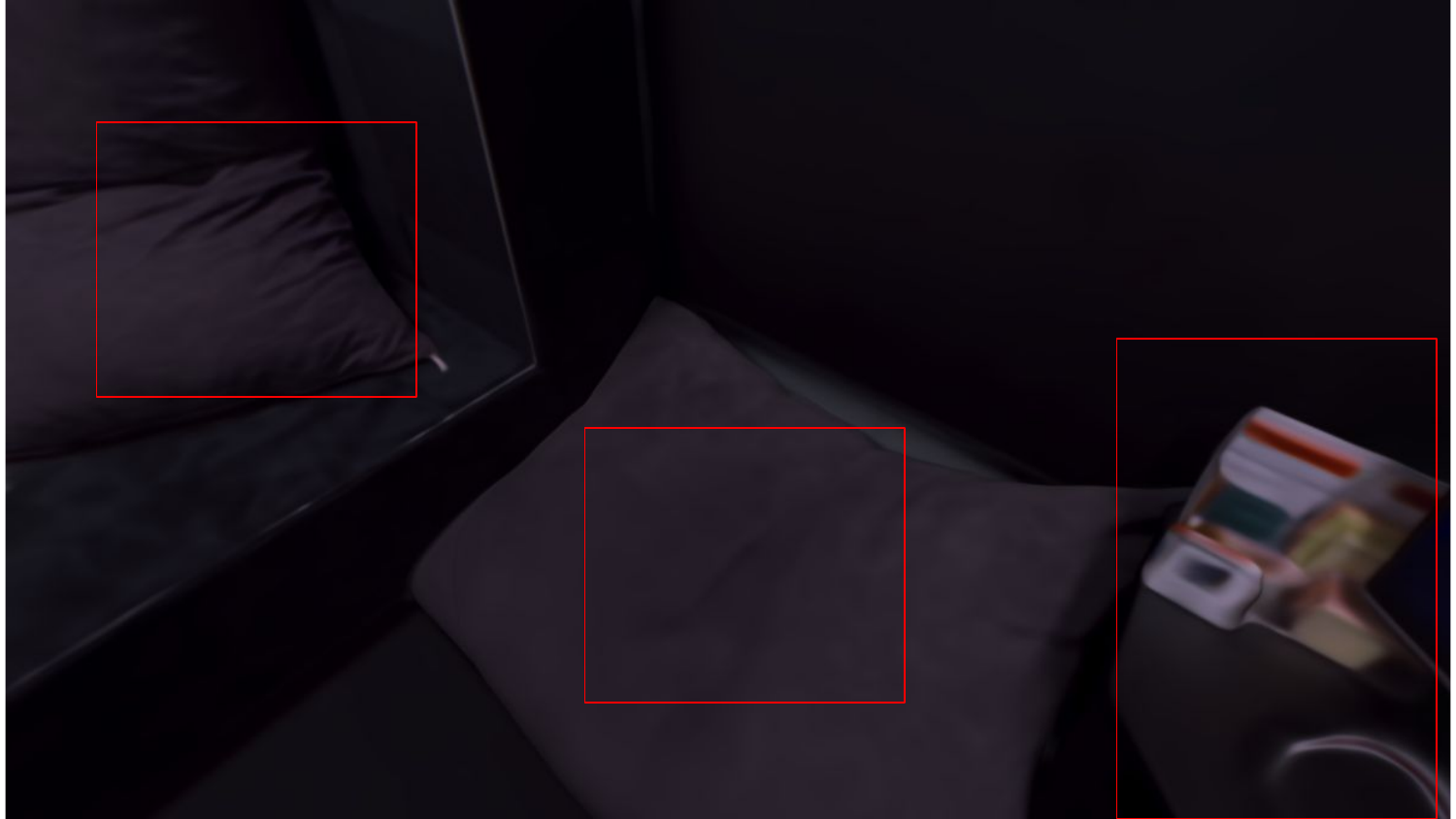} }
             &\raisebox{-0.5\height}{\includegraphics[width=.19\textwidth]{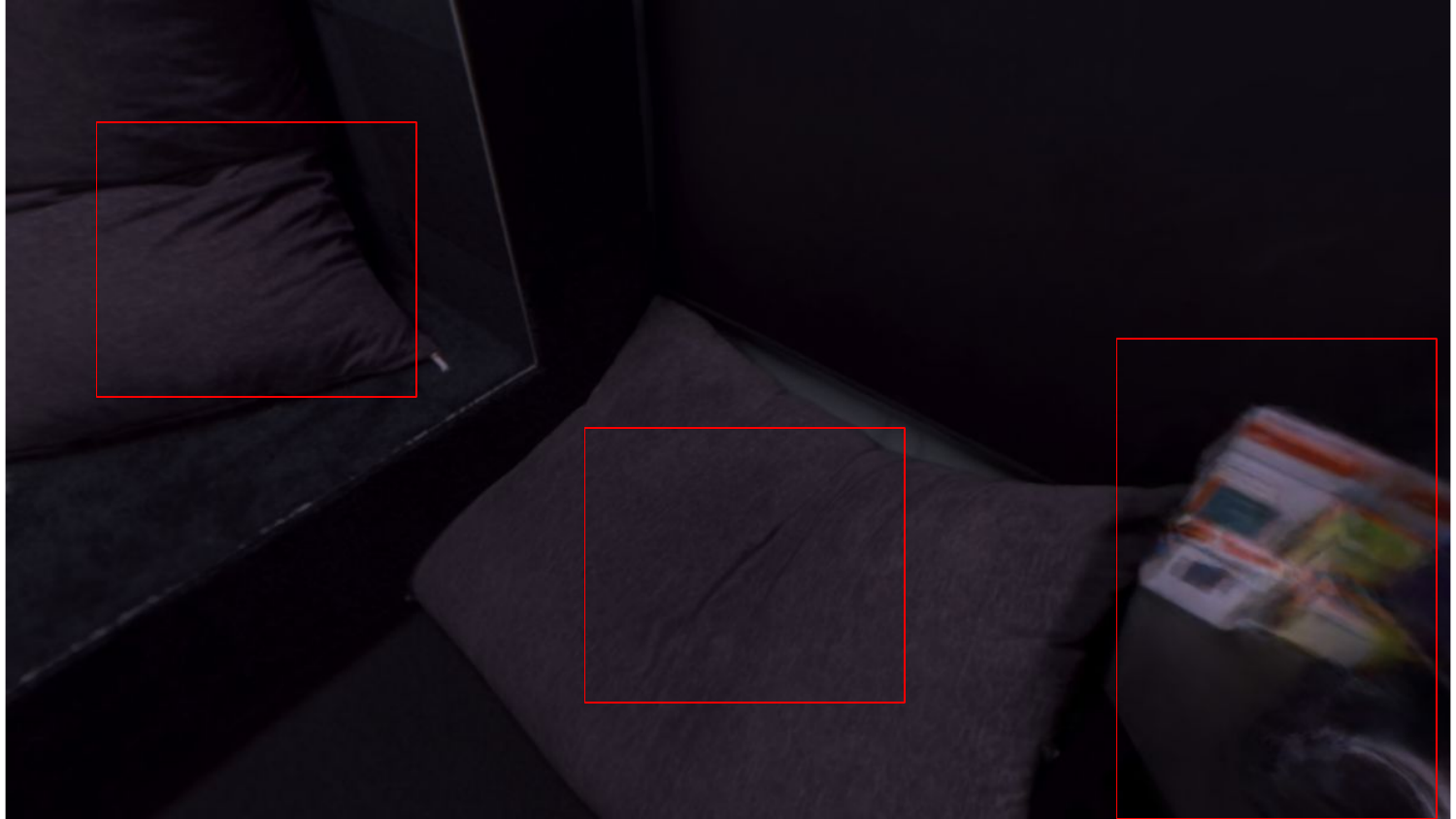} } & \raisebox{-0.5\height}{\includegraphics[width=.19\textwidth]{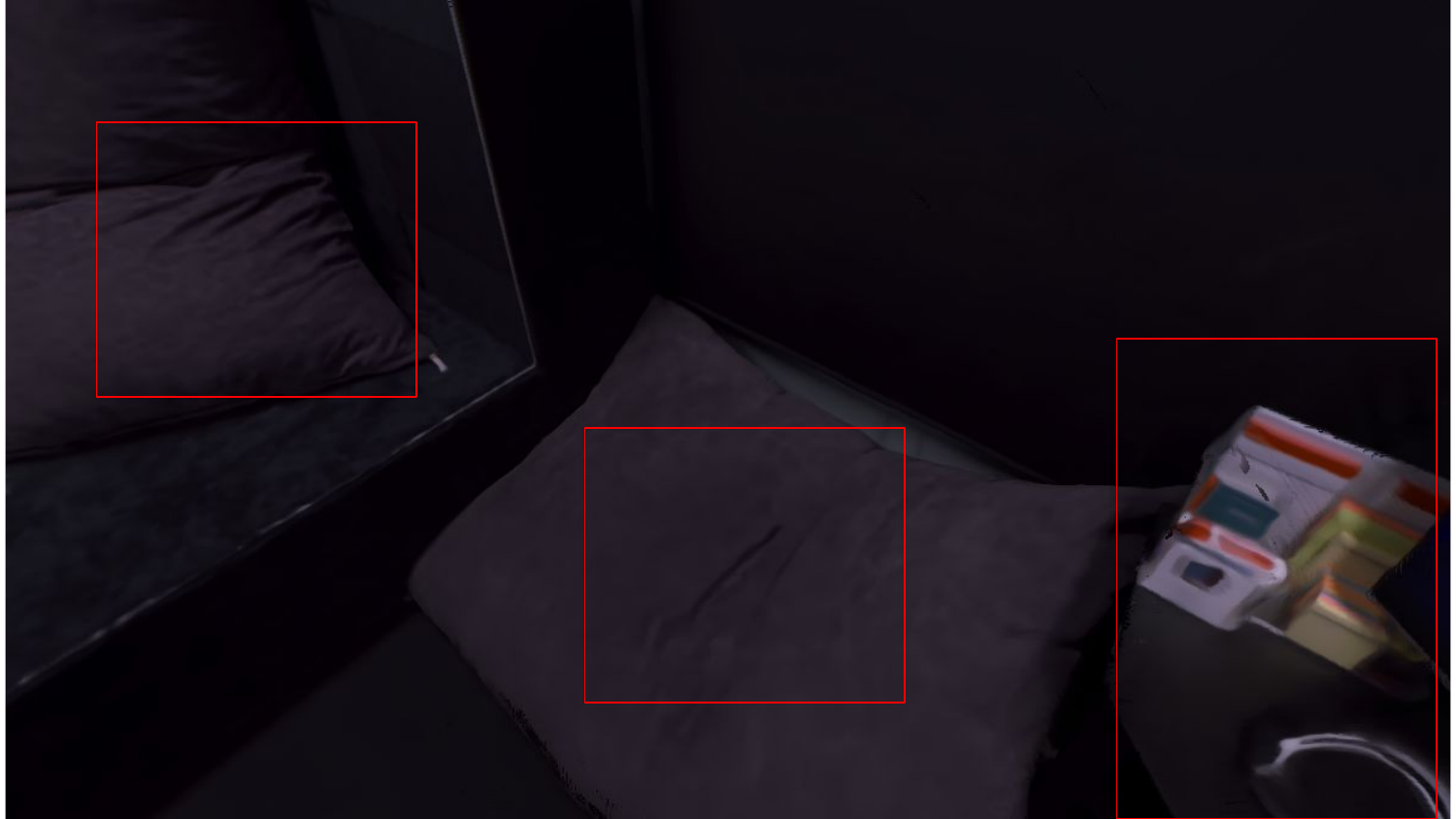} } & \raisebox{-0.5\height}{\includegraphics[width=.19\textwidth]{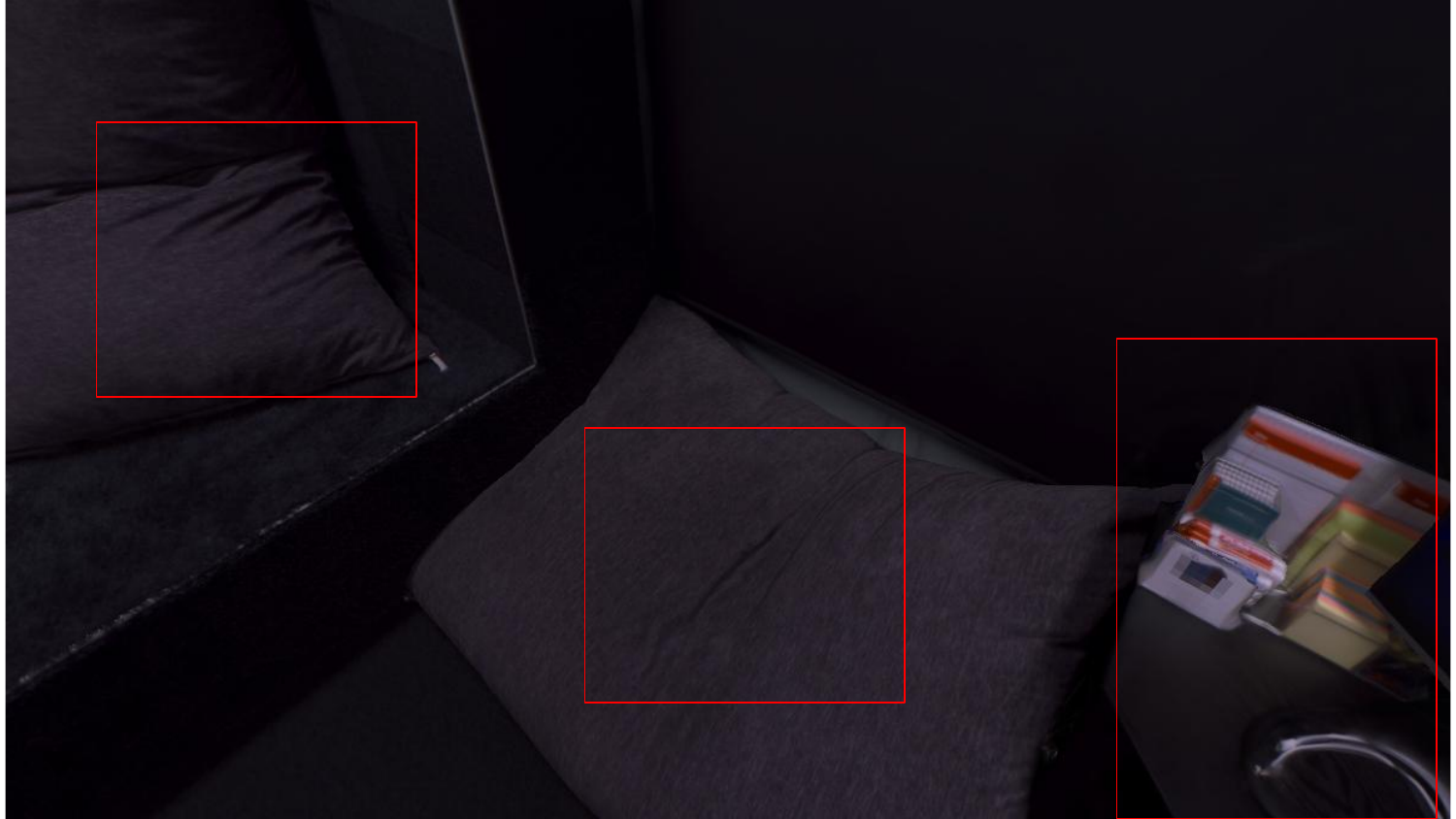} } \\
             \rotatebox[origin=b]{90}{ \texttt{Office 4}} & \raisebox{-0.5\height}{\includegraphics[width=.19\textwidth]{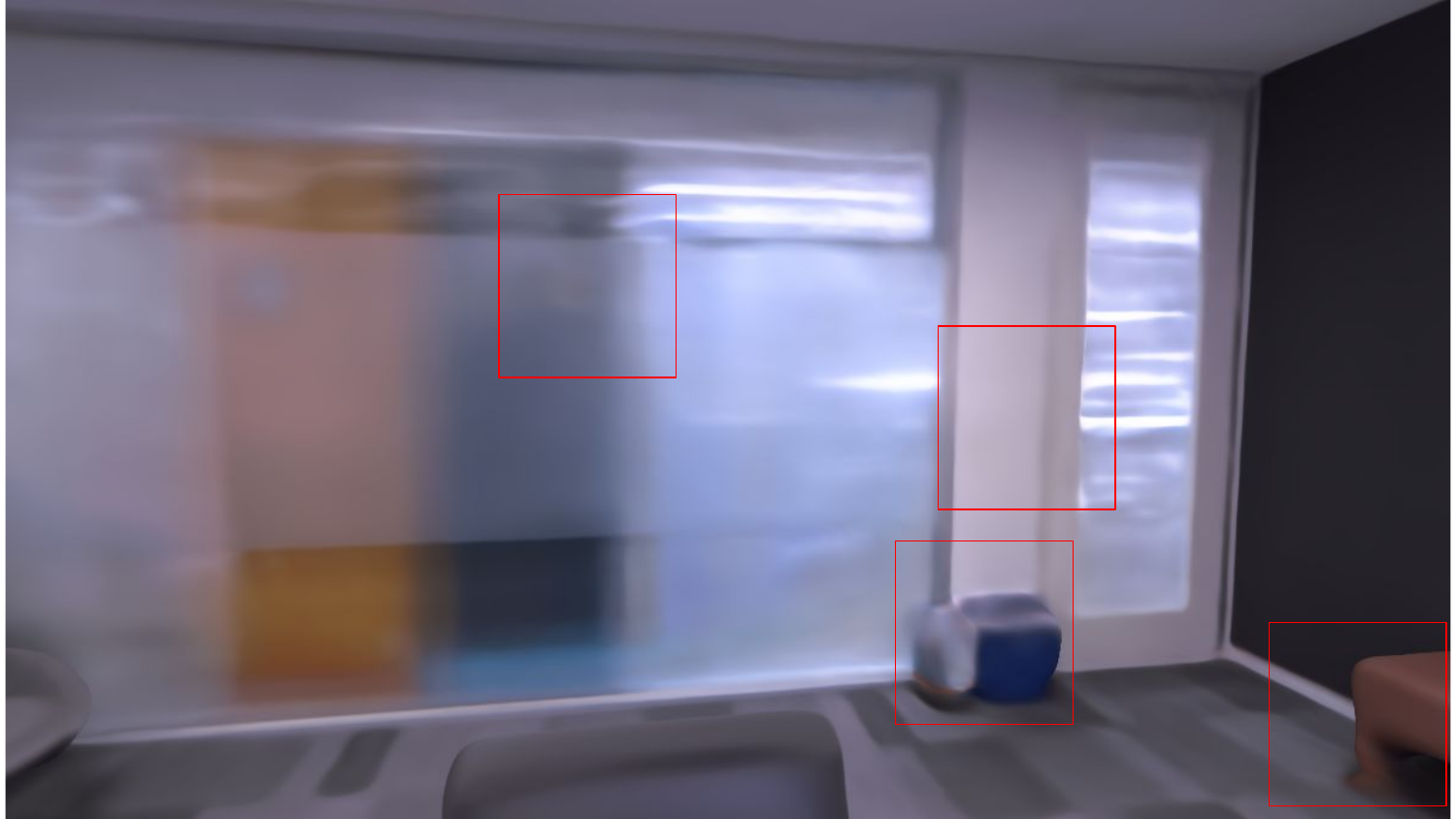} } & \raisebox{-0.5\height}{\includegraphics[width=.19\textwidth]{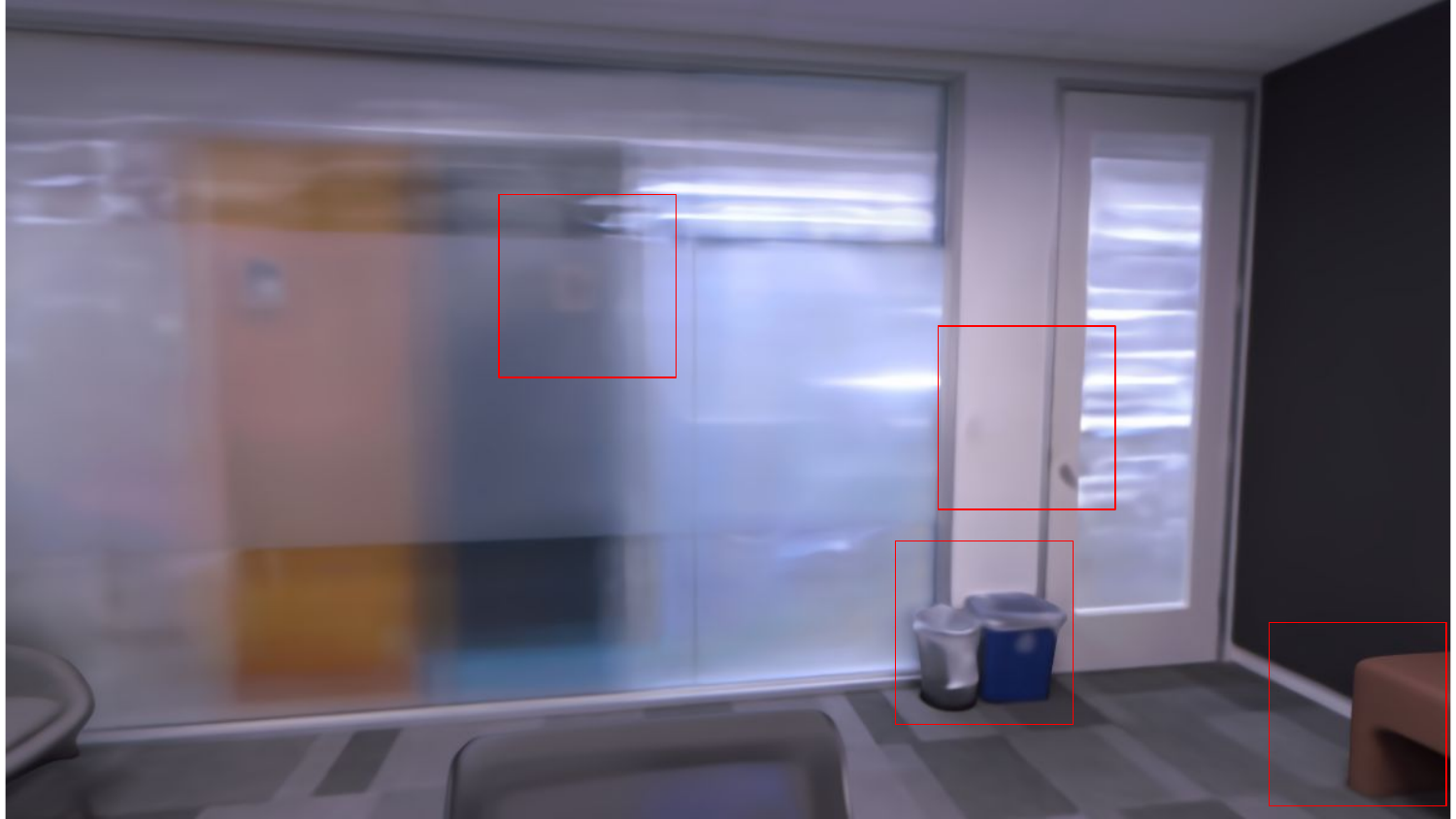} }
             &\raisebox{-0.5\height}{\includegraphics[width=.19\textwidth]{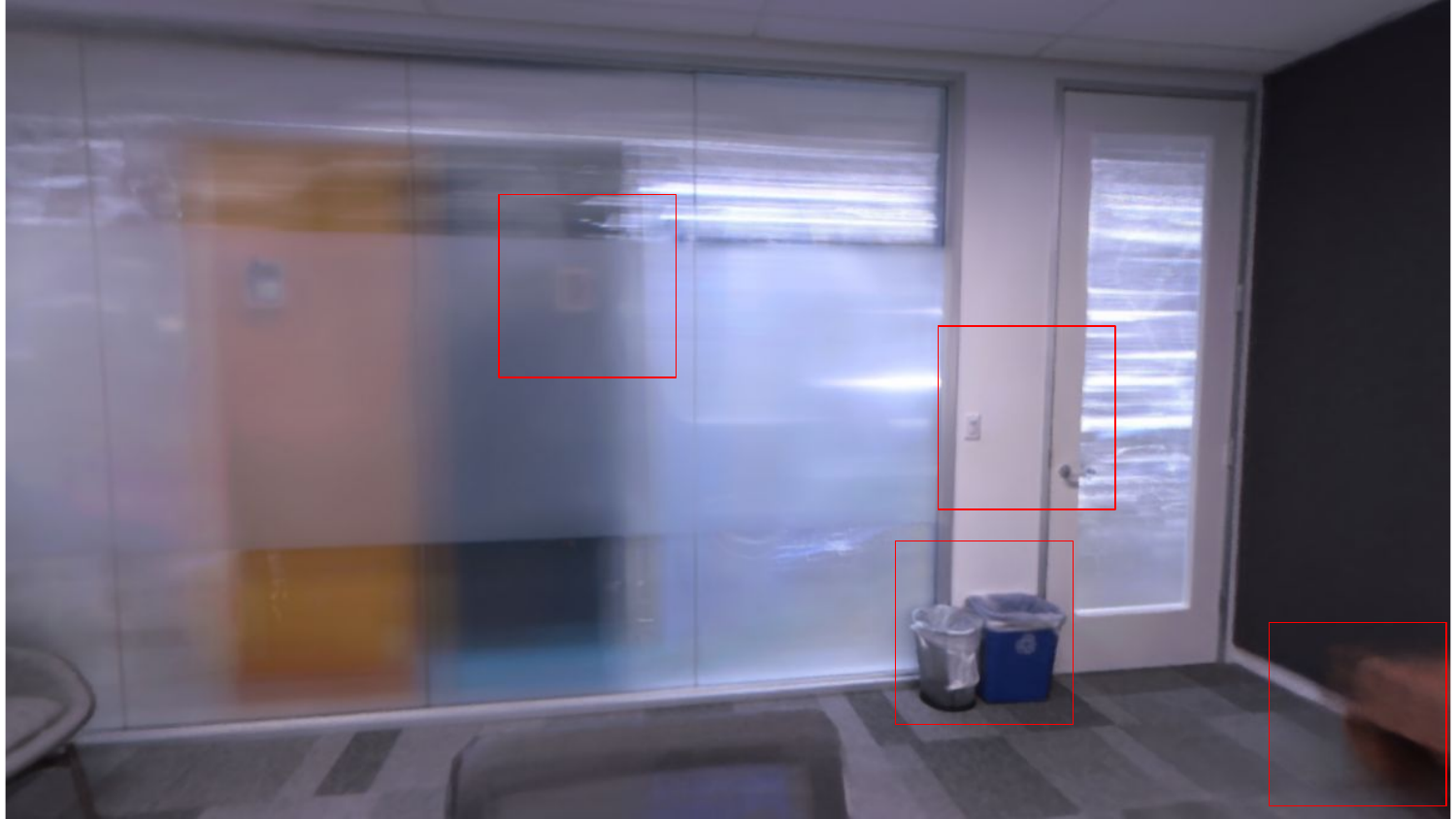} } & \raisebox{-0.5\height}{\includegraphics[width=.19\textwidth]{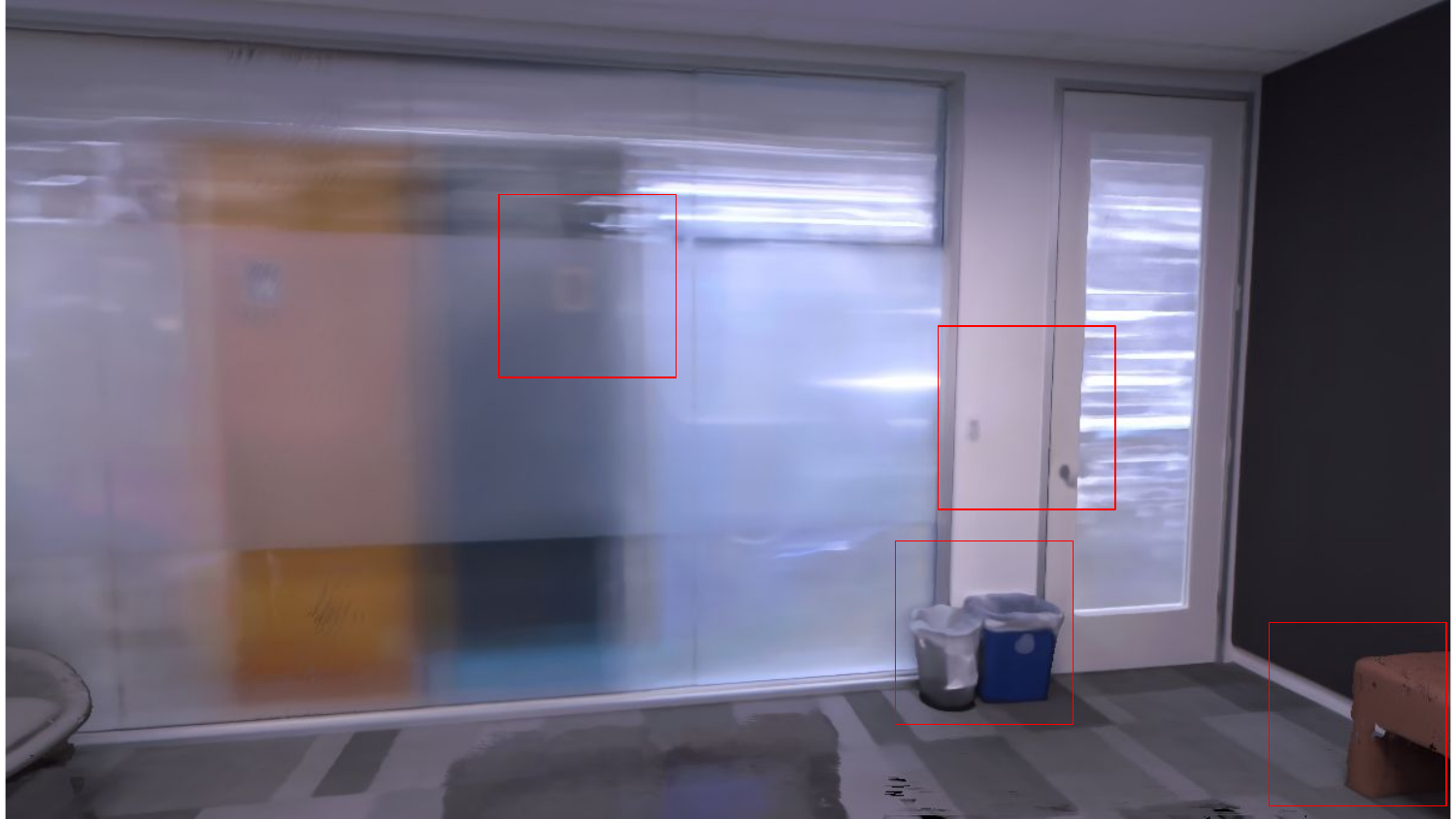} } & \raisebox{-0.5\height}{\includegraphics[width=.19\textwidth]{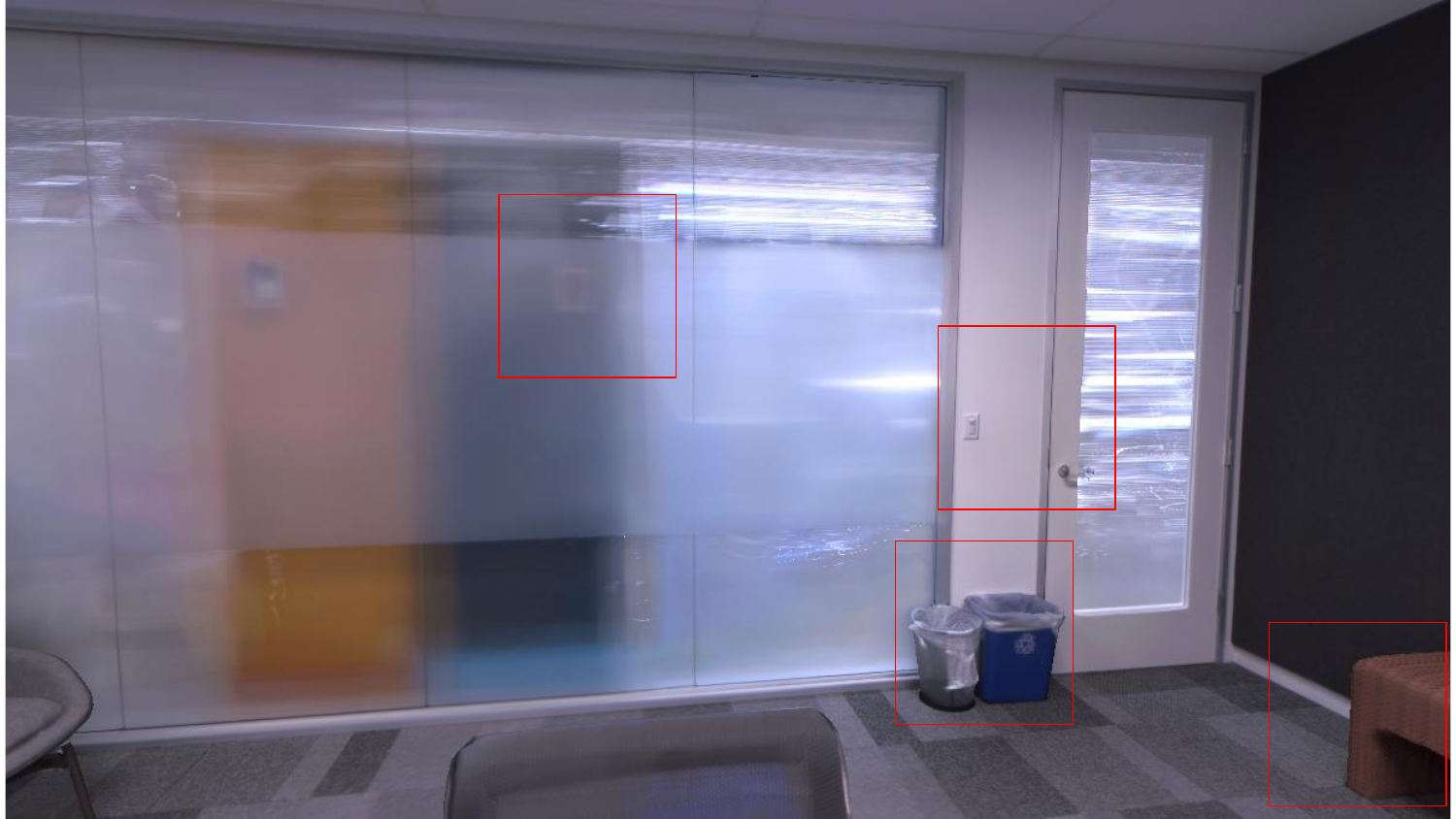} } \\
             \rotatebox[origin=b]{90}{ \texttt{0050}} & \raisebox{-0.5\height}{\includegraphics[width=.19\textwidth]{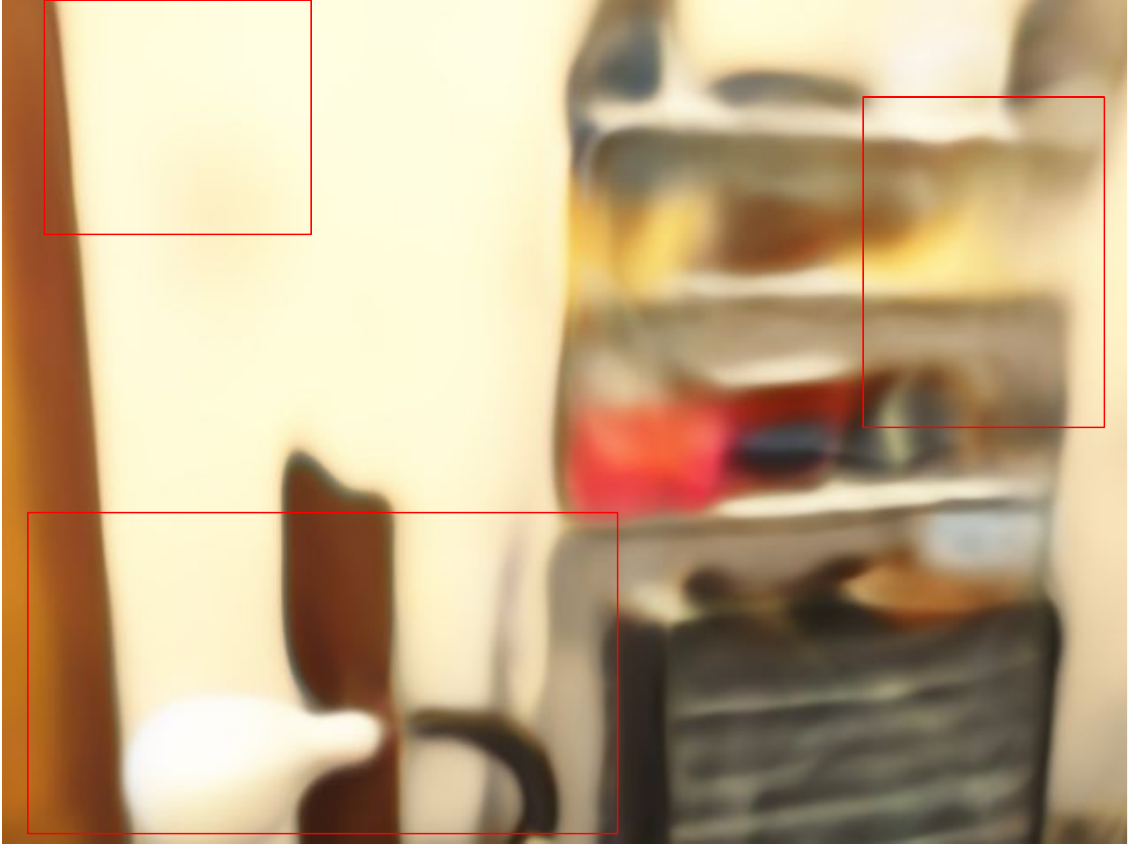} } & \raisebox{-0.5\height}{\includegraphics[width=.19\textwidth]{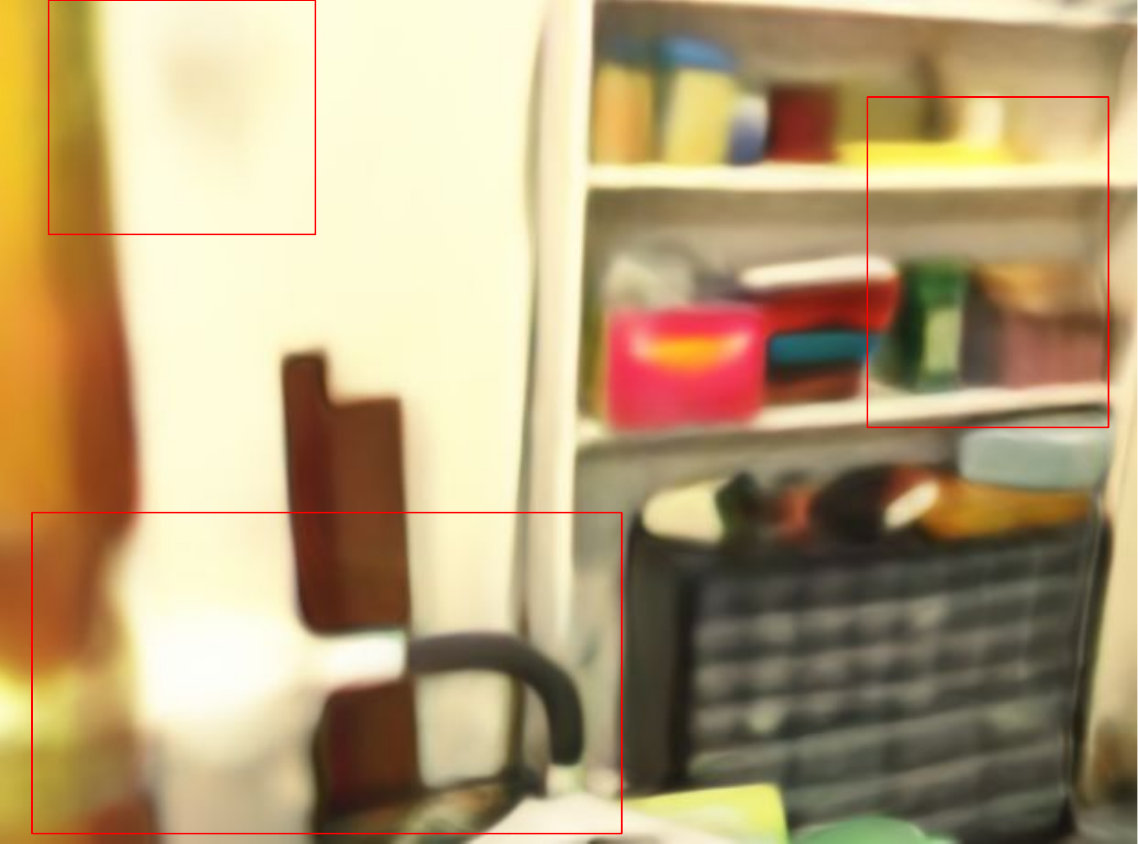} }
             &\raisebox{-0.5\height}{\includegraphics[width=.19\textwidth]{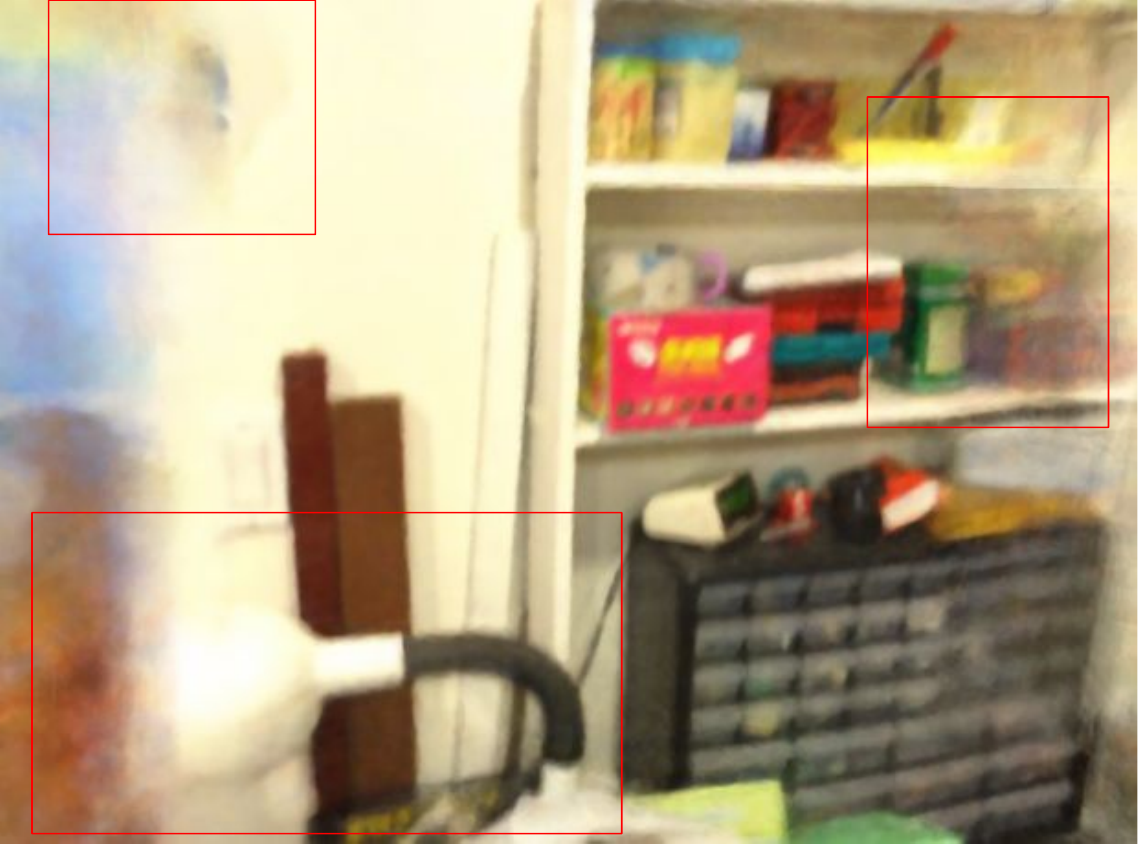} } & \raisebox{-0.5\height}{\includegraphics[width=.19\textwidth]{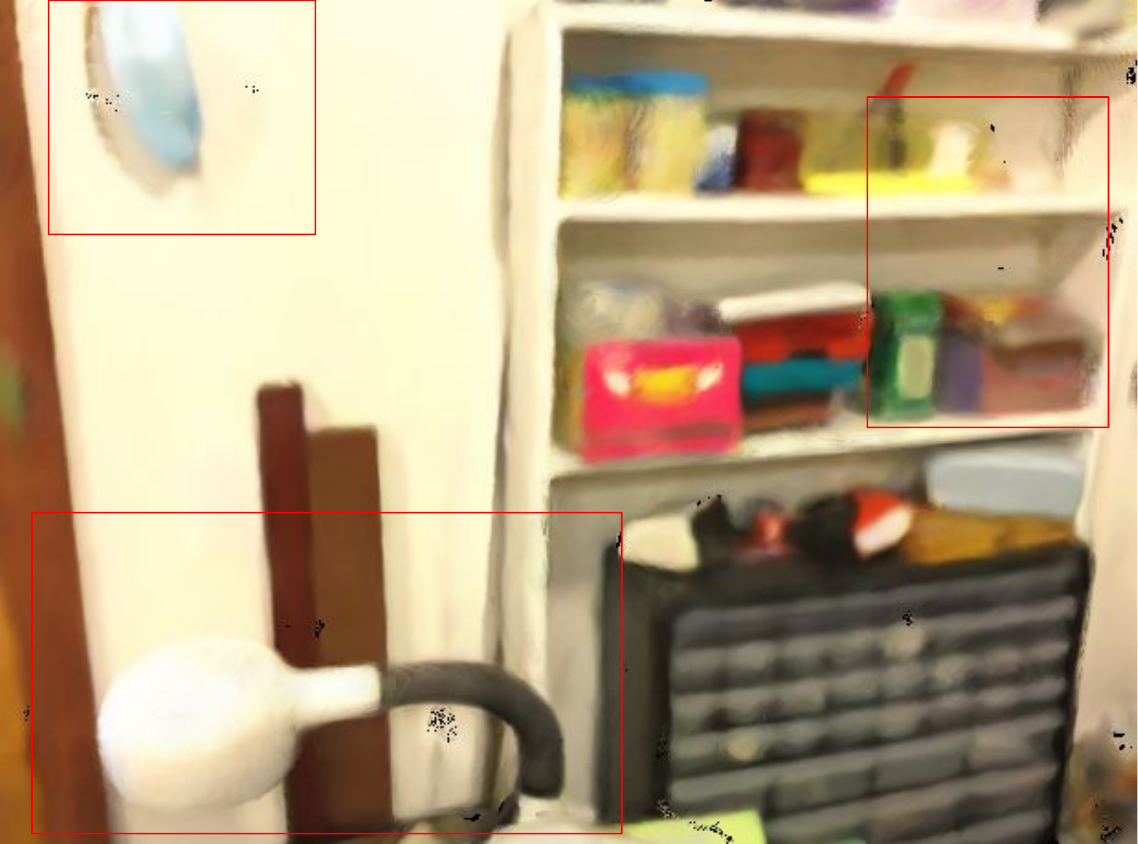} } & \raisebox{-0.5\height}{\includegraphics[width=.19\textwidth]{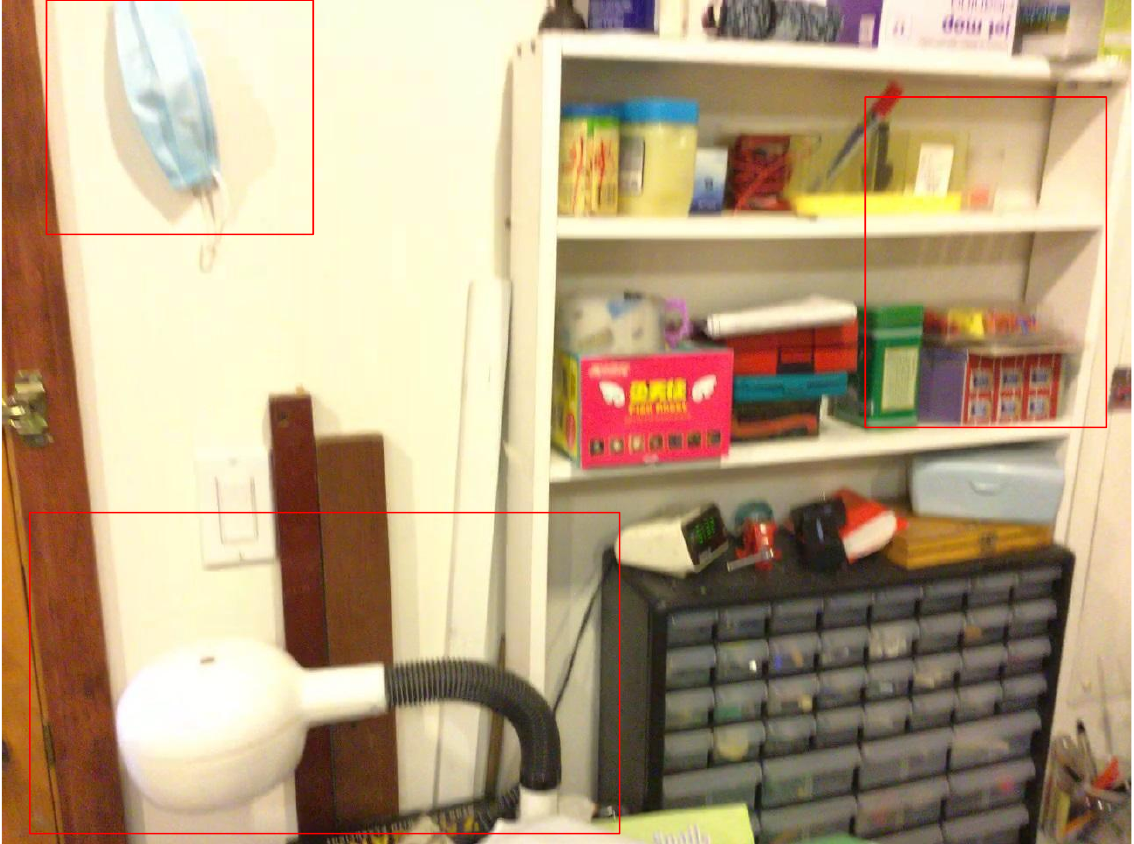} } \\
              \rotatebox[origin=b]{90}{ \texttt{0084}} & \raisebox{-0.5\height}{\includegraphics[width=.19\textwidth]{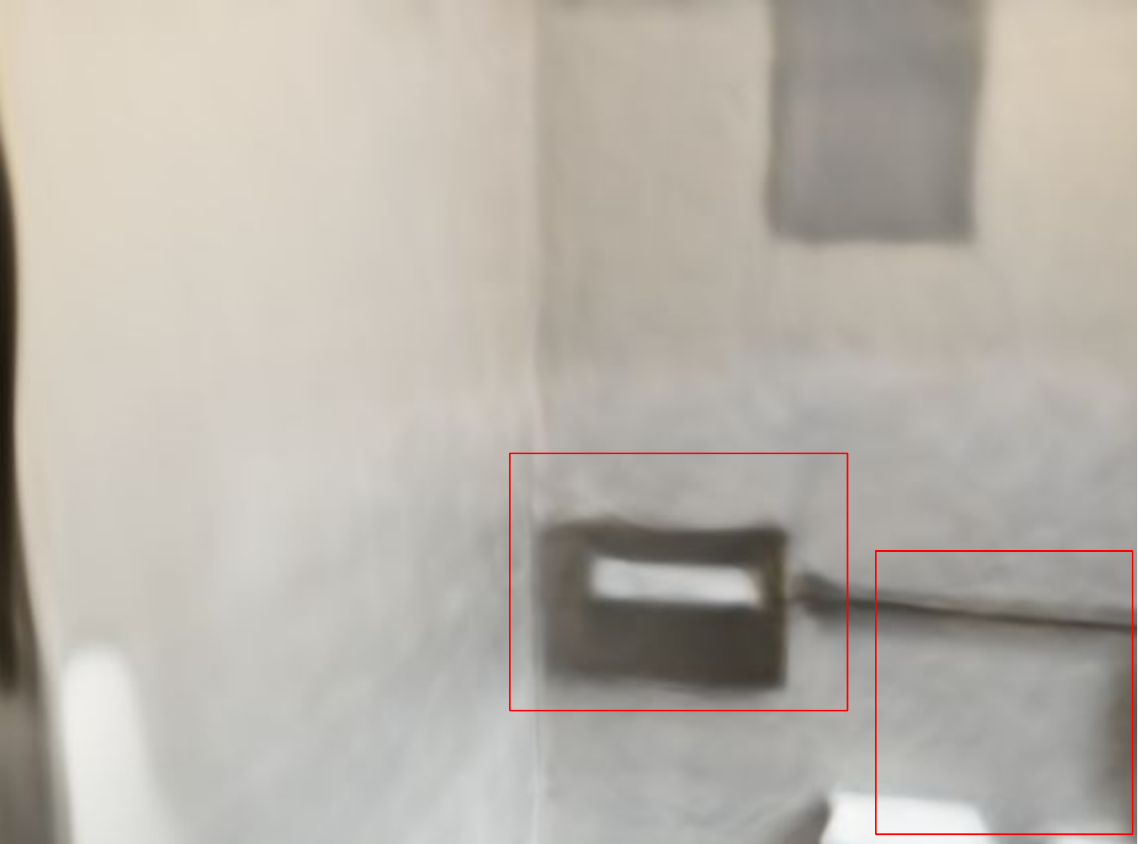} } & \raisebox{-0.5\height}{\includegraphics[width=.19\textwidth]{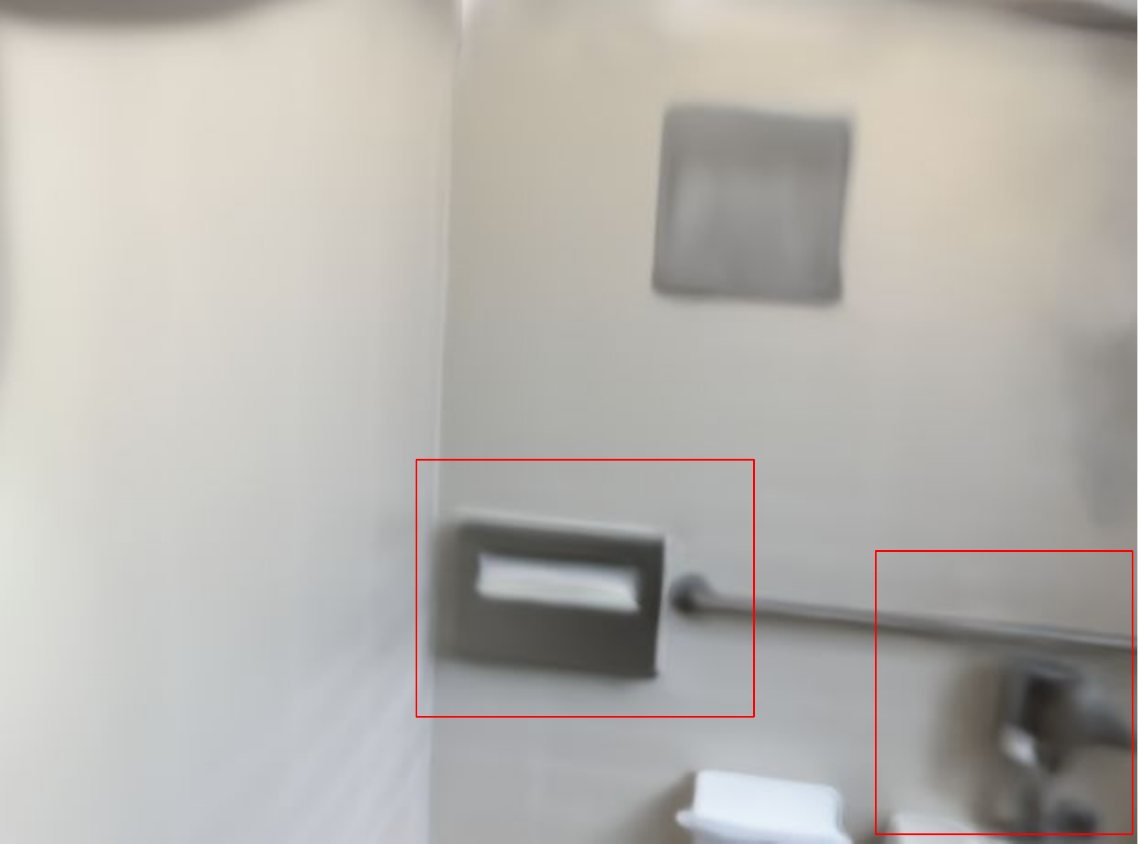} }
             &\raisebox{-0.5\height}{\includegraphics[width=.19\textwidth]{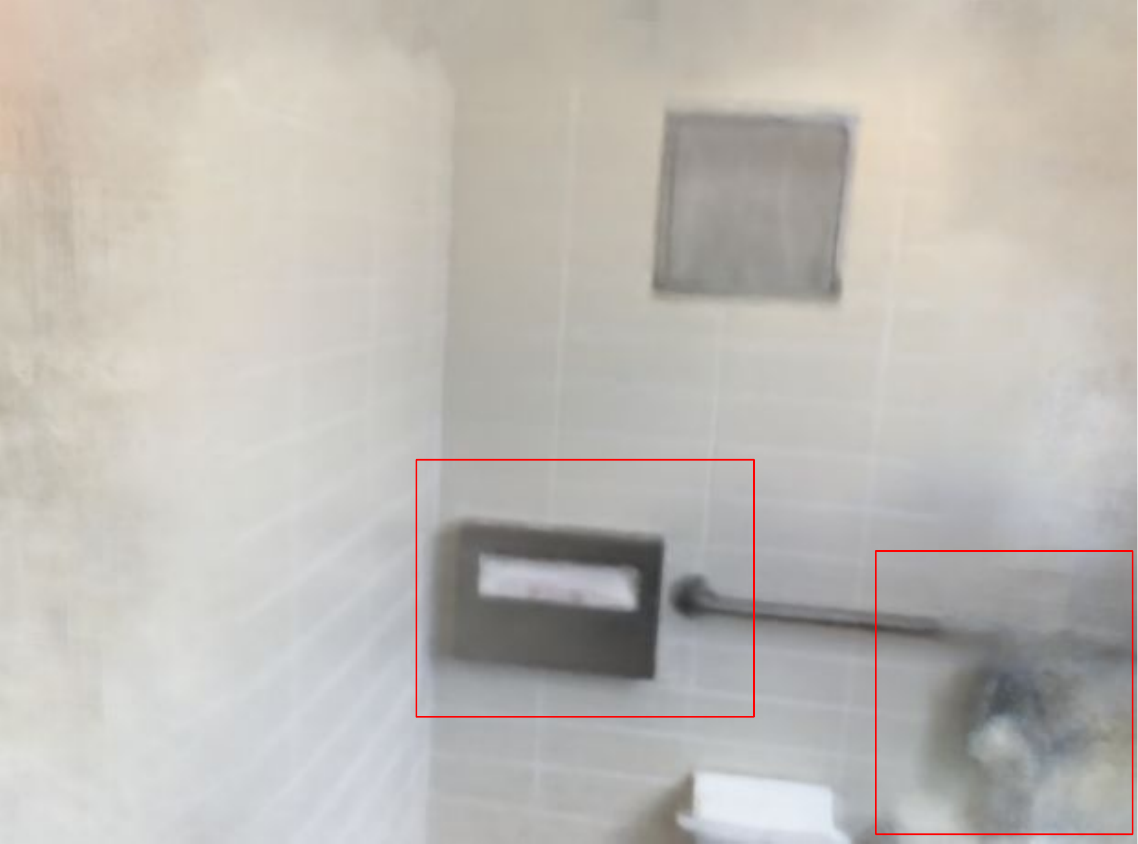} } & \raisebox{-0.5\height}{\includegraphics[width=.19\textwidth]{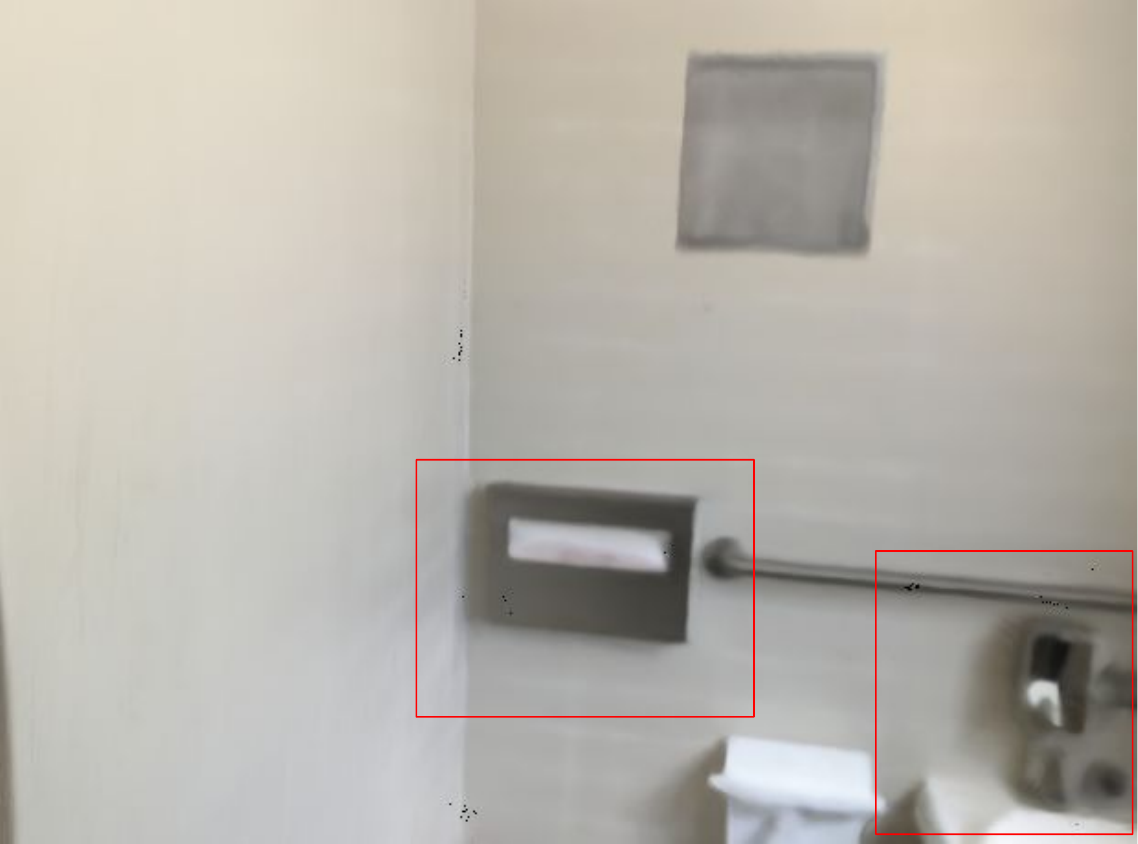} } & \raisebox{-0.5\height}{\includegraphics[width=.19\textwidth]{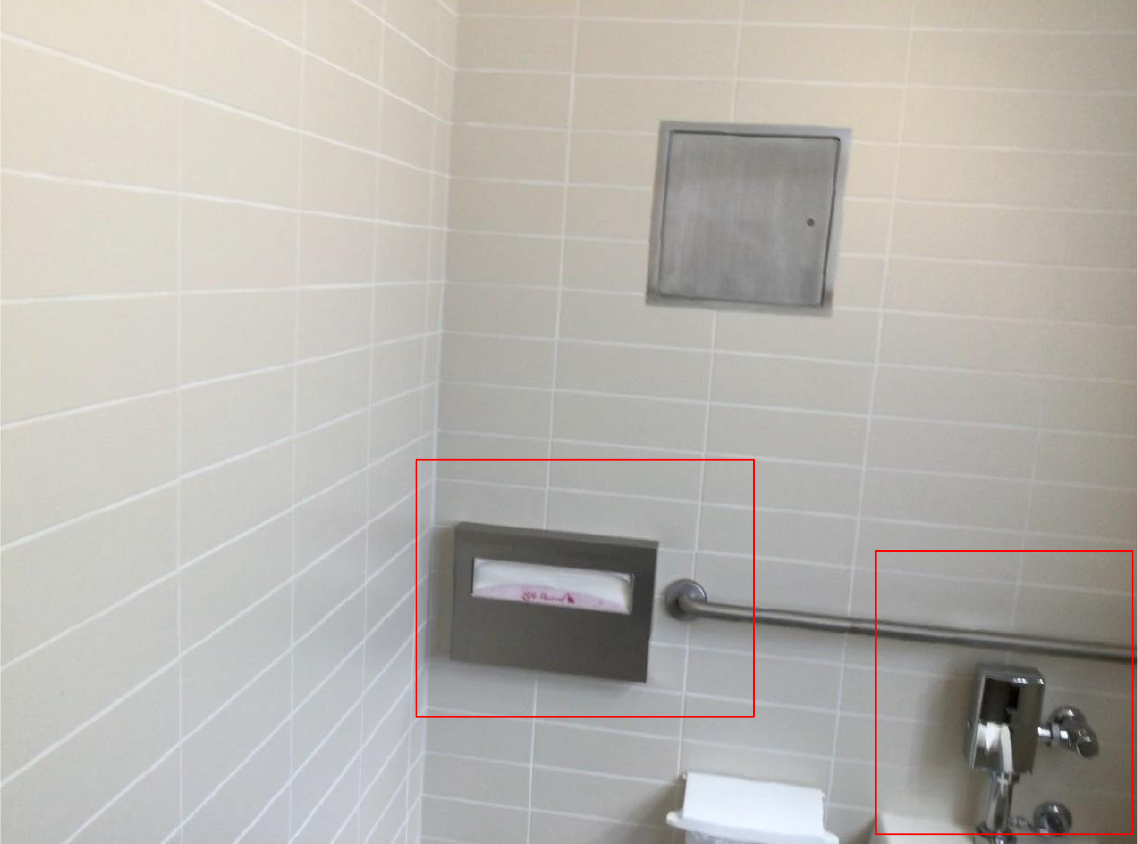} } \\
             \rotatebox[origin=b]{90}{ \texttt{0580}} & \raisebox{-0.5\height}{\includegraphics[width=.19\textwidth]{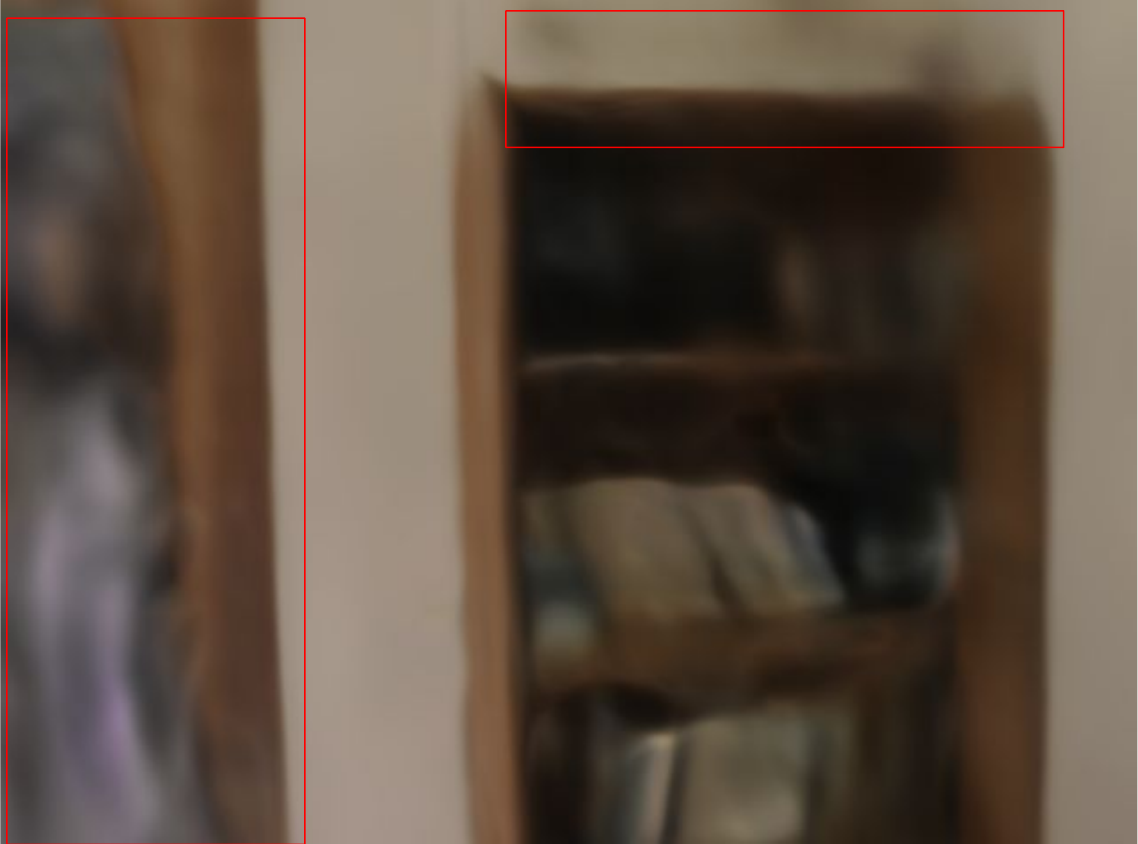} } & \raisebox{-0.5\height}{\includegraphics[width=.19\textwidth]{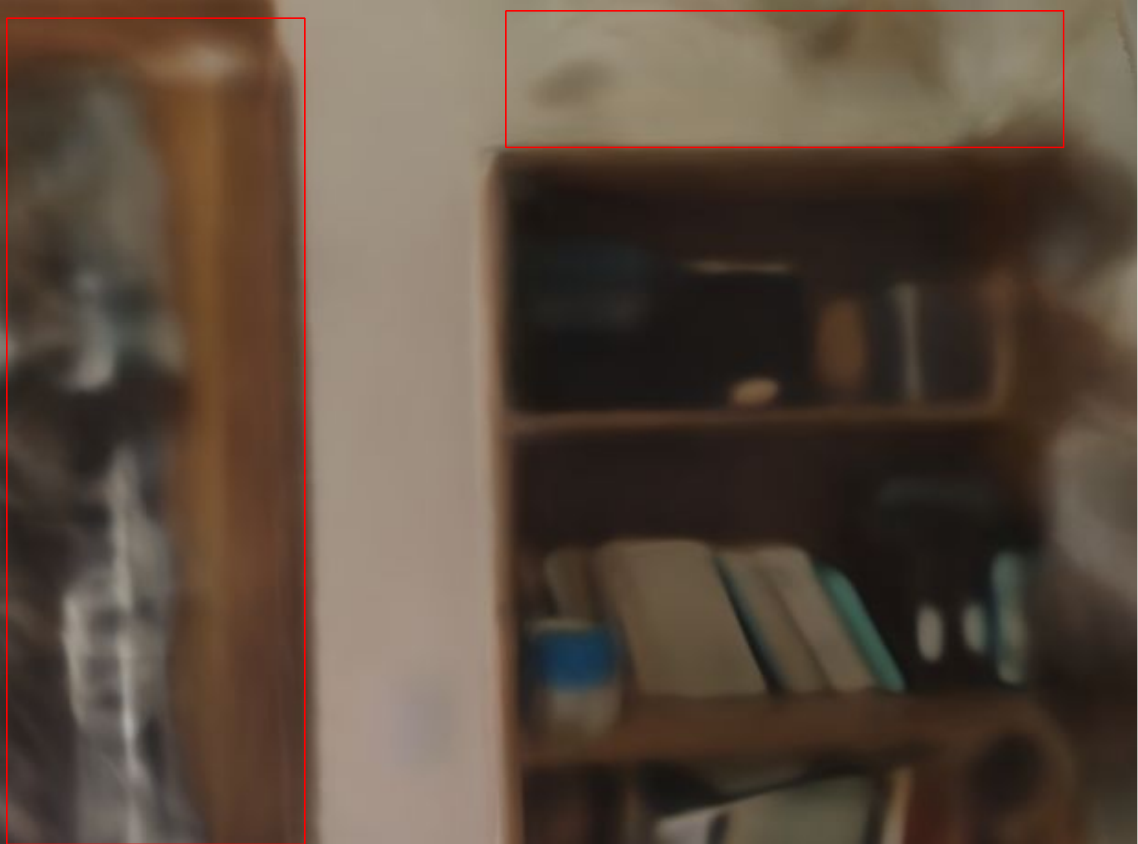} }
             &\raisebox{-0.5\height}{\includegraphics[width=.19\textwidth]{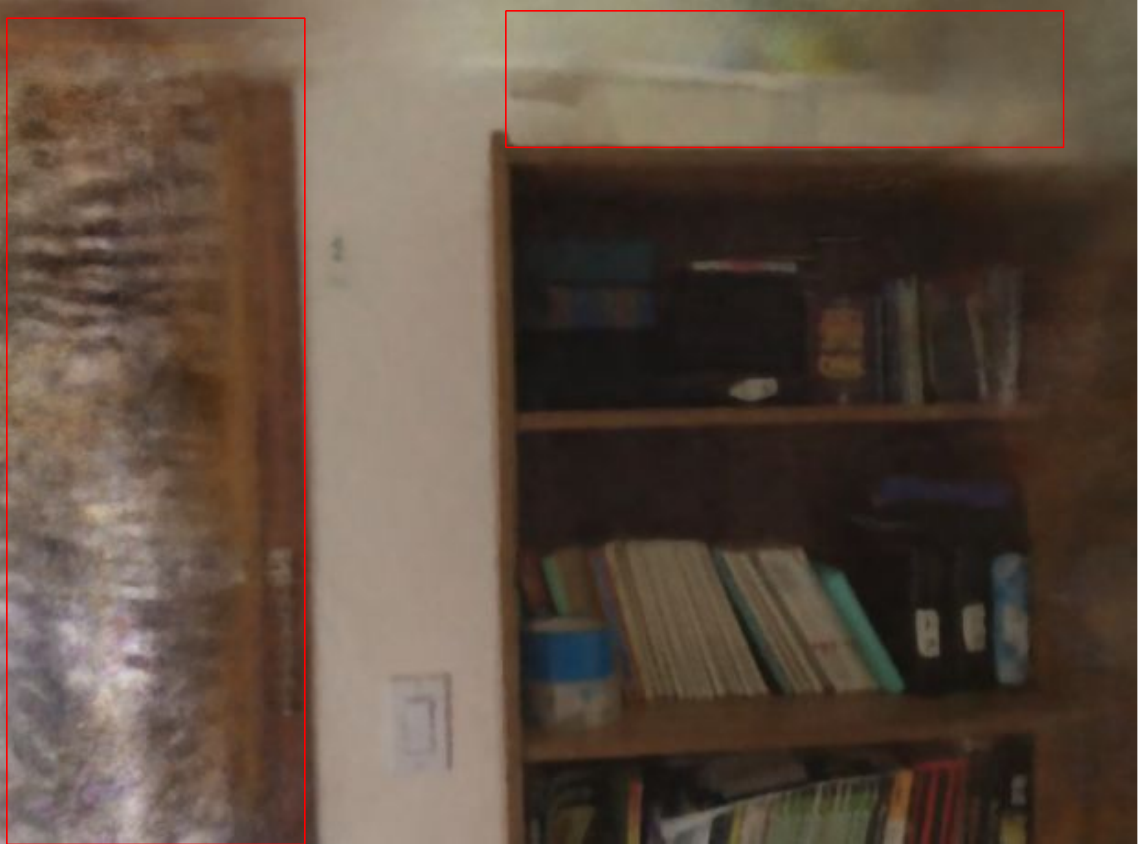} } & \raisebox{-0.5\height}{\includegraphics[width=.19\textwidth]{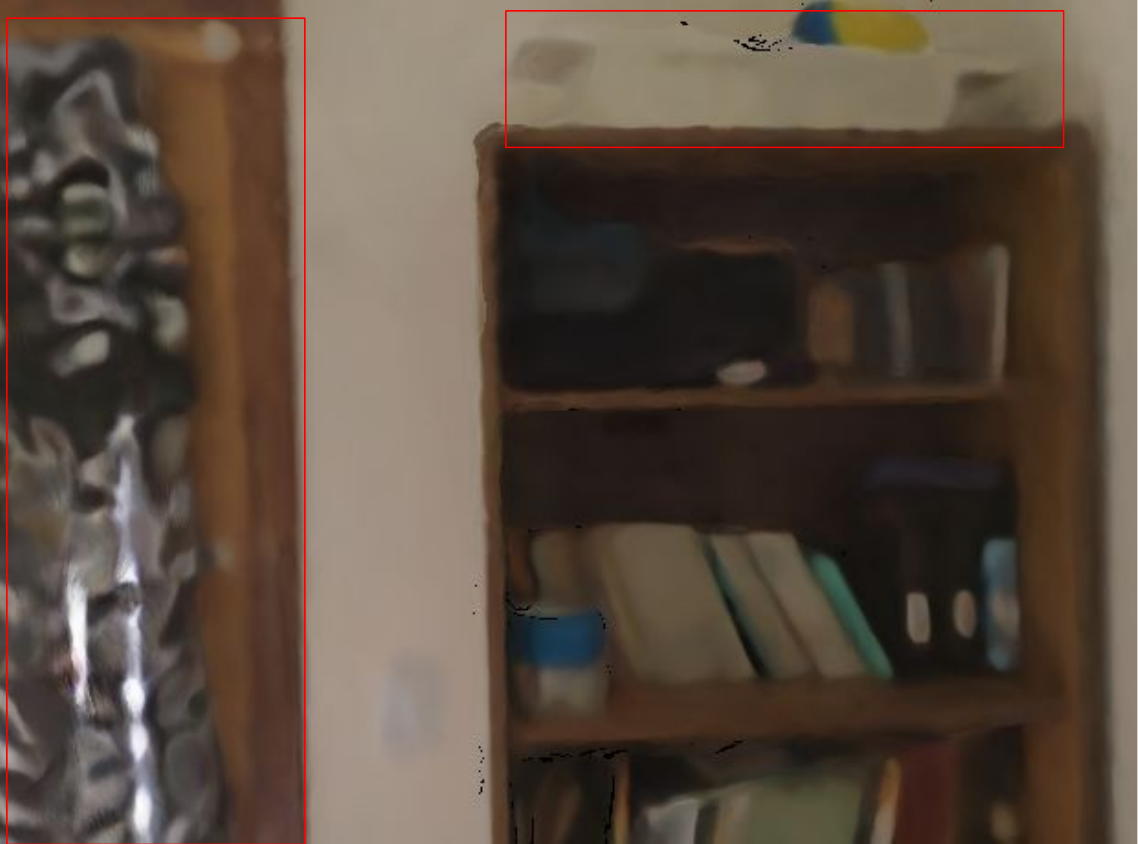} } & \raisebox{-0.5\height}{\includegraphics[width=.19\textwidth]{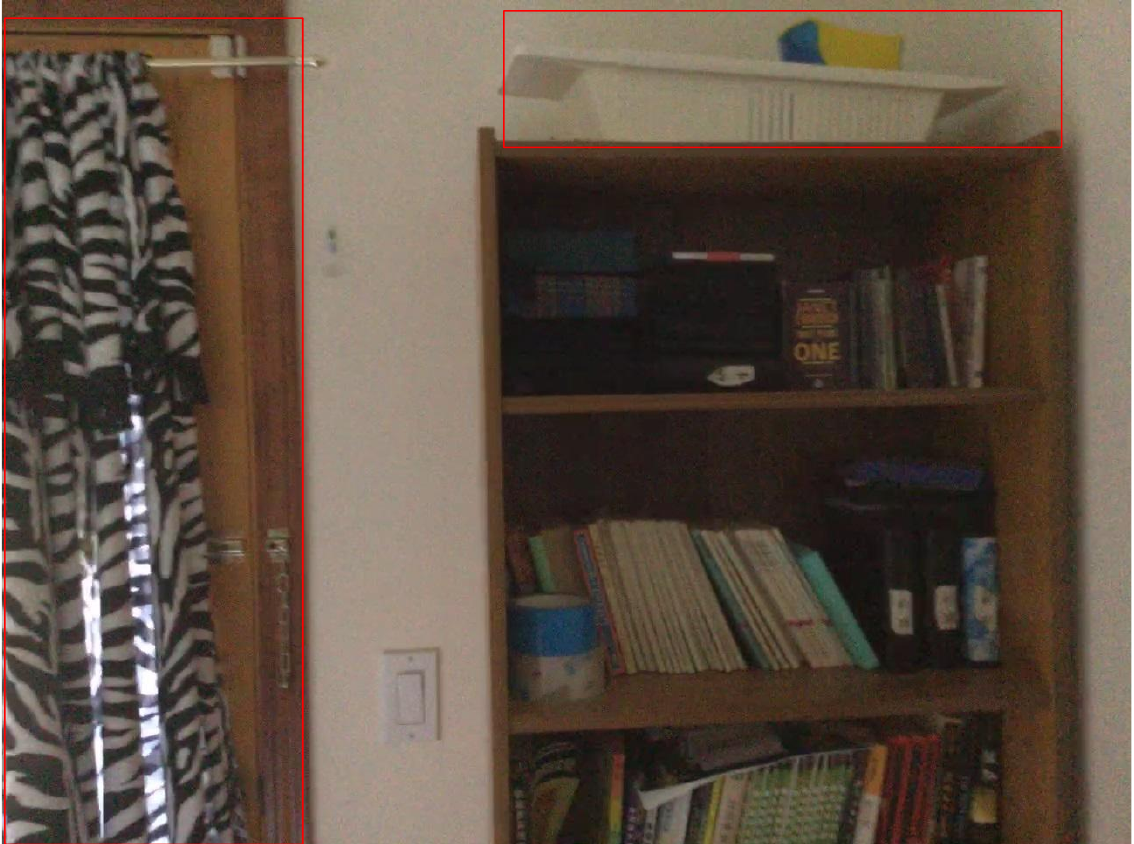} } \\
             & \textbf{Manhattan-} & \textbf{MonoSDF} & \textbf{Neuralangelo} & \textbf{VF-NeRF} & \textbf{Ground} \\
              & \textbf{SDF} &  &  & \textbf{(Ours)} & \textbf{truth} \\
        \end{tabular}
        \caption[Novel view synthesis qualitative results]{\textbf{Novel view synthesis qualitative results.}}
        \label{fig:view-synthesis}
    \end{figure*}

\Cref{fig:view-synthesis} presents qualitative comparisons of novel view synthesis for each scene on Replica and Scannet. \cref{tab:psnr} showcases the quantitative results for each individual scene in both datasets.

\setlength{\tabcolsep}{.5pt}

\setlength{\tabcolsep}{.5pt}

\end{document}